\PassOptionsToPackage{table}{xcolor}
\documentclass[pdflatex,sn-mathphys-ay]{sn-jnl}
\usepackage{multirow}

\makeatletter
\renewcommand{\paragraph}{%
  \@startsection{paragraph}{4}%
  {\z@}{0ex \@plus 1ex \@minus .2ex}{-1em}%
  {\normalfont\normalsize\bfseries}%
}
\makeatother

\usepackage{CJKutf8}
 
\usepackage[figuresright]{rotating}
\usepackage{amsmath,graphicx,fullpage}
\usepackage{booktabs}
\usepackage{multirow}
\usepackage{array}

\usepackage{booktabs}  % ?\usepackage{amsmath}   % 

\usepackage{amssymb}
\usepackage{pifont}    %  \ding{51} ??\ding{55} ?\usepackage{graphicx}  % 
\usepackage{longtable} % 
\usepackage{enumerate}
\usepackage{pdflscape}
\usepackage{color}
\usepackage{url}
\usepackage{float}
\newcolumntype{x}[1]{%
>{\centering\hspace{0pt}}p{#1}}%

\usepackage{color}
\usepackage{times}
\usepackage{helvet}
\usepackage{courier}
\usepackage{amsmath}
\usepackage{amssymb}
\usepackage{amsthm}
\usepackage{graphicx}
\usepackage{array}
\usepackage{booktabs}
\usepackage{amsopn}
\usepackage{amsmath,bm}
\usepackage{algorithm}
\usepackage{algorithmic}
\usepackage{enumerate}
\usepackage{color}
\usepackage{multirow}
\usepackage{bbding}
\usepackage{ragged2e}
\usepackage{threeparttable}
\usepackage{threeparttable} 
\usepackage{dsfont} 
\usepackage{makecell}
\usepackage{url}            % simple URL typesetting
\usepackage{multirow}
\usepackage{microtype}
\usepackage{cite}
\usepackage{outlines}
\usepackage{xcolor,colortbl}
\definecolor{tpamiblue}{rgb}{0.13333333333333333, 0, 1}
\usepackage[capitalize]{cleveref}
\crefname{section}{Sec.}{Secs.}
\Crefname{section}{Section}{Sections}

\crefname{figure}{Fig.}{Figs.}
\Crefname{figure}{Figure}{Figures}

\crefname{table}{Tab.}{Tabs.}
\Crefname{table}{Table}{Tables}

\crefname{algorithm}{Algo.}{Algos.}
\Crefname{algorithm}{Algorithm}{Algorithms}

\usepackage{tikz}
\usetikzlibrary{positioning, arrows.meta, calc}
\usetikzlibrary{shadows}
\usetikzlibrary{trees,positioning,shapes,shadows,arrows}
\usepackage[edges]{forest}
\definecolor{hiddendraw}{RGB}{205, 44, 36}

\tikzset{
  root/.style = {draw=blue!80!black, thick, fill=blue!30!white, rounded corners=4pt,
                 font=\sffamily\bfseries, text=blue!90!black, align=center,
                 minimum height=1cm, drop shadow={opacity=0.3}},
  category/.style = {draw=teal!80!black, fill=teal!20, rounded corners=4pt,
                     font=\sffamily\small, text width=2.5cm, minimum height=0.8cm,
                     align=center, inner sep=3pt},
  method/.style = {draw=orange!80!black, fill=orange!15, font=\footnotesize\sffamily,
                   text width=2.2cm, align=center, inner sep=2pt},
  technique/.style = {draw=violet!70!black, fill=violet!10, rounded corners=3pt,
                      font=\scriptsize\sffamily, text width=6cm, align=left, inner sep=4pt,
                      path fading=east},
  edge/.style = {draw=gray!50, thick, ->, >=latex}
}

\newcommand{\cmark}{\ding{51}}
\newcommand{\xmark}{\ding{55}}

\newcommand{\eg}{\emph{e.g.}}

\graphicspath{{figs/}}
\setkeys{Gin}{draft=false}
\definecolor{blue1}{RGB}{3, 4, 94}
\definecolor{blue2}{RGB}{2, 62, 138}
\definecolor{blue3}{RGB}{0, 119, 188}
\definecolor{blue4}{RGB}{0, 150, 199}
\definecolor{purple}{rgb}{0.48,0.12,0.64}

\definecolor{custommagenta}{RGB}{219, 79, 150}
\definecolor{custompurple}{RGB}{112, 81, 211}
\definecolor{customblue}{RGB}{80, 96, 221}
\definecolor{reviewerteal}{rgb}{0.0, 0.5, 0.5}
\definecolor{reviewersienna}{rgb}{0.62, 0.32, 0.18}
\definecolor{reviewercrimson}{rgb}{0.65, 0.10, 0.10}

\newif\ifmarkupversion
\markupversionfalse  % uncomment for clean submission version

\ifmarkupversion
  \colorlet{revblue}{blue}
  \colorlet{revgreen}{green!50!black}
  \colorlet{revpurple}{purple}
  \colorlet{revorange}{orange!90!black}
  \colorlet{revteal}{reviewerteal}
  \colorlet{revsienna}{reviewersienna}
  \colorlet{revcrimson}{reviewercrimson}
\else
  \colorlet{revblue}{black}
  \colorlet{revgreen}{black}
  \colorlet{revpurple}{black}
  \colorlet{revorange}{black}
  \colorlet{revteal}{black}
  \colorlet{revsienna}{black}
  \colorlet{revcrimson}{black}
\fi

\definecolor{darkblue}{rgb}{0, 0.08235, 0.43921}

\usepackage{fancyhdr}
\usepackage{xspace} 

\def\@onedot{%
    \ifx\@let@token.%
    \else.%
    \null% 
    \fi%
    \xspace%
}

\makeatletter
\DeclareRobustCommand\onedot{\futurelet\@let@token\@onedot}
\def\@onedot{\ifx\@let@token.\else.\null\fi\xspace}

\def\eg{\emph{e.g}\onedot}

\makeatother

\begin{document}

\title[Visual enhancement and 3D representation for underwater scenes: a review]{Visual enhancement and 3D representation for underwater scenes: a review}


\author*[1]{\fnm{Guoxi} \sur{ Huang}}\email{guoxi.huang@bristol.ac.uk}

\author[1]{\fnm{Haoran} \sur{Wang}}

\author[2]{\fnm{Brett } \sur{Seymour}}

\author[3]{\fnm{ Evan  } \sur{Kovacs}}
\author[4]{\fnm{John  } \sur{Ellerbroc}}
\author[5]{\fnm{ Dave  } \sur{Blackham}}
\author*[1]{\fnm{ Nantheera  } \sur{Anantrasirichai}}\email{n.anantrasirichai@bristol.ac.uk}

\affil[1]{\orgdiv{Visual Information Laboratory}, \orgname{University of Bristol}, \country{UK}}

\affil[2]{\orgdiv{Submerged Resources Center}, \orgname{National Park Service}, \country{USA}}

\affil[3]{\orgdiv{Marine Imaging Technologies}, \orgname{LLC}, \country{USA}}

\affil[4]{\orgdiv{Gates Underwater Products}, \orgname{Inc}, \country{USA}}

\affil[5]{\orgdiv{Esprit film and television Ltd}, \country{UK}}


\abstract{Underwater visual enhancement (UVE) and underwater 3D reconstruction pose significant challenges in computer vision and AI-based tasks due to complex imaging conditions in aquatic environments. Despite the development of numerous enhancement algorithms, a comprehensive and systematic review covering both UVE and underwater 3D reconstruction remains absent. To advance research in these areas, we present an in-depth review from multiple perspectives. First, we introduce the fundamental physical models, highlighting the peculiarities that challenge conventional techniques. We survey advanced methods for visual enhancement and 3D reconstruction specifically designed for underwater scenarios. The paper assesses various approaches from non-learning methods to advanced data-driven techniques, including Neural Radiance Fields and 3D Gaussian Splatting, discussing their effectiveness in handling underwater distortions. Finally, we conduct both quantitative and qualitative evaluations of state-of-the-art UVE and underwater 3D reconstruction algorithms across multiple benchmark datasets. Finally, we highlight key research directions for future advancements in underwater vision.}


\maketitle

\section*{Statements and Declarations}

\noindent \textbf{Competing Interests:} The authors declare no competing interests.

\paragraph*{Funding}
This work was funded by the EPSRC ECR International Collaboration Grants (EP/Y002490/1) and the UKRI MyWorld Strength in Places Programme (SIPF00006/1).

\paragraph*{Author contributions}
G.H.: Conceptualization, Visualization, Writing - original draft, Writing - review \& editing;  H.W.: Writing - original draft,  Writing - review \& editing;
B.S: Supervision, Writing - review \& editing; E.K.: Supervision, Writing - review \& editing; 
J.E.:Supervision, Writing - review \& editing;
D.B.:Supervision, Writing - review \& editing;
N.A.:Supervision, Funding acquisition, Writing - original draft,  Writing - review\& editing.

\paragraph{Data Availability} No datasets were generated or analyzed during the current study.

\newpage

\tableofcontents

\newpage

\section{Introduction}
\label{sec:introduction}

Underwater imaging plays an increasingly important role in scientific exploration, industrial inspection, and environmental monitoring. Because more than 70\% of the Earth's surface is covered by water, vast biological resources, geological structures, archaeological remains, and critical infrastructure remain hidden beneath oceans, seas, lakes, and rivers. Visual observation of these submerged environments is therefore essential for applications such as {\color{revgreen}marine biology \citep{shaker2023impact}, archaeology \citep{10600694}, geological surveying \citep{yaqoob2025advancing}, and the inspection of subsea assets \citep{rahnama2025subsea}}, including pipelines and offshore platforms. As the need for long-term ecosystem monitoring, resource management, hazard mitigation, and digital preservation continues to grow, underwater imaging has become a key enabling technology for both scientific understanding and operational decision-making.

Despite its importance, underwater visual sensing remains substantially more challenging than imaging in air. In addition to the practical difficulty and cost of acquiring data in submerged environments, underwater images are severely degraded by wavelength-dependent absorption, scattering, non-uniform illumination, suspended particles, and refractive distortions introduced by camera housings. These factors reduce contrast, distort colour, blur fine structures, and complicate reliable visual analysis. As a result, raw underwater imagery often lacks the visual fidelity required not only for human interpretation but also for downstream computer vision tasks.

For this reason, underwater visual enhancement has become an important and widely adopted pre-processing step. By improving visibility, restoring colour balance, and recovering local contrast, enhancement methods can make underwater data more suitable for subsequent analysis, including inspection, scene understanding, and 3D reconstruction. In practice, many underwater reconstruction pipelines benefit from enhanced inputs because improved image quality can facilitate feature detection, matching, camera pose estimation, and dense reconstruction. Consequently, underwater visual enhancement is not merely a perceptual refinement step, but often a practical means of improving the usability of underwater imagery for downstream geometric processing.

Alongside enhancement, underwater 3D reconstruction has also attracted substantial attention. Traditional 3D mapping pipelines frequently rely on photogrammetry, where Structure from Motion (SfM), visual Simultaneous Localization and Mapping (visual SLAM), and Multi-View Stereo (MVS) are used to estimate camera motion and recover scene geometry from overlapping images \citep{ZHANG2022100510,Teague2017Underwater,Storlazzi:vslam:2016}. Compared with laser scanning, photogrammetry is often more cost-effective for covering larger areas while preserving visually rich texture. More recently, advances in machine learning have introduced powerful alternatives for both enhancement and reconstruction. Deep networks have been used to learn underwater restoration from paired, synthetic, or unpaired data, while modern 3D representations such as Neural Radiance Fields (NeRF) \citep{mildenhall2020nerf} and 3D Gaussian Splatting (3DGS) \citep{kerbl:3Dgaussians:2023} have shown strong potential for high-fidelity scene modelling from image collections.

Nevertheless, underwater conditions continue to challenge both enhancement and reconstruction. Severe degradation, unstable illumination, domain shifts, and limited high-quality reference data restrict the robustness of current methods. Moreover, although enhancement as a pre-processing step is often effective in practice, it is not always neutral with respect to downstream geometry. Enhancement may alter image statistics, local textures, edge structures, or cross-view photometric consistency, which can in turn influence matching, calibration, or reconstruction quality. This does not diminish the practical value of two-stage pipelines; rather, it suggests that the interaction between enhancement and reconstruction deserves closer examination. In this review, we therefore consider underwater visual enhancement primarily as an enabling pre-processing component for downstream reconstruction, while also identifying the analysis of cumulative error in sequential enhancement--reconstruction pipelines as an important future research direction.

\subsection{Peculiarities of Underwater Environments}

The major factors that make underwater imagery fundamentally different from terrestrial imagery. These include light absorption and scattering, non-uniform illumination, dynamic water conditions, marine snow, optical and geometric distortions, and dynamic scenes. Importantly, these factors degrade not only visual appearance but also the reliability of feature extraction, correspondence estimation, calibration, and geometric reconstruction.

\begin{figure}[t!]
    \centering
    \includegraphics[height=3.38cm, width=0.32\linewidth]{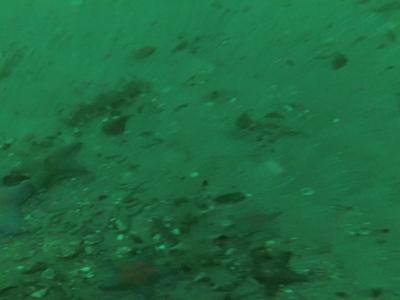}
    \includegraphics[height=3.38cm, width=0.32\linewidth]{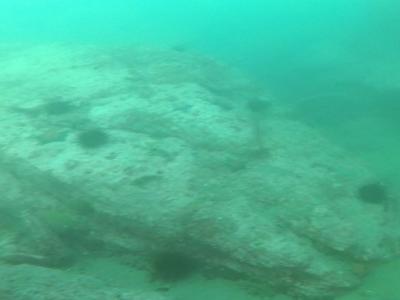}
    \includegraphics[height=3.38cm, width=0.32\linewidth]{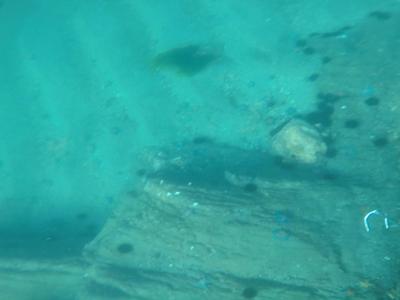}
    \caption{Examples of underwater images exhibiting wavelength-dependent colour casts and veiling effects~\citep{Liu:RealWorld:2020}.}
    \label{fig:diff_color_imgs}
\end{figure}

\paragraph{Light Absorption and Scattering}
Light attenuation in water is far more severe than in air and is governed by both absorption and scattering. Absorption, predominantly caused by water molecules and dissolved organic matter, is wavelength-dependent: red, orange, and yellow wavelengths are attenuated first, often giving underwater scenes a bluish or greenish appearance, as shown in \autoref{fig:diff_color_imgs}. In addition, suspended particles such as plankton and marine snow scatter the remaining light, creating a veil that reduces contrast and obscures fine details. These degradations make it significantly harder to detect reliable features and establish stable correspondences across views, both of which are essential for photogrammetric reconstruction. Correcting spectral imbalance and improving visibility are therefore central goals of underwater visual enhancement.

\begin{figure}[t!]
    \centering
    \includegraphics[height=2.38cm, width=0.24\linewidth]{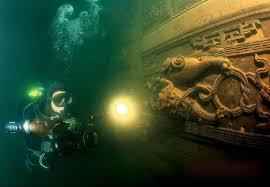} 
    \includegraphics[height=2.38cm, width=0.24\linewidth]{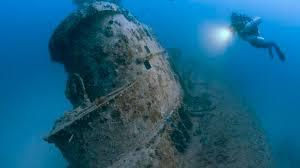}
    \includegraphics[height=2.38cm, width=0.24\linewidth]{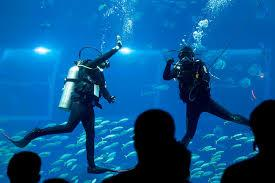}
    \includegraphics[height=2.38cm, width=0.24\linewidth]{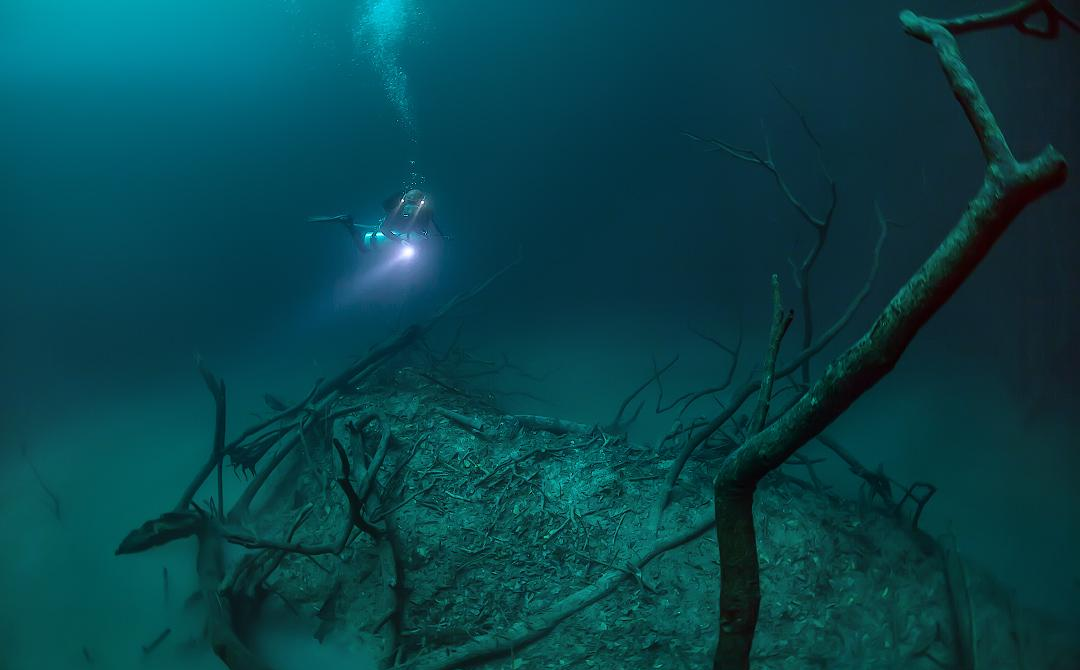} \\
    \includegraphics[height=2.38cm, width=0.24\linewidth]{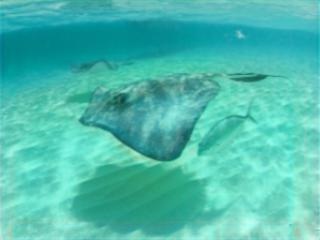}
    \includegraphics[height=2.38cm, width=0.24\linewidth]{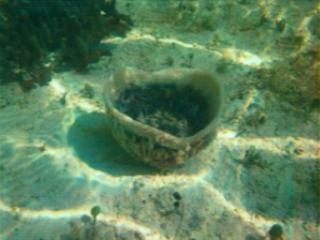}
    \includegraphics[height=2.38cm, width=0.24\linewidth]{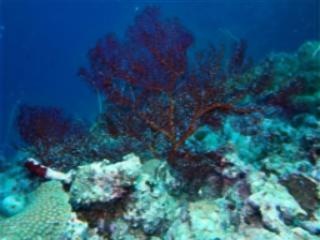}
    \includegraphics[height=2.38cm, width=0.24\linewidth]{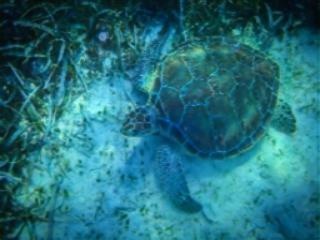}
    \caption{Example underwater images with non-uniform lighting: the top row shows images from the UIEB dataset~\citep{Li:Underwater:2020}, while the bottom row presents images from the LSUI dataset~\citep{peng2023u}.}
    \label{fig:nonuniform_light}
\end{figure}

\paragraph{Non-Uniform Illumination}
Lighting conditions underwater are often highly uneven, whether due to oblique sunlight, attenuation with depth, or artificial light sources mounted on cameras or vehicles. This leads to bright hotspots, dark shadow regions, and spatially varying colour responses within the same image. \autoref{fig:nonuniform_light} illustrates representative examples from the UIEB and LSUI datasets. Such non-uniform illumination complicates both enhancement and reconstruction: local contrast correction may over-amplify noise or saturate bright areas, while changing illumination patterns violate the photometric consistency assumptions commonly used in feature matching and multi-view geometry.

\begin{figure}[t!]
    \centering
    \includegraphics[width=\linewidth]{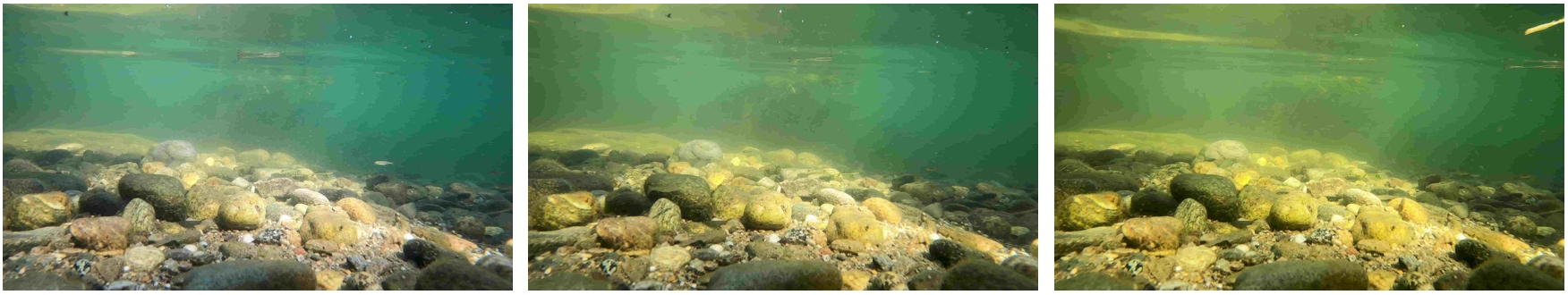}
    \caption{Examples of underwater images under dynamic environmental conditions~\citep{Xie:UVEB:2024}.}
    \label{fig:dynamic_env}
\end{figure}

\paragraph{Dynamic Water Conditions}
Oceanic and freshwater environments are rarely stable. Currents, waves, suspended matter, and biological activity produce rapid temporal changes in visibility and appearance, as shown in \autoref{fig:dynamic_env}. These fluctuations complicate frame-to-frame correspondence estimation and often lead to sparse or unstable reconstructions. For enhancement, temporal inconsistency may cause flickering or colour instability in video restoration. For reconstruction, dynamic appearance changes reduce the reliability of tracking, matching, and multi-view aggregation.

\begin{figure}[t!]
    \centering
    \includegraphics[height=2.38cm, width=0.24\linewidth]{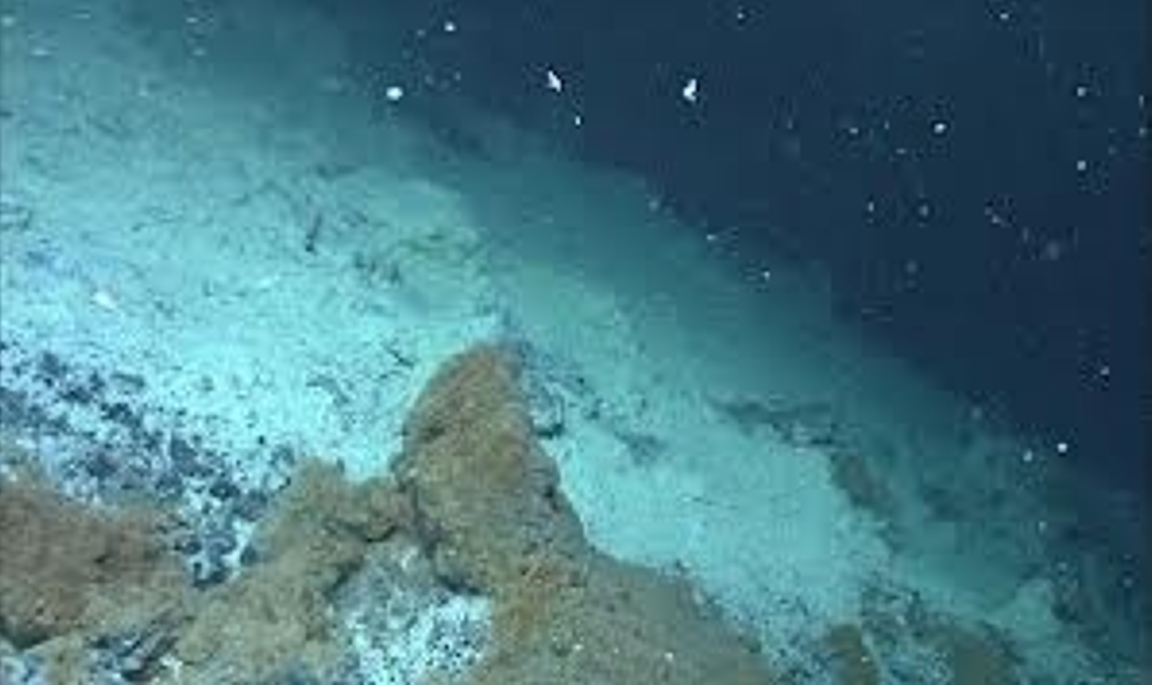}
    \includegraphics[height=2.38cm, width=0.242\linewidth]{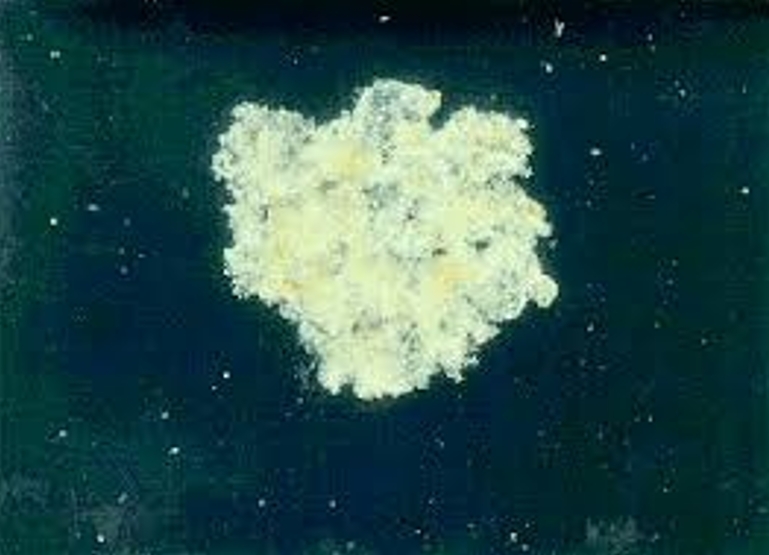}
    \includegraphics[height=2.38cm, width=0.24\linewidth]{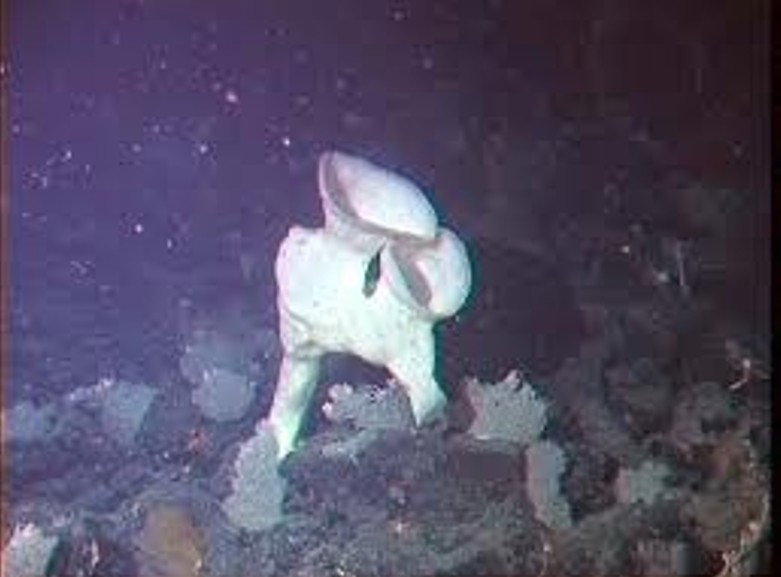}
    \includegraphics[height=2.38cm, width=0.24\linewidth]{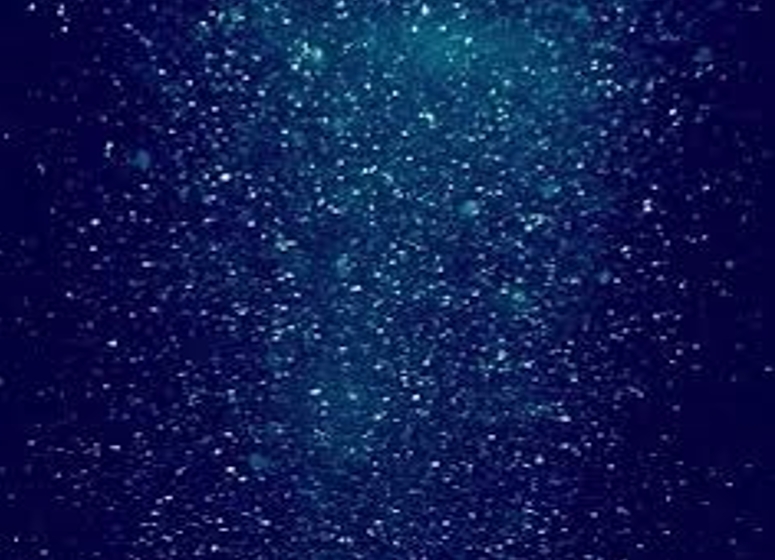}
    \caption{Examples of underwater images with marine snow~\citep{Banerjee:Elimination:2014}.}
    \label{fig:marine_snow}
\end{figure}

\paragraph{Marine Snow}
Marine snow consists of suspended particulate matter such as organic debris, phytoplankton shells, and other drifting material that appears as snow-like structures in underwater imagery. As illustrated in \autoref{fig:marine_snow}, these particles vary greatly in size, density, and reflectance, and can significantly degrade image clarity. Marine snow is particularly problematic for 3D reconstruction because it may create transient artefacts, false correspondences, and incomplete surface recovery \citep{Malyugina:beam:2025}. Effective enhancement methods must therefore suppress particulate interference while retaining scene details that are useful for later analysis and reconstruction.

\paragraph{Optical and Geometric Distortions}
Refraction at the interfaces between water, housing glass, and air causes projection behaviour that deviates significantly from the classic pinhole camera model. As a result, underwater images may exhibit nonlinear distortions that are not adequately captured by standard radial distortion correction alone. These distortions complicate camera calibration, feature matching, triangulation, and dense reconstruction, and may lead to systematic errors such as doming or bowling effects over large, approximately planar surfaces such as seafloors \citep{Wright2020UnderwaterPhotogrammetry}. Correcting refractive distortion is therefore essential for reliable underwater photogrammetry and 3D reconstruction.

\paragraph{Dynamic Scenes}
In addition to dynamic water conditions, many underwater scenes contain genuine scene motion, including swaying vegetation, drifting schools of fish, and moving particulate layers. Furthermore, the camera platform itself, such as an ROV or AUV, may experience unpredictable motion due to currents. These effects violate the static-scene assumptions underlying many SfM, SLAM, and MVS pipelines, and often generate spurious correspondences. Robust underwater reconstruction must therefore identify stable scene structure while remaining tolerant to motion-induced outliers.

\subsection{Motivation for Enhanced Underwater Imaging}

The demand for robust underwater imaging solutions is driven by a combination of scientific, industrial, and environmental requirements. In marine science and environmental monitoring, image quality directly affects the ability to identify species, assess habitats, estimate populations, and track ecological change over time. In underwater archaeology, clear visual observations and accurate reconstructions support non-invasive documentation of submerged cultural heritage, where incomplete or distorted imagery can lead to misinterpretation of structures or artefacts. In industrial inspection, enhanced imagery and reliable 3D models are essential for assessing the condition of offshore infrastructure under poor visibility and for identifying defects that may compromise operational safety.

Across these applications, underwater visual enhancement serves an important practical role by improving the usability of degraded imagery before downstream analysis. In particular, enhancement is often employed as a pre-processing step to support reconstruction pipelines that would otherwise struggle with low contrast, colour distortion, and reduced texture visibility. At the same time, the interaction between enhancement and reconstruction deserves closer attention, since improvements in perceptual quality do not always guarantee improvements in geometric accuracy. This review therefore considers both the practical value of enhancement-based pre-processing and the need to better understand its downstream effects.

\subsection{Scope of This Review and Literature Coverage}

This review provides a methodologically grounded synthesis of techniques for addressing degradation in underwater imagery and enabling reliable 3D reconstruction in subaqueous environments. {\color{revcrimson}
Works are selected based on their methodological contributions to imaging physics, visual enhancement, and geometric reconstruction, as well as their relevance, methodological clarity, and influence within the field; Selected conference papers are included to reflect fast-moving technical developments, while journal articles carry greater weight in the broader synthesis.
We exclude, or cite only peripherally, papers focused primarily on downstream detection, tracking, classification, or segmentation unless they directly help illustrate task utility, evaluation needs, or deployment constraints. Likewise, generic clear-medium computer vision and graphics methods are discussed only when they provide necessary conceptual background or representative context for underwater extensions, and works with insufficient methodological detail are not emphasized in the synthesis.}

We begin with underwater light propagation and image formation, establishing the physical basis for colour distortion, contrast attenuation, and visibility loss. Building on this, we examine underwater visual enhancement (UVE) methods spanning classical, physics-based, and learning-based approaches, followed by a survey of 3D reconstruction frameworks, including photogrammetry, structure-from-motion (SfM), visual SLAM, multi-view stereo (MVS), learning-based depth estimation, and neural scene representations such as NeRF and 3D Gaussian Splatting (3DGS). The review emphasizes the coupling between enhancement and reconstruction, particularly the role of UVE in influencing geometric accuracy and robustness. Datasets, evaluation protocols, and deployment constraints are also covered, with attention to cumulative error and pipeline-level robustness.
% \subsubsection{Image Enhancement Spectrum}
Methods for underwater enhancement can be broadly grouped into three categories:
\begin{itemize}
    \item \textbf{Physics-Based Models}: methods that explicitly model underwater attenuation and scattering \citep{berman2017diving,Li:Underwater:2021} to restore colour balance and scene contrast;
    \item \textbf{Traditional Image Processing}: classical approaches such as histogram equalization, Retinex-based enhancement, and contrast adjustment, which are often computationally lightweight but may be less robust under severe degradation;
    \item \textbf{Deep Learning}: data-driven methods, including CNN- and GAN-based models, that learn mappings from degraded to enhanced images using paired, synthetic, or unpaired supervision.
\end{itemize}
% \subsubsection{3D Reconstruction Pathways}
We also examine how mainstream 3D reconstruction pipelines are adapted to underwater conditions:
\begin{itemize}
    \item \textbf{SfM and MVS}: extensions of terrestrial photogrammetry that address refractive effects, unstable correspondences, and inconsistent lighting;
    \item \textbf{Learning-Based 3D}: neural depth estimation, volumetric reconstruction, and representation learning methods adapted to underwater inputs;
    \item \textbf{Sequential Pipelines}: practical workflows in which enhanced images are used as inputs to downstream reconstruction systems, together with discussion of their benefits and limitations.
\end{itemize}

\subsection{Contributions}

This review is motivated by the rapid growth of underwater imaging research and by the limitations of existing surveys, which often focus narrowly on either image restoration or 3D mapping. In contrast, we provide a broader and more connected perspective that examines underwater visual enhancement together with underwater 3D reconstruction, while preserving the practical viewpoint that enhancement is frequently used as a pre-processing stage. The main contributions of this paper are as follows:
\begin{itemize}
    \item \textbf{A unified taxonomy of underwater enhancement and reconstruction methods}: we organize the literature across physics-based, traditional, and learning-based paradigms, and summarize their assumptions, strengths, and limitations;
    \item \textbf{A review of enhancement for downstream reconstruction}: we examine how underwater visual enhancement is used in practice as a pre-processing step for 3D reconstruction, and discuss when and why it is beneficial;
    \item \textbf{Open challenges and future directions}: we highlight unresolved issues including domain generalization, benchmark design, robustness under severe degradation, and the cumulative error that may arise in sequential enhancement and reconstruction pipelines.
\end{itemize}

This review is intended to provide a structured reference for researchers and practitioners working across underwater image restoration, geometric reconstruction, and real-world deployment.

\subsection{Organization of the Paper}

The remainder of this paper is organized as follows. Section~\ref{sec:underwater_physics} reviews the physics of underwater light propagation and image formation. Section~\ref{sec:underwater_enhancement} surveys underwater visual enhancement methods, spanning non-learning, data-driven, and hybrid approaches. Section~\ref{sec:underwater_3d_recon} discusses underwater 3D reconstruction techniques, including traditional photogrammetry as well as recent methods based on NeRF and 3D Gaussian Splatting. Section~\ref{sec:Benchmarking} presents pipeline-level evaluation, including dataset context, evaluation criteria, and reconstruction case studies. Finally, Section~\ref{sec:conclusion} concludes the paper and outlines future research directions.

\begin{figure}[H]
\centering
\includegraphics[width=1.\linewidth]{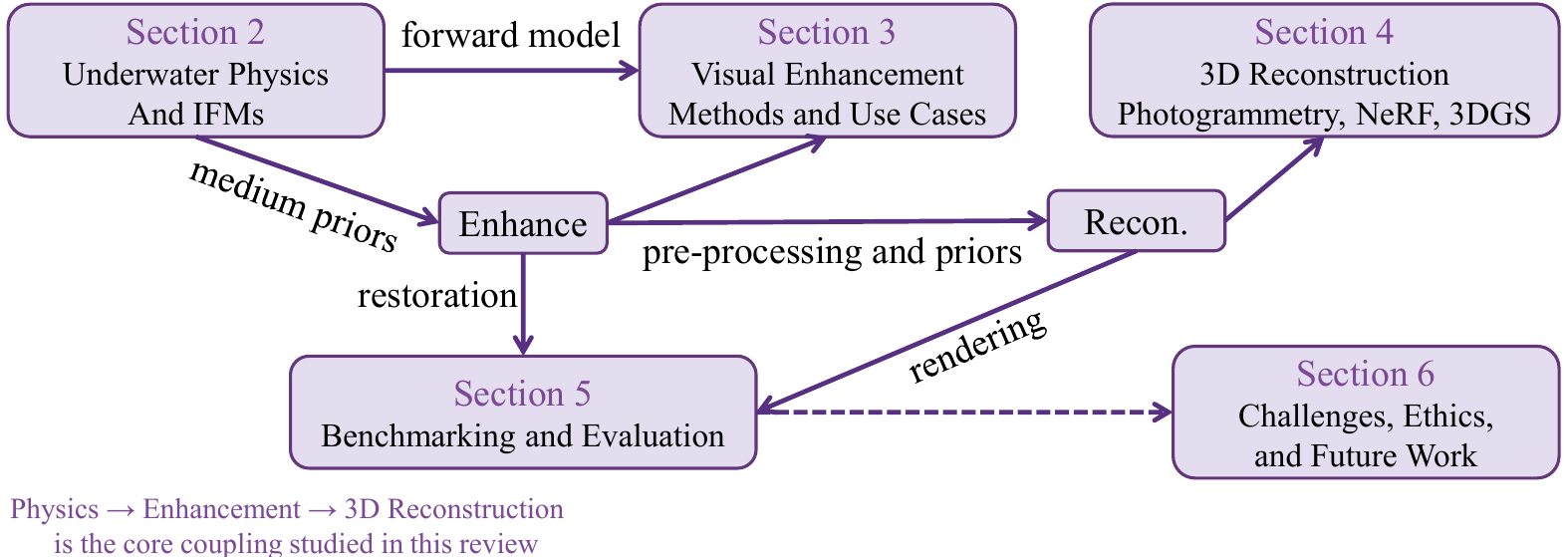}
\caption{{\color{revpurple}Roadmap of the paper. The review is organized around the idea that underwater physics provides the forward model, visual enhancement improves image quality for both visual inspection and downstream processing, and modern reconstruction methods such as photogrammetry, NeRF, and 3D Gaussian Splatting can often benefit from enhanced inputs.}}
\label{fig:review_roadmap}
\end{figure}

Figure~\ref{fig:review_roadmap} provides a compact roadmap of the paper structure and the methodological relationships emphasized throughout this review. In particular, underwater physics defines the degradation process, visual enhancement improves image quality for human interpretation and downstream processing, and modern reconstruction methods can often benefit from enhanced inputs, while future work is needed to better understand the cumulative effects of sequential pipelines.

\noindent Overall, the field of underwater visual enhancement and 3D reconstruction is entering a period of rapid development. By addressing the coupled optical, computational, and geometric challenges of subaqueous environments, future research can enable safer, more scalable, and more informative exploration of the underwater world.

\section{Underwater Light Propagation and Image Formation}
\label{sec:underwater_physics}

Understanding the physical principles of underwater light propagation is fundamental to developing effective image enhancement and 3D reconstruction algorithms. Compared to terrestrial imaging, underwater photography encounters significantly more complex distortions arising from wavelength-dependent absorption, scattering by suspended particles, refractive effects at media interfaces, and non-uniform illumination. This section provides a detailed overview of these phenomena, highlights the Jaffe--McGlamery underwater image formation model (IFM) and its simplified variants, and discusses specialized calibration procedures required in underwater imaging.

%-------------------------------------------------------------------------
\subsection{Absorption and Attenuation of Light}
When light travels through water, its intensity decays exponentially due to both absorption and scattering. The Beer--Lambert law describes the {attenuation} of light as a function of the propagation distance:
\begin{equation}
\label{eq:beerlambert}
I_d(x) \;=\; J(x)\,e^{-\beta(\lambda)\,d(x)},
\end{equation}
where \(I_d(x)\) is the {direct transmission component} of the observed intensity at pixel \(x\).
 \(J(x)\) represents the ideal intensity from the object (i.e., what would be measured in a clear medium without attenuation).
 \(\beta(\lambda)\) is the wavelength-dependent {total attenuation coefficient}, combining absorption and scattering effects.
 \(d(x)\) is the object-to-camera distance.

Because attenuation coefficients vary across the visible spectrum, the usable color bandwidth narrows with increasing depth. 
\begin{table}[!t]
    \centering
    \scriptsize
    \setlength{\tabcolsep}{2.9mm}
    \renewcommand{\arraystretch}{1.4}
    \caption{Penetration depths of different light wavelengths in clear seawater}
    \begin{tabular}{@{}c|c|c@{}}
        \hline
        \textbf{Light Color} & \textbf{Wavelength (nm)} & \textbf{Approx. Penetration Depth} \\
        \hline
        Ultraviolet (UV) & $<$400 & $<$5\,m \\
        \hline
        Blue Light & 400--500 & 50--100\,m \\
        \hline
        Green Light & 500--550 & 30--50\,m \\
        \hline
        Yellow Light & 550--600 & $\sim$20\,m \\
        \hline
        Red Light & 600--700 & $<$5\,m  \\
        \hline
        Near-Infrared (NIR) & $>$700 &  $<$1\,m \\
        \hline
    \end{tabular}
    \label{tab:light_penetration}
\end{table}
As shown in \autoref{tab:light_penetration}, blue and green wavelengths penetrate more deeply in clear seawater, while red and near-infrared wavelengths attenuate rapidly.
Strong attenuation of red wavelengths frequently imparts a bluish or greenish cast to underwater images.

% This law encapsulates how red wavelengths (\(\sim 600\)--\(700\)\,nm) typically attenuate more quickly than green or blue (\(\sim 400\)--\(500\)\,nm), thereby imparting the familiar bluish or greenish appearance in underwater scenes.

%-------------------------------------------------------------------------
\subsection{Scattering Phenomena}
In addition to absorption, \emph{scattering} drastically impacts underwater imagery. Particulates (e.g., silt, algae) can deviate light from its path, reducing clarity and contrast. Scattering is typically divided into:
\begin{itemize}
    \item \textbf{Forward scattering}: Light is deflected at small angles, leading to blurring.
    \item \textbf{Backscattering}: Light is scattered back toward the camera, creating a haze-like effect that reduces image contrast.
\end{itemize}
While both types degrade image quality, backscattering often proves more detrimental, as it adds a veil of background illumination. Although many haze-removal methods in atmospheric imaging~\citep{he2010single} inspire underwater dehazing solutions, the scattering coefficients and spectral absorption underwater can differ substantially.

%-------------------------------------------------------------------------

%-------------------------------------------------------------------------
\subsection{Jaffe--McGlamery Underwater Image Formation Model}
\label{subsec:jaffe_mcglamery_models}

\begin{figure}[t]
    \centering
    \includegraphics[width=.8\linewidth]{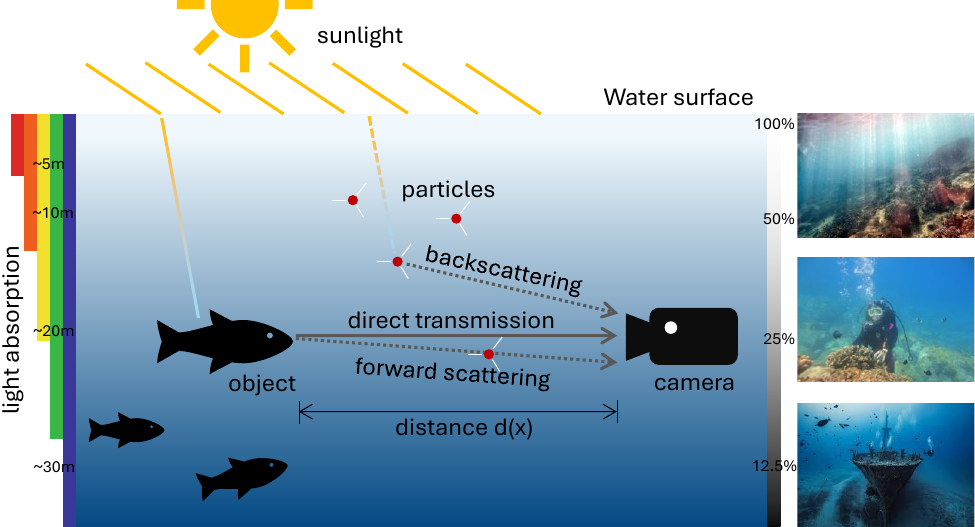}
    \caption{Jaffe--McGlamery underwater IFM, depicting light absorption and the selective attenuation of underwater illumination. The diagram highlights the effects of direct transmission, forward scattering, and backscattering caused by suspended particles, all of which influence image quality. The color gradient illustrates the depth-dependent absorption of light, while the side images demonstrate varying levels of underwater visibility at different depths}
    \label{fig:uie_imaging}
\end{figure}

A widely accepted image formation model (IFM) for underwater optical imaging was introduced by Jaffe and McGlamery~\citep{mcglamery1980computer,jaffe1990computer}, offering a comprehensive framework to describe how light interacts with water and suspended particulates. As illustrated in \autoref{fig:uie_imaging}, the model decomposes the observed signal into direct transmission $I_d$, forward scattering $I_f$, and backscatter $I_b$, corresponding respectively to signal-preserving object radiance, blur induced by small-angle scattering, and veil-like background illumination accumulated along the camera ray. These three components collectively determine the total irradiance recorded by the camera sensor and are dictated by factors such as water turbidity, imaging depth, and the dominant light wavelengths.

{\color{revorange}
We retain this classical formulation here only as the minimum physical background needed for the later discussions of simplified restoration models, revised IFMs, and physics-guided NeRF/3DGS methods.
}

Mathematically, the Jaffe--McGlamery model often expresses the captured intensity \(I(x)\) of a pixel location \(x\) as:
\begin{equation}
I(x) = I_d(x) + I_f(x) + I_b(x),
\label{eq:three_compo}
\end{equation}
where \( I_d(x) \) is the {direct transmission} from the scene \( I_f(x) \) is the {forward-scattered} term, and \( I_b(x) \) is the {backscattered} term.

\paragraph{Transmission Map and Attenuation}
The direct transmission component decays exponentially with distance, following the Beer--Lambert law:
\begin{equation}
I_d(x) = J(x)\, T(x),
\end{equation}
where \(J(x)\) is the scene radiance from the object and $T(x)$ is the transmission function:
\begin{equation}
T(x) = e^{-\beta(\lambda)\,d(x)},
\label{eq:transmission_fn}
\end{equation}
where \(\beta(\lambda) = a(\lambda) + b(\lambda)\) denotes the \emph{total} attenuation coefficient (absorption \(a(\lambda)\) plus scattering \(b(\lambda)\)), and \(d(x)\) is the object-camera distance.

\paragraph{Forward- and Backscattering Components}
Forward scattering ($I_f$) introduces blur by deviating a portion of the light rays, whereas backscattering ($I_b$) adds an additional haze-like illumination:
\begin{equation}
\begin{aligned}
I_f(x) &= \int_0^{d(x)} J(x)\, S_f(s)\, e^{-\beta(\lambda) s}\, ds, \\
I_b(x) &= \int_0^{d(x)} L(\lambda) \, S_b(s)\, e^{-\beta(\lambda) s}\, ds,
\end{aligned}
\end{equation}
where \(S_f(s)\) and \(S_b(s)\) represent phase functions describing the angular distribution of forward and backscattered light, respectively, and \(L(\lambda)\) denotes the ambient light.
% While these formulations are accurate, they can be computationally expensive in practice~\citep{mcglamery1980computer}.

%-------------------------------------------------------------------------
\subsection{Simplified Underwater IFMs}
\label{ssec:IFMs}
Considering that the full Jaffe--McGlamery IFM is often too complex for real-time or large-scale applications, many practical systems adopt simplified assumptions~\citep{bryson2016true,schechner2004clear}.

\paragraph{Simplified Jaffe--McGlamery model}
As $I_d(x) \gg I_f(x)$, the forward-scattering term $I_f(x)$ can be negligible. Further assuming a homogeneous medium with a constant \(\beta(\lambda)\), and approximating the backscattering phase function \(S_b\) as isotropic (thus treated as constant), leads to a simpler integral form:

\begin{equation}
\begin{aligned}
I_b(x) &= \int_{0}^{d(x)} L_b\, S_b\, e^{-\beta(\lambda)s}\, ds \\
       &= L(\lambda)\, S_b \int_{0}^{d(x)} e^{-\beta(\lambda)s}\, ds \\
       &= L(\lambda)\, S_b\, \frac{1 - e^{-\beta(\lambda)\,d(x)}}{\beta(\lambda)} \\
       &= A(\lambda)\,\bigl[1 - T(x)\bigr],
\end{aligned}
\end{equation}
where \(A(\lambda) = \frac{L(\lambda)\, S_b}{\beta(\lambda)}\) is the spatially invariant ambient light.
Thus, a widely used \emph{simplified} Jaffe--McGlamery IFM for the observed intensity $I$ becomes:
\begin{equation}
I(x) = J(x)\,T(x) \;+\; A(\lambda)\,\bigl[\,1 - T(x)].
\label{eq:simplified_jaffe}
\end{equation}
This simplified IFM captures the essential interplay between direct attenuation and backscatter while omitting the more complex forward-scattering integral. Despite its approximations, it remains effective for many underwater imaging tasks, especially when water clarity is moderate and the scene is relatively close to the camera.

\paragraph{Atmospheric Scattering Model (ASM)}
Considering water-induced degradation is similar to haze in aerial images, several works~\citep{Chiang:Underwater:2012,Peng:Underwater:2017,peng2018generalization,Li:Underwater:2021,schechner2004clear,berman2017diving,Carlevaris-Bianco:Initial:2010,DrewsJr:Transmission:2013,lu2015contrast} also treat underwater image formation as an extension of the atmospheric scattering model (ASM)~\citep{narasimhan2002vision,narasimhan2003interactive,Tan:Visibility:2008,Fattal:Single:2008,Narasimhan:Chromatic:2000}. The simplified mathematical representation is given by:
\begin{equation}
I(x) = J(x) T(x) + A \bigl[1 -T(x) \bigr].
\label{eq:asm}
\end{equation}
Compared to the simplified Jaffe--McGlamery model in Eq.~\eqref{eq:simplified_jaffe}, the ASM assumes that the ambient illumination remains constant across the spectrum.
The function of ASM is to remove the veiling effect, similar to dehazing, but it cannot correct color cast issues.
In shallow water regions, we can assume that the attenuation rate of all wavelengths is consistent. Therefore, ASM can achieve a similar effect to Eq.~\eqref{eq:simplified_jaffe} in shallow underwater scenes (1--5 m). However, in underwater scenes beyond 5 m, ASM-based images tend to exhibit a noticeable green or blue color cast.

\paragraph{RGB Channel-Based ASM}
For practical applications, this model is often expressed in terms of the RGB channels~\citep{Fattal:Single:2008, Tarel:Fast:2009,peng2018generalization}:
\begin{equation}
\begin{aligned}
I^c(x) &= J^c T^c(x) +  A^c \bigl[1 - T^c(x) \bigr],  \\
T^c(x) &= e^{-\beta^c\,d(x)}, \quad \text{for} \quad c \in \{R,G,B\}.
\end{aligned}
\label{eq:rgb_asm}
\end{equation}
Here, the coefficients $\beta_{R} \gg \beta_{G} > \beta_{B}$ indicate that red light is absorbed more rapidly than green and blue light, leading to the characteristic blue-green appearance of underwater images. 
By applying a simple transformation to Eq.~\eqref{eq:rgb_asm}, we can calculate the scene radiance $J^{c}(x)$ by:
\begin{equation}
    J^{c}(x)=\frac{I^{c}(x)-A^{c}}{{T}^c(x)}+A^{c}
    \label{eq:rgg_asm_tran}
\end{equation}
The RGB ASM can be considered an intermediate-complexity physical model between the simplified Jaffe--McGlamery model (\autoref{eq:simplified_jaffe}) and ASM (\autoref{eq:asm}). It reduces the dependence on the wavelength $\lambda$ by leveraging RGB channels while addressing the color distortion issue that ASM fails to handle.

{\color{revpurple}
\paragraph{Revised underwater image formation model}
A key refinement, particularly relevant for modern restoration and neural rendering, is the revised underwater image formation model of \citet{akkaynak2018revised}. Instead of sharing one attenuation term between object radiance decay and backscatter accumulation, the revised model separates the direct-signal attenuation and the backscatter growth:
\begin{equation}
\begin{aligned}
I^c(x) &= D^c(x) + B^c(x), \\
D^c(x) &= J^c(x)\, e^{-\beta_D^c d(x)}, \\
B^c(x) &= B_\infty^c \left(1 - e^{-\beta_B^c d(x)}\right),
\end{aligned}
\label{eq:revised_ifm}
\end{equation}
where \(\beta_D^c\) and \(\beta_B^c\) denote distinct wideband coefficients for direct transmission and backscatter, respectively. Compared with the simplified models in Eqs.~\eqref{eq:simplified_jaffe}--\eqref{eq:rgb_asm}, this formulation avoids conflating two different physical processes and therefore provides a more suitable starting point for SeaThru-style restoration, depth-aware color recovery, and underwater NeRF/3DGS pipelines that explicitly model the medium.
}

In the next section, we build on these physical insights to survey underwater image enhancement methods, including purely physics-based restoration, histogram-based techniques, and Retinex-based corrections. Comprehending these foundational methods provides a basis for later analysis of more advanced, learning-centric pipelines.

\section{Underwater Visual Enhancement}
\label{sec:underwater_enhancement}

% Following the physical models and calibration procedures discussed in Section~\ref{sec:underwater_physics}, 
% we now examine  {underwater image enhancement} methods aimed at mitigating color casts, haze effects, 
% and scattering artifacts. 
Improving underwater imagery involves numerous challenges, including color 
inconsistencies due to selective wavelength absorption, light scattering from suspended particulates, and 
viewpoint-dependent refraction effects. This section provides a detailed literature survey of the 
existing approaches, spanning from  {traditional statistical-based} to  {data-driven deep-learning} methods. 
While some algorithms rely on simplified assumptions (\eg, uniform attenuation), others incorporate domain 
knowledge or advanced neural architectures to handle in-situ complexities. We group the methods according to 
their underlying strategies and highlight open research problems for future development.

{\color{revgreen}
As summarized later in Table~\ref{tab:data_driven_uie_timeline}, underwater enhancement has evolved from prior- and IFM-driven restoration toward increasingly data-driven paradigms that must simultaneously handle visual quality, domain shift, and downstream task utility.
}

\newpage

\begin{figure}[htbp]
  \centering
  \resizebox{1.\textwidth}{!}{%
  
  % Local TikZ style definitions
  \tikzset{
    root/.style={draw=blue!50!black, fill=blue!50!black, text=white, rounded corners=4pt, font=\bfseries\small, align=center, inner sep=6pt},
    category/.style={draw=blue!50!black, thick, fill=blue!3, rounded corners=3pt, font=\bfseries\small, align=center, inner sep=5pt},
    method/.style={draw=black!60, fill=white, rounded corners=2pt, font=\small, align=center, inner sep=4pt},
    technique/.style={font=\scriptsize, align=left, text=black!80, inner sep=1pt} % 文献列表左对齐
  }
  
  \begin{forest}
    for tree={
      grow'=east,             % 树向右生长
      parent anchor=east,     % 父节点连线起点
      child anchor=west,      % 子节点连线终点
      anchor=west,            % 节点自身的对齐锚点
      edge={draw=blue!50!black, thick, rounded corners=4pt}, % 高级圆角海军蓝连线
      edge path={
        \noexpand\path[\forestoption{edge}]
        % +(10pt,0) 意味着共用一段 10pt 的主干线，然后再上下分叉，避免线条杂乱
        (!u.parent anchor) -- +(10pt,0) |- (.child anchor) 
        \forestoption{edge label};
      },
      l sep=0.7cm, % 层级之间的水平间距 (根据需要可微调)
      s sep=0.2cm, % 兄弟节点之间的垂直间距
    }
  [Underwater \\ Visual\\Enhancement, root
    [Non-Learning\\Methods, category
      [Statistical\\Methods, method
        [Gray world assumption\\Histogram Equalization\\White Balance\\\dots, technique]]
      [IFM-Based\\Methods, method
        [{\protect\citet{he2010single}, \protect\citet{Yang:Low:2011}}\\
         {\protect\citet{Chiang:Underwater:2012}, \protect\citet{Wen:Single:2013}}\\
         {\protect\citet{DrewsJr:Transmission:2013}, \protect\citet{Galdran:Automatic:2015}}\\
         {\protect\citet{peng2018generalization}, \protect\citet{Liang:GUDCP:2022}}\\
         \dots, technique]]
      [Retinex-Based\\Methods, method
        [{\protect\citet{Joshi:Quantification:2008}}\\
         {\protect\citet{Funt:Retinex:2000}}\\
         {\protect\citet{Fu:retinexbased:2014}}\\
         {\protect\citet{Zhuang:Bayesian:2021}}\\
         \dots, technique]]
      [Fusion-Based\\Methods, method % 加上了换行，让排版更紧凑
        [{\protect\citet{Mertens:Exposure:2009}, \protect\citet{Khan:Underwater:2016}}\\
         {\protect\citet{Ancuti:Color:2018}, \protect\citet{MohdAzmi:Naturalbased:2019}}\\
         {\protect\citet{Zhang:Single:2019}, \protect\citet{Sethi:Fusion:2019}}\\
         {\protect\citet{Mohan:Underwater:2020}, \protect\citet{Zhang:Underwater:2023}}\\
         \dots, technique]]
    ]
    [Deep Learning\\Methods, category
      [CNNs, method
        [{\protect\citet{Shin:Estimation:2016}, \protect\citet{Li:Underwater:2020}}\\
         {\protect\citet{Li:Underwater:2021}, \protect\citet{Wang:UIEC^2Net:2021}}\\
         {\protect\citet{Jiang:Underwater:2022}, \protect\citet{Guo:Underwater:2024}}\\
         {\protect\citet{Wang:UIERL:2024}}\\
         \dots, technique]]
      [Transformers, method
        [{\protect\citet{Wang:Uformer:2022}, \protect\citet{Shen:UDAformer:2023}}\\
         {\textcolor{blue!70!black}{\protect\citet{wen2024waterformer}}, \protect\citet{peng2023u}}\\
         {{\textcolor{blue!70!black}{\protect\citet{wen2024waterformer}}, \protect\citet{peng2023u}}}\\
         {\protect\citet{Khan:Phaseformer:2024}, \protect\citet{Khan:Spectroformer:2024}}\\
         \dots, technique]]
      [Mamba, method
        [{\protect\citet{Guan:WaterMamba:2024}, \protect\citet{Lin_2024_ACCV}}\\
         {\protect\citet{Zhang2024MambaUIE}}\\
         {\protect\citet{Chen:MUIR:2024}}\\
         \dots, technique]]
      [Diffusion Models, method
        [{\protect\citet{LU2023103926}, \protect\citet{Lu:speed:2024}}\\
         {\protect\citet{Xia:patch:2025}, \protect\citet{CHEN2025129274}}\\
         {\protect\citet{DU2025125271}}\\
         \dots, technique]]
      [Limited / No Paired\\Supervision, method
        [{\protect\citet{Li:WaterGAN:2017}, \protect\citet{anwar2020diving}}\\
         {\protect\citet{zhu2023unsupervised}, \protect\citet{jiang2023perception}}\\
         {\textcolor{blue!70!black}{\protect\citet{jiang2022two}}, \protect\citet{huang2023contrastive}, \protect\citet{wen2025ssduie}}\\
         {{\textcolor{blue!70!black}{\protect\citet{jiang2022two}}, \protect\citet{huang2023contrastive}, \protect\citet{wen2025ssduie}}}\\
         {\textcolor{teal!70!black}{\protect\citet{kapoor2023domain}}}\\
         \dots, technique]]
    ]
    [Hybrid Methods, category
      [{Scene priors, decomposition,\\and synthetic-real bridging}, method
        [{\textcolor{blue!70!black}{\protect\citet{LI2020107038}, \protect\citet{chen2020perceptual}}}\\
         {\textcolor{blue!70!black}{\protect\citet{mu2023generalized}, \protect\citet{liu2021ipmgan}}}\\
         {\textcolor{blue!70!black}{\protect\citet{Zhang2024MambaUIE}, \protect\citet{yan2023hybrur}}}\\
         {\textcolor{blue!70!black}{\protect\citet{wen2023syreanet}, \protect\citet{Zhang:Underwater:2025}}}\\
         \textcolor{blue!70!black}{\dots}, technique]]
    ]
  ]
  \end{forest}%
  }
  \caption{Taxonomy of selected underwater image and video enhancement works across non-learning and learning paradigms. Table~\ref{tab:data_driven_uie_timeline} provides a complementary chronological view of the rise of deep-learning methods in UIE.}
  \label{fig:masked_tree}
\end{figure}

\subsection{Conventional Methods}

% Traditional single-image  {dehazing} and color correction approaches often inherit techniques from 
%  {atmospheric} image restoration, adapting them to underwater conditions by modifying the physical light-attenuation model~\citep{he2010single,schechner2004clear,narasimhan2002vision}.
\subsubsection{Statistical Approaches}
\begin{figure}[t]
  \centering

  % --- Row 1 ---
  \begin{minipage}[b]{0.49\linewidth}
    \centering
    \includegraphics[height=2.5cm,width=3.8cm]{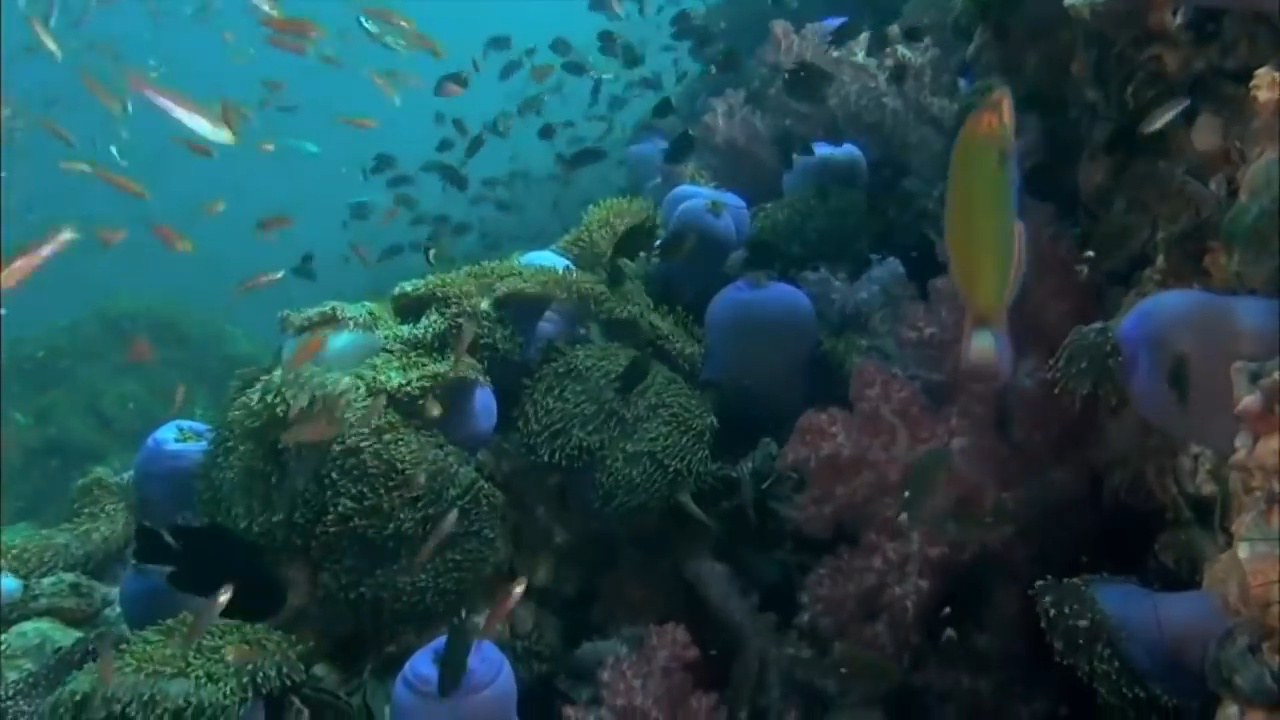}\hspace{2mm}%
    \includegraphics[height=2.5cm,width=3.8cm]{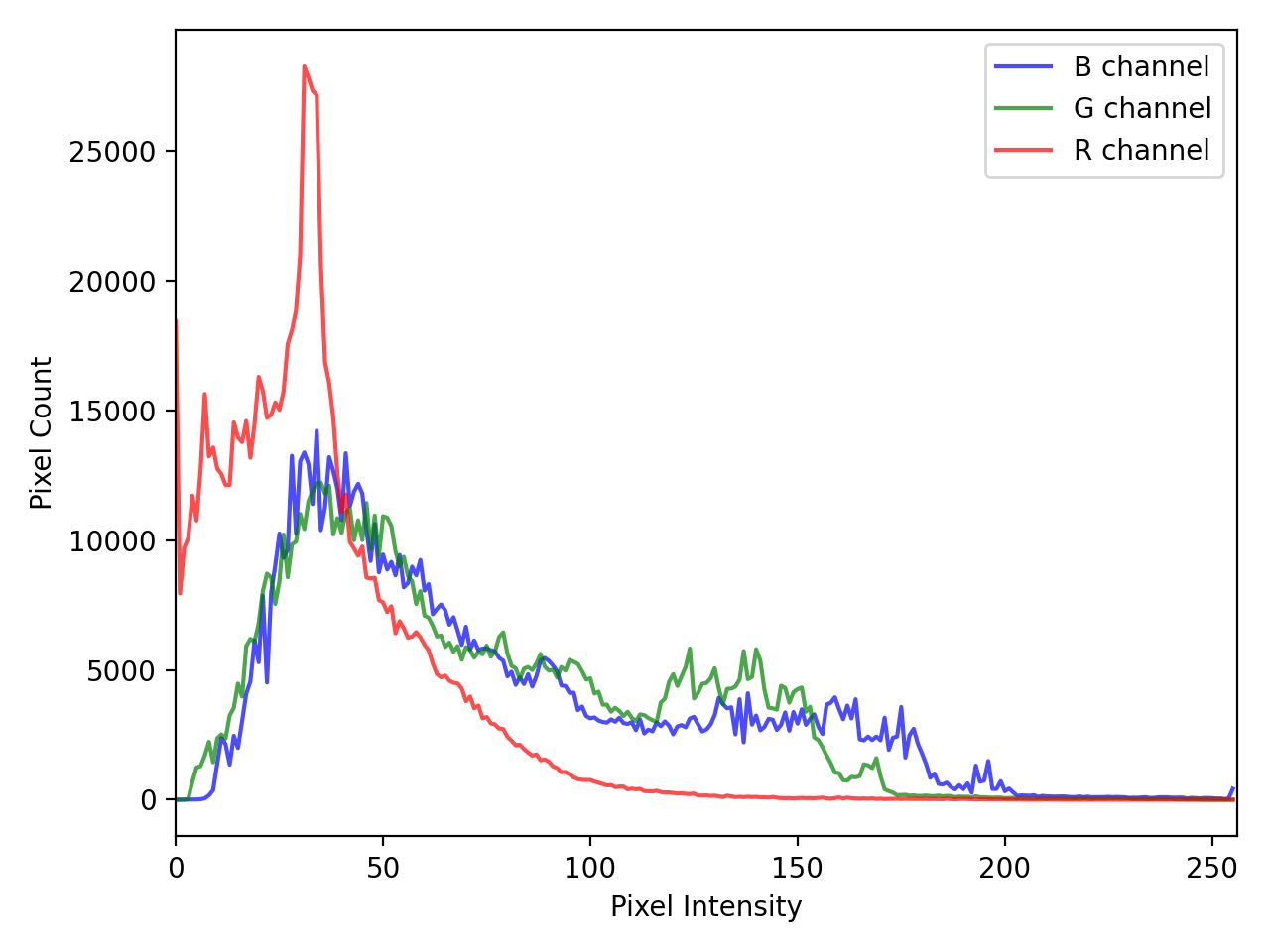}\\
    \small (a) Original image
  \end{minipage}\hfill
  \begin{minipage}[b]{0.49\linewidth}
    \centering
    \includegraphics[height=2.5cm,width=3.8cm]{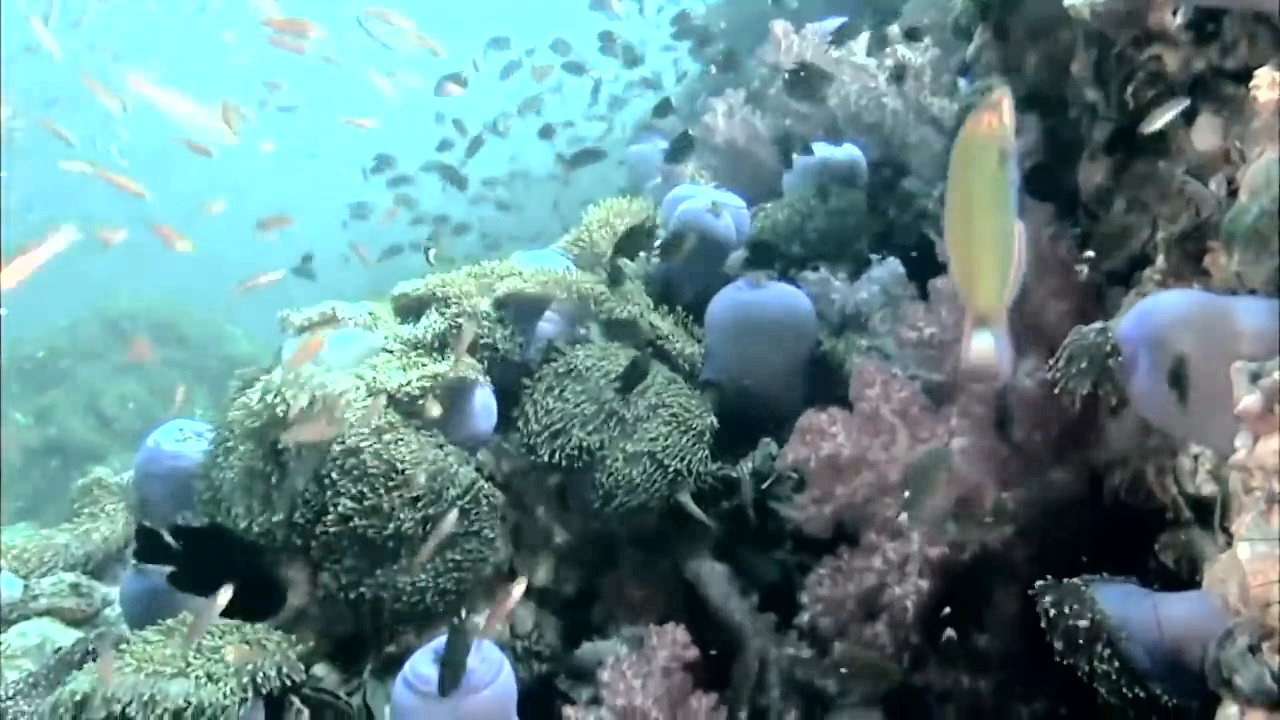}\hspace{2mm}%
    \includegraphics[height=2.5cm,width=3.8cm]{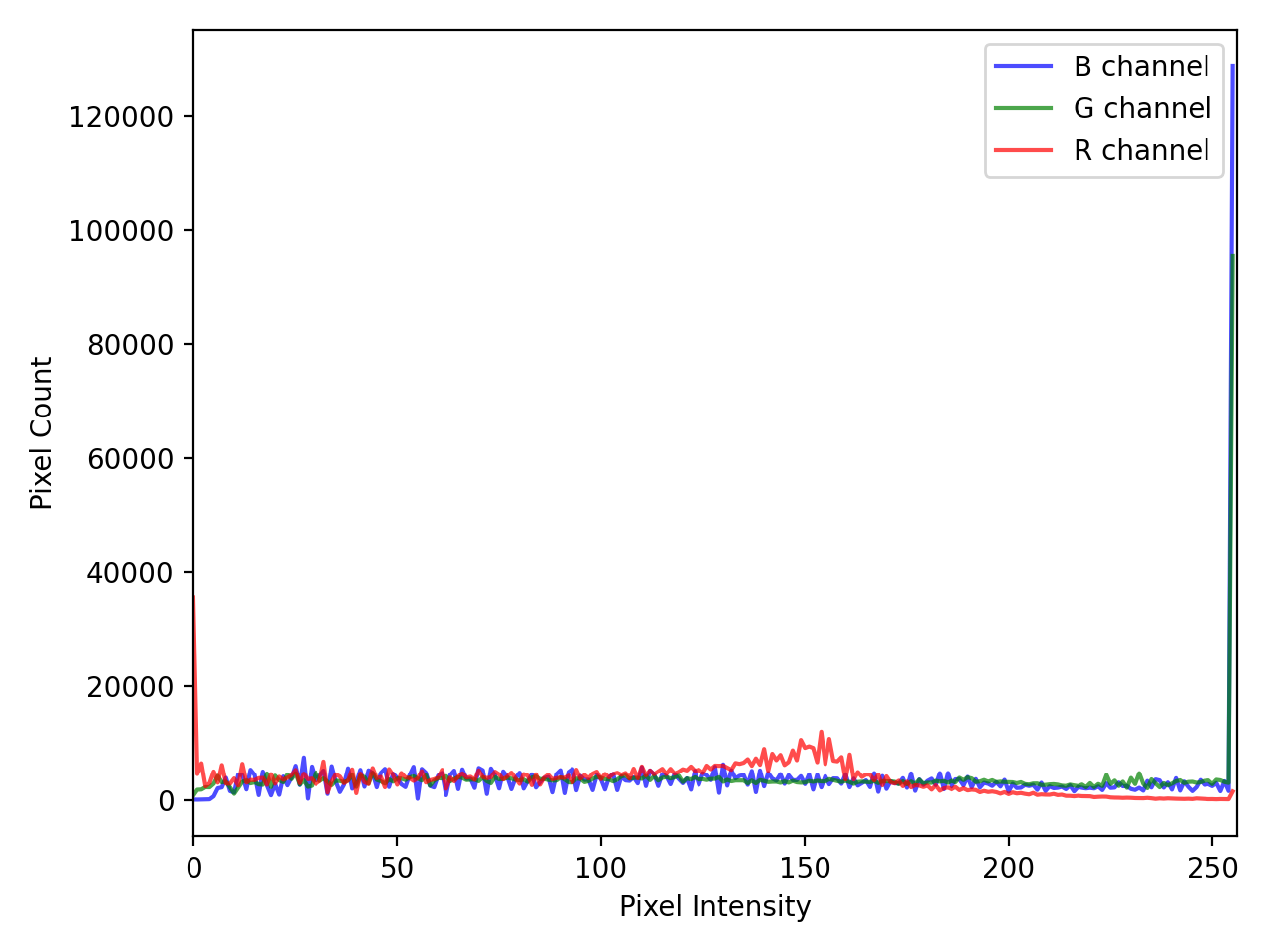}\\
    \small (b) Output with HE
  \end{minipage}

  \vspace{0.6em}

  % --- Row 2 ---
  \begin{minipage}[b]{0.49\linewidth}
    \centering
    \includegraphics[height=2.5cm,width=3.8cm]{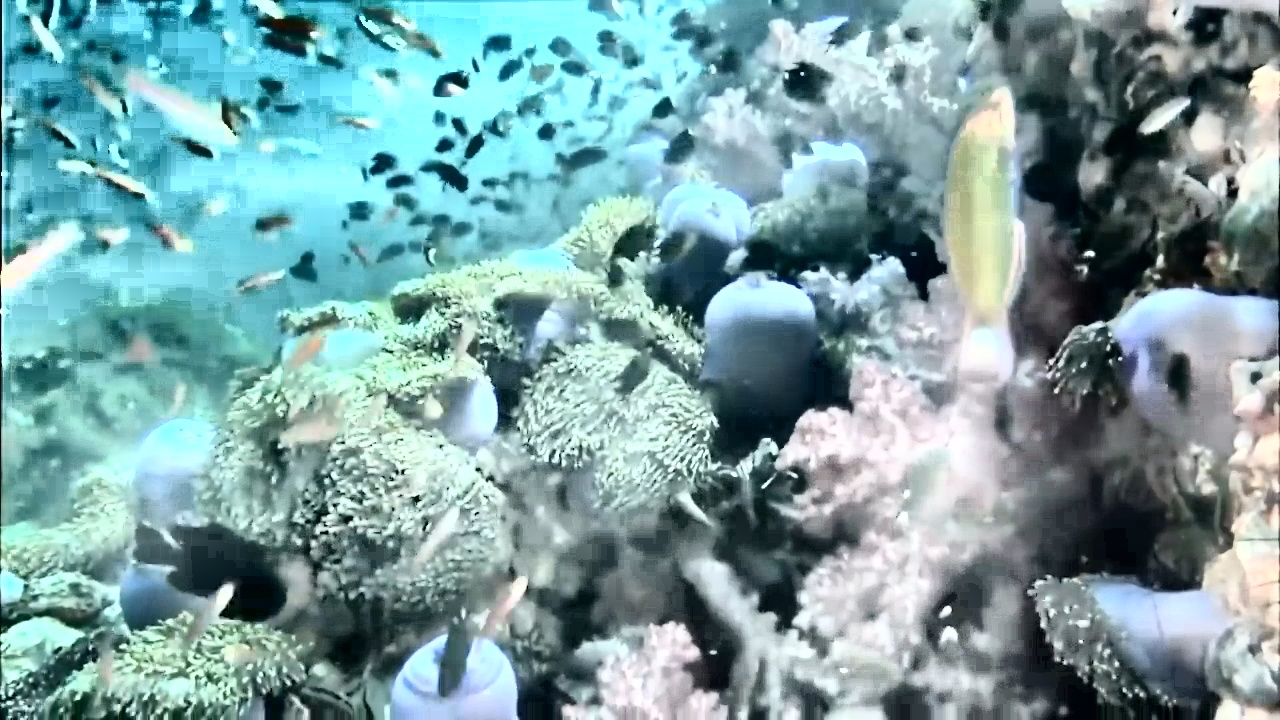}\hspace{2mm}%
    \includegraphics[height=2.5cm,width=3.8cm]{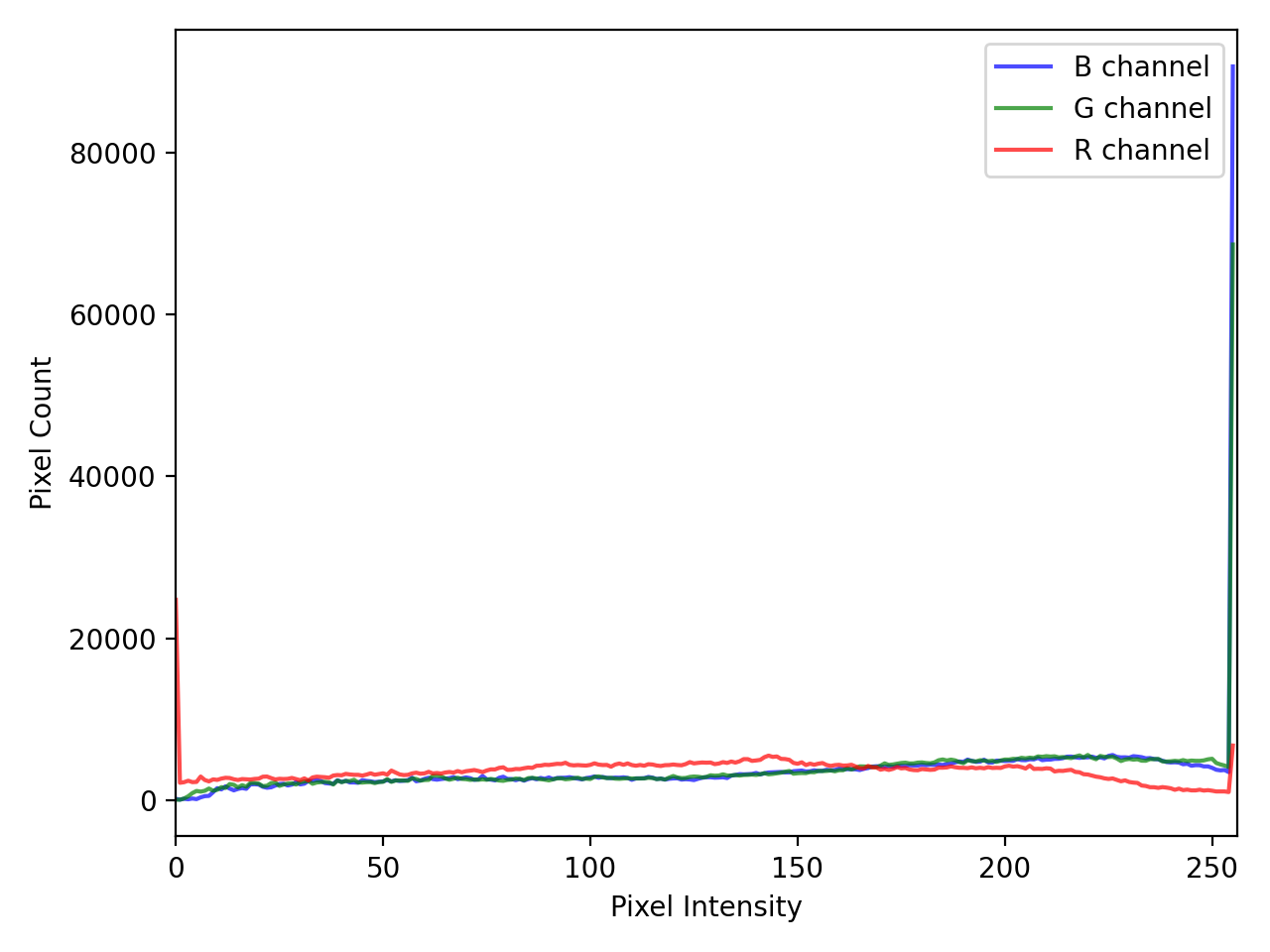}\\
    \small (c) Output with AHE
  \end{minipage}\hfill
  \begin{minipage}[b]{0.49\linewidth}
    \centering
    \includegraphics[height=2.5cm,width=3.8cm]{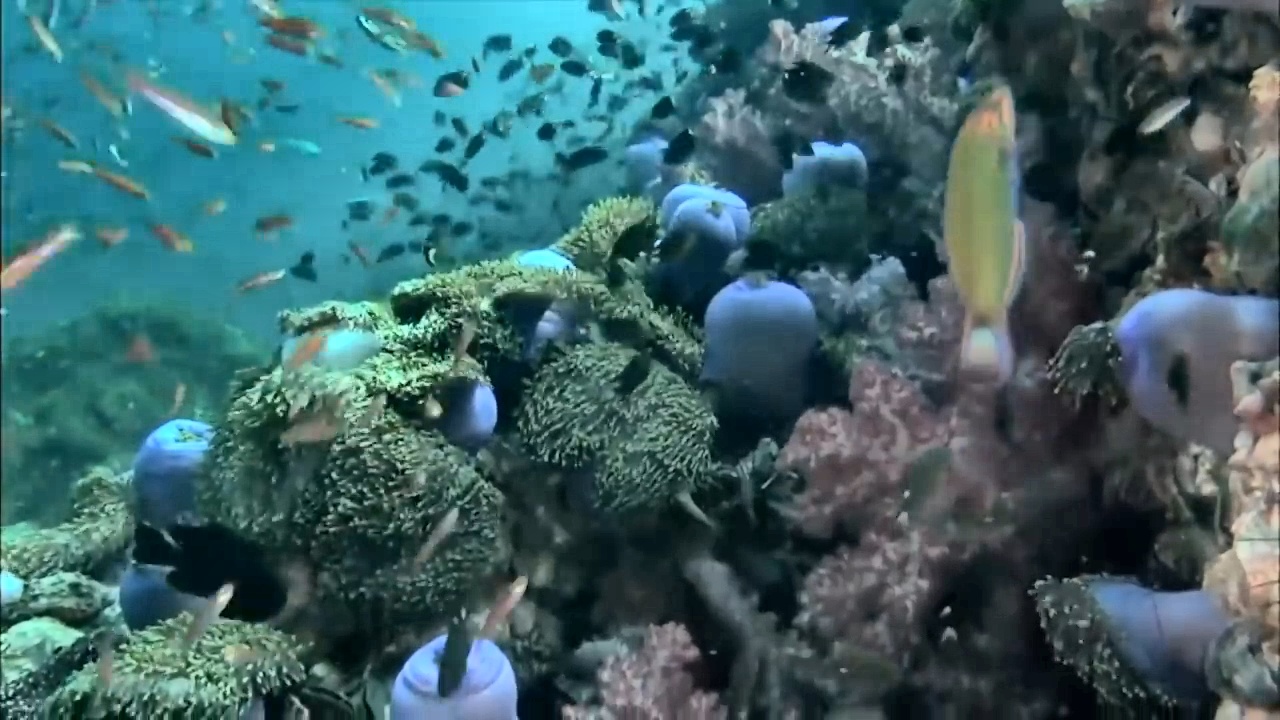}\hspace{2mm}%
    \includegraphics[height=2.5cm,width=3.8cm]{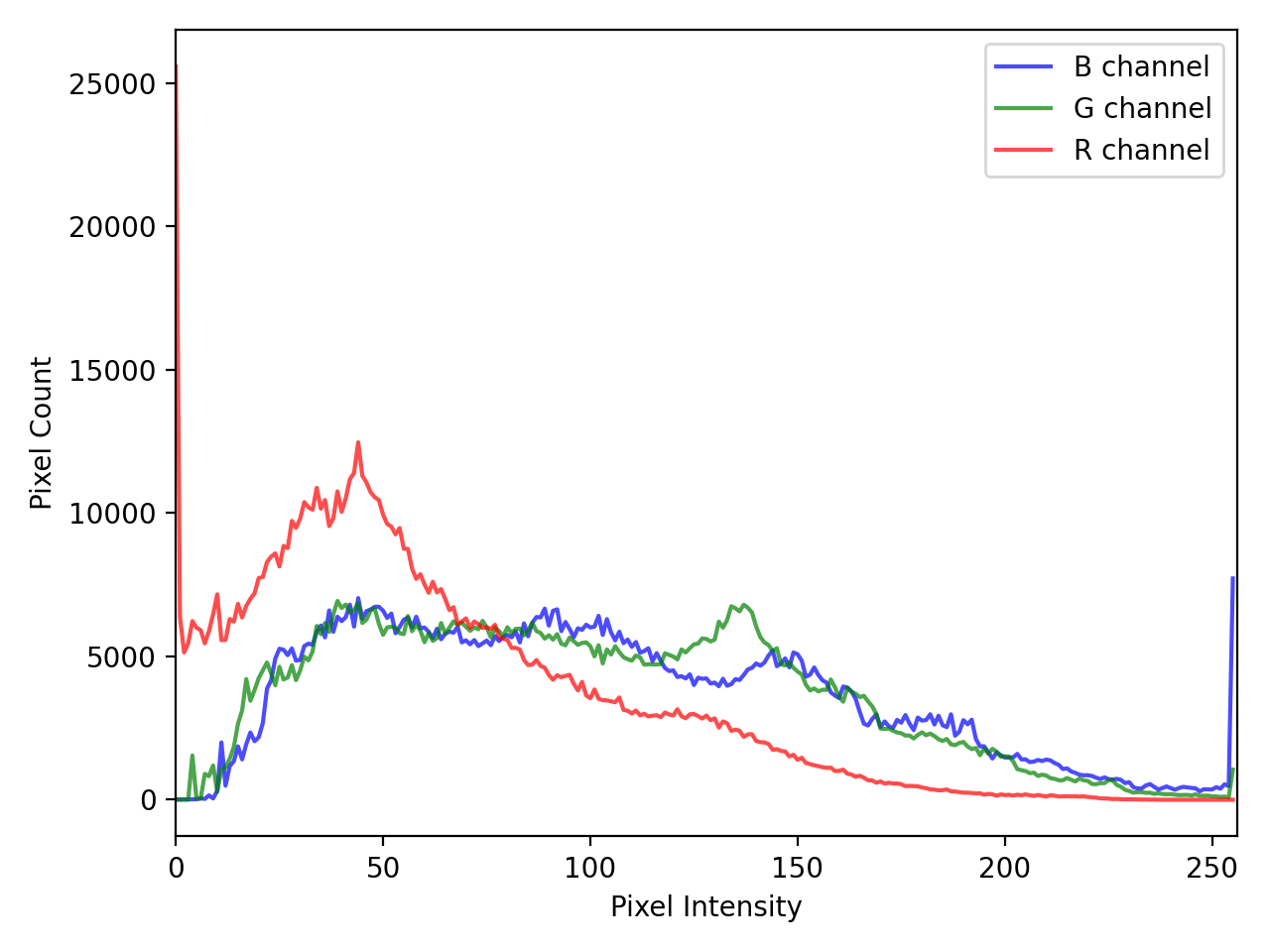}\\
    \small (d) Output with CLAHE
  \end{minipage}

  \caption{Comparison of different histogram equalization techniques applied to the underwater image}
  \label{fig:clahe_compa}
\end{figure}

\paragraph{Histogram Equalization}
The simplest approaches to enhancing underwater imagery involve histogram stretching, similar to contrast enhancement for images in a clear medium. \autoref{fig:clahe_compa} presents the enhanced underwater images using various histogram equalization techniques, including histogram equalization (HE), adaptive histogram equalization (AHE)~\citep{Pizer:Adaptive:1987}, and contrast-limited adaptive histogram equalization (CLAHE)~\citep{Pizer:Contrastlimited:1990}.  These methods enhance contrast in underwater imagery to varying degrees, with AHE and CLAHE providing more localized adjustments. CLAHE is a typical baseline in this category.
% \begin{figure}[t!]
%     \centering
%     \begin{subfigure}{0.49\linewidth}
%         \includegraphics[height=2.5cm, width=3.8cm]{original.jpg}
%         \includegraphics[height=2.5cm, width=3.8cm]{original_hist.png}
%         \caption{Original image} 
%     \end{subfigure}
%     \begin{subfigure}{0.49\linewidth}
%         \includegraphics[height=2.5cm, width=3.8cm]{global_he.jpg}
%         \includegraphics[height=2.5cm, width=3.8cm]{global_he_hist.png}
%         \caption{Output with HE} 
%     \end{subfigure}\\
%     \begin{subfigure}{0.49\linewidth}
%         \includegraphics[height=2.5cm, width=3.8cm]{ahe_interpolation.jpg}
%         \includegraphics[height=2.5cm, width=3.8cm]{ahe_interpolation_hist.png}
%         \caption{Output with AHE} 
%     \end{subfigure}
%         \begin{subfigure}{0.49\linewidth}
%         \includegraphics[height=2.5cm, width=3.8cm]{clahe.jpg}
%         \includegraphics[height=2.5cm, width=3.8cm]{clahe_hist.png}
%         \caption{Output with CLAHE} 
%     \end{subfigure}
%     \caption{Comparison of different histogram equalization techniques applied to the underwater image.}
%     \label{fig:clahe_compa}
% \end{figure}

% Although computationally light, purely histogram-driven approaches often lack the
% spatial adaptivity to handle backscatter or patch-wise variations in clarity. 

Approaches for underwater scenes are usually slightly more advanced, incorporating channel compensation, as different wavelengths of light affect the image appearance differently.
Many works rely on global or local histogram adjustments to address color bias. For instance, 
\citet{Zhou:Underwater:2023} propose sub-histogram equalization across multiple intervals, 
while \citet{zhang2024pixel} incorporate pixel-level gradient constraints for 
channel-specific stretching. Although computationally light, purely histogram-driven approaches often lack 
the spatial adaptivity to handle backscatter or patchwise variations in clarity.

\subsubsection{IFM-based Methods}
Traditional prior-based underwater image enhancement methods often adapt single-image RGB  {dehazing} schemes from atmospheric context to underwater conditions by modifying them for wavelength-dependent attenuation. 
In the single RGB underwater image enhancement methods based on RGB,  
the recovered image can be obtained using Eq.~\eqref{eq:rgg_asm_tran}, where the transmission map $T(x)=$ and ambient light $A^c$ are unknown variables. Therefore, we can use some prior assumptions to estimate the two variables.

{\color{revorange}
For many IFM-based methods, the estimation pipeline can be summarized compactly as
\begin{equation}
\hat{A}^c = \Phi_A(I), \qquad \hat{T}^c = \Phi_T(I), \qquad
\hat{J}^c(x) = \frac{I^c(x)-\hat{A}^c}{\max(\hat{T}^c(x),\epsilon)} + \hat{A}^c,
\label{eq:ifm_estimation_pipeline}
\end{equation}
where \(\Phi_A\) and \(\Phi_T\) denote the background-light and transmission estimators induced by a specific prior. Most classical variants therefore differ less in the reconstruction formula itself than in how they parameterize \(\hat{A}^c\) and \(\hat{T}^c\).
}
% A commonly adopted physical model for underwater image formation 
% can be written as:
% \begin{equation}
%     I_c(\mathbf{x}) \;=\; J_c(\mathbf{x}) \,e^{-\alpha_c\,d(\mathbf{x})}\;+\;B_c\,\bigl(1 - e^{-\alpha_c\,d(\mathbf{x})}\bigr),
%     \label{eq:uw_formation}
% \end{equation}
% where $I_c(\mathbf{x})$ is the observed pixel intensity in color channel $c \in \{r,g,b\}$, $J_c(\mathbf{x})$ is 
% the ideal (unknown) true scene color, $\alpha_c$ is the attenuation coefficient for channel $c$, $d(\mathbf{x})$ 
% is the scene depth, and $B_c$ denotes the background light intensity.
\paragraph{Transmission Map Estimation with Dark Channel Prior (DCP)}
Originally introduced for atmospheric haze removal \citep{he2010single}, DCP estimates
 the transmission map $T(x)$ and was later applied to underwater 
imagery by adjusting $\beta^c$ in Eq.~\eqref{eq:rgb_asm} to account for color-selective absorption~\citep{peng2018generalization}. DCP exploits the observation that, in 
most local patches $\Omega(\mathbf{x})$, at least one color channel tends to have near-zero intensity in clear 
scenes. Formally, the dark channel is:
\begin{equation}
J^{RGB}_{\text {dark }}({x})=\min _{c \in\{R, G, B\}}\left[\min _{{y} \in \Omega({x})}\left(J^{c}({y})\right)\right] \approx 0.
    \label{eq:dark_channel}
\end{equation}
Through DCP (\autoref{eq:dark_channel}) and the RGB ASM (\autoref{eq:rgb_asm}), the transmission map can be estimated simply by:
\begin{equation}
\begin{aligned}
\tilde{T}({x})=1-\min _{c}\left[\min _{{y} \in \Omega({x})}\left(\frac{I^{c}({y})}{A^{c}}\right)\right],
\end{aligned}
\label{eq:dcp_transmission}
\end{equation}
where $A^c$ is a constant value selected from one of the farthest and haziest pixels in the input image. 
After the transmission map is estimated, the recovered image can be calculated using Eq.~\eqref{eq:rgg_asm_tran}.
% An issue occurred by DCP's transmission map estimation~\eqref{eq:dcp_transmission} is that different color channels share the same transmission map, \ie, $ T^c(x) := \tilde{T}({x})$ for $c \in \{R, G, B\})$. 
DCP, as a simple prior assumption, performs well in atmospheric haze environments. However, due to its overly simplistic assumption, it cannot be directly applied to other scenarios, such as sandstorms and underwater turbid images. Below, we outline some notable issues with DCP:
1) A key issue with DCP's transmission map estimation~\eqref{eq:dcp_transmission} is its assumption of uniform transmission across color channels (\textit{i.e.}, \( T^c(x) := \tilde{T}({x}) \) for \(c \in \{R, G, B\}\))~\citep{Galdran:Automatic:2015}.
This renders DCP ineffective in addressing wavelength-dependent color casts, a limitation that has emerged as a key research focus in subsequent studies based on DCP. 
2) Meanwhile, DCP tends to overestimate the transmission map in certain regions, leading to color distortions and halo artifacts around edges, according to~\citet{Huang:Visibility:2014,Huang:Advanced:2015}.
3) In underwater photography with artificial light sources, the light intensity decreases with distance, which is the opposite of the assumption about ambient light $A$ in DCP, according to~\citet{peng2018generalization} and \citet{Peng:Underwater:2017}.
A follow-up study~\citep{Chao:Removal:2010} directly applies DCP without modifications to underwater image processing, but the resulting visual quality shows limited improvement. Nevertheless, the authors highlight that the normalized image (\( I^c / A \)) can mitigate the impact of wavelength-dependent color absorption in underwater images.

\begin{figure}[t!]
    \centering
    \includegraphics[width=0.5\linewidth]{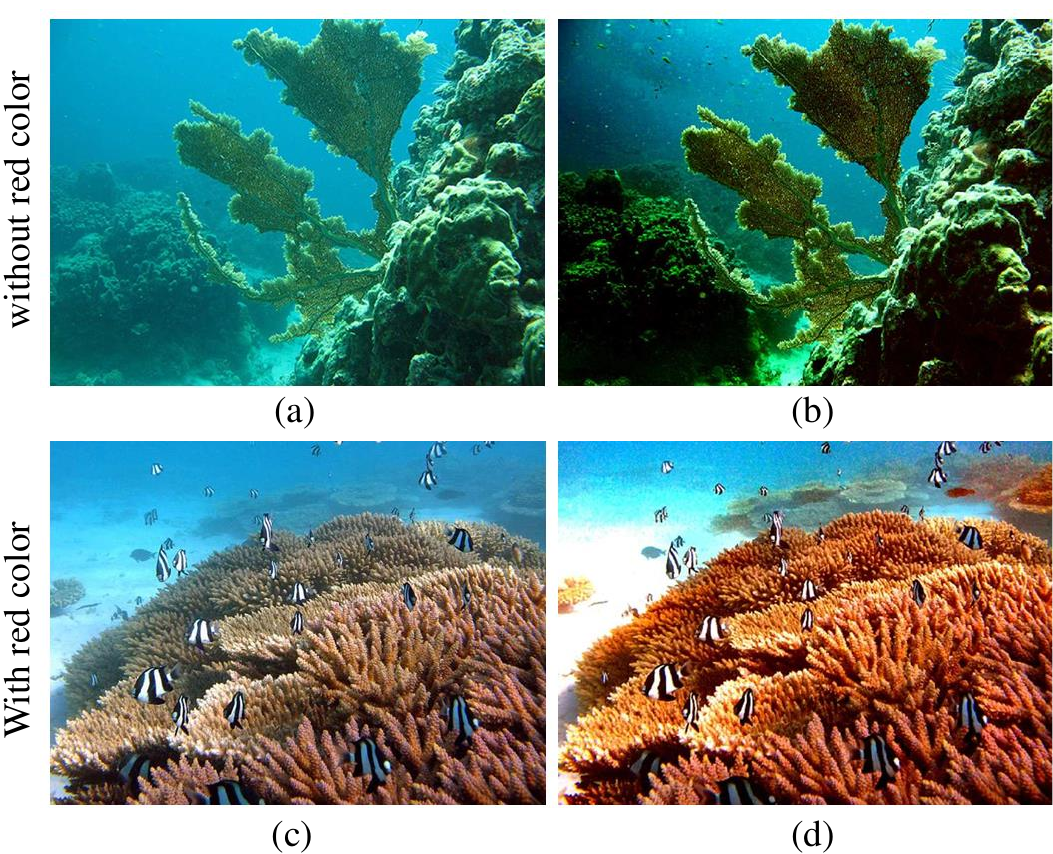}
    \caption{Enhanced underwater images using MIP \citep{Carlevaris-Bianco:Initial:2010}. Images (a) and (c) are the input images with and without red color, respectively, while images (b) and (d) are their corresponding enhanced results. MIP fails in images with blue-green dominant light, leading to issues such as loss of details
}
    \label{fig:Bianco_DCP_vis}
\end{figure}
\begin{figure}[t!]
    \centering
    \includegraphics[width=1.\linewidth]{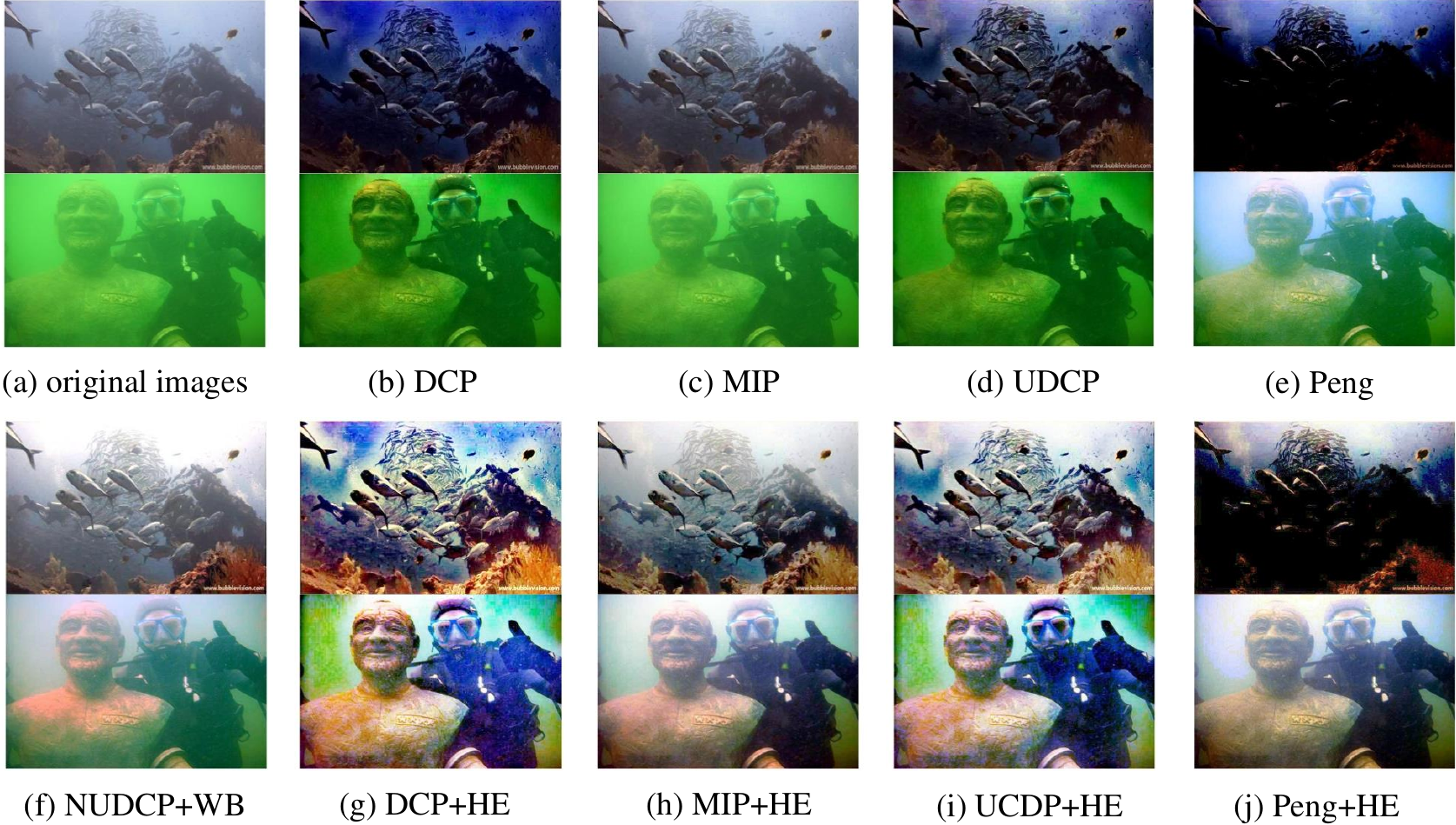}
    \caption{Comparative results of various underwater image enhancement methods. (a) Original images, (b) DCP~\citep{he2010single}, (c) MIP~\citep{Carlevaris-Bianco:Initial:2010}, (d) UDCP~\citep{DrewsJr:Transmission:2013}, (e) Peng~\citep{Peng:Underwater:2017}, (f) NUDCP + WB (white balance)~\citep{Song:Enhancement:2020}, (g) DCP + HE (histogram equalization), (h) MIP + HE, (i) UDCP + HE, and (j) Peng + HE}
    \label{fig:dcp_var_comparision}
\end{figure}

 
\paragraph{DCP variants for Underwater}
% Because real-world underwater scenes often exhibit spatially varying attenuation, multiple methods incorporate 
% rough geometry estimates to guide color recovery. 
Inspired by DCP, \citet{Carlevaris-Bianco:Initial:2010} proposed the maximum intensity prior (MIP) to estimate a coarse depth estimation by leveraging the difference between the red channel and the blue-green channels, $\max_ {{x} \in \Omega} I^{R}({x})-\max_{{x} \in \Omega, c \in{\{B, G\}}} I^{c}({x})$. However, since it relies entirely on the existence of red light, this algorithm is not applicable in deep-sea environments where red light is absent, as shown in \autoref{fig:Bianco_DCP_vis}.
As red light attenuates much faster than blue and green light, the smallest value among the RGB channels is always in the red channel in deep water scenes. Consequently, the DCP in RGB channels, termed $\text{DCP}_{RGB}$, becomes merely a zero map, which leads to an erroneous transmission map and results in poor restoration, as shown in \autoref{fig:dcp_var_comparision}. To address this problem, research works, such as those proposed by \citet{Wen:Single:2013}, \citet{DrewsJr:Transmission:2013} and \citet{Emberton:Hierarchical:2015}, calculate the dark channel based only on the blue and green channels, termed $\text{DCP}_{GB}$. For instance, \citet{DrewsJr:Transmission:2013} and \citet{Drews:Underwater:2016} propose Underwater Dark Channel Prior (UDCP) by focusing 
on the blue and green channels, typically dominant underwater. Later, \citet{Liang:GUDCP:2022} proposed a generalized method of Underwater Dark Channel Prior
(GUDCP), estimating image transmission from multiple spectral profiles of different water types, enhancing its robustness across varied underwater conditions.

\citet{Galdran:Automatic:2015} proposed the Red Channel Prior, which mitigates the erroneous transmission estimation caused by the small values in the red channel by inverting the values of the red channel. Additionally, it computes separate transmission maps for the three color channels: \(T^R(x)\), \(T^G(x)\), and \(T^B(x)\).
% However, like \citet{Carlevaris-Bianco:Initial:2010}, it remains ineffective in deep-sea environments dominated by blue-green wavelengths.
% A further 
% extension,  {Water Type Desensitized Restoration}~\citep{LI2020107038,Li:Underwater:2021}, uses the 
% model in \eqref{eq:uw_formation} along with an automatically inferred ater type?index to adapt $\alpha_c$ 
% region by region. This allows the method to handle varying turbidity or particulate concentrations across the 
% image, ensuring more stable color/contrast restoration.
\citet{Chiang:Underwater:2012} proposed a hybrid method combining wavelength compensation (to address color distortion from depth-dependent absorption) and dehazing (to reduce scattering effects) by modeling how longer 
wavelengths (e.g., red) attenuate more rapidly than shorter ones (e.g., blue/green). Their semi-inverse approach 
estimates an approximate $\alpha_c$ for each channel using:
$ \beta_c \;=\;\bigl(a_c + b_c\bigr)\,\beta\bigl(d(\mathbf{x})\bigr)$,
where $a_c$ and $b_c$ capture absorption and scattering effects for color channel $c$, and $\beta(d)$ modulates 
these based on depth or water type.
The estimated transmission map (TM) based on DCP exhibits block-like artifacts, resulting in a halo effect and blurred edges, even when soft matting~\citep{Levin:ClosedForm:2008} is applied to mitigate this issue.
To eliminate the halo effect and preserve boundaries, \citet{Yang:Low:2011} and \citet{Gibson:Investigation:2012} proposed the Median DCP. In short, this method replaces $\min _{{y} \in \Omega({x})}$ in Eq.~\eqref{eq:dcp_transmission} with $\mathrm{med}_{{y} \in \Omega({x})}$, where $\mathrm{med}$ represents a median filter.

\begin{figure}[t!]
    \centering
    \includegraphics[width=0.7\linewidth]{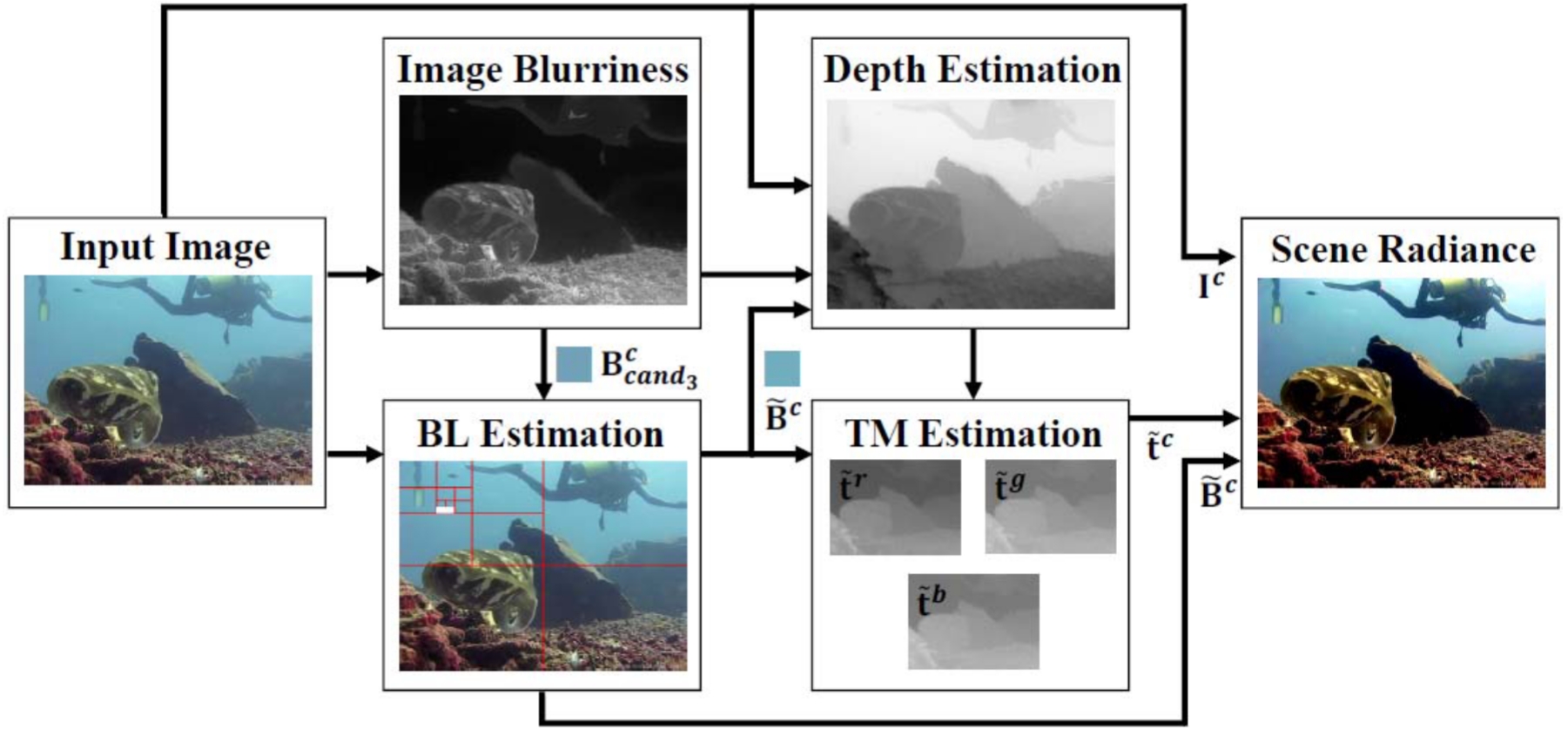}
    \caption{Illustration of an integrated framework, combining light absorption modeling and image blurriness estimation for UVE~\citep{Peng:Underwater:2017}}
    \label{fig:diagram_peng2017}
\end{figure}

In deep underwater scenes where sunlight is weak or absent, artificial light becomes the dominant illuminant. Under such conditions, points closer to the camera appear brighter, while those further away in the background appear darker. This illumination pattern directly contradicts the assumption of DCP.  This means simply relying on color information for transmission estimation is not enough. Therefore, \citet{Peng:Single:2015} proposed a method to estimate the transmission map and scene depth based on the level of blurriness in the scene, considering that objects further away from the camera exhibit a blurrier appearance due to the scattering effect. This approach effectively restores underwater images that deviate from the assumptions of DCP- or MIP-based methods, as it does not rely on color channel information for underwater scene depth estimation. In their sequel~\citep{Peng:Underwater:2017}, the authors further integrated light absorption and image blurriness information to estimate the transition map, the flowchart of which is presented in \autoref{fig:diagram_peng2017}.

%  clear underwater images (i.e., images with minimal distortion), DCP, , dark channel $J^{RGB}$
However, previous DCP-based or MIP-based methods often fail to estimate the background light $A$ in complex underwater conditions. 
\citet{Song:Enhancement:2020} argue that the poor performance of previous DCP-based methods stems from the overly aggressive assumption that $J^{RGB}_\text{dark} = 0$. To improve the accuracy of background light and transmission map estimation, the authors conducted a statistical analysis on 500 high-quality underwater images (i.e., images with minimal distortion) and found that the actual value is closer to $J_\text{dark} = 0.1$. Based on this finding, they proposed the New Underwater Dark Channel Prior (NUDCP), which adopts a less aggressive darkness assumption and leverages high-quality underwater images to mitigate artificial lighting distortions.
Additionally, their two-step enhancement approach (Restoration + Color Correction) further improves contrast, visibility, and color fidelity. However, the real underwater dataset used in their study, limited to 500 images, may not generalize well to all underwater conditions.

% To improve background light estimation in diverse underwater scenes, \citet{Song:Enhancement:2020}  introduce a statistical correction based on 500 real underwater images. For transmission map refinement, they proposed the New Underwater 
% propose the new underwater dark channel prior (NUDCP) that utilizes statistical pixel analysis to address non-uniform illumination, resolving the color distortion and detail loss caused by previous DCP-based methods?reliance on uniform lighting assumptions. 

% the statistic of clear and high resolution (HD) underwater images, then a scene depth map based on the underwater light attenuation prior and an adjusted reversed saturation map are applied to compensate and modify the coarse transmission map of the red channel.

To facilitate a clearer comparison and understanding of various IFM-based UVE methods, \autoref{tab:ifm_underwater_methods} summarizes the formulas used for background light and transmission map estimation. Additionally, \autoref{fig:dcp_var_comparision} presents qualitative comparisons of different MIP- and DCP-based approaches. As illustrated in the figure, these IFM-based methods struggle to effectively correct color distortion when not supplemented with additional color correction techniques such as histogram equalization or white balance.

So far, IFM-based methods, including those not previously mentioned~\citep{Liu:Underwater:2016, Liang:GUDCP:2022, peng2018generalization, li2016single, Wang:Single:2018}, have struggled to directly correct color distortion in images captured in blue-green seawater. In most cases, additional color correction is needed as a post-processing step to mitigate this issue. This limitation has led to the emergence of fusion- and Retinex-based methods, which offer improved color correction capabilities compared to purely IFM-based approaches. Overall, these single-image priors prioritize efficiency, requiring limited computation beyond local patch 
statistics. Although assumptions like horizontally homogeneous water or minimal forward scattering can be 
restrictive, the resulting simplicity still offers a favorable blend of speed and effectiveness in moderately 
challenging conditions.
% As scene radiance recovered from a degraded image often involves a color cast using Eq. (5), it often leads to an even stronger color cast. 
% Dire

% For underwater images, \citet{peng2018generalization} introduced per-channel attenuation to estimate a spatially 
% varying transmission $t(\mathbf{x})$, then solved for $J_c(\mathbf{x})$.
% Build up DCP, \citet{Liu:Underwater:2016} further cooperate color consistant loss for more accurate 

\begin{sidewaystable}[thp]
    \small
    \centering
    \renewcommand{\arraystretch}{2.2}
    \setlength{\tabcolsep}{20pt}
    \caption{Formulas for Estimation of background light (BL) and transmission map (TM) in underwater visual enhancement methods}
    \begin{tabular}{l | c c}
        \toprule
        Method & BL Estimation $(A  \text{ or } {A}^c$)  & TM Estimation $(\tilde{T} \text{ or } \tilde{T}^c)$  \\
        \midrule
        \citet{Chao:Removal:2010} & $I^c \big( \arg \max_x p(x) \big)$ & $\tilde{T}(x) = 1 - \min_c \Big( \min_{y \in \Omega(x)} \frac{I^c(y)}{{A}^c} \Big)$ \\
        \midrule
        \citet{Carlevaris-Bianco:Initial:2010} & $I^c \big( \arg \min_x \tilde{T}(x) \big)$ & $\tilde{T}(x) = D_\mathrm{MIP}(x) + (1 - \max_x D_\mathrm{MIP}(x))$ \\
        \midrule
        \citep{Yang:Low:2011} & $I^c \big( \arg \max_{x \in p_{0.1\%}} (\sum_c I^c(x)) \big)$ & $\tilde{T}(x) = 1 - \min_c \Big( \operatorname{med}_{y \in \Omega(x)} \frac{I^c(y)}{A^c} \Big)$ \\
        \midrule
        \multirow{2}{*}{\citep{Chiang:Underwater:2012}} & \multirow{2}{*}{$I^c \big( \arg \max_x I_{dark}^c(x) \big)$} & $\tilde{T}^R(x) = 1 - \min_k \Big( \min_{y \in \Omega(x)} \frac{I^c(y)}{A^c} \Big)$,  \\
        & & $\tilde{T}^c = (\tilde{T}^R)^{\frac{\beta^c}{\beta^R}}$ \\
        \midrule
        \multirow{2}{*}{\citep{Wen:Single:2013}} & \multirow{2}{*}{$I^c \Big( \arg \min_x \big( I_{dark}^R(x) - \max_{c'} (I_{dark}^{c'}(x)) \big) \Big)$} & $\tilde{T}^c(x) = 1 - \min_{c'} \Big( \min_{y \in \Omega(x)} \frac{I^{c'}(y)}{A^{c'}} \Big)$,  \\
        & &  $\tilde{T}^R = (\tau \max_y I^R(y))$, $\tau = \frac{\operatorname{avg}_x (\tilde{T}^c(x))}{\operatorname{avg}_x \big( \max_{y \in \Omega(x)} I^R(y) \big)}$ \\
        \midrule
        \citep{DrewsJr:Transmission:2013} & $I^c \big( \arg \max_x p(x) \big)$ & $\tilde{T}(x) = 1 - \min_k \Big( \min_{y \in \Omega(x)} \frac{I^R(y)}{A^R} \Big)$ \\
        \midrule
        \multirow{2}{*}{\citep{Galdran:Automatic:2015}} & \multirow{2}{*}{$I^c \big( \arg \min_{x \in p_{10\%}} I^R(x) \big)$} & $\tilde{T}^c(x) = 1 - \min \Big( \frac{\min_{y \in \Omega(x)} (1 - I^R(y))}{1 - A^R}, \frac{\min_{y \in \Omega(x)} I^G(y)}{A^G},$  \\
        & & $\frac{\min_{y \in \Omega(x)} I^B(y)}{A^B} \Big)$ \\
        \midrule
        \multirow{2}{*}{\citep{Zhao:Deriving:2015} } & \multirow{2}{*}{$I^c \big( \arg \max_{x \in p_{0.1\%}, c'} \lvert I^R(x) - I^{c'}(x) \rvert \big)$ } & $\tilde{T}^R(x) = 1 - \min_c \Big( \min_{y \in \Omega(x)} \frac{I^c(y)}{A^c} \Big)$,  \\
        & & $\tilde{T}^c = (\tilde{T}^R)^{\frac{\beta^c}{\beta^R}}$ \\
        \midrule
        \citep{Peng:Single:2015} & $\frac{1}{\lvert p_{0.1\%} \rvert} \sum_{x \in p_{0.1\%}} I^c(x)$ & $\tilde{T}(x) = F_s(P_{blr}(x))^\dagger$ \\
        \bottomrule
    \end{tabular}

    \footnotetext{$p$ denotes the gray-level pixel intensity of observed image I.}
    \footnotetext{$c \in \{R,G,B\}$ and $ c' \in \{G,B\}$}
    \label{tab:ifm_underwater_methods}
\end{sidewaystable}

% \subsubsection{Scene Depth Priors and Light Attenuation Modeling}
% Many prior-based algorithms revolve around the Beer--Lambert attenuation law, with the 
%  {simplified Jaffe--McGlamery model} serving as a baseline (\cf 
% Sec.~\ref{subsec:jaffe_mcglamery_models}). For instance, 
% \textbf{Drews \etal}~\citep{drews2016underwater} approximate the distance between camera and scene by analyzing 
% blurriness cues. \textbf{Song \etal}~\citep{song2020enhancement} propose a  {rapid scene-depth estimation} 
% method to handle variations in scattering coefficients. However, strong assumptions about homogeneous media or 
% constant ambient illumination sometimes limit their applicability to real-world scenarios.

\subsubsection{Retinex-based Methods}
Beyond the dark-channel variants,  {Retinex}-based approaches
decompose images into reflectance and illumination components, thereby tackling non-uniform lighting. 
For example, \citet{Hou:Benchmarking:2020} show that pre-estimating a color restoration map 
significantly helps with strong color cast removal. More recent approaches adopt multi-color space decompositions 
(\eg,  {HSV} or  {Lab} domains) to handle severe color shift, as in the  {UIEC\,$^2$-Net} 
\citep{Wang:UIEC^2Net:2021} or the  {WCID} approach~\citep{chen2021underwater}, both enhancing results for a variety 
of underwater conditions. Despite such progress, prior-based methods sometimes produce oversaturated reds or 
overcorrected backgrounds in complex scenes.

%------------------------------------------------------------

\subsubsection{Fusion-based Methods}
Fusion-based methods aim to integrate the strengths of multiple enhancement techniques to overcome the limitations of individual approaches when restoring degraded underwater images. In these methods, separate processes such as color correction, contrast enhancement, and dehazing are first applied to generate different "views" of the input image. Then, using various fusion strategies, these complementary results are combined into a single enhanced image with improved visibility, natural colors, and better contrast.

% Fusion-based underwater image enhancement methods leverage multiple sources of information to generate visually enhanced results. These methods integrate contrast correction, color compensation, and dehazing techniques by fusing multiple images. The fusion process selectively retains the most informative features, mitigating the limitations of individual enhancement approaches.

In earlier work, \citet{Mertens:Exposure:2009} introduced a straightforward exposure fusion technique to enhance underexposed images by compensating for insufficient illumination. Their method utilizes a Laplacian pyramid-based multi-scale fusion approach to blend multiple images with different exposure levels, producing a well-exposed final image. Later, \citet{Ancuti:Single:2013} extended this technique to the dehazing task by first generating two enhanced versions of the input image, one through white balance correction and the other via contrast enhancement. These two images were then fused multiple times using weight maps to produce a haze-free output. In work~\citep{Ancuti:Enhancing:2012,Ancuti:Color:2018}, a similar approach was adopted and applied to underwater image and video enhancement tasks. Furthermore, the authors introduced a temporal consistency filtering method to reduce noise in videos while preserving edge details. 
% \begin{equation}
% J(x) = \sum_{i=1}^{N} W_i(x) I_i(x),
% \end{equation}
% where \( J(x) \) is the final enhanced image, \( I_i(x) \) represents different processed versions, and \( W_i(x) \) are adaptive weight maps. This method effectively enhances brightness and fine details, but it relies on multiple input images or pre-enhanced versions.
Instead of using multi-scale Laplacian fusion, \citet{Vasamsetti:Wavelet:2017} adopted discrete wavelet transform (DWT) to decompose images into low-frequency and high-frequency components. The low-frequency component represents global brightness and color information, while the high-frequency component captures edges and texture details. Additionally, contrast enhancement was achieved using Euler-Lagrange Variational Optimization, effectively preventing excessive sharpening.

Due to the lack of local contrast and color adjustment in white balance, \citet{Garg:Underwater:2018} applied CLAHE to enhance local contrast and used percentile stretching on pixel intensities to restore attenuated colors. Additionally, the authors blended the enhanced results from the RGB color space with those from the HSV color space to further improve color restoration. \citet{Ghani:Underwater:2014,AbdulGhani:Enhancement:2015} used a similar process but additionally employed Rayleigh stretching to preserve edge details. In their subsequent work \citep{AbdulGhani:Automatic:2017}, the authors iteratively adjusted the histogram of the V (value) channel toward a Rayleigh distribution to enhance brightness and contrast while preserving color relationships in the HSV color space.

\subsection{Data-Driven Approaches}
\label{subsec:data_driven}

\subsubsection{CNN-Based Methods}

With the surge of deep learning,  {Convolutional Neural Networks} (CNNs) have shown promise in 
 {underwater image restoration}. Particularly when ground-truth or realistic synthetic data are available, supervised learning approaches are employed. Typical pipelines predict a per-pixel  {transmission map} or 
color correction from raw underwater inputs in an  {end-to-end} manner. Examples include:
     \citet{Li:Underwater:2020} constructed a dataset of aligned underwater/tank images 
    and used a CNN to predict color-corrected outputs via channel-wise attenuations.
    \citet{jiang2020novel} developed a multi-branch network with specialized modules 
    for color balancing and detail preservation, supervised by a curated dataset of synthetic pairs.
Recently, \citet{rao2023deep} proposed an end-to-end framework, integrating a color compensation module with an enhancement module. The first module extracts features from brightness and colors separately and merges them back using Probabilistic Volume Aggregation with simple MLP layers.\citet{Ju:Towards:2025} exploit a synthetic dataset to remove marine snow in the multi-scale Fourier domain using a simple U-Net architecture.

While these data-driven methods excel in typical underwater scenes, collecting fully {paired ground truth} 
is notoriously difficult. Hence, many rely on {synthetic data} generation pipelines 
(e.g., {Blender}, {Unreal Engine}) or {weakly supervised} setups. 
\citet{Li:WaterGAN:2017} introduced {WaterGAN} to synthesize underwater images for 
training restoration networks, enabling robust real-world inference. Similarly,  
\citet{Hou:Benchmarking:2020} compile extensive real datasets (U48, EUVP, UFO-120) with approximate reference 
targets, fostering deeper CNN training.

%\subsubsection{Underwater Denoising and Low-Light Enhancement}
In many scenarios, image sequences are captured under extreme low-light conditions with high ISO noise. 
\citet{fu2022unsupervised} incorporate {homology-based constraints} for 
denoising and color balancing, whereas \citet{liu2019underwater} adopt a deep residual 
framework to handle both shot noise and scattering artifacts. Recent {transformer-based} 
architectures ~\citep{peng2023u} further push performance, especially in 
extremely low signal-to-noise ratio conditions, by modeling global contexts across entire frames. Still, 
{noise statistics} vary drastically between different water types, raising open questions about 
{generalization} across diverse diving environments.

%------------------------------------------------------------
{\color{revblue}
\subsubsection{Transformer-Based Methods}
\label{subsec:transformers}

Transformer-based UIE has evolved into a major branch of learning-based enhancement because self-attention is well suited to the spatially non-uniform degradations of underwater imagery. In contrast to purely convolutional models, hierarchical or windowed transformers can aggregate long-range contextual cues that stabilise global colour correction while still preserving local structures through multi-scale feature fusion. This is particularly valuable underwater, where attenuation, backscatter, turbidity, and illumination often vary across the same frame rather than acting as a single global degradation. Building on the broader success of vision transformers and Swin-style hierarchical attention~\citep{liu2021swin}, recent UIE methods increasingly use transformer blocks to jointly model local texture recovery and scene-level colour consistency.
}

{\color{revblue}
One prominent line follows U-shape or hierarchical encoder--decoder designs that inject attention into otherwise U-Net-like enhancement pipelines. AutoEnhancer~\citep{tang2022autoenhancer} explores this direction through transformer-on-U-Net architecture search, while U-shape Transformer~\citep{peng2023u} provides a representative end-to-end design for multi-scale feature aggregation and robust colour correction. UDAformer~\citep{Shen:UDAformer:2023} strengthens this family with dual attention that combines channel self-attention and shifted-window pixel self-attention, and the adaptive group attention-based multiscale cascade transformer~\citep{huang2022underwater} further emphasises coordinated local--global restoration across scales. Related efforts such as underwater image enhancement using a pre-trained transformer~\citep{boudiaf2022underwater} and DAUT~\citep{badran2023daut} show how transfer from stronger transformer priors or depth-aware conditioning can further improve recovery in challenging scenes. Taken together, these methods establish the main transformer recipe in UIE: preserve the strong local-detail pathways of encoder--decoder restoration while using attention to better address non-uniform colour casts and contrast loss.
}

{\color{revblue}
More recent work moves beyond generic hierarchical attention toward underwater-specific transformer modules. WaterFormer~\citep{wen2024waterformer} is a representative example that combines global--local transformer reasoning with an explicit environment adaptor, allowing enhancement behaviour to be conditioned on prevailing water characteristics rather than assuming a single mapping for all domains. This design is especially relevant for robustness across different water types and turbidity levels. In parallel, UIE-Convformer~\citep{wang2024uie} blends convolutional inductive biases with transformer reasoning, X-CAUNet~\citep{pramanick2024x} explicitly models cross-colour channel interactions, Phaseformer~\citep{Khan:Phaseformer:2024} introduces phase-based attention to better recover structures under severe degradation, and the Globally Deformable Information Selection Transformer~\citep{zhuang2024globally} highlights flexible feature selection under spatially heterogeneous underwater distortions. These architectures indicate a clear shift from simply importing generic vision transformers toward designing transformer blocks that reflect underwater colour coupling, multi-scale scattering effects, and environment-dependent degradation.
}

{\color{revblue}
The transformer family is also expanding toward more task-aware and physics-aware variants. Reinforced Swin-Convs Transformer~\citep{ren2022reinforced} couples enhancement with super-resolution for degraded sensing imagery, TAFormer~\citep{li2025taformer} injects transmission-aware priors into transformer restoration, Histoformer~\citep{peng2025histoformer} uses histogram-guided modelling to correct global tone statistics efficiently, and UIE-SFIFormer~\citep{zhou2025uiesfiformer} combines physical guidance with spatial--frequency interaction. These extensions suggest that transformer-based UIE is no longer a single architectural template, but a spectrum of models that can incorporate transmission cues, colour-statistics priors, or physical guidance. At the same time, transformer-based enhancement remains computationally demanding and typically more data-hungry than lightweight CNN baselines. For practical deployment on low-power underwater platforms such as ROVs and AUVs, an important open issue is how to retain cross-domain robustness and stable restoration quality without incurring prohibitive memory, latency, or temporal-consistency costs.
}

{\color{revteal}
From a physics-aware viewpoint, the attraction of transformers is not only larger receptive fields. Global attention helps relate distant regions that share the same water mass, illumination trend, or attenuation pattern, making it easier to correct spatially varying colour loss and backscatter than with purely local convolution alone. This is particularly useful when artificial lighting or turbidity affects only part of the frame. The trade-off is that more expressive global modelling can also alter low-level gradients and radiometric ordering more aggressively, so reconstruction-oriented use still benefits from structure-aware losses and cross-view consistency constraints.
}

{\color{revsienna}
From an engineering perspective, this usually places transformer-based UIE on the offline or high-end near-online side of the deployment spectrum. Reported latency, FLOPs, and memory vary too widely across datasets and hardware to support a fair numeric ranking, but the practical trend is that full transformer restoration is rarely the safest choice for tightly power-bounded SLAM or navigation front-ends.
}

%------------------------------------------------------------
\subsubsection{Mamba-based Methods}
\label{subsec:Mamba_models}

{\color{revteal}
Recently, state space models, known as `Mamba,' have emerged as a linear alternative to Transformers \citep{Gu2024Mamba}. Their relevance to underwater enhancement is not merely architectural novelty: they offer a practical way to capture long-range dependencies caused by spatially varying attenuation and illumination while avoiding the quadratic memory growth of full attention. This efficiency is potentially valuable for field systems that must process high-resolution frames under limited onboard memory. The Mamba blocks assembled in a UNet-like architecture were first introduced by \citet{ruan2024vm}. MUIR \citep{Chen:MUIR:2024} incorporates depth estimation into the framework. UWMamba \citep{Guan:WaterMamba:2024} combines visual state space to capture long-range global features and convolution to capture local, detailed features. PixMamba \citep{Lin_2024_ACCV} improves the clarity and sharpness of fine details by merging two branches: pixel-level and patch-level, both based on Mamba architectures. BEM~\citep{huang2026bayesian} models the one-to-many mapping relations between input and targets by integrating vision Mamba into Bayesian neural networks. Relative to heavier transformers, these models offer a more attractive compromise between context modelling and efficiency, although most current implementations still target high-quality offline enhancement rather than tightly power-bounded AUV or ROV deployment.
}

{\color{revsienna}
Among global-context UIE families, compact Mamba variants are therefore one of the more plausible candidates for near-online robotic preprocessing. Even so, the current literature still reports them mainly in high-quality offline settings, and clearer platform-level reporting will be needed before practitioners can judge whether a given model is realistic for embedded underwater deployment.
}

%------------------------------------------------------------
\subsubsection{Diffusion Model-based Methods}
\label{subsec:DMs}

Diffusion models are among the most effective generative AI techniques, having demonstrated their ability to create realistic, high-resolution images and videos \citep{Anantrasirichai2025AI}. These techniques have also gained traction in underwater enhancement. The Denoising Diffusion Probabilistic Model (DDPM) was utilized in \citep{LU2023103926, Lu:speed:2024}, which required training on paired datasets. This method has been adapted to a patch-based approach \citep{Xia:patch:2025} to better capture local information and achieve higher resolution. Alternatively, UIEDP \citep{DU2025125271} utilizes a pre-trained diffusion model to circumvent the scarcity of paired underwater datasets. This method integrates natural image priors where clean in-air images provide balanced color distributions and rich structural details, which can be transferred into underwater restoration priors. Similarly, BDMUIE \citep{CHEN2025129274} uses prior information from clean air images in Bayesian diffusion models, effectively merging top-down (prior) and bottom-up (data-driven) information distributions.

{\color{revteal}
These models are attractive when degradation is severe because strong generative priors can recover missing contrast and plausible colour statistics. However, this same flexibility creates a tension with geometry-sensitive downstream use. Iterative denoising remains substantially heavier than one-pass CNN, Mamba, or GAN inference, and independently restored frames can drift in colour or local detail in ways that hurt key-point repeatability, pose estimation, or multi-view consistency. In practice, diffusion-based UIE is therefore better suited to offline post-processing, archival restoration, or human-facing visualisation than to real-time navigation, SLAM, or calibration-critical SfM stages unless additional structure, temporal, or cross-view constraints are imposed.
}

{\color{revsienna}
For the same reason, diffusion is currently better understood as an offline restoration or visualization tool than as a front-end for real-time robotic reconstruction. Without strong temporal or geometric constraints, per-frame diffusion enhancement can improve appearance while still changing the very cues that a SLAM, SfM, NeRF, or 3DGS pipeline depends on for stable optimization.
}

\subsubsection{Learning Under Limited or No Paired Supervision}
\label{subsec:GAN_contrastive}
\label{subsec:ssda_methods}

Paired clean references are particularly difficult to obtain underwater because the same scene is rarely captured with identical geometry, illumination, turbidity, and depth before and after degradation. Consequently, a large part of learning-based UIE has shifted toward settings with limited supervision, including unpaired translation, weak supervision, semi-supervised learning, and domain adaptation. We group these methods together because they address a shared problem: how to learn a robust restoration mapping when synthetic supervision is incomplete and real deployment conditions differ from the training distribution.

One major route relies on adversarial learning and cycle-consistent translation. WaterGAN~\citep{Li:WaterGAN:2017} synthesizes underwater imagery from in-air data to create realistic training pairs, while \citet{fabbri2018enhancing} use GAN-based restoration to improve perceptual quality under real degradation. Domain-adversarial and unpaired translation frameworks such as \citet{uplavikar2019all} further reduce the gap between synthetic and real underwater domains, and cycle-consistent variants such as U-CycleGAN~\citep{anwar2020diving}, the multi-scale formulation of \citet{desai2021ruig}, and the model-driven UW-CycleGAN~\citep{yan2023uw} show how adversarial mapping can be constrained to preserve colour trends and structure without requiring exact pixel-aligned references. Their main limitation is that global appearance transfer can still hallucinate textures or over-correct colours when the source and target domains only partially overlap.

Another line of work seeks to stabilise weakly supervised learning through representation regularisation rather than relying only on adversarial discrimination. Twin adversarial contrastive learning~\citep{liu2022twin} uses contrastive consistency to retain structure during enhancement, while the perception-driven framework without paired supervision of \citet{jiang2023perception} explicitly optimizes perceptual objectives when reference images are unavailable. Content-style disentanglement methods such as \citet{zhu2023unsupervised} and double-order contrastive formulations such as \citet{yin2024unsupervised} further separate degradations from scene content, aiming to reduce hallucination and make no-paired-supervision training less brittle. Compared with plain GAN translation, these models generally place more emphasis on preserving semantic layout and suppressing over-enhancement.

{\color{revblue}
A third branch explicitly targets semi-supervised and domain-adaptive transfer. The two-step framework of \citet{jiang2022two} decomposes degradation adaptation and enhancement refinement, while \citet{chen2022domain} and \citet{bing2023domain} adapt features across domains through content-style separation or in-air-to-underwater transfer. Semi-UIR~\citep{huang2023contrastive} introduces contrastive semi-supervised learning with a reliable bank to stabilise optimisation on real data, and \citet{wen2025ssduie} further combine synthetic supervision with cross-domain feature alignment for real-world UIE. These methods are especially important underwater because synthetic data remain useful for controllable supervision, yet model performance often collapses unless the training objective also adapts to the statistics of real water types and turbidity conditions.
}

{\color{revgreen}
Domain-adversarial learning provides a particularly direct solution path for this transfer problem. In addition to early domain-adversarial formulations such as \citet{uplavikar2019all}, \citet{kapoor2023domain} show that aligning latent features across source and target domains during enhancement training can improve real-world robustness without assuming abundant paired underwater annotations.
}

Taken together, these methods show that underwater UIE increasingly depends not only on the backbone architecture, but also on how supervision is constructed when exact references are absent. Some recent approaches go one step further by coupling limited-supervision learning with explicit physical or model-based components; we discuss those physics-guided designs separately in \autoref{subsec:hybrid_models} to avoid conflating supervision strategy with physical/model coupling.

{\color{revgreen}
Table~\ref{tab:data_driven_uie_timeline} complements the taxonomy in \autoref{fig:masked_tree} by summarising the data-driven UIE methods discussed in \cref{subsec:data_driven} in chronological order, making the progression from early GAN/CNN models to transformer-, Mamba-, diffusion-, and domain-adaptive methods more explicit.}

{\color{revgreen}
\begingroup
\scriptsize
\setlength{\LTleft}{0pt}
\setlength{\LTright}{0pt}
\setlength{\LTpre}{6pt}
\setlength{\LTpost}{6pt}
\setlength{\LTcapwidth}{\textwidth}
\setlength{\tabcolsep}{2pt}
\renewcommand{\arraystretch}{1.15}
\begin{longtable}{@{}p{3.0cm}p{2.50cm}p{7.45cm}p{2.55cm}@{}}
\caption{Chronological summary of data-driven underwater image enhancement methods, highlighting the shift from early deep-learning models toward more recent transformer-, Mamba-, diffusion-, and domain-adaptive approaches.}
\label{tab:data_driven_uie_timeline}\\
\toprule
\textbf{Method} & \textbf{Family / Setting} & \textbf{Contribution} & \textbf{Remarks} \\
\midrule
\endfirsthead
\multicolumn{4}{c}{\tablename~\thetable{} -- Continued from previous page}\\
\toprule
\textbf{Method} & \textbf{Family / Setting} & \textbf{Contribution} & \textbf{Remarks} \\
\midrule
\endhead
\midrule
\multicolumn{4}{r}{Continued on next page}\\
\endfoot
\bottomrule
\endlastfoot
\citet{Li:WaterGAN:2017} & GAN / synthetic paired & Synthesizes realistic underwater imagery from in-air data to create restoration training pairs for subsequent enhancement models. & Data generation; colour correction \\
\citet{fabbri2018enhancing} & GAN / perceptual enhancement & Uses generative adversarial restoration to improve the perceptual quality of real degraded underwater images. & Perceptual restoration \\
\citet{liu2019underwater} & CNN / residual & Applies a deep residual framework to suppress noise and scattering artefacts while recovering colour and contrast. & Denoising + enhancement \\
\citet{uplavikar2019all} & GAN / domain-adversarial & Learns domain-invariant enhancement across diverse water types through domain-adversarial training. & Cross-domain robustness \\
\citet{Li:Underwater:2020} & CNN / supervised & Introduces the UIEB benchmark and a benchmark-driven supervised enhancement framework built from candidate restorations and learned fusion. & Benchmark-driven UIE \\
\citet{jiang2020novel} & CNN / supervised & Develops a multi-branch network for coordinated colour balancing and detail preservation using curated paired supervision. & Colour-detail balance \\
\citet{anwar2020diving} & Cycle-consistent / unpaired & Represents the paired-free cycle-consistent restoration line that constrains translation without exact pixel-aligned references. & Unpaired restoration \\
\citet{desai2021ruig} & GAN / generation & Generates realistic underwater imagery to narrow the gap between synthetic training data and real degradations. & Data generation; domain bridging \\
\citet{fu2022unsupervised} & CNN / unsupervised & Uses homology-based constraints to regularize denoising and colour balancing without paired supervision. & Severe turbidity; overlaps with hybrid discussion \\
\citet{tang2022autoenhancer} & Transformer / NAS & Searches transformer-on-U-Net enhancement architectures automatically to improve restoration design under underwater degradation. & Architecture search \\
\citet{huang2022underwater} & Transformer / multiscale & Uses adaptive group attention and a multiscale cascade design for coordinated local-global enhancement. & Multi-scale attention \\
\citet{boudiaf2022underwater} & Transformer / pre-trained & Transfers a pre-trained transformer prior into underwater enhancement to improve restoration under difficult conditions. & Transfer learning \\
\citet{ren2022reinforced} & Transformer / task-aware & Couples enhancement with super-resolution in a reinforced Swin-Convs transformer for degraded sensing imagery. & Joint enhancement + SR \\
\citet{liu2022twin} & Contrastive / weak supervision & Introduces twin adversarial contrastive learning to preserve structure during paired-free enhancement. & Contrastive consistency \\
\citet{jiang2022two} & Semi-supervised / domain-adaptive & Decomposes degradation adaptation and enhancement refinement into a two-step synthetic-to-real transfer framework. & Two-step transfer \\
\citet{chen2022domain} & Domain adaptation & Separates content and style to adapt enhancement models across domains with differing underwater appearance statistics. & Content-style separation \\
\citet{peng2023u} & Transformer / U-shape & Provides an end-to-end U-shape transformer with strong multi-scale aggregation and robust colour correction. & Overall enhancement \\
\citet{Shen:UDAformer:2023} & Transformer / dual-attention & Combines channel self-attention and shifted-window pixel attention to improve underwater restoration. & Dual attention \\
\citet{badran2023daut} & Transformer / depth-aware & Injects depth-aware conditioning into a U-shape transformer to better handle scene-dependent degradation. & Depth-aware conditioning \\
\citet{LU2023103926} & Diffusion / paired & Adapts DDPM to underwater restoration with paired supervision for high-fidelity generative enhancement. & Paired-data dependent \\
\citet{yan2023uw} & GAN / model-driven & Constrains CycleGAN-style restoration with model-driven colour and structure cues to reduce artefacts. & Unpaired restoration \\
\citet{jiang2023perception} & Weak supervision / perceptual & Optimizes perceptual objectives for underwater enhancement without requiring paired references. & Perception-driven \\
\citet{zhu2023unsupervised} & Disentanglement / unpaired & Separates scene content and degradation style to stabilise unsupervised underwater enhancement. & Content-style disentanglement \\
\citet{bing2023domain} & Domain adaptation / transfer & Transfers in-air priors to underwater enhancement through deep domain adaptation. & In-air to underwater \\
\citet{huang2023contrastive} & Semi-supervised & Uses a reliable-bank contrastive strategy to stabilise semi-supervised underwater restoration on real data. & Reliable bank \\
\citet{kapoor2023domain} & Domain-adversarial & Aligns latent source and target features during training to improve real-world UIE robustness. & Cross-domain robustness \\
\citet{rao2023deep} & CNN / end-to-end & Integrates colour compensation with enhancement through coupled feature extraction and aggregation. & Generalised colour compensation \\
\citet{wen2024waterformer} & Transformer / environment-adaptive & Couples global-local transformer reasoning with an environment adaptor to condition restoration on water characteristics. & Water-type adaptation \\
\citet{wang2024uie} & Transformer / convformer & Blends convolutional inductive bias with transformer reasoning for balanced local and global restoration. & Conv-transformer hybrid \\
\citet{pramanick2024x} & Transformer / channel-aware & Explicitly models cross-colour channel interactions for underwater enhancement. & Colour coupling \\
\citet{zhuang2024globally} & Transformer / deformable selection & Uses globally deformable information selection to adapt feature usage under spatially heterogeneous distortions. & Flexible feature selection \\
\citet{Chen:MUIR:2024} & Mamba / depth-aware & Extends Mamba-based restoration with integrated depth estimation to improve underwater enhancement. & Depth-aware Mamba \\
\citet{Guan:WaterMamba:2024} & Mamba / visual state space & Combines state-space global modelling with local convolutions for underwater enhancement. & Long-range + local detail \\
\citet{Lin_2024_ACCV} & Mamba / dual-branch & Uses pixel-level and patch-level Mamba branches to improve clarity and fine-detail recovery. & Sharpness/detail recovery \\
\citet{Lu:speed:2024} & Diffusion / efficient & Speeds up DDPM-based underwater enhancement toward more practical real-time deployment. & Real-time oriented \\
\citet{yin2024unsupervised} & Contrastive / disentanglement & Uses double-order contrastive disentanglement to improve unpaired underwater enhancement. & Stronger regularisation \\
\citet{Ju:Towards:2025} & CNN / synthetic marine snow & Removes marine snow in the multi-scale Fourier domain using synthetic supervision and a lightweight U-Net backbone. & Marine snow removal \\
\citet{Khan:Phaseformer:2024} & Transformer / phase-aware & Introduces phase-based attention to recover structures under severe underwater degradation. & Structure recovery \\
\citet{li2025taformer} & Transformer / physics-aware & Injects transmission-aware priors into transformer-based restoration. & Transmission-aware \\
\citet{peng2025histoformer} & Transformer / histogram-guided & Uses histogram modelling to improve efficient global tone correction in underwater images. & Tone statistics \\
\citet{zhou2025uiesfiformer} & Transformer / spatial-frequency & Combines physical guidance with spatial-frequency interaction for underwater enhancement. & Physics-aware transformer \\
\citet{Xia:patch:2025} & Diffusion / patch-based & Uses patch-based diffusion to capture local details more accurately at higher resolution. & High-resolution local detail \\
\citet{DU2025125271} & Diffusion / prior-guided & Transfers diffusion priors from clean natural images to reduce dependence on paired underwater training data. & Diffusion prior \\
\citet{CHEN2025129274} & Diffusion / Bayesian & Combines air-image priors with Bayesian diffusion to merge top-down and bottom-up restoration cues. & Top-down + bottom-up \\
\citet{wen2025ssduie} & Semi-supervised / domain-adaptive & Combines synthetic supervision with cross-domain feature alignment for real-world underwater enhancement. & Real-world transfer \\
\citet{huang2026bayesian} & Mamba / Bayesian & Models one-to-many enhancement mappings with Bayesian neural networks and Vision Mamba. & Uncertainty-aware \\
\end{longtable}
\endgroup
}

%------------------------------------------------------------
\subsection{Hybrid Approaches}\label{subsec:hybrid_models}

{\color{revblue}
Hybrid approaches in underwater UIE explicitly learned restoration with underwater imaging knowledge, i.e., physics-guided UIE. Whereas \cref{subsec:ssda_methods} focuses on how enhancement models learn when paired references are scarce, this subsection focuses on what is coupled into the model: underwater IFMs, transmission or background-light estimation, scene priors, decomposition constraints, and controllable synthetic-real training mechanisms. The common goal is to move beyond a purely black-box mapping and instead embed medium-aware structure into the restoration process.
}

{\color{revblue}
One major line couples networks with scene priors or physically motivated restoration terms. Underwater scene prior inspired enhancement~\citep{LI2020107038} uses scene priors to guide restoration across images and videos, while perceptual underwater image enhancement with deep learning and physical priors~\citep{chen2020perceptual} combines learned enhancement with physically informed constraints on color and structure. Related designs such as the generalized physical-knowledge-guided dynamic model~\citep{mu2023generalized} and IPMGAN~\citep{liu2021ipmgan} further show how physical priors can be injected into the objective, network dynamics, or adversarial training process. Across these methods, the hybrid element lies in reintroducing underwater imaging knowledge into optimization rather than relying solely on a data-driven image-to-image transform.
}

{\color{revblue}
A second line emphasizes decomposition and explicitly constrained architectures. Instead of predicting an enhanced image directly, these methods decompose the problem into physically meaningful components such as transmission, background light, or scene radiance, then optimize them jointly. The homology-driven framework of \citet{fu2022unsupervised} illustrates how domain knowledge can regularize unsupervised restoration under severe turbidity, while Mamba-UIE~\citep{Zhang2024MambaUIE} uses a physically constrained multi-branch formulation to estimate medium-related variables and scene content together. Hybrur~\citep{yan2023hybrur} likewise combines physical modeling with neural restoration in an unsupervised setting, and wavelet-guided designs such as \citet{Zhang:Underwater:2025} show that hybrid coupling can also occur through reflectance or frequency-domain priors. These methods make clear that hybrid UIE is not only about mixing data sources; it also includes model-level and objective-level coupling.
}

{\color{revblue}
A third line focuses on synthetic-real bridging under physical guidance. Here the aim is to exploit controllable degradations from simulation or physical modeling while still anchoring training to real underwater imagery. SyreaNet~\citep{wen2023syreanet} is a representative example of this direction rather than its sole defining method: it uses physically controllable synthetic degradations together with real images to reduce the gap between modeled water conditions and deployment scenes. In this sense, synthetic-real bridging complements the limited/no-paired-supervision strategies in \cref{subsec:ssda_methods}, but we discuss it here because the defining feature is not merely the supervision regime; it is the explicit physical or model-coupled design that governs how supervision is constructed. Across these strands, physics-guided and model-coupled UIE offers a promising route to better cross-domain stability~\citep{LI2020107038,Hou:Benchmarking:2020}, although its effectiveness still depends on how well the chosen priors and decomposition assumptions match real underwater conditions.
}

%------------------------------------------------------------
\subsection{Evaluation and Benchmark}
\label{subsec:eval_benchmarks}

\begin{table*}[t]
    \centering
    \scriptsize
    \renewcommand{\arraystretch}{1.3}
    \setlength{\tabcolsep}{1pt}
    \caption{Summary of publicly available underwater image enhancement and restoration datasets with their key properties. ``\#~Images'' gives the approximate number of samples. ``Real/Synth.'' specifies whether data are captured in real water or generated synthetically via simulation/rendering. ``Paired?'' indicates whether ground-truth/reference images are provided}
    \label{tab:underwater_datasets_extended}
    
    \resizebox{\textwidth}{!}{%
    \begin{tabular}{l c c c c c}
        \toprule
        \textbf{Dataset} & \textbf{Year} & \textbf{\#~Images} & \textbf{Real/Synth.} & \textbf{Paired?} & \textbf{Download} \\
        \midrule
        UIEB~\citep{Li:Underwater:2020}
          & 2019
          & 890 
          & Real 
          & Approx. paired 
          & \href{https://li-chongyi.github.io/proj_benchmark.html}{Link} \\

        EUVP~\citep{Islam:Fast:2020}
          & 2020 
          & 4,414 real + 6,850 syn.
          & Both 
          & Paired (syn.) / Unpaired (real) 
          & \href{http://irvlab.cs.umn.edu/resources/euvp-dataset}{Link} \\

        U45~\citep{li2019fusion}
          & 2019
          & 45 
          & Real 
          & Unpaired 
          & \href{https://github.com/IPNUISTlegal/underwater-test-dataset-U45-}{Link} \\

        UFO-120~\citep{islam2020simultaneous}
          & 2020 
          & 1,550 
          & Real 
          & Unpaired
          & \href{https://irvlab.cs.umn.edu/resources/ufo-120-dataset}{Link} \\

        RUIE~\citep{Liu:RealWorld:2020}
          & 2020 
          & 4230
          & Real 
          &  Unpaired
          & \href{https://github.com/dlut-dimt/Realworld-Underwater-Image-Enhancement-RUIE-Benchmark/tree/master}{Link} \\

        WaterGAN~\citep{Li:WaterGAN:2017}
          & 2017 
          & 10K+ (synthetic) 
          & Synthetic 
          & Paired (rendered pairs) 
          & \href{https://github.com/kskin/WaterGAN}{Link} \\

        Sea-Thru~\citep{Akkaynak:seathrue:2019}
          & 2019 
          & 1,114 
          & Real 
          & Partial (depth-registered) 
          & \href{https://www.kaggle.com/datasets/colorlabeilat/seathru-dataset}{Link} \\

        SQUID~\citep{berman2020underwater}
          & 2020
          & Various 
          & Real 
          & Unpaired 
          & \href{https://csms.haifa.ac.il/profiles/tTreibitz/datasets/ambient_forwardlooking/index.html}{Link} \\

        OceanDark~\citep{marques2020l2uwe}
          & 2020
          & 183 
          & Real 
          & Unpaired 
          & \href{https://sites.google.com/view/oceandark/home}{Link} \\

        LNRUD~\citep{ye2022underwater}
          & 2022
          & 50,000  
          & Synthetic 
          & Paired 
          & \href{https://github.com/Ephemeral182/UWNR}{Link} \\
          
        LSUI~\citep{peng2023u}
          & 2023
          & 4,279 
          & Real 
          & Paired 
          & \href{https://drive.google.com/file/d/10gD4s12uJxCHcuFdX9Khkv37zzBwNFbL/view}{Link} \\

        UVEB~\citep{Xie:UVEB:2024}
          & 2024
          & 453,000
          & Real 
          & Paired 
          & \href{https://github.com/yzbouc/UVEB}{Link} \\

        \bottomrule
    \end{tabular}
    }

        % RUIE 
        %   & 2019
        %   & Various 
        %   & Real 
        %   & Unpaired 
        %   & \href{https://csms.haifa.ac.il/profiles/tTreibitz/datasets/ambient_forwardlooking/index.html}{Link} \\
        % SUIM 
        %   & 2020 
        %   & 1,525 
        %   & Real 
        %   & Unpaired 
        %   & \href{https://irvlab.cs.umn.edu/resources/suim-dataset}{Link} \\

        % UIQS 
        %   & 2017 
        %   & 60 real scenes 
        %   & Real 
        %   & Unpaired 
        %   & \textit{(Research pages)} \\

    % \vspace{0.5em} % 
    % \begin{flushleft}
    % \scriptsize
    % \textbf{Notes}:
    % \begin{itemize}
    %     \item \textbf{UIEB} provides approximate GT references.  
    %     \item \textbf{EUVP}, \textbf{U45}, and \textbf{UFO-120} provide {unpaired real} and 
    %      {paired synthetic} subsets for training and evaluating various enhancement methods. 
    %     \item \textbf{WaterGAN} offers fully synthetic data, simulating scattering/attenuation for 
    %     ground-truth supervision. 
    %     \item \textbf{Sea-Thru} includes depth measurements to facilitate physically accurate color 
    %     correction.
    %     % \item \textbf{SUIM} is primarily used for segmentation tasks but also relevant for enhancement benchmarking. 
    %     % \item \textbf{RUIE} and \textbf{UIQS} contain diverse real-world samples but do {not} provide strictly paired images.
    % \end{itemize}
    % \end{flushleft}
\end{table*}

A key challenge in underwater enhancement is the scarcity of reliable ground-truth pairs. Several datasets attempt to mitigate this issue, including UIEB~\citep{Li:Underwater:2020}, EUVP~\citep{Islam:Fast:2020}, UVEB~\citep{Xie:UVEB:2024}, and LSUI~\citep{peng2023u}. However, in UIEB the pairs are not fully registered, which introduces misalignment between degraded inputs and reference images, thereby limiting the reliability of supervised learning.

The EUVP dataset~\citep{Islam:Fast:2020} provides over 12K paired and 8K unpaired samples, where the paired references are generated using CycleGAN. U45~\citep{Hou:Benchmarking:2020} serves as a test set of 45 real-world underwater images degraded by color casts, low contrast, and haze-like effects. 
OceanDark~\citep{marques2020l2uwe} is composed of 183 underwater images of 1280 x 720 pixels captured by video cameras located in profound depths using artificial lighting.
LNRUD~\citep{ye2022underwater} contains 50000 clean images and 50000 corresponding underwater images synthesized from 5000 real underwater scene images. A comprehensive dataset summary is provided in \autoref{tab:underwater_datasets_extended}.
% Large-scale synthetic datasets~\citep{Li:WaterGAN:2017, Olson:Synthetic:2018} approximate scattering with the Jaffe--McGlamery model, varying attenuation coefficients to simulate different water types.

{\color{revpurple}
For benchmarking and training design, it is useful to distinguish three dataset regimes. First, \emph{fully synthetic} resources such as WaterGAN and LNRUD offer controllable degradations and clean supervision, but inevitably simplify real underwater variability. Second, \emph{pseudo-paired or model-based generated} resources use physics-inspired generation or restoration targets to construct approximate pairs; representative examples include the scene-prior-driven framework of \citet{LI2020107038} and the synthetic-real bridging strategies discussed in Section~\ref{subsec:hybrid_models}. Third, \emph{real unpaired or approximately paired} collections such as UIEB, EUVP, Sea-Thru, LSUI, and UVEB better reflect deployment conditions, but often trade away exact correspondence or strict radiometric ground truth. This distinction matters because reported gains are often tied as much to the supervision regime as to the model architecture itself.
}
While these resources have significantly advanced the field, the lack of standardized evaluation protocols and the inherent variability of underwater conditions continue to impede fair comparison across methods. For a comprehensive benchmark of deep learning-based approaches, we refer readers to \citet{CONG2021337}.

{\color{revgreen}
For readers interested in task-driven evaluation beyond enhancement quality, Fish4Knowledge~\citep{spampinato2014typhoon} and the Underwater Change Detection Dataset~\citep{radolko2016dataset} provide useful complementary resources for fish monitoring, behaviour analysis, change detection, and moving-object segmentation. These datasets are not enhancement benchmarks in the strict sense, but they help illustrate whether an enhancement method preserves motion cues, target boundaries, and background stability well enough for adjacent video-analysis settings.
}

\subsubsection{Evaluation Metrics}
\label{sssec:evaluation}

The assessment of image enhancement quality is non-trivial, particularly in underwater scenarios where distortions are complex and subjective human perception often diverges from pixel-wise fidelity measures. In this section, we review both general-purpose image quality metrics and underwater-specific evaluation criteria.

\begin{itemize}
\itemsep2em
\item \textbf{MSE and PSNR.}
The most widely used distortion-based criteria are Mean Squared Error (MSE) and its logarithmic counterpart, Peak Signal-to-Noise Ratio (PSNR). MSE computes the average squared difference between a reference image $x$ and a test image $y$, expressed as
\begin{equation}
    \text{MSE} = \frac{1}{N}\sum_{i=1}^{N}(x_i - y_i)^2,
\end{equation}
where $N$ denotes the number of pixels. PSNR is subsequently derived as
\begin{equation}
    \text{PSNR} = 10 \log_{10}\frac{L^2}{\text{MSE}},
\end{equation}
with $L$ representing the dynamic range of pixel intensities (\eg, 255 for 8-bit images).  
Despite their mathematical simplicity, clear interpretation, and popularity in optimization contexts, these measures exhibit weak correlation with human visual perception~\citep{wang2009mean}. They treat all pixel errors equally and ignore spatial structures, making them inadequate for perceptual quality assessment.

\item \textbf{SSIM.}
\citet{wang2002universal} introduced the Structural Similarity Index (SSIM), which evaluates luminance, contrast, and structural similarity between local patches. Formally, SSIM is defined as
\begin{equation}
\begin{split}
\text{SSIM}(x,y) &= l(x,y)\cdot c(x,y)\cdot s(x,y) \\
&= \left(\frac{2\mu_x\mu_y+C_1}{\mu_x^2+\mu_y^2+C_1}\right)
   \left(\frac{2\sigma_x\sigma_y+C_2}{\sigma_x^2+\sigma_y^2+C_2}\right)
   \left(\frac{\sigma_{xy}+C_3}{\sigma_x\sigma_y+C_3}\right),
\end{split}
\end{equation}
where $\mu$, $\sigma$, and $\sigma_{xy}$ denote local means, standard deviations, and cross-covariance, respectively. Constants $C_1$, $C_2$, and $C_3$ stabilize the computation. SSIM has been demonstrated to align more closely with the human visual system by emphasizing structural integrity rather than pixel accuracy.

\item \textbf{PCQI.}
The Patch-based Contrast Quality Index (PCQI)~\citep{wang2015patch} extends perceptual assessment by explicitly modeling three independent components within local patches: mean intensity, contrast change, and structural distortion. It is formulated as
\begin{equation}
\text{PCQI} = q_i(x,y) \cdot q_c(x,y) \cdot q_s(x,y),
\end{equation}
where $q_i$, $q_c$, and $q_s$ quantify luminance, contrast, and structure fidelity, respectively. Compared with global metrics, PCQI better reflects local contrast variations but incurs higher computational complexity.

\item \textbf{UCIQE.}
For underwater imagery, \citet{yang2015underwater} proposed the Underwater Color Image Quality Evaluation (UCIQE) index, which operates in the perceptually uniform CIELab color space. UCIQE linearly combines chroma dispersion, luminance contrast, and average saturation as
\begin{equation}
\text{UCIQE} = C_1\sigma_c + C_2 con_l + C_3 \mu_s,
\end{equation}
where $\sigma_c$ is the standard deviation of chroma, $con_l$ the luminance contrast, and $\mu_s$ the mean saturation. UCIQE is widely used because it can be computed without reference images and captures several degradations that are common underwater; however, like other no-reference metrics, it reflects only a limited proxy of perceptual quality and should not be interpreted as a substitute for reference-based fidelity measures when reliable ground truth is available. {\color{revteal}Its components can also reward over-saturated or contrast-stretched outputs, so a high UCIQE score does not necessarily imply physically plausible colour recovery or stable multi-view geometry.}

\item \textbf{UIQM.}
Another underwater-specific measure, the Underwater Image Quality Measure (UIQM)~\citep{panetta2015human}, explicitly incorporates aspects of the human visual system (HVS) without requiring ground-truth images. UIQM integrates three components: the Underwater Image Colorfulness Measure (UICM), the Underwater Image Sharpness Measure (UISM), and the Underwater Image Contrast Measure (UIConM):
\begin{equation}
\text{UIQM} = c_1 \cdot \text{UICM} + c_2 \cdot \text{UISM} + c_3 \cdot \text{UIConM},
\end{equation}
where the coefficients $c_1$, $c_2$, and $c_3$ can be adjusted depending on whether color fidelity, sharpness, or contrast is prioritized. UIQM is practical when no reference image exists, but it can still reward over-sharpening or colour shifts that are undesirable for restoration fidelity or downstream vision tasks. {\color{revteal}This is especially important for robotics and reconstruction, where inflated sharpness or colourfulness scores can coincide with degraded feature repeatability, altered radiometry, or unstable correspondences.}

\item \textbf{UIF.}
More recently, \citet{Zhou:Underwater:2023} proposed the Underwater Image Fidelity (UIF) index, which aims to measure color consistency in underwater scenes by analysing the distribution of pixel intensities in multiple sub-interval histograms. Specifically, UIF divides the chroma channel into $K$ intervals and computes the fidelity by evaluating deviations between the reference-free enhanced image and an idealized uniform distribution. Formally, UIF can be expressed as
\begin{equation}
    \text{UIF} = \sum_{k=1}^{K} w_k \cdot \left(1 - \frac{|h_k - \bar{h}_k|}{\bar{h}_k + \epsilon}\right),
\end{equation}
where $h_k$ is the observed histogram value in the $k^{th}$ sub-interval, $\bar{h}_k$ is the expected reference value under an idealized distribution, $w_k$ is a weighting factor reflecting the perceptual importance of the interval, and $\epsilon$ is a small constant to prevent division by zero.  

By quantifying chroma distribution consistency, UIF provides a more fine-grained evaluation of underwater color distortions than global statistics such as UCIQE or UIQM. Nevertheless, the correlation with subjective human perception is still imperfect, highlighting the need for task-driven or perceptual-learning-based evaluation frameworks in underwater image enhancement.

\item \textbf{\texorpdfstring{{\color{revpurple}Learned underwater IQA and ranking}}{Learned underwater IQA and ranking}}.
{\color{revpurple}
Recent work has started to replace hand-crafted no-reference criteria with learned ranking models. Underwater Ranker~\citep{guo2023underwater}, for example, learns pairwise preferences over underwater image quality and can be used both to compare outputs and to guide enhancement. This direction is important because subjective preference, restoration fidelity, and downstream task utility are not equivalent objectives: an image preferred by a human observer may still alter geometric cues, detection boundaries, or colour constancy in ways that hurt analysis. Learned ranking therefore complements, rather than replaces, full-reference metrics, classical no-reference metrics, and task-driven evaluation.
}

\item \textbf{\texorpdfstring{{\color{revteal}Task-driven geometry and robotics metrics}}{Task-driven geometry and robotics metrics}}.
{\color{revteal}
For robotics and 3D reconstruction, enhancement quality should also be judged by whether the restored images preserve the cues needed by downstream geometry pipelines. Relevant indicators include feature repeatability and inlier matching rates, track length, reprojection or pose-estimation error, and downstream SLAM robustness or drift. These measures are often more meaningful than UIQM or UCIQE when the target application is SfM, visual SLAM, COLMAP, or neural rendering, because they test whether enhancement helps or harms correspondence stability and geometric consistency rather than only whether the image appears vivid~\citep{Malyugina:beam:2025,vrochidis2025underwater3d}.
}

{\color{revsienna}
For joint enhancement--reconstruction pipelines, geometry-aware reporting should go further still. Useful targets include Chamfer Distance or Cloud-to-Cloud (C2C) distance for reconstructed geometry, Absolute Trajectory Error (ATE) or Relative Pose Error (RPE) for camera-motion quality, and track-stability or reprojection statistics for correspondence reliability. The current literature rarely reports these together with classical UIE metrics, which makes cross-domain assessment difficult and should itself be regarded as a major open challenge.
}

\end{itemize}

\begin{table*}[t]
\centering
\small
\setlength{\tabcolsep}{6pt}
\caption{Summary of commonly used evaluation metrics for image enhancement}
\label{tab:metrics_summary}
\begin{tabular}{lccc}
\toprule
\textbf{Metric} & \textbf{Reference Required} & \textbf{Perceptual-inspired} & \textbf{Underwater-specific} \\
\midrule
MSE / PSNR & \cmark & \xmark & \xmark \\
SSIM & \cmark & \cmark & \xmark \\
PCQI & \cmark & \cmark & \xmark \\
UCIQE & \xmark & \cmark & \cmark \\
UIQM & \xmark & \cmark & \cmark \\
UIF & \xmark & \cmark & \cmark \\
{\color{revpurple}Underwater Ranker} & {\color{revpurple}\xmark} & {\color{revpurple}\cmark} & {\color{revpurple}\cmark} \\
\bottomrule
\end{tabular}
\end{table*}

{\color{revpurple}
In summary, different evaluation metrics serve complementary purposes depending on the application. Full-reference measures such as PSNR, SSIM, and PCQI remain more reliable whenever trustworthy aligned references exist, because they directly quantify fidelity to a target image. No-reference measures such as UCIQE, UIQM, UIF, or learned underwater rankers remain useful in real underwater deployments where references are unavailable. However, neither family is sufficient on its own: reference metrics can miss human preference and task relevance, while no-reference metrics can reward visually pleasing but geometrically or semantically harmful alterations. \autoref{tab:metrics_summary} therefore should be read as a toolbox rather than a leaderboard, and later benchmarking results should be interpreted together with downstream utility and cross-domain robustness.
}

{\color{revteal}
For reconstruction-oriented settings, that toolbox should explicitly include task-driven reporting. Feature-matching success, pose accuracy, correspondence stability across views, and SLAM robustness can reveal failure modes that no-reference image-quality scores miss entirely, especially when colour vividness is achieved at the expense of geometric consistency.
}

\begin{sidewaystable}[thp]
    \caption{Quantitative comparisons on the UIEB-R90, UIEB-C60, U45, and UCCS datasets in terms of PSNR, SSIM, UIQM, and UCIQE. For each column, the best, second-best, and third-best results across all methods are highlighted with \textcolor{red}{red}, \textcolor{orange}{orange}, and \textcolor{yellow}{yellow} backgrounds, respectively. A dash `--' indicates that the corresponding method did not report this metric in the original publication.}
    \label{tab:result_UIE}
    % \vspace{-6pt}
    \centering
    \scriptsize
    \renewcommand{\arraystretch}{1.63}
    \setlength{\tabcolsep}{12pt}
    \begin{tabular}{@{}lcccccccc}
    \toprule
        \multirow{2}{*}{\small Method} & \multicolumn{2}{c}{UIEB-R90} & \multicolumn{2}{c}{UIEB-C60}   &  \multicolumn{2}{c}{U45} &  \multicolumn{2}{c}{UCCS} \\
        \addlinespace[-2pt]
        \cmidrule(lr){2-3} \cmidrule(lr){4-5} \cmidrule(lr){6-7} \cmidrule(lr){8-9}
        \addlinespace[-2pt]
        & PSNR $\uparrow$ & SSIM $\uparrow$ & UIQM $\uparrow$ & UCIQE $\uparrow$ & UIQM $\uparrow$ & UCIQE $\uparrow$ & UIQM $\uparrow$ & UCIQE $\uparrow$  \\
        \midrule
        \multicolumn{9}{l}{Deterministic Models} \\
        \midrule
        WaterNet~\citep{Li:Underwater:2020}  
        & 21.04 & 0.860 & 2.399 
        &  0.591 
        & -- & -- 
        & 2.275 & 0.556 \\

        Ucolor~\citep{Li:Underwater:2021} 
        & 20.13 & 0.877 
        & 2.482 & 0.553  
        & 3.148 & 0.586  
        & 3.019 & 0.550 \\

        FiveA$+$~\citep{jiang2023five} 
        & 23.06 & 0.911 
        & -- & -- & -- & -- & -- & -- \\

        HCLR-Net~\citep{zhou2024hclr} 
        &  24.99 & 0.925 
        & 2.695 &  0.587 
        & 3.103 &  0.610 
        & 3.045 &  0.579 \\

        Restormer~\citep{zamir2022restormer} 
        & 23.82 & 0.903 
        & 2.688 & 0.572 
        & 3.097 & 0.600 
        & 2.981 & 0.542  \\

        CECF~\citep{cong2024underwater} 
        & 21.82 & 0.894 
        & -- & -- & -- & -- & -- & --  \\

        U-Shape~\citep{peng2023u} 
        & 20.39 & 0.803  
        & 2.730 & 0.560 
        & 3.151 & 0.592 
        & -- & -- \\

        PixMamba~\citep{Lin_2024_ACCV} 
        & 23.59 & 0.921  
        &  2.868 &  0.586  
        & -- & -- 
        & 3.053 &  0.561 \\

        WaterMamba~\citep{Guan:WaterMamba:2024} 
        & 24.72 &  0.931 
        & 2.835 & 0.582 
        & -- & -- 
        & 3.057 & 0.555\\

        X-CAUNET~\citep{pramanick2024x} 
        & 22.30 & 0.908  
        & 2.683 & 0.564  
        & -- & -- 
        & 2.922 & 0.541 \\

        Convformer~\citep{wang2024uie} 
        & 23.13 & 0.904  
        & 2.684 & 0.572 
        & -- & -- 
        & 2.946 & 0.555 \\

        WFI2-Net~\citep{zhao2024wavelet} 
        & 23.86 & 0.873 
        & -- & -- 
        & 3.181 &  0.619 
        & -- & --  \\

        Phaseformer~\citep{Khan:Phaseformer:2024} 
        &  25.98 &  0.928  
        & -- & -- 
        &  4.491 & -- 
        & --  & --  \\

        GS-Transformer~\citep{zhuang2024globally} 
        & 24.42 & 0.861 
        & -- & -- & -- & -- & -- & -- \\
        \midrule
        \multicolumn{9}{l}{Generative Models} \\
        \midrule
        UIE-DM~\citep{tang2023underwater}  
        & 23.03 & 0.910 
        & -- & -- & -- & -- & -- & --\\ 

        FUnIEGAN~\citep{Islam:Fast:2020} 
        & 19.12 & 0.832 
        & 2.867 & 0.556 
        & 2.495 & 0.545  
        &  3.095 & 0.529\\

        PUGAN~\citep{cong2023pugan} 
        & 22.65 & 0.902 
        & 2.652 & 0.566 
        & -- & -- 
        & 2.977 & 0.536 \\ 

        PUIE-MP~\citep{fu2022uncertainty} 
        & 21.05 & 0.854 
        & 2.524 & 0.561  
        & 3.169 & 0.569  
        & 2.758 & 0.489 \\

        Semi-UIR~\citep{huang2023contrastive} 
        & 22.79 & 0.909 
        & 2.667 & 0.574 
        & 3.185 & 0.606 
        &  3.079 & 0.554 \\

        DCGF~\citep{zhang2024dcgf} 
        & -- & -- & -- & -- & -- & -- 
        & 1.377 &  0.609 \\

        DiffWater~\citep{guan2023diffwater} 
        & 20.97 & 0.895 
        &  4.655 & 0.433 
        &  4.730 & 0.462 
        & -- & -- \\ 

        BEM~\citep{huang2026bayesian} 
        &  25.62 &  0.940 
        &  2.931 & 0.567 
        &  3.406 &  0.620  
        &  3.224 &  0.561 \\
        
    \bottomrule
    \end{tabular}
\end{sidewaystable}

\subsubsection{Performance Evaluation of various Enhancement methods}
We summarize the performance of recent enhancement methods across several benchmark datasets and evaluation metrics. As shown in \autoref{tab:result_UIE}, the models are grouped into deterministic and generative categories, and their quantitative results are reported on the UIEB-R90, UIEB-C60, U45, and UCCS test sets using PSNR, SSIM, UIQM, and UCIQE. {\color{revteal}All values are taken from the original publications rather than reproduced here under one unified experimental protocol}.

With the rapid progress of generative modeling in recent years and its broad application in image-related tasks, generative approaches have increasingly shown advantages over deterministic models. In particular, generative methods tend to achieve stronger perceptual quality, as reflected by consistently competitive or superior performance on perceptual metrics such as UIQM and UCIQE across multiple datasets. This trend is especially evident on challenging benchmarks (e.g., UIEB-C60 and U45), where diffusion- and uncertainty-aware models demonstrate a clear ability to enhance global appearance and color fidelity.

{\color{revpurple}
At the same time, these metric trends should be interpreted cautiously. Strong UIQM or UCIQE scores do not by themselves imply better reference fidelity, temporal stability, or better suitability for downstream analysis, which is why the quantitative comparisons in \autoref{tab:result_UIE} are best read together with the task-oriented discussion in Section~\ref{subsec:downstream_tasks}.
}

In contrast, deterministic models generally perform well on distortion-based metrics such as PSNR and SSIM, especially on relatively constrained datasets. However, their performance on perceptual metrics is often less consistent, suggesting limitations in modeling the inherent ambiguity of underwater image degradation. Overall, these results highlight the growing importance of generative modeling for underwater image enhancement, particularly in scenarios where perceptual quality and visual realism are critical.

% \subsubsection{Objective Metrics}
% Evaluation is typically based on full-reference metrics such as PSNR, SSIM~\citep{wang2004image} and LPIPS~\citep{zhang2018unreasonable}, or information fidelity measures such as IFC and VIF~\citep{mittal2012making}, complemented by no-reference or reduced-reference alternatives. To address underwater-specific challenges, \citet{Wang:UIEC^2Net:2021} proposed the UIQM metric, which integrates factors including color fidelity, contrast, and sharpness. 
% \citet{yang2015underwater} proposed the UCIQE metric, a no-reference measure that evaluates underwater image quality by linearly combining chroma, saturation, and contrast. Compared with generic image quality metrics, UCIQE is specifically designed to capture the color distortions and reduced visibility characteristic of underwater scenes.

% More recent efforts include UIF~\citep{Zhou:Underwater:2023}, which employs multi-interval sub-histograms to assess underwater color consistency. Nevertheless, aligning objective scores with human perception remains challenging, motivating the development of task-driven evaluation protocols.

%------------------------------------------------------------
\subsection{Discussion and Open Challenges}

{\color{revorange}
This subsection focuses on bottlenecks that remain specific to underwater image enhancement. Shared issues such as benchmark design, supervision scarcity across videos and multi-view data, and broader evaluation strategy are synthesized later in \autoref{subsec:cross_cutting_challenges} and \autoref{subsec:future_datasets_models_metrics}.
}

{\color{revblue}
\paragraph{Generalization Across Water Types and Depth Ranges}
Many learning-based models are still tuned to a narrow distribution of water optics or illumination conditions, partly because supervised training often optimizes a one-to-one mapping tied to a specific dataset. However, recent methods now treat this issue as a concrete methodological direction rather than only an open challenge. As discussed in \autoref{subsec:ssda_methods}, learning under limited or no paired supervision uses synthetic-to-real transfer, reliable-bank or pseudo-label regularization, consistency constraints, and feature alignment to reduce performance drops across water types and turbidity levels. Even so, robust generalization under severe shifts in depth, particulate concentration, and illumination remains difficult, especially when appearance correction must transfer across water types without disturbing structure that later processing depends on.
}

{\color{revsienna}
The mismatch becomes especially severe when models trained on synthetic or controlled-tank data are deployed in field conditions with strong turbidity, spatially non-uniform caustics, drifting marine snow, or unstable artificial lighting. These effects are hard to approximate with idealized water models, yet they strongly influence whether restored images remain useful for later multi-view matching and reconstruction.
}

{\color{revorange}
\paragraph{Low-Light, Turbulence, and Temporal Stability}
Severe noise, forward scattering, and non-stationary particulate clutter remain central UIE-specific bottlenecks, especially in deep or murky water. Methods that improve single-frame contrast may still introduce flicker, unstable colour correction, or detail hallucination when applied to video. Lightweight restoration that jointly improves low-light visibility, suppresses turbulence-related artefacts, and preserves frame-to-frame consistency is therefore still needed for {autonomous underwater vehicles} (AUVs), {ROVs}, and other resource-constrained deployments.
}

\paragraph{Task-Oriented Utility of Enhancement}\label{subsec:downstream_tasks}
{\color{revgreen}
Enhancement utility depends on the intended task, not only on visual appeal. In operational settings such as underwater surveillance, inspection, teleoperation, and photogrammetry, low contrast, colour cast, and backscatter can obscure small targets, destabilise background models, and weaken situational awareness or measurement reliability~\citep{rout2024surveillance,garciagarcia2020background}. Related application studies likewise show that image-analysis pipelines benefit when enhancement or pre-processing recovers discriminative structure rather than merely producing vivid colours~\citep{helan2006object,bazeille2006automatic}, while more recent reviews of underwater object detection underline the importance of dataset bias, degraded visibility, and task-oriented robustness in practical machine-processing settings~\citep{fu2023rethinking}. Motion-analysis methods are especially sensitive to temporal consistency, boundary preservation, and stable background modelling under flicker and dynamic underwater clutter~\citep{nissar2026human,kapoor2024principal,kapoor2025graph}, and tracking or long-duration surveillance makes similar demands on local visibility and temporal stability~\citep{zhang2024fishtracking,humbert2023octopus}.
}

{\color{revgreen}
For this review, the main implication is that enhancement methods intended for machine processing should preserve structure, radiometric consistency, and temporal stability more carefully than methods aimed mainly at perceptual vividness. These properties are also the ones most likely to benefit reconstruction-oriented pipelines, where reliable correspondences, stable geometry-relevant cues, and frame-to-frame consistency matter directly. Broader questions of dataset coverage and evaluation for such machine-oriented settings are revisited in \autoref{subsec:cross_cutting_challenges} and \autoref{subsec:future_datasets_models_metrics}. Readers interested in wider downstream underwater surveillance or motion-analysis settings may also refer to the surveys of \citet{rout2024surveillance} and \citet{garciagarcia2020background}.
}

{\color{revteal}
\paragraph{Perceptual Quality, Geometric Consistency, and Deployment Constraints}
For reconstruction-oriented pipelines, better-looking enhancement is not automatically better geometry. Mild physical-model correction or conservative structure-preserving restoration often perturbs local gradients and cross-view radiometry less than aggressive black-box generation, whereas GAN- and diffusion-style enhancement can improve visibility while also hallucinating textures, shifting colours, or changing contrast in ways that destabilise feature matching, bundle adjustment, or neural-rendering supervision~\citep{Malyugina:beam:2025,vrochidis2025underwater3d}. Lightweight CNN and Mamba-style restorers often lie between these extremes: they can improve contrast and denoising with less stylistic drift, but they still require structure-aware losses if the output is later consumed by SfM, SLAM, COLMAP, or pose estimation. In calibration-critical stages, raw images or only lightly corrected images combined with explicit physical or refractive camera models may therefore be preferable to strong learned enhancement; stronger restoration is often safer later in offline dense reconstruction, novel-view rendering, or operator-facing visualisation~\citep{Wright2020UnderwaterPhotogrammetry,sedlazeck2009rov,vrochidis2025colormap}.
}

{\color{revsienna}
This is a genuine negative-transfer risk rather than a minor caveat. A method may detect more corner-like responses or produce visually sharper frames while still reducing the number of repeatable, view-consistent correspondences that survive geometric verification across time and viewpoint changes. In other words, single-view perceptual gain and multi-view geometric reliability are related but not interchangeable objectives.
}

{\color{revteal}
Exact FLOPs, latency, and memory are reported inconsistently across underwater enhancement papers and hardware settings. \autoref{tab:uie_geometry_deployment} therefore summarizes typical operational tendencies rather than a strict cross-paper benchmark.
}

\begin{table*}[t]
\centering
{\color{revteal}
\small
\renewcommand{\arraystretch}{1.12}
\setlength{\tabcolsep}{4pt}
\caption{{\color{revteal}Geometry-consistency and deployment implications of major underwater enhancement families.}}
\label{tab:uie_geometry_deployment}
\begin{tabular}{p{2.4cm}p{4.2cm}p{2.2cm}p{2.5cm}p{3.1cm}}
\toprule
\textbf{Family} & \textbf{Typical effect on geometry / key-points} & \textbf{Typical compute / memory} & \textbf{Deployment suitability} & \textbf{Typical best-use scenario} \\
\midrule
IFM-based / mild physics-guided correction & Usually preserves edges and radiometric ordering best when correction is conservative; lowest risk for calibration-sensitive stages & Low to moderate & Front-end or offline & Preprocessing for SfM, SLAM, pose estimation, metric reconstruction \\
Lightweight CNN / residual CNN & Often improves contrast and denoising with limited hallucination; geometry impact usually moderate and controllable & Low to moderate & Front-end / near-online feasible for compact models & Navigation support, inspection, operator assistance, robust preprocessing \\
Transformer-based UIE & Strong global colour correction for spatially varying attenuation, but can alter local gradients if trained aggressively & High memory and latency & Mostly offline; near-online only on high-end GPUs & High-quality offline enhancement across variable water conditions \\
Mamba-based UIE & Better efficiency than full attention with useful long-range context; geometry preservation depends on how strongly appearance is remapped & Moderate & Near-online for compact variants; otherwise offline or mixed & A compromise between context modelling and deployment cost \\
GAN / unpaired translation & Most prone to style drift, hallucinated texture, or cross-view colour inconsistency despite strong perceptual gains & Moderate inference; training is heavy & Mostly offline or operator-facing unless explicitly lightweight & Visual inspection, perceptual enhancement, domain translation \\
Diffusion-based UIE & Can recover severe degradations, but iterative inference and generative priors make geometry-relevant cues less predictable if unconstrained & High to very high & Offline only in current practice & Challenging restoration, archival enhancement, human-facing outputs \\
\bottomrule
\end{tabular}
}
\end{table*}

\section{3D Reconstruction for Underwater Scenes}
\label{sec:underwater_3d_recon}

Underwater 3D reconstruction is challenged by scattering, absorption, refraction at the housing interface, low-contrast imagery, illumination inconsistency, and platform motion. Together, these factors make feature extraction, correspondence estimation, and geometry recovery substantially less stable than in clear-medium scenes. Practical systems must also balance reconstruction fidelity against computational limits in applications such as inspection, navigation, and field mapping.

{\color{revgreen}
These challenges matter across marine science, archaeology, inspection, and seafloor mapping, and they have shaped a clear methodological progression. Underwater 3D reconstruction has moved from calibration-heavy photogrammetry and SLAM toward learning-based underwater MVS, NeRF, 3D Gaussian Splatting, and physics-guided hybrid formulations that model the water medium more explicitly.
}

{\color{revpurple}
Within this progression, visual enhancement remains relevant because reconstruction quality depends not only on geometry estimation but also on how underwater image degradation is handled. In this review, underwater physics provides the forward model, UIE provides image-domain priors, and reconstruction methods either rely on these assumptions implicitly or encode them explicitly.
}

This section therefore first covers photogrammetry and underwater MVS within the photogrammetry part (\autoref{subsec:photogrammetry}, \autoref{subsec:uw_end2end_mvs}), then discusses NeRF (\autoref{subsec:nerf}), 3D Gaussian Splatting (\autoref{subsec:3dgs}), and finally the main limitations and open challenges for robust underwater scene reconstruction (\autoref{sec:Discussion and Open Challenges}).

\subsection{Photogrammetry}
\label{subsec:photogrammetry}

\noindent\textbf{Fundamental Principles.}
Photogrammetry reconstructs 3D structures from overlapping 2D images by identifying correspondences across viewpoints and solving for camera poses and scene geometry. Key modules include: 1) feature extraction and matching, which identify local features (corners, edges, or learned descriptors) robust to changes in viewpoint or illumination; 2) camera pose estimation via SfM, which incrementally or globally determines camera extrinsic/intrinsic parameters such that reprojected correspondences align in 3D; 3) dense reconstruction with MVS, which estimates detailed geometry by matching pixels across multiple images, typically using techniques such as patch-based stereo or plane sweeping; and 4) meshing and texturing, which convert the resulting point cloud into a polygon mesh and map images onto the surface to retain realism.

Underwater photogrammetry modifies these steps to handle color distortion, low contrast, and refraction effects. For instance, robust matching frequently requires color normalization or contrast enhancement in a preprocessing pipeline.

\subsubsection{Photogrammetry Approaches for Underwater Scenes.}

\begin{figure}[ht]
\centering
\includegraphics[width=1.\linewidth]{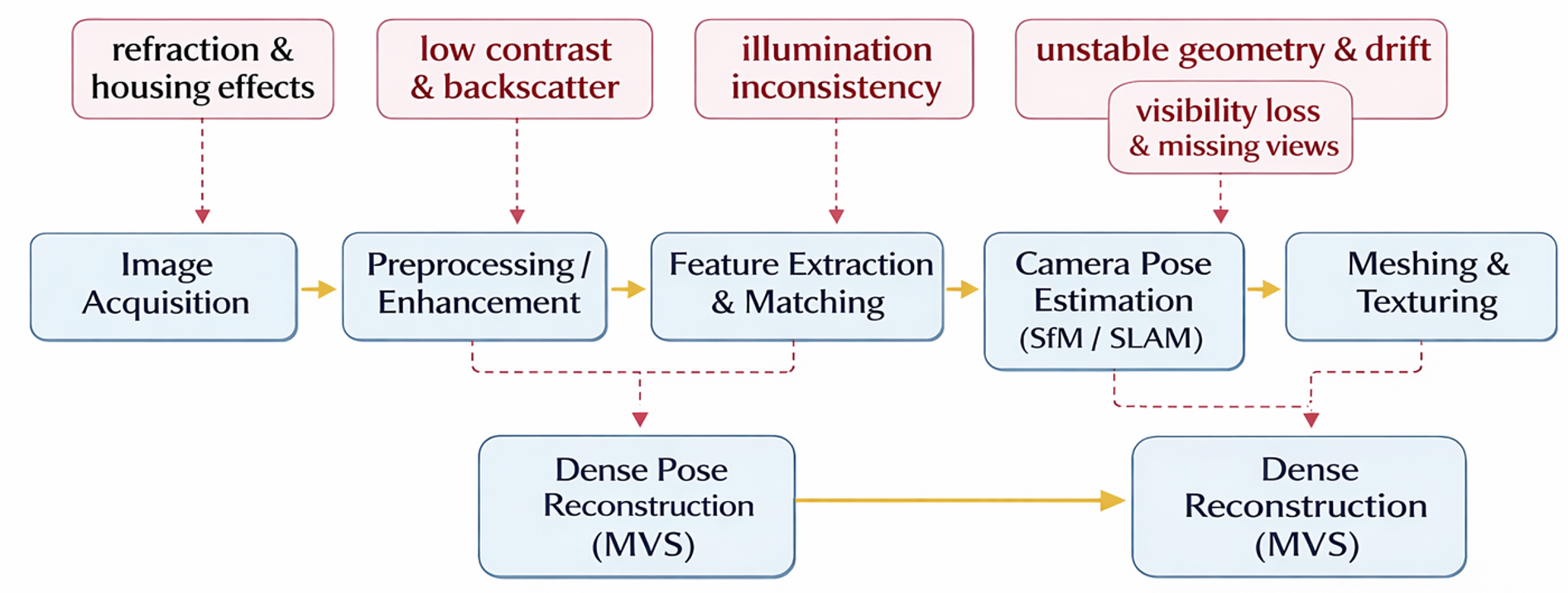}

\caption{{\color{revpurple}Conceptual flow of underwater photogrammetry. Compared with clear-medium pipelines, underwater processing must explicitly handle refraction, low contrast, backscatter, and illumination inconsistency across acquisition, matching, pose estimation, and dense reconstruction.}}
\label{fig:uw_photogrammetry_flow}
\end{figure}

A large body of literature \citep{Prado:3D:2020, ZHANG2022100510} has tackled photogrammetry in the subaqueous setting. For instance, \citet{Prado:3D:2020} developed a specialized pipeline for capturing circalittoral rocky shelves, demonstrating that terrain classification can enhance region-based matching accuracy. Similarly, \citet{Wright2020UnderwaterPhotogrammetry} assessed the accuracy of SfM pipelines for archaeology-related tasks, such as mapping submerged shipwrecks. Their findings indicated that SfM often outperformed Real-Time Kinematic (RTK) surveying for localized applications. In another study, ~\citet{Nocerino:coral:2020} incorporated Multi-View Stereo (MVS) to refine the reconstructed geometry following the initial SfM step, effectively mitigating the bowling effect, where large continuous surfaces appear erroneously curved.

Common issues in underwater SfM revolve around stable pose estimation when the scene lacks strong textures or includes extensive repeated patterns (e.g., sandy seabed). Bowling or {doming} arises from marginal pose constraints \citep{Wright2020UnderwaterPhotogrammetry}, where small camera rotations or inaccurate correspondences accumulate into large global errors. Although advanced bundle adjustment approaches can alleviate some drift, the presence of refraction or local lighting variations often complicates the inherent assumption of a single pinhole projection model. To address these problems, specialized calibration techniques have been introduced. Researchers sometimes use multi-camera rigs with known baselines, allowing for direct stereo matching that is more robust to color anomalies. {\color{revteal}Others implement \emph{flat port} or \emph{dome port} calibrations, explicitly modeling the glass boundary through which images are captured. The distinction matters: flat ports generally introduce stronger refraction and violate central pinhole assumptions more severely, while hemispherical dome ports often yield better photogrammetric behavior in practice and can approach a single-viewpoint model only when the lens is well centered and aligned with the dome geometry~\citep{Menna:FlatVsDome:2017,She:UnderwaterDomes:2022}. Even so, both configurations can bias metric geometry if refractive interfaces are ignored, which is why refractive calibration or ray-based modeling remains important for accurate underwater photogrammetry~\citep{sedlazeck2009rov}.}

\paragraph{Bowling Effects and Remedies}
As noted, large uniform terrains, such as expansive sandy seafloors or regions covered with short vegetation, can hinder robust alignment in SfM due to a lack of high-frequency features. This often leads to degenerate configurations, resulting in reconstructions with exaggerated curvature or bowl-shaped deformations \citep{Wright2020UnderwaterPhotogrammetry, Samboko:evaluating:2022}. To mitigate these issues, several strategies have been proposed.
One common approach is to introduce artificial markers with known geometry or color-coded patterns, which serve as reliable anchor points for SfM \citep{Wright2020UnderwaterPhotogrammetry, WITTMANN2024100072}. Divers or automated systems can place these markers strategically to enhance feature matching. Another effective method involves leveraging trajectory constraints. When an autonomous underwater vehicle (AUV) or remotely operated vehicle (ROV) logs inertial or acoustic data, these measurements can be integrated into the reconstruction pipeline to minimize drift and improve overall stability.
Additionally, mesh regularization techniques \citep{Aubram2013} are often employed as a post-processing step following standard multi-view stereo (MVS). By enforcing surface smoothness or incorporating planar constraints, these methods help reduce spurious curvature in the final model. In cases where partial depth measurements are available, incorporating data from sonar or short-range LiDAR can further stabilize the photogrammetry pipeline \citep{Istenic2019}. These local depth cues provide additional constraints on scale and geometry, effectively reducing global distortions in the reconstructed scene.

\subsubsection{Real-Time Visual SLAM}
Where offline SfM reconstructs a scene after collecting images, visual SLAM attempts to solve camera localization (odometry) and mapping on the fly. Underwater robots can deploy SLAM for navigation, obstacle avoidance, or on-the-spot mapping. According to \citet{Storlazzi:vslam:2016}, visual SLAM can exceed the resolution of large-scale LiDAR or side-scan sonar data, which typically provide coarser point clouds at a broader scale. This advantage is crucial for tasks like surveying coral polyps, delicate rock formations, or subtle archaeological relics. However, achieving real-time performance requires carefully chosen features or deep learningased front ends robust to turbidity and color distortions. Some pipelines also incorporate acoustic or inertial measurements for multi-sensor fusion, offsetting the difficulties introduced by water's optical properties.

\subsubsection{\texorpdfstring{{\color{revblue}End-to-End Underwater MVS}}{End-to-End Underwater MVS}}\label{subsec:uw_end2end_mvs}

{\color{revblue}
Recent work has begun to explore end-to-end underwater MVS as an alternative to purely sequential SfM-plus-MVS pipelines. The main idea is not merely to enhance frames before reconstruction, but to make underwater-aware correspondence learning and dense geometry inference part of the reconstruction model itself.
}

\begin{figure*}[ht]
\centering
\includegraphics[width=0.98\textwidth]{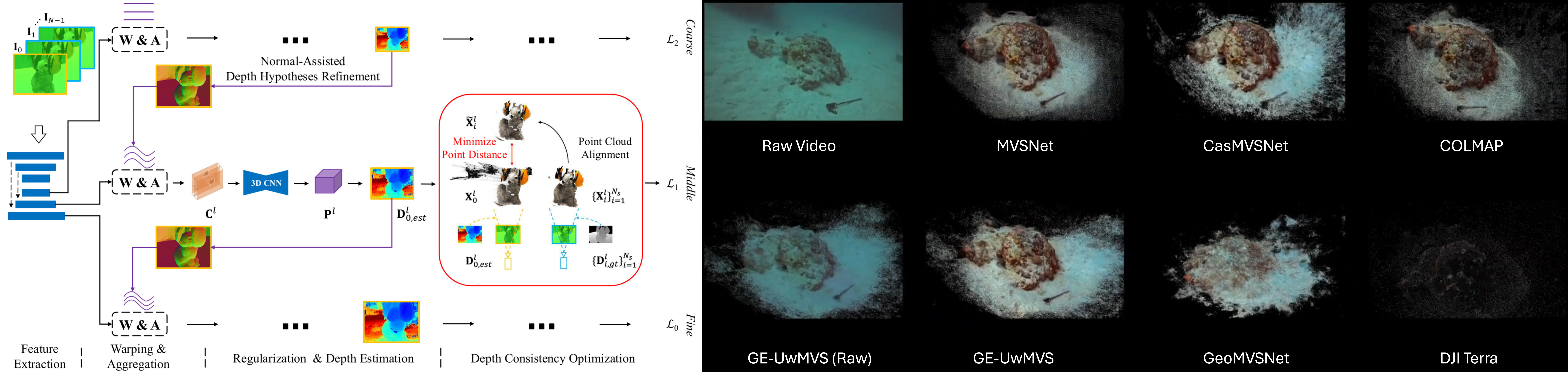}
\caption{{
(a) End-to-end underwater MVS framework, GE-UwMVS~\citep{yang2025uwmvs}. (b) Dense reconstruction comparison against representative MVS baselines, including COLMAP~\citep{schonberger2016structure}, MVSNet~\citep{yao2018mvsnet}, CasMVSNet~\citep{gu2020casmvsnet}, GeoMVSNet~\citep{zhang2023geomvsnet}, and the commercial software DJI Terra~\citep{dji_terra}.}}
\label{fig:uwmvs}
\end{figure*}

{\color{revblue}
In particular, \citet{yang2025uwmvs} propose an end-to-end underwater multi-view stereo framework for dense scene reconstruction that learns geometry directly from underwater image sets while explicitly accounting for appearance degradations caused by the medium. The importance of this work is less that it establishes a mature family of methods, and more that it shows underwater MVS can be formulated as a learned dense reconstruction problem rather than as a fixed photogrammetric pipeline with enhancement only attached upstream. In this sense, it is an early but important attempt to incorporate underwater appearance distortion into the dense matching and depth inference stages themselves. Overall, end-to-end underwater MVS remains a small but important direction for closing the loop from visual enhancement to dense geometry recovery under real underwater conditions.
}

\subsection{Neural Radiance Fields (NeRF) }
\label{subsec:nerf}

\begin{figure}[htp]
    \centering
    \includegraphics[width=1.\linewidth]{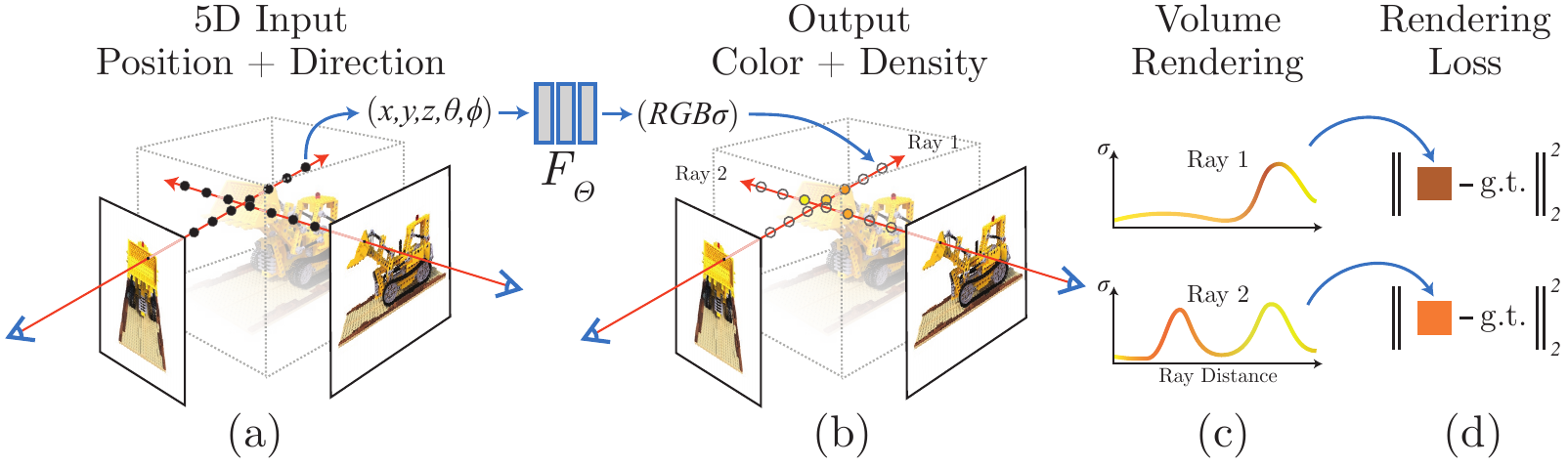}
    \caption{Illustration of NeRF and its differentiable rendering process. It involves sampling 5D coordinates (position and direction) along camera rays (a), using an MLP to produce color and density (b), and rendering these into an image (c). The differentiable function allows optimization by minimizing differences between rendered and actual images (d)~\citep{mildenhall2020nerf}}
    \label{fig:enter-label}
\end{figure}
Neural Radiance Fields (NeRF), introduced by \citet{mildenhall2020nerf}, have brought a paradigm shift to 3D reconstruction and novel view synthesis. Unlike traditional representations such as voxel grids or explicit point clouds, NeRF encodes scene appearance and geometry through a Multi-Layer Perceptron (MLP). Given a 3D point \(\mathbf{x}\) and a viewing direction \(\mathbf{d}\), NeRF predicts color \(c(\mathbf{x}, \mathbf{d})\) and volume density \(\sigma(\mathbf{x})\), allowing a continuous representation of the underlying scene.

\subsubsection{Principles and Volume Rendering}

The main idea behind NeRF is grounded in the volume rendering equation \citep{Kajiya:rendering:1986}. A ray is parameterized as:
\[
\mathbf{r}(t) = \mathbf{o} + t \mathbf{d},
\]
where \(\mathbf{o}\) is the camera center and \(\mathbf{d}\) is the unit viewing direction. The radiance along the ray is accumulated as:
\[
C(\mathbf{r}) = \int_{t_n}^{t_f} T(t) \,\sigma(\mathbf{r}(t)) \, c(\mathbf{r}(t), \mathbf{d}) \, dt,
\]
where 
\[
T(t) = \exp\Bigl(-\int_{t_n}^{t} \sigma(\mathbf{r}(s))\, ds \Bigr)
\]
represents the transmittance from \(t_n\) to \(t\). NeRF approximates this integral by sampling discrete points along the ray and summing their contributions.

To train NeRF, one minimizes the discrepancy between synthesized pixels and real observed pixels in the input images. A common objective is the mean squared error (MSE) between the rendered color \(C_i\) and the corresponding ground truth \(\hat{C}_i\):
\[
L = \sum_{i}\,\bigl\|C_i - \hat{C}_i\bigr\|^2.
\]
Through gradient-based optimization, the MLP learns both the geometry (\(\sigma\)) and appearance (\(c\)).

\noindent\textbf{Comparison with Traditional 3D Reconstruction.}
Conventional 3D reconstruction methods, such as Structure-from-Motion (SfM) and Multi-View Stereo (MVS), rely on geometric feature matching and explicit estimation of depth maps or point clouds \citep{snavely2006photo, seitz2006mvs}. While they can yield accurate geometry with sufficient texture and baseline, they typically require clear, feature-rich images and separate steps for dense reconstruction. Photometric Stereo \citep{woodham1980photometric} is adept at capturing surface normals under controlled lighting, but it lacks flexibility for unstructured, real-world scenes. 

By contrast, NeRF directly learns an implicit volumetric function from raw images, often resulting in superior novel view synthesis. However, it can be computationally demanding, and extracting explicit geometry (e.g., meshes) is less straightforward. NeRF also tends to require many images around the subject for optimal training, though newer variants reduce that data requirement.

\begin{table*}[t]
\centering
\caption{High-level comparison of NeRF features}
\renewcommand{\arraystretch}{1.1}
\begin{tabular}{@{}lll@{}}
\toprule
\textbf{Aspect} & \textbf{NeRF Advantages} & \textbf{NeRF Disadvantages} \\
\midrule
Representation   & Implicit, continuous 3D model        & Complex neural inference \\
View Synthesis   & Photorealistic novel views           & High computational cost   \\
Geometry         & Learned implicitly                   & No direct mesh extraction \\
Generalization   & Uses raw image supervision           & Often data-hungry         \\
\bottomrule
\end{tabular}
\label{tab:nerf_compare}
\end{table*}
\subsubsection{NeRF Variants}
Rather than exhaustively surveying the very large clear-medium NeRF literature, we retain here only the generic variants that best contextualize the later underwater discussion. The most relevant background directions concern efficiency, robustness to unconstrained capture, and dynamic/video modeling. The next subsection then focuses on the genuinely underwater methodological divergence.

\paragraph{Efficient NeRF representations}
One major line of work reduces NeRF's heavy training and rendering cost through more compact scene encodings. Instant-NGP \citep{mueller2022instant} uses multi-resolution hash grids to cut training time from hours to seconds, Plenoxels \citep{fridovich2022plenoxels} replace the large MLP with directly optimized sparse voxels and spherical harmonics, and TensoRF \citep{chen2022tensorf} factorizes the radiance field into low-rank tensor components for compact storage and faster optimization. These methods are representative because they define the efficiency baseline against which later underwater NeRF systems must still be judged: once medium-aware rendering is added, training cost and memory usage remain a practical bottleneck.

\paragraph{Unconstrained and pose-robust NeRF}
Another relevant direction relaxes NeRF's assumption of uniformly captured, well-posed image sets. NeRF-W \citep{martinbrualla2021nerfw} introduces appearance latents and transient components to absorb lighting variation and occluders, while BARF \citep{lin2021barf} and UP-NeRF \citep{kim2023upnerf} explicitly tackle uncertain or missing camera poses during optimization. These variants matter for underwater sensing because real deployments often combine viewpoint drift, radiometric inconsistency, and transient distractors rather than the carefully calibrated capture conditions assumed by the original NeRF.

\paragraph{Dynamic / video NeRF}
Dynamic NeRF variants extend the radiance field to time-varying scenes through deformation fields or temporal consistency models. Representative examples include NeRFies \citep{park2021nerfies} and D-NeRF \citep{pumarola2021dnerf} for deformable scenes, together with Video-NeRF \citep{xian2021videonerf} and Neural Radiance Flow \citep{du2020neuralflow} for temporally coherent 4D rendering. This direction is especially relevant underwater because moving organisms, marine snow, flicker, and platform motion all challenge the static-scene assumption; indeed, \citet{gough2025aquanerf} note that generic dynamic NeRF methods remain insufficient for high-frequency underwater disturbances without more explicit disturbance-aware modeling.

Taken together, these generic NeRF variants mainly address efficiency, robustness, or temporal modeling in clear-medium settings. Underwater NeRF methods inherit these concerns, but they must additionally account for medium-aware rendering, radiometric degradation, and tighter coupling between restoration and reconstruction.

\subsubsection{Underwater NeRF Applications}
{\color{revpurple}
We organize existing underwater NeRF methods by their modeling choices. Compared with clear-medium NeRF variants, the underwater extensions usually modify the rendering equation, introduce explicit medium parameters, or couple restoration and reconstruction within the same optimization. To make this design space more explicit, we first summarize the main underwater NeRF families through compact methodological equations and then map representative methods onto these routes.
}

\begin{figure}[H]
\centering
\resizebox{0.85\linewidth}{!}{%
\begin{tikzpicture}[
    >=Stealth,
    node distance=1.1cm,
    thick,
    % 定义学术配色
    main_color/.style={blue!50!black}, 
    border_color/.style={black!60},    
    bg_color/.style={blue!3},          
    % 定义基础样式
    base_style/.style={
        rounded corners=2pt, 
        line width=0.8pt,
        inner sep=8pt,
        align=center,
        font=\small
    },
    % 定义左侧 Base NeRF 样式
    main_node/.style={
        base_style,
        draw=blue!50!black, 
        fill=blue!3,
        minimum width=5.5cm,
    },
    % 定义右侧子分支样式
    sub_node/.style={
        base_style,
        draw=black!60,      
        fill=white,
        minimum width=7.2cm,
        font=\scriptsize,
        % 左侧装饰条 (Accent Bar)
        postaction={
            path picture={
                \fill[blue!50!black] ([xshift=2pt]path picture bounding box.north west) rectangle (path picture bounding box.south west);
            }
        }
    }
]

% 1. 放置节点
% 根节点在左侧
\node[main_node] (base) at (0,0) {
    \textbf{Base NeRF} \\[0.8ex]
    $C(\mathbf{r})=\int T(t)\sigma(\mathbf{r}(t))c(\mathbf{r}(t),\mathbf{d})\,dt$ \\[0.5ex]
    \textcolor{black!70}{\scriptsize Volume rendering formulation}
};

% ==================== 主要修改部分开始 ====================
% 将右侧节点的中心坐标从 7.2 移到了 8.5，给连线留出充足的水平空间
\coordinate (right_center) at (8.5, 0); 
% ==========================================================

\node[sub_node] (f2) at ($(right_center) + (0, 0.85)$) {
    \textbf{F2: Joint photometric correction} \\[0.5ex]
    $L=L_{\mathrm{render}}+\lambda_{\mathrm{corr}}L_{\mathrm{photo/corr}}$ \\[0.3ex]
    \textit{WaterNeRF, UWNeRF}
};

\node[sub_node, above=of f2] (f1) {
    \textbf{F1: Medium disentanglement} \\[0.5ex]
    $C(\mathbf{r})=C_{\mathrm{obj}}(\mathbf{r})+C_{\mathrm{med}}(\mathbf{r})$ \\[0.3ex]
    \textit{ScatterNeRF, SeaThru-NeRF}
};

\node[sub_node] (f3) at ($(right_center) + (0, -0.85)$) {
    \textbf{F3: Dynamic / distractor-aware} \\[0.5ex]
    $C(\mathbf{r})=m_sC_s(\mathbf{r})+m_dC_d(\mathbf{r})$ \\[0.3ex]
    \textit{AquaNeRF}
};

\node[sub_node, below=of f3] (f4) {
    \textbf{F4: Restoration-aware rendering} \\[0.5ex]
    $L=L_{\mathrm{render}}+\lambda_{\mathrm{rest}}L_{\mathrm{rest}}$ \\[0.3ex]
    \textit{WaterHE-NeRF}
};

% ==================== 主要修改部分开始 ====================
% 分支交汇点：将主干线的长度设为 1.0cm (原为 0.6cm)
% 这样剩下的空间将留给进入右侧节点的带有箭头的水平线
\coordinate (branch) at ($(base.east) + (1.0, 0)$); 
% ==========================================================

\draw[main_color, line width=1pt] (base.east) -- (branch); % 主干线

\foreach \i in {f1, f2, f3, f4} {
    \draw[main_color, line width=1pt, rounded corners=4pt, ->] 
        (branch) |- (\i.west); % 分支线
}

\end{tikzpicture}%
}
\caption{{\color{revpurple}Methodological design space of underwater NeRF variants. The main families diverge through medium disentanglement, joint photometric correction, dynamic distractor handling, and restoration-aware rendering.}}
\label{fig:uw_nerf_design_space}
\end{figure}

{\color{revpurple}
\noindent\textbf{Medium disentanglement and revised-IFM-guided rendering.}
\[
C(\mathbf{r}) = C_{\mathrm{obj}}(\mathbf{r}) + C_{\mathrm{med}}(\mathbf{r}),
\]
where \(C_{\mathrm{obj}}(\mathbf{r})\) captures scene radiance transported along the ray and \(C_{\mathrm{med}}(\mathbf{r})\) captures medium-dependent terms such as backscatter, attenuation, or water colour. ScatterNeRF~\citep{ramazzina2023scatternerf} demonstrates the basic idea of disentangling scattering effects from scene content in participating media, which carries over naturally to underwater settings. SeaThru-NeRF~\citep{levy2023seathru} is the first dedicated underwater NeRF to build this idea on the revised underwater image formation model~\citep{akkaynak2018revised}, described in~\autoref{eq:revised_ifm}, introducing per-ray backscatter, attenuation-density, and medium-colour branches so that object radiance and water-medium appearance are modeled separately. SP-SeaNeRF~\citep{CHEN2024104025} extends this line by adding learnable illumination embeddings and an explicit degradation simulation step, improving sharpness under non-uniform lighting. The main strength of this family is that it grounds novel-view synthesis in a physically interpretable forward model; its main limitation is the added optimization complexity and the need to separate medium effects from sparse observations.
}

{\color{revpurple}
\noindent\textbf{Photometric correction coupled with reconstruction.}
\[
L = L_{\mathrm{render}} + \lambda_{\mathrm{corr}}L_{\mathrm{photo/corr}},
\]
where the optimization is driven jointly by radiance-field reconstruction and a correction term that stabilizes colour or illumination across views. A second family therefore couples radiometric correction with geometry estimation rather than treating enhancement as an external pre-processing stage. WaterNeRF~\citep{Sethuraman:WaterNeRF:2023} combines underwater light transport modeling with optimal-transport-based colour correction to stabilize view-to-view appearance. UWNeRF~\citep{10656460} integrates photometric correction directly into the NeRF rendering process, allowing the network to infer clearer 3D point appearance while preserving cues that may be discarded by independent enhancement. This direction is important because it explicitly links UIE-style colour recovery to multi-view consistency, instead of assuming that the two can be optimized separately.
}

{\color{revpurple}
\noindent\textbf{Dynamic scenes, floaters, and distractors.}
\[
C(\mathbf{r}) = m_s C_s(\mathbf{r}) + m_d C_d(\mathbf{r}), \qquad m_s + m_d = 1,
\]
where the rendering is decomposed into static and disturbance-related contributions, or equivalently biased toward a dominant surface to suppress floating clutter. Dynamic underwater disturbances remain a central difficulty because classical NeRF assumes a static scene. UWNeRF~\citep{10656460} differentiates between static and dynamic components using motion masks as a secondary mechanism, while AquaNeRF~\citep{gough2025aquanerf} reduces the impact of floaters and moving objects by enforcing a single dominant surface along each ray and maintaining medium transmittance with a Gaussian weighting scheme. These strategies reduce artifact accumulation around fish or suspended particles, but their effectiveness still depends on the quality of masks or ray-level visibility assumptions.
}

\begin{table*}[t]
\centering
{\color{revpurple}
\scriptsize
\renewcommand{\arraystretch}{1.12}
\setlength{\tabcolsep}{3pt}
\caption{{\color{revpurple}Summary of representative underwater NeRF methods.}}
\label{tab:underwater_nerf_summary}
\resizebox{\textwidth}{!}{%
\begin{tabular}{p{1.9cm}p{0.95cm}p{2.9cm}p{2.4cm}p{2.3cm}p{2.3cm}p{1.9cm}p{2.0cm}}
\toprule
\textbf{Method} & \textbf{Primary family} & \textbf{Core underwater modeling idea} & \textbf{Relation to physics / IFM} & \textbf{Relation to enhancement} & \textbf{Dynamic / distractor handling} & \textbf{Strength} & \textbf{Limitation} \\
\midrule
ScatterNeRF \citep{ramazzina2023scatternerf} & F1 & Separates scene content from scattering medium in participating media & Medium-aware rendering, but not underwater-specific IFM & Indirect; improves visibility through disentanglement & Not explicit & General scattering disentanglement & Limited underwater specificity \\
SeaThru-NeRF \citep{levy2023seathru} & F1 & Per-ray backscatter, attenuation, and medium colour branches & Explicitly based on revised underwater IFM & Jointly restores colour while reconstructing & Static-scene assumption & Physically interpretable underwater rendering & Higher model complexity and data sensitivity \\
SP-SeaNeRF \citep{CHEN2024104025} & F1 & SeaThru-NeRF plus illumination embeddings and degradation simulation & Physics-guided with learnable lighting factors & Joint enhancement and reconstruction & Limited dynamic handling & Better sharpness under non-uniform illumination & More parameters and training overhead \\
UWNeRF \citep{10656460} & F2 & Integrates photometric correction into NeRF geometry learning & Physics-aware radiometric rendering & Strong coupling of correction and reconstruction & Motion masks for dynamic regions & Preserves geometry cues better than separate pre-enhancement & Depends on mask quality and robust pose estimates \\
WaterHE-NeRF \citep{zhou2023waterhe} & F4 & Water-ray matching field with Retinex-style colour correction & Implicit physical prior plus restoration field & Explicit restoration-aware rendering & Not a primary focus & Connects colour recovery to view synthesis & Restoration assumptions can bias geometry \\
AquaNeRF \citep{gough2025aquanerf} & F3 & Single-surface-per-ray rendering to suppress floaters & Medium transmittance modeled along ray & Indirect enhancement through cleaner rendering & Explicit floater suppression & More robust static-object reconstruction in clutter & May oversimplify complex multi-layer scenes \\
WaterNeRF \citep{Sethuraman:WaterNeRF:2023} & F2 & Couples underwater transport modeling with colour-consistency correction & Physics-aware transport plus colour stabilization & Jointly optimizes appearance correction and reconstruction & Not a primary focus & Better view-to-view photometric stability & Additional coupled losses and optimization burden \\
\bottomrule
\end{tabular}}
}
\end{table*}

{\color{revpurple}
\noindent\textbf{Enhancement-aware and restoration-aware rendering.}
\[
L = L_{\mathrm{render}} + \lambda_{\mathrm{rest}}L_{\mathrm{rest}},
\]
where restoration-aware terms or auxiliary fields are written directly into the radiance-field optimization rather than applied as a separate post-processing stage. Some methods therefore introduce restoration cues more explicitly into the radiance-field parameterization. WaterHE-NeRF~\citep{zhou2023waterhe}, for example, augments NeRF with a water-ray matching field derived from Retinex-style reasoning, aiming to recover colour and illumination while reconstructing geometry. Relative to standard NeRF variants, these models are closer to joint enhancement-reconstruction systems, but they also inherit the risk that aggressive restoration assumptions may distort the very correspondences needed for geometry estimation.
}

\par
{\color{revpurple}
\noindent Table~\ref{tab:underwater_nerf_summary} maps representative methods to four shorthand families: F1 = medium disentanglement / revised-IFM-guided rendering, F2 = photometric correction coupled with reconstruction, F3 = dynamic / distractor-aware rendering, and F4 = restoration-aware rendering.
}

\subsection{3D Gaussian Splatting}
\label{subsec:3dgs}

\par
\begin{figure}[H]
    \centering
    \includegraphics[width=1.\linewidth]{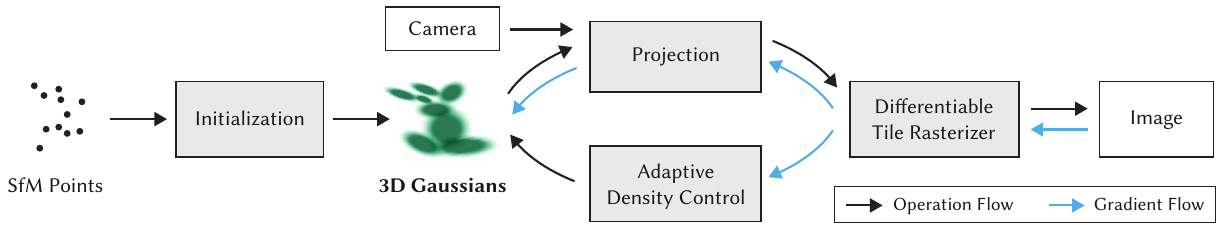}
    \caption{Overview of the 3D Gaussian Splatting pipeline. Starting from a sparse SfM point cloud, an initial set of 3D Gaussians is constructed. These Gaussians are iteratively refined through projection, adaptive density control, and differentiable tile-based rasterization, where gradients are propagated to update their parameters. This optimization procedure enables efficient training while maintaining high fidelity, and the final representation supports real-time rendering and interactive navigation across diverse scenes~\citep{kerbl:3Dgaussians:2023}
 }
    \label{fig:3dgs_pipeline}
\end{figure}

3D Gaussian Splatting (3DGS)~\citep{kerbl:3Dgaussians:2023} is an explicit scene representation in which a 3D scene is modeled by Gaussian primitives rather than an implicit volume. Each Gaussian primitive $\mathcal{G}_i$ is parameterized by a center position $\mu_i$, a covariance matrix $\Sigma_i$, an opacity $\alpha_i$, and a view-dependent colour $c_i$ often decoded from spherical harmonics. For differentiable optimization, the covariance is typically factorized into rotation and scaling components:
\begin{equation}
\Sigma_i = R_i S_i S_i^{T} R_i^{T}.
\end{equation}
For rendering, the 3D Gaussian is projected to the image plane through a viewing transformation $W$. Under a local affine approximation, its 2D covariance can be written as
\begin{equation}
\Sigma'_i = J W \Sigma_i W^{T} J^{T},
\end{equation}
where $J$ is the Jacobian of the projection around the current viewing configuration. The contribution of each projected Gaussian is then accumulated by alpha compositing:
\begin{equation}
C = \sum_{i} c_i\, \alpha'_i \prod_{j=1}^{i-1} (1 - \alpha'_j), \qquad
\alpha'_i = \alpha_i \exp\!\left(-\tfrac{1}{2} (x - \mu'_i)^{\!T} (\Sigma'_i)^{-1} (x - \mu'_i)\right).
\end{equation}
Here, $\mu'_i$ denotes the projected 2D mean of the Gaussian and $\alpha'_i$ is its effective blending weight at pixel $x$. In practice, optimization alternates between updating Gaussian parameters and adapting the Gaussian density, which enables fast convergence and real-time rendering but can still require careful control to avoid redundant primitives or loss of fine detail.

3DGS offers several advantages:
\begin{itemize}
    \item \textbf{Speed of Convergence:} The explicit nature of 3DGS enables faster training and more accurate estimation of scene geometry and color with fewer samples compared to methods such as NeRF.
    \item \textbf{Real-Time Rendering Potential:} Once trained, 3DGS supports real-time rendering by directly rasterizing Gaussian points using GPU-based alpha blending.
    \item \textbf{Adaptability:} Leveraging spherical harmonics allows 3DGS to capture appearance variations under complex lighting conditions. However, additional modeling is needed to handle phenomena such as reflection or refraction.
\end{itemize}

\subsubsection{3DGS Variants}
Since the introduction of 3DGS \citep{kerbl:3Dgaussians:2023}, several enhancements have been proposed to extend its capabilities. Some techniques that could potentially be used for underwater scenes are described here:

    \paragraph{Anti-Aliasing} Mip-Splatting \citep{yu2024mip} addresses aliasing artifacts that emerge when the sampling rate changes--a shortcoming primarily due to the vanilla 3DGS's lack of 3D frequency constraints and its reliance on a 2D dilation filter. By integrating 3D smoothing with 2D mip filters, this method enables aliasing-free rendering. Similarly, \citet{yan2024multi} report that aliasing becomes pronounced in low-resolution renderings when the pixel size drops below the Nyquist threshold. They propose a multi-scale rendering strategy using levels-of-detail (LOD) and mipmap techniques that synthesize larger Gaussians for low-resolution outputs by aggregating smaller Gaussians from high-resolution inputs.
    \paragraph{Deblurring} \citet{chen2024deblur} observe that novel view synthesis from the original 3DGS degrades significantly with blurred input images, which could possibly be one of the challenges for underwater 3D reconstruction.. To mitigate this, they introduce a physically based model that approximates the camera trajectory by incorporating pseudo-camera poses along the path and blending the corresponding images to simulate motion blur. Additionally, \citet{lee2024deblurring} propose an alternative variant that specifically addresses defocusing blur, further enhancing performance in challenging imaging conditions .
    \paragraph{4D Rendering.
    } The original 3DGS is designed for static scenes, limiting its ability to capture dynamic inputs. To overcome this, 4DGS \citep{wu20244d} employs an encoder-decoder deformation network \citep{yang2024deformable3dgs} that uses both timestamps and Gaussian center coordinates to predict deformed positions and covariance matrices. In a related approach, \citet{lin2024gaussian} introduces the Gaussian-flow method, which uses a dual-domain deformation model to estimate deformed attributes for efficient 4D scene rendering. However, while effective for modeling slow motion, these techniques struggle with fast dynamics and unpredictable trajectories.
    \paragraph{Density Control} 3DGS improves the representation of the Gaussian point cloud through an adaptive density control strategy. Abnormal average 2D gradients upon projection reveal regions of under-reconstruction, prompting the subdivision or duplication of Gaussians based on size. Nevertheless, \citet{zhang2024pixel} note that this approach can introduce needle-like and blurred artifacts in sparse initial point density areas. They propose using the total coverage pixel count across different views as the weighting metric--rather than solely the number of viewpoints--to achieve more detailed reconstructions in regions with repetitive textures.
{\color{revpurple}
These clear-medium variants mainly improve splatting as a generic rendering and optimization framework. The more distinct underwater divergence begins in the next subsection, where the literature is better organized by medium modeling, structural priors, disturbance handling, and restoration coupling.
}

\subsubsection{Underwater 3DGS Applications}
Despite advances that have enabled 3DGS to generate more accurate and real-time reconstructions, it still faces challenges in underwater environments. The inherent design of the original 3DGS focuses on representing geometric features and does not account for the scattering characteristics of the medium. Furthermore, underwater images are affected by complex optical phenomena--such as light absorption, backscattering, and motion-induced blur--that further complicate reconstruction.

{\color{revpurple}
\citet{Yang:comparison:2024} compare vanilla NeRF and vanilla 3DGS on underwater captures and report a pattern that broadly motivates later underwater-specific splatting methods: 3DGS is attractive for fast and sharp rendering in static scenes, but it needs additional modeling to cope with underwater physics, sparse baselines, and dynamic distractors. We therefore organize underwater 3DGS variants by design choice rather than chronology. Similar to the underwater NeRF overview above, this subsection first summarizes the main underwater splatting families through compact methodological equations and then maps representative methods onto these routes.
}

{\color{revteal}
Compared with NeRF, this limitation is also representational. NeRF's volumetric rendering can more naturally absorb distributed attenuation and scattering along a ray, whereas standard 3DGS represents the scene with explicit Gaussian particles whose colour and opacity can easily entangle true geometry with water-column effects. In turbid water, backscatter and marine snow may therefore be baked into floating splats or unstable opacity fields unless the renderer explicitly separates medium and object transport. This is one reason why underwater 3DGS papers increasingly add medium-aware transmittance, distractor masks, temporal degradation models, or restoration-coupled optimisation rather than relying on clear-medium splatting alone.
}

\begin{figure}[H]
\centering
\resizebox{0.85\linewidth}{!}{%
\begin{tikzpicture}[
    >=Stealth,
    node distance=1.1cm,
    thick,
    main_color/.style={purple},
    border_color/.style={black!60},
    bg_color/.style={purple!4},
    base_style/.style={
        rounded corners=2pt,
        line width=0.8pt,
        inner sep=8pt,
        align=center,
        font=\small
    },
    main_node/.style={
        base_style,
        draw=purple,
        fill=purple!4,
        minimum width=5.4cm,
    },
    sub_node/.style={
        base_style,
        draw=black!60,
        fill=white,
        minimum width=7.3cm,
        font=\scriptsize,
        postaction={
            path picture={
                \fill[purple] ([xshift=2pt]path picture bounding box.north west) rectangle (path picture bounding box.south west);
            }
        }
    }
]

\node[main_node] (base) at (0,0) {
    \textbf{Base 3DGS} \\[0.8ex]
    $\hat{I}(\mathbf{p})=\sum_i w_i(\mathbf{p})\alpha_i c_i$ \\[0.5ex]
    \textcolor{black!70}{\scriptsize Explicit Gaussian rasterization}
};

\coordinate (right_center) at (7.3,0);

\node[sub_node] (f2) at ($(right_center) + (0, 0.85)$) {
    \textbf{F2: Structure-/prior-guided stabilization} \\[0.5ex]
    $L=L_{\mathrm{render}}+\lambda_dL_{\mathrm{depth}}+\lambda_sL_{\mathrm{struct}}$ \\[0.3ex]
    \textit{Z-Splat, SeaFree-GS}
};

\node[sub_node, above=of f2] (f1) {
    \textbf{F1: Medium-aware splatting} \\[0.5ex]
    $\hat{I}(\mathbf{p})=\sum_i w_i(\mathbf{p})T_i^{\mathrm{med}}c_i+C_{\mathrm{back}}(\mathbf{p})$ \\[0.3ex]
    \textit{UW-GS, SeaSplat}
};

\node[sub_node] (f3) at ($(right_center) + (0, -0.85)$) {
    \textbf{F3: Dynamic / disturbance-aware splatting} \\[0.5ex]
    $\hat{I}_t(\mathbf{p})=m_s\hat{I}^{\mathrm{static}}_t(\mathbf{p})+m_d\hat{I}^{\mathrm{disturb}}_t(\mathbf{p})$ \\[0.3ex]
    \textit{MarineSTD-GS}
};

\node[sub_node, below=of f3] (f4) {
    \textbf{F4: Restoration-coupled splatting} \\[0.5ex]
    $L=L_{\mathrm{render}}+\lambda_{\mathrm{rest}}L_{\mathrm{rest}}$ \\[0.3ex]
    \textit{RecGS, R-Splatting}
};

\coordinate (branch) at ($(base.east) + (0.6, 0)$);
\draw[main_color, line width=1pt] (base.east) -- (branch);
\foreach \i in {f1, f2, f3, f4} {
    \draw[main_color, line width=1pt, rounded corners=4pt, ->]
        (branch) |- (\i.west);
}
\end{tikzpicture}%
}
\caption{{\color{revpurple}Methodological design space of underwater 3D Gaussian Splatting variants. The main families diverge through medium-aware splatting, structure- and prior-guided stabilization, dynamic disturbance modeling, and restoration-coupled optimization.}}
\label{fig:uw_3dgs_design_space}
\end{figure}

{\color{revpurple}
\noindent\textbf{Medium-aware splatting and medium-object transport.}
\[
\hat{I}(\mathbf{p})=\sum_i w_i(\mathbf{p})T_i^{\mathrm{med}}c_i + C_{\mathrm{back}}(\mathbf{p}),
\]
where medium-aware transmittance and backscatter terms are added on top of Gaussian colour and opacity so that the renderer separates object appearance from water-column effects. This is the main direction in UW-GS~\citep{wang2024uw}, SeaSplat~\citep{yang2024seasplat}, and WaterSplatting~\citep{li2024watersplatting}, all of which explicitly model attenuation, backscatter, or object-medium transport within splatting. Gaussian Splashing~\citep{mualem2024gaussian} and Aquatic-GS~\citep{liu2024aquatic} follow the same broad idea by treating the water medium as part of the rendering problem rather than as noise to be ignored. Relative to standard 3DGS variants, this family is the closest analogue to revised-IFM-aware underwater NeRFs, but it also increases optimization complexity and parameter coupling.
}

{\color{revpurple}
\noindent\textbf{Structure- and prior-guided stabilization.}
\[
L=L_{\mathrm{render}}+\lambda_dL_{\mathrm{depth}}+\lambda_sL_{\mathrm{struct}},
\]
where depth, edge, semantic, or smoothness priors are injected into splatting to stabilize geometry under degraded appearance. Z-Splat~\citep{qu2024z} improves limited-baseline underwater reconstruction by fusing sonar information with RGB splats, while SeaFree-GS~\citep{Liu:SeaFree:2025}, RestorGS~\citep{Qiao:RestoreGS:2025}, RUSplatting~\citep{jiang2025rusplatting}, SWAGSplatting~\citep{jiang2025swagsplatting}, and 3D-UIR~\citep{yuan2025threeduir} further exploit pseudo-depth, edge-aware losses, semantic cues, or smoothness constraints. UW-GS also uses normalized pseudo-depth from Depth Anything~\citep{Yang:depthanything:2024} as a secondary mechanism. Compared with generic 3DGS regularization, these underwater variants use priors not merely for denoising but to stabilize geometry when attenuation and colour shifts weaken correspondence reliability.
}

{\color{revpurple}
\noindent\textbf{Dynamic disturbances, marine snow, and temporal degradations.}
\[
\hat{I}_t(\mathbf{p}) = m_s \hat{I}^{\mathrm{static}}_t(\mathbf{p}) + m_d \hat{I}^{\mathrm{disturb}}_t(\mathbf{p}), \qquad m_s + m_d = 1,
\]
where the observation at time \(t\) is decomposed into static structure and disturbance-related terms, or equivalently filtered by masks and temporal degradation models. Dynamic disturbances arise not only from moving fish or suspended particles, but also from underwater-specific temporal effects such as caustics and flickering. MarineSTD-GS~\citep{Liu:Spatiotemporal:2025} explicitly models temporal degradations in the underwater image formation process, while UW-GS uses motion masks to reduce marine distractors as a secondary component. These methods differ from clear-medium dynamic 3DGS by treating marine clutter and temporal lighting instability as part of the observation model itself.
}

{\color{revsienna}
This also raises a coupling question for integrated pipelines: front-end marine-snow removal or dehazing can help dynamic reconstruction only if it remains temporally stable. If enhancement is applied independently to each frame and changes particle appearance, motion boundaries, or flicker statistics inconsistently, it can disrupt motion masks, erase useful temporal cues, or confuse the dynamic components of a 3DGS or NeRF backend instead of supporting them.
}

{\color{revpurple}
\noindent\textbf{Restoration-coupled splatting.}
\[
L=L_{\mathrm{render}}+\lambda_{\mathrm{rest}}L_{\mathrm{rest}},
\]
where restoration cues are written directly into splatting optimization rather than applied as a separate pre-processing stage. RecGS~\citep{zhang2024recgs} improves perceptual consistency by suppressing caustics with low-pass filtering and recurrent training, and R-Splatting~\citep{huang2025fromrestoration} fuses multiple restoration outputs into a single 3DGS model to address cross-view illumination shifts. AtlantisGS~\citep{Yi:AtlantisGS:2025} can be read as a more aggressive quality-oriented continuation of this route, pushing rendering quality further through stronger Gaussian optimization. These methods highlight a central point of this review: underwater 3D reconstruction increasingly depends on how enhancement cues are integrated, not simply on how well a vanilla renderer fits degraded images.
\par
}

{\color{revteal}
The main engineering trade-off relative to underwater NeRF is therefore clear: 3DGS offers faster training and much stronger interactive rendering once the representation is optimized, but NeRF-style volumetric models remain conceptually better matched to participating media and ray-wise scattering. In practice, underwater 3DGS compensates by introducing explicit medium terms, priors, or distractor suppression, whereas underwater NeRF variants can often encode those effects more directly in volumetric rendering at the cost of heavier optimisation.
}

\par
\begin{table*}[t]
\centering
{\color{revpurple}
\scriptsize
\renewcommand{\arraystretch}{1.12}
\setlength{\tabcolsep}{2.5pt}
\caption{{\color{revpurple}Summary of representative underwater 3D Gaussian Splatting methods.}}
\label{tab:underwater_3dgs_summary}
\resizebox{\textwidth}{!}{%
\begin{tabular}{>{}p{1.75cm}>{}p{1.1cm}>{}p{2.65cm}>{}p{2.3cm}>{}p{2.2cm}>{}p{2.2cm}>{}p{1.75cm}>{}p{1.9cm}}
\toprule
\textbf{Method} & \textbf{Primary family} & \textbf{Core underwater modeling idea} & \textbf{Relation to physics / IFM} & \textbf{Relation to enhancement} & \textbf{Dynamic / distractor handling} & \textbf{Strength} & \textbf{Limitation} \\
\midrule
Z-Splat \citep{qu2024z} & F2 & Extends splats along depth and fuses sonar with RGB under limited baselines & No explicit water-medium model & Indirect; focuses on geometry support & Not a primary focus & Helps missing-cone and sparse-baseline cases & Limited colour/medium fidelity \\
UW-GS \citep{wang2024uw} & F1 & Color-appearance model plus physics-guided density control & Explicit scattering-aware modeling & Produces cleaner rendered views through joint modeling & Motion masks plus pseudo-depth as secondary mechanisms & Strong joint gains in rendering and geometry & Depends on pseudo-depth and mask quality \\
SeaSplat \citep{yang2024seasplat} & F1 & Physically grounded underwater IFM inside splatting & Explicit medium-aware rendering & Generates enhanced renderings during reconstruction & Static-scene assumption & Fast rendering with improved visibility & Limited handling of moving objects \\
WaterSplat \citep{li2024watersplatting} & F1 & Separate transmittance for objects and surrounding medium & Strong physics-guided transmittance modeling & Joint restoration and rendering & Limited explicit dynamics & Competitive quality with real-time rendering & Added model complexity \\
RecGS \citep{zhang2024recgs} & F4 & Recurrent training with caustic suppression & Weak physical prior & Restoration-coupled via caustic removal & Indirect temporal consistency only & Better perceptual stability & Does not explicitly model water medium \\
R-Splatting \citep{huang2025fromrestoration} & F4 & Fuses multiple restoration outputs into one splat model & Uses restoration cues more than explicit IFM & Strong restoration-coupled reconstruction & Handles illumination variation across views & Improves geometric fidelity under varying lighting & Sensitive to upstream restoration quality \\
MarineSTD-GS \citep{Liu:Spatiotemporal:2025} & F3 & Integrates temporal degradations such as caustics and flicker & Physics-guided temporal degradation model & Indirect enhancement through temporal correction & Explicit spatiotemporal degradation modeling & Better robustness to dynamic illumination & Larger model and training cost \\
\bottomrule
\end{tabular}}
}
\end{table*}

{\color{revpurple}
\noindent Table~\ref{tab:underwater_3dgs_summary} maps representative methods to four shorthand families: F1 = medium-aware splatting, F2 = structure-/prior-guided stabilization, F3 = dynamic / disturbance-aware splatting, and F4 = restoration-coupled splatting.
}

\subsection{Performance Evaluation of Underwater NeRF/3DGS Models}
\label{ssec:end2end}

This subsection summarizes reported performance and visual comparisons of representative underwater NeRF- and 3DGS-based reconstruction models. Rather than treating these results as a unified reproduced benchmark, we use them to compare how existing models couple rendering, restoration, medium modeling, and dynamic-scene handling under underwater degradation.

{\color{revpurple}
\autoref{tab:underwater_comparison} compiles representative reported results for NeRF- and 3DGS-based methods on the SeaThru-NeRF dataset. All values are taken from the original publications rather than reproduced here under one unified experimental protocol, so the table should be read as a literature-level synthesis rather than as a strict fair-play leaderboard. It is most useful for illustrating trends: early physics-guided models such as UW-GS already outperform vanilla 3DGS by a clear margin, and later methods such as AtlantisGS~\citep{Yi:AtlantisGS:2025} report further gains in rendering quality and efficiency. The specific ranking should therefore be interpreted together with each method's assumptions about dynamics, medium modeling, priors, preprocessing, and training setup.
}

%\begin{sidewaystable}[t]
\begin{sidewaystable}[thp]
\centering
\caption{Quantitative comparison of NeRF- and Splatting-based methods on the SeaThru-NeRF dataset. All values are literature-reported results from the original publications, included here for indicative comparison rather than as a unified reproduced benchmark.}
% \resizebox{\textwidth}{!}{
\renewcommand{\arraystretch}{1.9}
\begin{tabular}{lccccccccccccccc}
\toprule
\multirow{2}{*}{Scene Method} &
\multicolumn{3}{c}{Curacao} &
\multicolumn{3}{c}{Panama} &
\multicolumn{3}{c}{IUI-Reasea} &
\multicolumn{3}{c}{Japanese-Redsea} &
\multicolumn{3}{c}{Average} \\
\cmidrule(lr){2-4} \cmidrule(lr){5-7} \cmidrule(lr){8-10} \cmidrule(lr){11-13} \cmidrule(lr){14-16}
 & PSNR$\uparrow$ & SSIM$\uparrow$ & LPIPS$\downarrow$ 
 & PSNR$\uparrow$ & SSIM$\uparrow$ & LPIPS$\downarrow$
 & PSNR$\uparrow$ & SSIM$\uparrow$ & LPIPS$\downarrow$
 & PSNR$\uparrow$ & SSIM$\uparrow$ & LPIPS$\downarrow$
 & PSNR$\uparrow$ & SSIM$\uparrow$ & LPIPS$\downarrow$ \\
\midrule
\multicolumn{16}{c}{\textbf{\textit{NeRF-based methods}}} \\ \hline
MIP-360 \citep{Barron:Mip-NeRF360:2022}     & 28.23 & 0.683 & 0.571 & 18.32 & 0.556 & 0.595 & 19.62 & 0.624 & 0.492 & 19.55 & 0.510 & 0.520 & 21.93 & 0.593 & 0.545 \\
Instant-NGP \citep{mueller2022instant}      & 27.91 & 0.707 & 0.385 & 23.46 & 0.636 & 0.426 & 20.63 & 0.558 & 0.603 & 23.23 & 0.655 & 0.357 & 23.81 & 0.639 & 0.443 \\
UWNeRF \citep{10656460}                     & 30.03 & 0.828 & 0.238 & 23.75 & 0.687 & 0.263 & 25.81 & 0.853 & 0.183 & 22.70 & 0.624 & 0.348 & 25.57 & 0.748 & 0.258 \\
SeaThru-NeRF \citep{levy2023seathru}        & 29.92 & 0.856 & 0.298 & 26.90 & 0.789 & 0.330 & 25.89 & 0.754 & 0.353 & 21.75 & 0.735 & 0.337 & 26.11 & 0.784 & 0.330 \\
ZipNeRF \citep{Barron:Zip:2023}             & 29.93 & 0.938 & 0.124 & 32.34 & 0.956 & 0.064 & 29.35 & 0.899 & 0.106 & 23.45 & 0.883 & 0.136 & 28.77 & 0.919 & 0.108 \\ \hline
\multicolumn{16}{c}{\textbf{\textit{Splatting-based methods}}} \\ \hline
3DGS \citep{kerbl:3Dgaussians:2023}                    & 30.97 & 0.936 & 0.193 & 30.80 & 0.919 & 0.196 & 23.13 & 0.877 & 0.240 & 21.97 & 0.867 & 0.202 & 26.72 & 0.900 & 0.208 \\
WildGaussian \citep{kulhanek2024wildgaussians} & 29.52 & 0.880 & 0.313 & 24.94 & 0.756 & 0.395 & 28.34 & 0.870 & 0.177 & 22.08 & 0.839 & 0.312 & 26.22 & 0.836 & 0.299 \\
SeaSplat \citep{yang2024seasplat}           & 30.30 & 0.900 & 0.190 & 28.76 & 0.900 & 0.150 & 26.67 & 0.870 & 0.210 & 22.70 & 0.870 & 0.180 & 27.61 & 0.885 & 0.183 \\
WA-GS \citep{Fan:waater:2025}               & 28.29 & 0.900 & 0.158 & 30.07 & 0.938 & 0.084 & 30.43 & 0.891 & 0.186 & 23.17 & 0.864 & 0.153 & 27.99 & 0.898 & {0.145} \\
3D-UIR \citep{yuan2025threeduir}            & 30.98 & 0.907 & 0.187 & 29.82 & 0.900 & 0.181 & 28.30 & 0.841 & 0.252 & 23.37 & 0.857 & 0.187 & 28.12 & 0.876 & 0.202 \\
UW-GS \citep{wang2024uw}                    & 31.77 & 0.943 & 0.144 & 31.79 & 0.936 & 0.116 & 28.65 & 0.933 & 0.125 & 23.05 & 0.860 & 0.190 & 28.82 & 0.918 & 0.144 \\
RUSplatting \citep{jiang2025rusplatting}    & 30.96 & 0.932 & 0.161 & 31.87 & 0.934 & 0.140 & 29.80 & 0.929 & 0.185 & 24.54 & 0.871 & 0.181 & 29.29 & 0.917 & 0.167 \\
RestorGS \citep{Qiao:RestoreGS:2025} & 31.95 & 0.944 & 0.055 & 30.79 & 0.932 & 0.046 & 29.97 & 0.952 & 0.028 & 24.05 & 0.882 & 0.071 & 29.19 & 0.928 & 0.050 \\
WaterSplatting \citep{li2024watersplatting} & 32.67 & 0.957 & 0.110 & 31.49 & 0.948 & 0.075 & 29.39 & 0.910 & 0.180 & 25.20 & 0.904 & 0.113 & 29.19 & 0.930 & 0.120 \\
SWAGSplatting \citep{jiang2025swagsplatting}& 31.84 & 0.941 & 0.161 & 31.82 & 0.939 & 0.139 & 30.37 & 0.936 & 0.180 & 24.52 & 0.889 & 0.177 & 29.64 & 0.926 & 0.164 \\
R-Splatting \citep{huang2025fromrestoration}& 32.98 & 0.956 & 0.163 & 32.52 & 0.930 & 0.107 & 30.15 & 0.947 & 0.105 & 24.03 & 0.868 & 0.211 & 29.92 & 0.925 & 0.147 \\
AtlantisGS \citep{Yi:AtlantisGS:2025}       & 33.27 & 0.949 & 0.112 & 32.35 & 0.953 & 0.076 & 31.42 & 0.935 & 0.201 & 26.59 & 0.916 & 0.108 & 30.91 & 0.938 & 0.124
\\

\bottomrule
\end{tabular}
% }
\label{tab:underwater_comparison}
\end{sidewaystable}
%\end{sidewaystable}

\begin{figure}[ht]
    \centering
    \includegraphics[width=\linewidth]{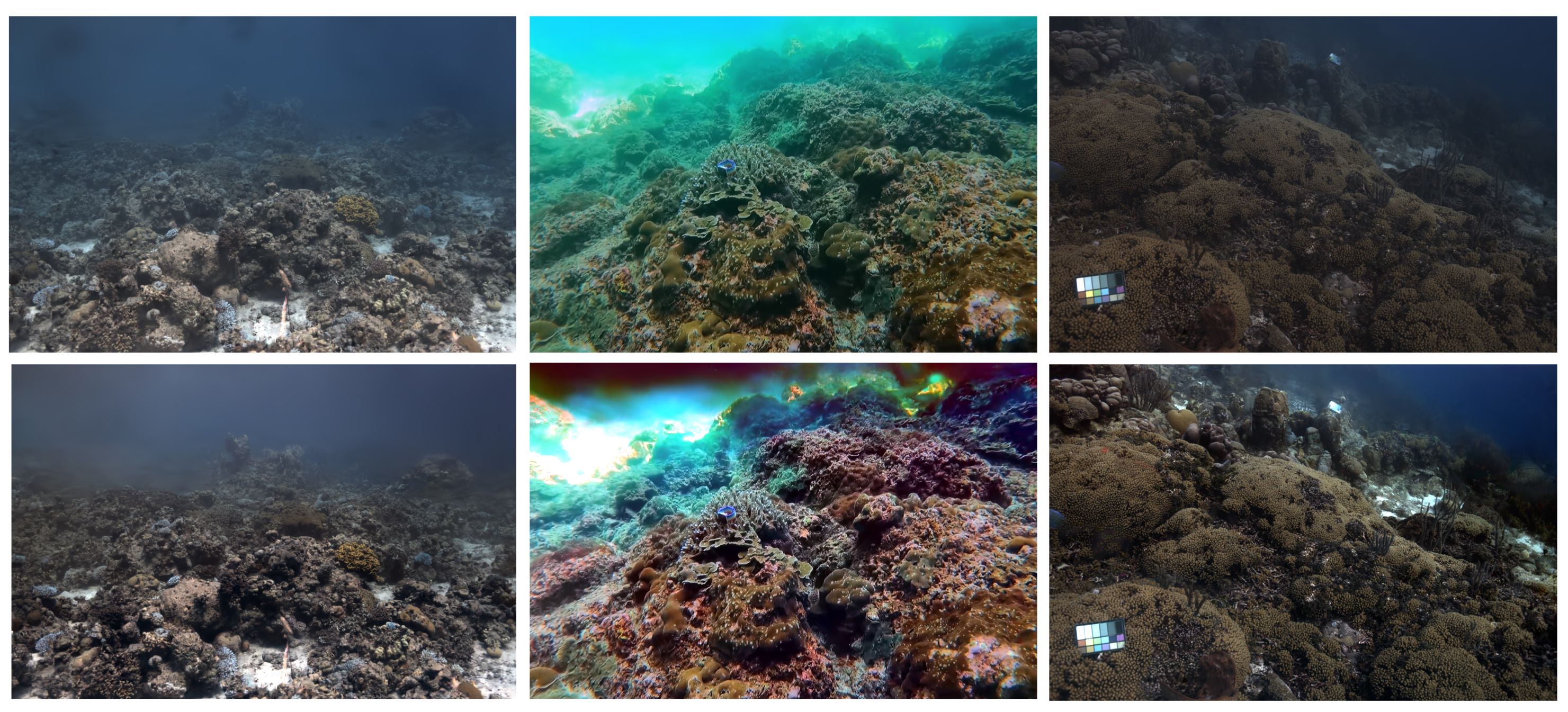}
    \caption{Visualization of rendered images (top) and the estimated clean images (bottom) using UW-3DGS~\citep{wang2024uw}}
    \label{fig:uw3dgs_vis1}
\end{figure}

Representative underwater scene reconstruction models increasingly integrate visual recovery with medium-aware rendering, rather than treating enhancement as a separate post-processing step. \autoref{fig:uw3dgs_vis1} presents the rendered images and the estimated clean images produced by UW-3DGS~\citep{wang2024uw}, where the water-medium component is removed from the rendered result.

\begin{figure}[ht]
    \centering
    \includegraphics[width=\linewidth]{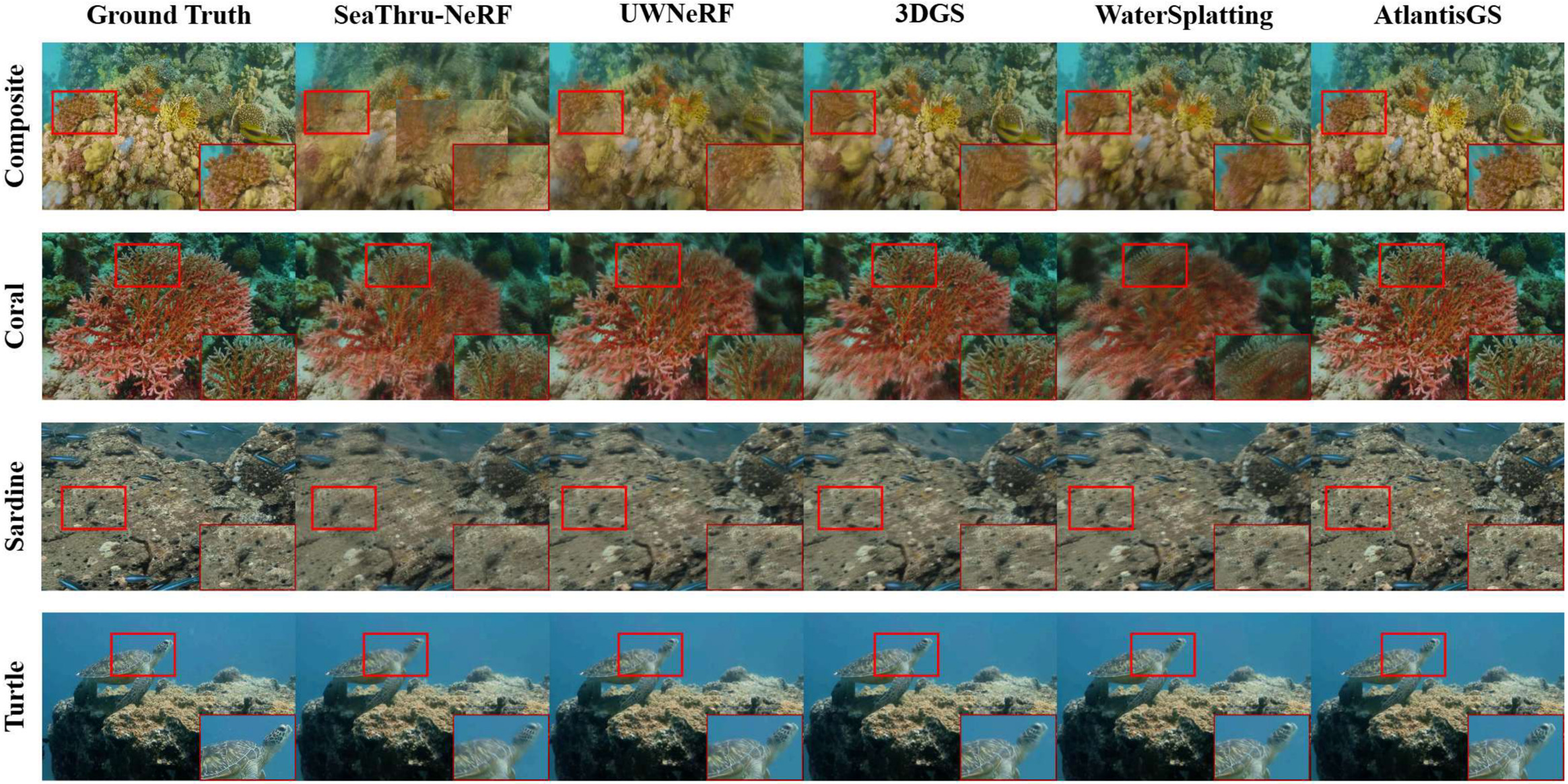}
    \caption{Underwater scene rendering comparison on the UWNeRF dataset, reproduced from Fig.~5 of AtlantisGS~\citep{Yi:AtlantisGS:2025}. From left to right: ground truth, SeaThru-NeRF, UWNeRF, 3DGS, WaterSplatting, and AtlantisGS. The comparison highlights that traditional 3DGS and NeRF methods with proposal sampling struggle with semi-transparent underwater media.}
    \label{fig:uw3dgs_vis_results}
\end{figure}

{\color{revpurple}
\autoref{fig:uw3dgs_vis_results} provides an illustrative sparse-view rendering comparison on the UWNeRF dataset among SeaThru-NeRF, UWNeRF, vanilla 3DGS, WaterSplatting, and AtlantisGS. The main review-level conclusion is not that one method is universally ``best'', but that underwater reconstruction models benefit when medium effects, sparse-view constraints, and restoration cues are handled within the rendering process. In this comparison, AtlantisGS recovers sharper local structure and more coherent appearance under semi-transparent underwater media, whereas clear-medium or less medium-aware baselines tend to blur fine details or retain colour and opacity inconsistencies.
}

{\color{revsienna}
It is also important to note that most current benchmarking still reports rendering quality or image-domain fidelity rather than true joint geometry-and-trajectory outcomes. Metrics such as Chamfer distance, Cloud-to-Cloud distance, ATE, or RPE are still rarely reported together with enhancement-oriented measures, which makes fair cross-domain assessment of integrated pipelines an important open challenge.
}

Even as NeRF/3DGS-based approaches promise photorealistic reconstructions, they remain computationally demanding, especially for large-scale underwater surveys. Processing thousands of images from extensive sites, such as reefs spanning hundreds of meters, is far beyond the typical usage scenario in small object scans or single-room reconstructions. Practical considerations include:
    \textit{Data Collection Overlap:} Achieving consistent coverage in a turbid environment can be challenging. If certain areas are poorly visible or have drastically different color casts, the optimization might fail.
   \textit{Long Training Times:} While faster NeRF variants exist, training can still take hours or days for high-resolution scenes. 3DGS training time is shorter, but generating the initial point cloud, generally through COLMAP, still takes hours. Real-time or near-real-time feedback is typically unattainable in standard setups.
  \textit{Refraction and Partial Occlusions:} The integrated assumption that each pixel ray corresponds to a linear path in Euclidean space is flawed with thick camera housings or highly refractive ports. Additional geometry modeling is required to handle these complexities.

Despite these hurdles, NeRFs/3DGSs and their successors represent a key direction for future underwater 3D modeling, given their ability to produce physically consistent volumetric reconstructions that merge geometry with realistic appearance.

{\color{revsienna}
For dynamic scenes, however, the interaction with front-end enhancement remains delicate. Dynamic NeRF and 3DGS backends need temporally stable cues to separate moving objects, marine snow, and illumination fluctuations; front-end enhancement helps only when it is disturbance-aware and temporally consistent. Independent per-frame beautification can instead erase motion cues or hallucinate temporally inconsistent structure, pushing enhancement and dynamic reconstruction to work at cross-purposes rather than synergistically.
}

\subsection{Hybrid and Multi-Sensor Systems}

While the preceding subsections focus primarily on optical data and neural or geometric reconstructions, there is a growing trend toward fusing optical imagery with other sensor modalities. Sonar or acoustic cameras can provide robust wide-area scans even in turbid conditions, albeit at reduced resolution. Short-range lasers or structured light can yield accurate depth near the camera. By merging these complementary datasets, reconstructions can become both more extensive (covering large areas in rough resolution) and more detailed (where optical data is available, it refines geometry and color). 

Examples of multi-sensor approaches might incorporate:
\begin{itemize}
    \item \textbf{Acoustic Bathymetry + Photogrammetry:} Large-scale mapping using side-scan or multibeam sonar plus local photogrammetry for high-detail in key areas \citep{Lacka2024}.
    \item \textbf{Forward-Looking Sonar + Visual SLAM:} Robotics platforms that rely on sonar for broad obstacle detection, combined with visual-inertial SLAM for fine-scale corridor or hull inspection \citep{Rahman:SVIn2:2019, Cheng2022, Zhang:Integration:2024}.
    \item \textbf{RGB-D Fusion:} Underwater variants of Kinect-like sensors or scanning lasers to collect partial depth maps~\citep{Anwer:Underwater:2017,Lu:Depth:2017}, integrated into a more general volumetric or point-based pipeline. 
\end{itemize}
Though each sensor type introduces unique calibration complexities (particularly with refraction), the synergy can mitigate the classic photogrammetry pitfalls (e.g., lack of features, severe scattering).

\subsection{Discussion and Open Challenges}
\label{sec:Discussion and Open Challenges}
{\color{revorange}
This subsection focuses on challenges that remain specific to underwater reconstruction itself. Shared issues such as supervision scarcity, benchmark coverage, and cross-domain evaluation are synthesized later in \autoref{subsec:cross_cutting_challenges} and \autoref{subsec:future_datasets_models_metrics}; here the emphasis is on geometry-specific failure modes and deployment bottlenecks.
}

{\color{revteal}\noindent\textbf{Refraction Modeling.}
Explicitly handling the multi-layer interfaces in camera housings or free-floating cameras is essential for accurate geometry. Classical SfM/MVS pipelines, COLMAP-style calibration, vanilla NeRF, vanilla 3DGS, and most learned MVS methods still inherit pinhole-like or central-camera assumptions by default. In underwater deployments, that approximation is most fragile for flat-port housings, while dome-port systems can sometimes be closer to a single-viewpoint model if alignment is good. Methods that incorporate refractive calibration or ray-tracing through flat ports, curved domes, or multiple media transitions can reduce systematic distortions, but require more complex solvers~\citep{sedlazeck2009rov,DIAMANTI2024105985}. Underwater neural renderers such as SeaThru-NeRF, WaterHE-NeRF, SeaSplat, and WaterSplatting partly move in this direction by coupling rendering with medium-aware optics~\citep{levy2023seathru,zhou2023waterhe,yang2024seasplat,li2024watersplatting}, yet a robust and widely adoptable solution for metric field use is still missing.}

{\color{revorange}
\noindent\textbf{Domain Adaptation and Robust Learning.}
For reconstruction, robust learning matters because appearance shifts must be handled without breaking multi-view consistency. Domain adaptation, synthetic-to-real simulation, and geometry-aware self-supervision are therefore useful when they stabilize correspondence and depth across viewpoints rather than merely improving isolated views. Broader shared issues of data scarcity, supervision, and evaluation are discussed later in \autoref{subsec:cross_cutting_challenges} and \autoref{subsec:future_datasets_models_metrics}.
}

\noindent\textbf{Dynamic Scene Reconstruction.}
Many oceanic scenes contain dynamic elements, from small fish to shifting vegetation. While some tasks focus on static structures (like coral reefs, ship hulls, or archaeological remains), others may explicitly aim to capture dynamic processes--e.g., ecological interactions or pipeline flow. Generalizing approaches such as 4D Gaussian Splatting or dynamic NeRF to handle partial occlusions, swirling particulates, or flickering illumination remains challenging.

{\color{revteal}\noindent\textbf{Efficient Rendering and Interactivity.}
NeRF training times can be lengthy, while classical SfM can be slow or memory-intensive for large sites. In operational contexts--e.g., an ROV exploring a deep shipwreck--an interactive interface could provide immediate feedback on coverage or areas needing more data. Achieving near real-time or incremental updates for underwater 3D scenes requires optimizing every stage of the pipeline: from feature matching and bundle adjustment to volumetric or splat-based rendering. This also means distinguishing genuinely deployable online tools from offline post-processing pipelines: many high-performing enhancement, NeRF, and 3DGS systems remain better suited to post-dive reconstruction than to frame-by-frame robotic navigation.
}

\noindent\textbf{Enhanced Physical Modeling.}
Both classical photogrammetry and neural approaches can benefit from better integration of the physical laws governing underwater light transport. For instance, embedding the {Jaffe-McGlamery} model \citep{mcglamery1980computer} or the {revised model} \citep{akkaynak2018revised} into cost functions might more accurately tease apart geometry from color attenuation. Similarly, scattering phenomena could be parameterized for each camera viewpoint based on distance or angle, significantly boosting reconstruction fidelity in murky or partially lit conditions.

\noindent\textbf{Towards Autonomous Large-Scale Survey.}
A key ambition is to perform wide-area mapping of underwater environments autonomously at high resolution. Potentially, a swarm of AUVs or ROVs could coordinate to gather photometric data from multiple vantage points, merging the partial reconstructions. This could yield a holistic map of reefs or canyons spanning kilometers. Achieving stable stitching of partial reconstructions from many vantage points or vehicles, each with its own dynamic lighting, remains a complex challenge, especially if there is little ground truth or fixed reference.

{\color{revorange}
To conclude the methodological review before benchmarking, \autoref{tab:cross_domain_summary} summarizes the main strengths, limitations, and best-use scenarios of the principal UIE and 3D reconstruction families, together with their typical impact on downstream geometry.
\par
}

\begin{table*}[ht]
\centering
{\color{revorange}
\scriptsize
\renewcommand{\arraystretch}{1.12}
\setlength{\tabcolsep}{4pt}
\caption{{\color{revorange}Summary of major underwater enhancement and reconstruction method families.}}
\label{tab:cross_domain_summary}
\resizebox{\textwidth}{!}{%
\begin{tabular}{p{2.2cm}p{3.0cm}p{2.7cm}p{2.7cm}p{2.7cm}p{3.2cm}}
\toprule
\textbf{Family} & \textbf{Core idea} & \textbf{Strengths} & \textbf{Limitations} & \textbf{Best-use scenario} & \textbf{Impact on 3D reconstruction} \\
\midrule
Classical UIE & Histogram, Retinex, DCP/ASM-style priors, and fusion heuristics & Fast, interpretable, light-weight, often easy to deploy & Limited robustness, brittle assumptions, framewise inconsistency & On-board preview, moderate distortions, shallow-water correction & Can improve feature visibility, but may also amplify colour bias or mismatch views \\
Learning-based UIE & CNN, GAN, transformer, Mamba, and diffusion restoration & Strong visual quality and data-driven adaptation & Data hungry, risk of hallucination or domain overfitting & Complex appearance distortions with sufficient training data & Helpful when structural fidelity is preserved; harmful if geometry cues are altered \\
Physics-guided UIE & Combines deep models with IFM or revised-IFM constraints & Better physical plausibility and cross-water robustness & Requires medium assumptions or auxiliary priors & Real scenes with limited labels and strong water-medium effects & Often safer for downstream geometry because enhancement remains physically constrained \\
Photogrammetry & SfM/MVS with calibration, matching, bundle adjustment, and meshing & Mature, interpretable geometry, low-cost capture & Sensitive to low contrast, refraction, and repetitive texture & Static scenes with sufficient overlap and calibration & Strong geometry when correspondences are reliable; deteriorates rapidly under degraded imagery \\
NeRF & Implicit volumetric scene representation with differentiable rendering & High-quality novel views and joint geometry/appearance learning & Slow training, high compute, difficult explicit geometry extraction & Medium-sized scenes prioritizing view synthesis and appearance fidelity & Can absorb enhancement and medium modeling into one optimizer, but requires stable supervision \\
3DGS & Explicit Gaussian representation with fast rasterization & Real-time rendering, rapid convergence, strong visual detail & Needs robust initialization and additional modeling for underwater physics & Interactive rendering and faster reconstruction loops & Benefits strongly from medium-aware splatting and restoration-aware priors \\
Integrated enhancement + reconstruction & Joint or tightly coupled restoration and geometry estimation & Better correspondence preservation than naive pre-enhancement and more faithful rendering than degraded-only training & Higher modeling complexity and stronger data/optimization requirements & Challenging scenes where appearance degradation directly harms geometry & Most promising route for closing the UIE-to-3D loop under real underwater conditions \\
\bottomrule
\end{tabular}}
}
\end{table*}

\section{Pipeline-Level Evaluation}
\label{sec:Benchmarking}
{\color{revpurple}
This section summarizes public dataset context, evaluation criteria, and pipeline-level case studies for underwater 3D reconstruction. It is intended as a review-oriented evaluation discussion rather than a claim of a new standalone benchmark contribution. We focus on two practical reconstruction settings: (1) direct reconstruction without enhancement and (2) a two-stage pipeline that first enhances underwater images and then reconstructs geometry. Model-level reported results for existing underwater NeRF/3DGS methods are discussed earlier in \autoref{ssec:end2end}; here the comparisons are included to show how different image-processing choices affect geometry, correspondence quality, and visible failure modes.
}

{\color{revpurple}
We rely on public datasets wherever possible. Publicly accessible underwater 3D scene datasets remain scarce, as summarized in \autoref{tab:underwater_3d_datasets}; many contain only a small number of scenes or images per scene, which in turn limits how broadly any comparison can be generalized.
} {\color{revteal}The most commonly used evaluation criteria for NeRF-based and 3DGS-based methods are based on the visual quality of image reconstruction (also see Secion \ref{sssec:evaluation}). Rendering quality is computed using full-reference metrics such as PSNR, SSIM, and LPIPS~\citep{zhang2018unreasonable}. Other metrics for 3D modeling are also used, including feature matching success rate, Absolute Trajectory Error (such as that used by \citet{Malyugina:beam:2025}), Relative Pose Error (RPE), Chamfer Distance (CD) measuring the dissimilarity between two finite point sets, Cloud-to-Cloud (C2C) distance, F-score combining precision and recall of reconstructed points, and dynamic point tracking metrics, including occlusion accuracy (OA) measuring the binary accuracy of occlusion predictions, and average Jaccard (AJ) evaluating tracking and occlusion prediction jointly~\citep{doersch2023tapir}.}
\par
\begin{table*}[ht]
    \centering
    \small
    \renewcommand{\arraystretch}{1.2}
    \setlength{\tabcolsep}{2.5pt}
    \caption{\textbf{Summary of publicly available underwater 3D scene datasets and their key characteristics} 
    ``Total \#~Images'' refers to the total number of raw images in each dataset. }
    \label{tab:underwater_3d_datasets}
    
    \begin{tabular}{l c c c c c c}
        \toprule
        \textbf{Dataset} & \textbf{Year} & \textbf{\#~Scenes} & \textbf{Total \# images} &  \textbf{Add. Info.} & \textbf{Resolution} & \textbf{Download} \\
        \midrule
          UWBundle~\citep{Skinner:2016ab} &  2016 & 1 & 32 & None & 1K & \href{https://github.com/kskin/UWbundle/tree/master/data}{Link} \\
          SeaThru~\citep{levy2023seathru} & 2023&  4 & 88 &  Pose & 1K & \href{https://sea-thru-nerf.github.io/}{Link} \\
        NUSR~\citep{10656460}
          & 2024 
          & 4 
          & 82 
          & Motion mask, Pose
          & 1K -- 2K
          &  \href{https://drive.google.com/file/d/1AErUPPKwQ1cOwTF0ORVd6J91Od_Brw8z/view}{Link} \\
          BVI-Coral~\citep{bvi-coral} & 2024 & 23 & $>$2000 & None & 1K & \href{https://zenodo.org/records/11093417}{Link}\\
          S-UW~\citep{wang2024uw} & 2024 & 4 & 96 & Pose & 1K & \href{https://zenodo.org/records/14835876}{Link} \\
          Submerged3D~\citep{jiang2025rusplatting} & 2025 & 4 & 80 & Pose & 1K & \href{https://zenodo.org/records/15482420}{Link} \\
          \bottomrule
    \end{tabular}
\end{table*}

\subsection{Pipeline-Level Reconstruction Case Studies}
\label{ssec:pure3D}
{\color{revpurple}
As an illustrative baseline setting, we consider reconstruction pipelines that do not explicitly correct underwater degradations before geometry estimation. The purpose is not to claim a new baseline benchmark, but to make the effect of later enhancement-aware pipelines easier to interpret.
}

\begin{figure}[ht]
  \centering

  % --- Row 1 ---
  \begin{minipage}[b]{0.32\linewidth}
    \centering
    \includegraphics[height=3.38cm,width=\linewidth]{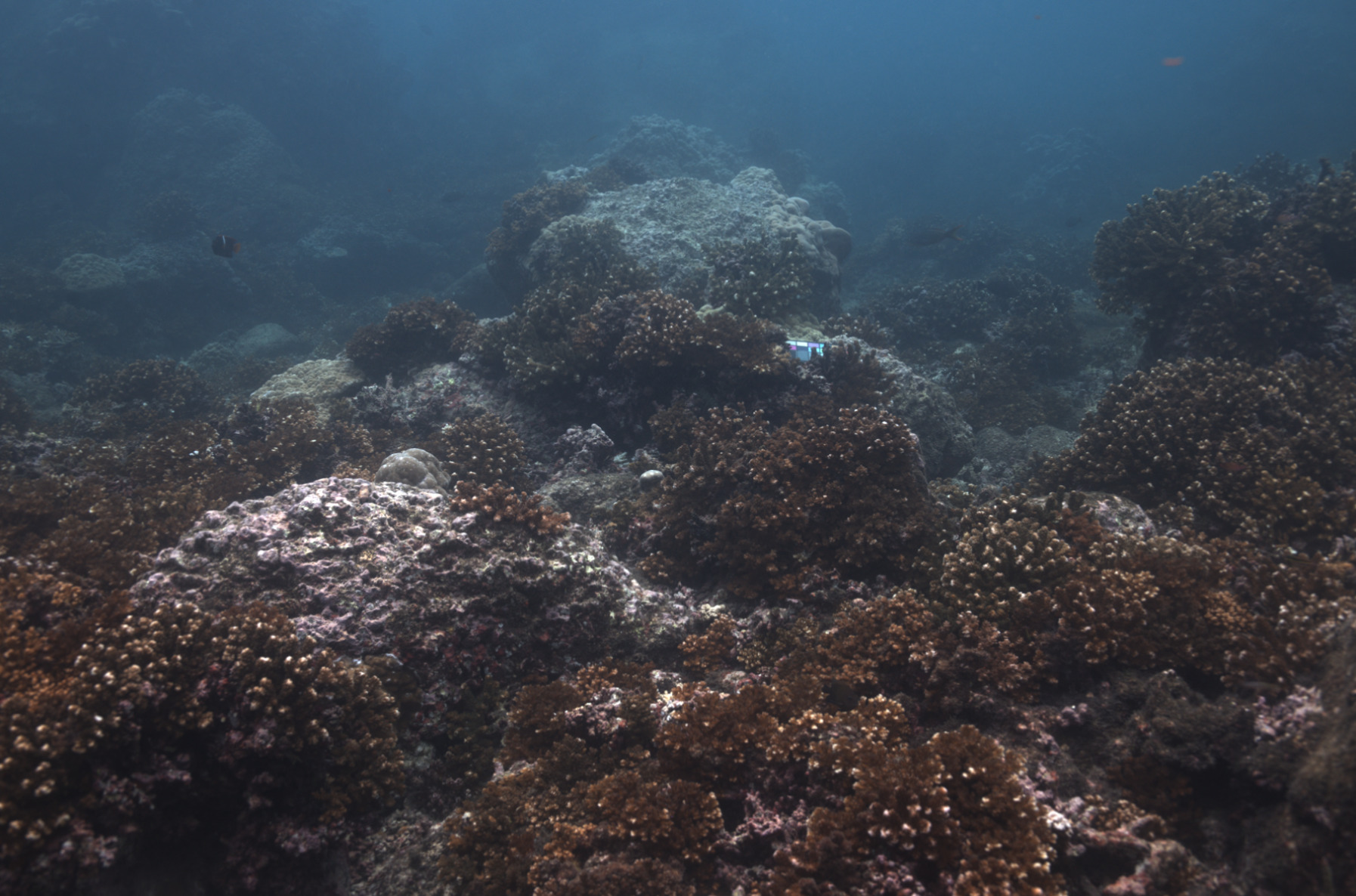}\\
    \small (a)
  \end{minipage}\hfill
  \begin{minipage}[b]{0.32\linewidth}
    \centering
    \includegraphics[height=3.38cm,width=\linewidth]{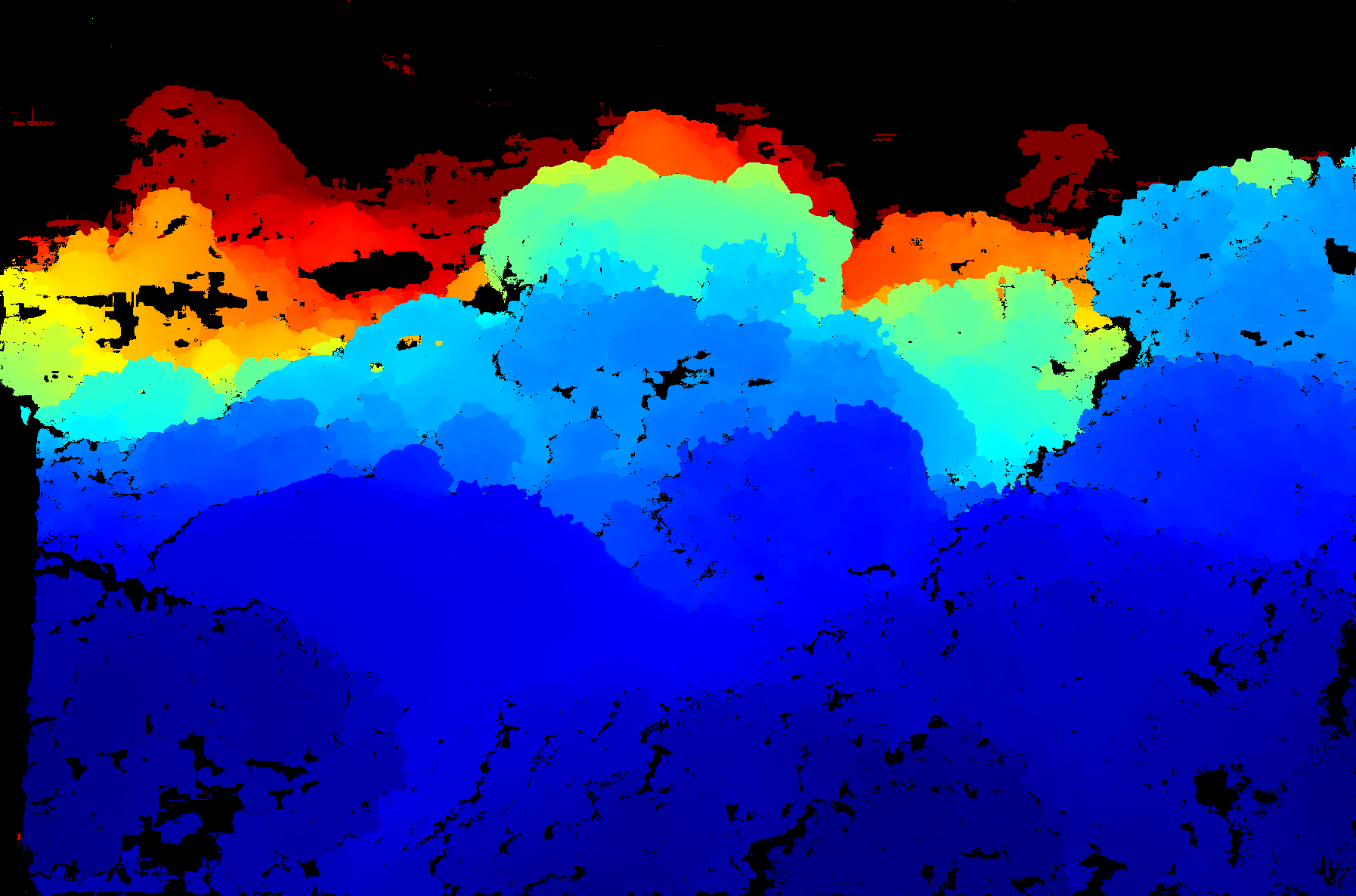}\\
    \small (b)
  \end{minipage}\hfill
  \begin{minipage}[b]{0.32\linewidth}
    \centering
    \includegraphics[height=3.38cm,width=\linewidth]{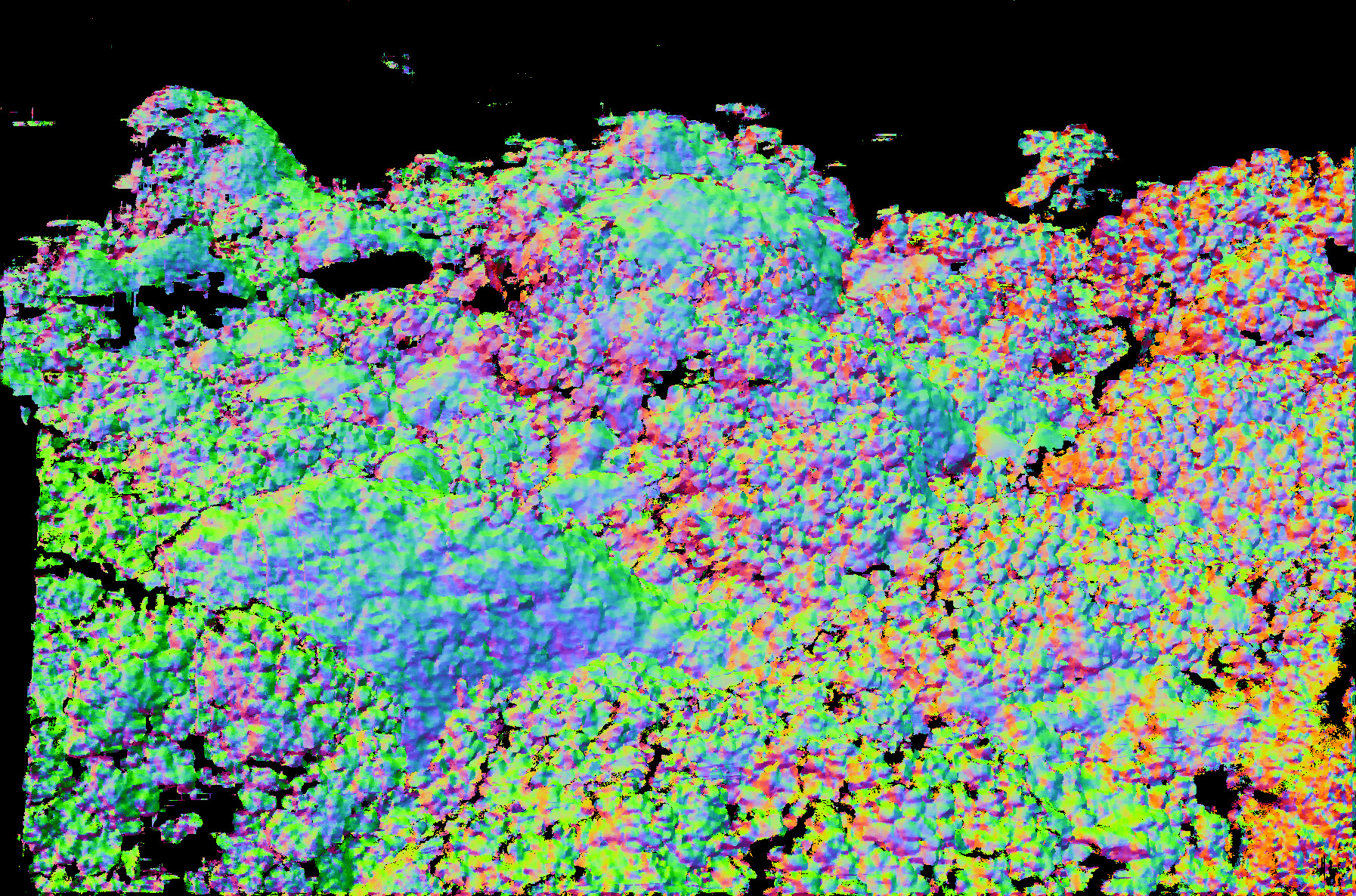}\\
    \small (c)
  \end{minipage}

  \vspace{0.5em}

  % --- Row 2 ---
  \begin{minipage}[b]{0.32\linewidth}
    \centering
    \includegraphics[height=3.38cm,width=\linewidth]{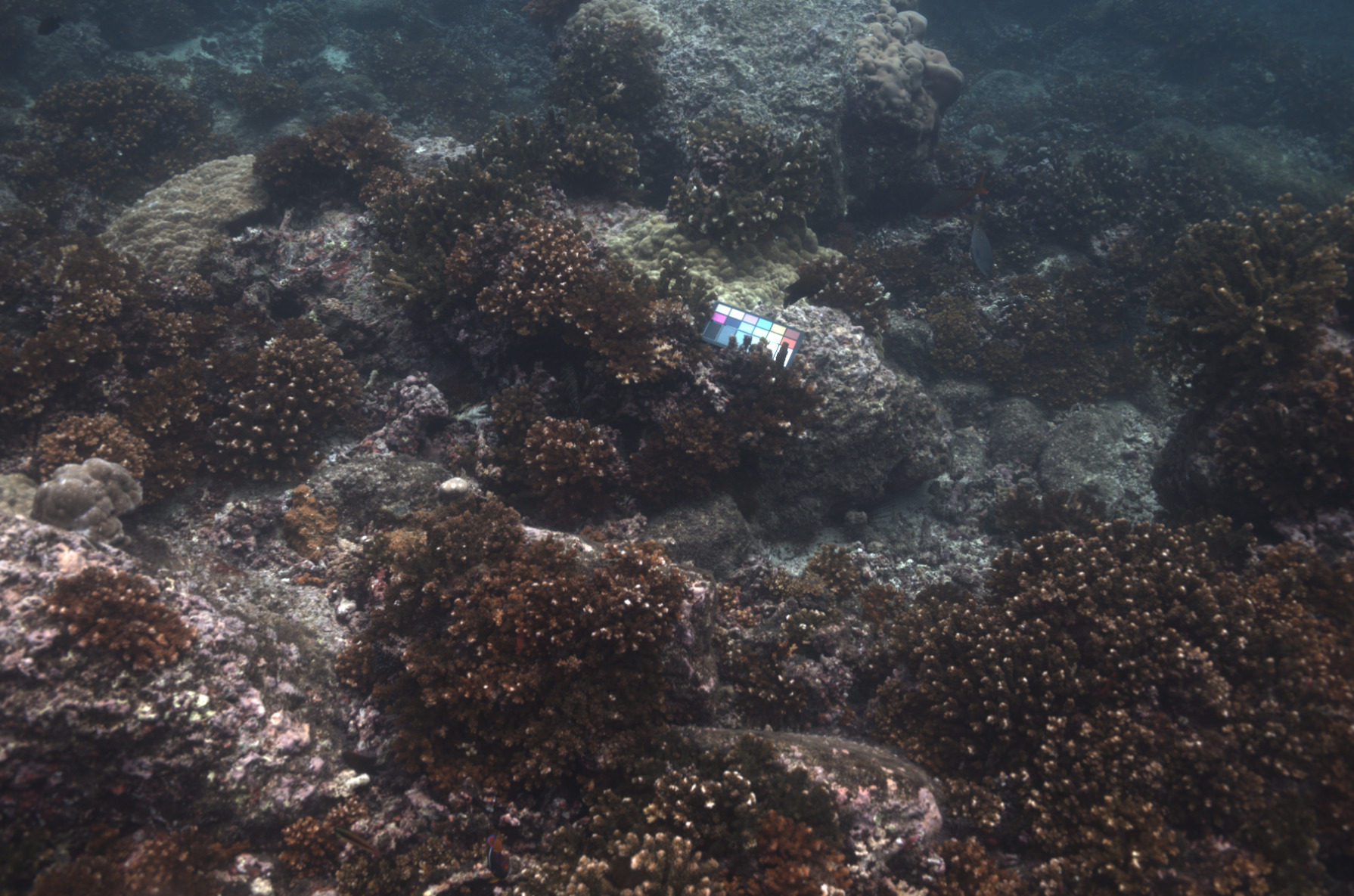}\\
    \small (d)
  \end{minipage}\hfill
  \begin{minipage}[b]{0.32\linewidth}
    \centering
    \includegraphics[height=3.38cm,width=\linewidth]{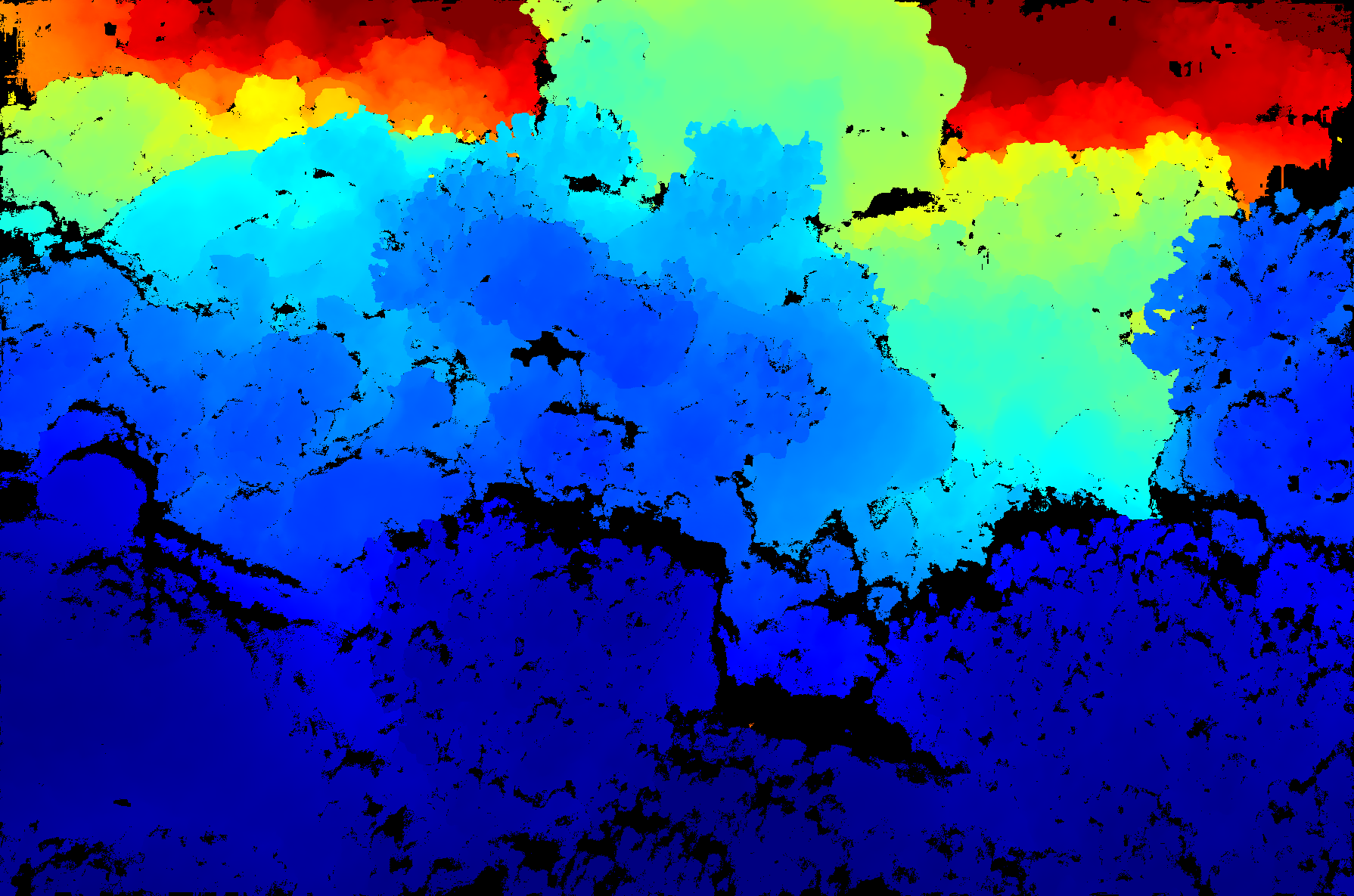}\\
    \small (e)
  \end{minipage}\hfill
  \begin{minipage}[b]{0.32\linewidth}
    \centering
    \includegraphics[height=3.38cm,width=\linewidth]{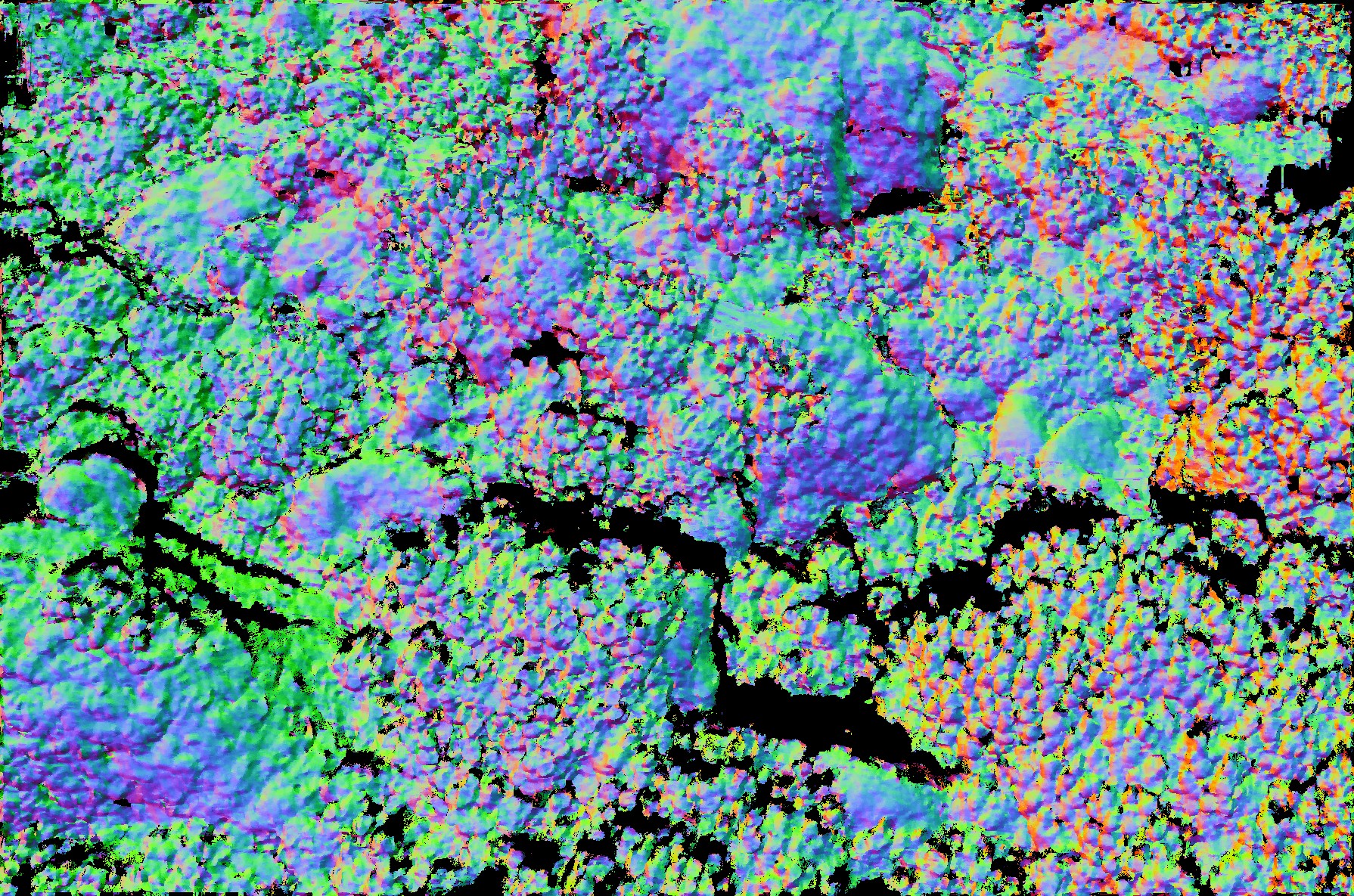}\\
    \small (f)
  \end{minipage}

  \vspace{0.5em}

  % --- Row 3 ---
  \begin{minipage}[b]{0.32\linewidth}
    \centering
    \includegraphics[height=3.38cm,width=\linewidth]{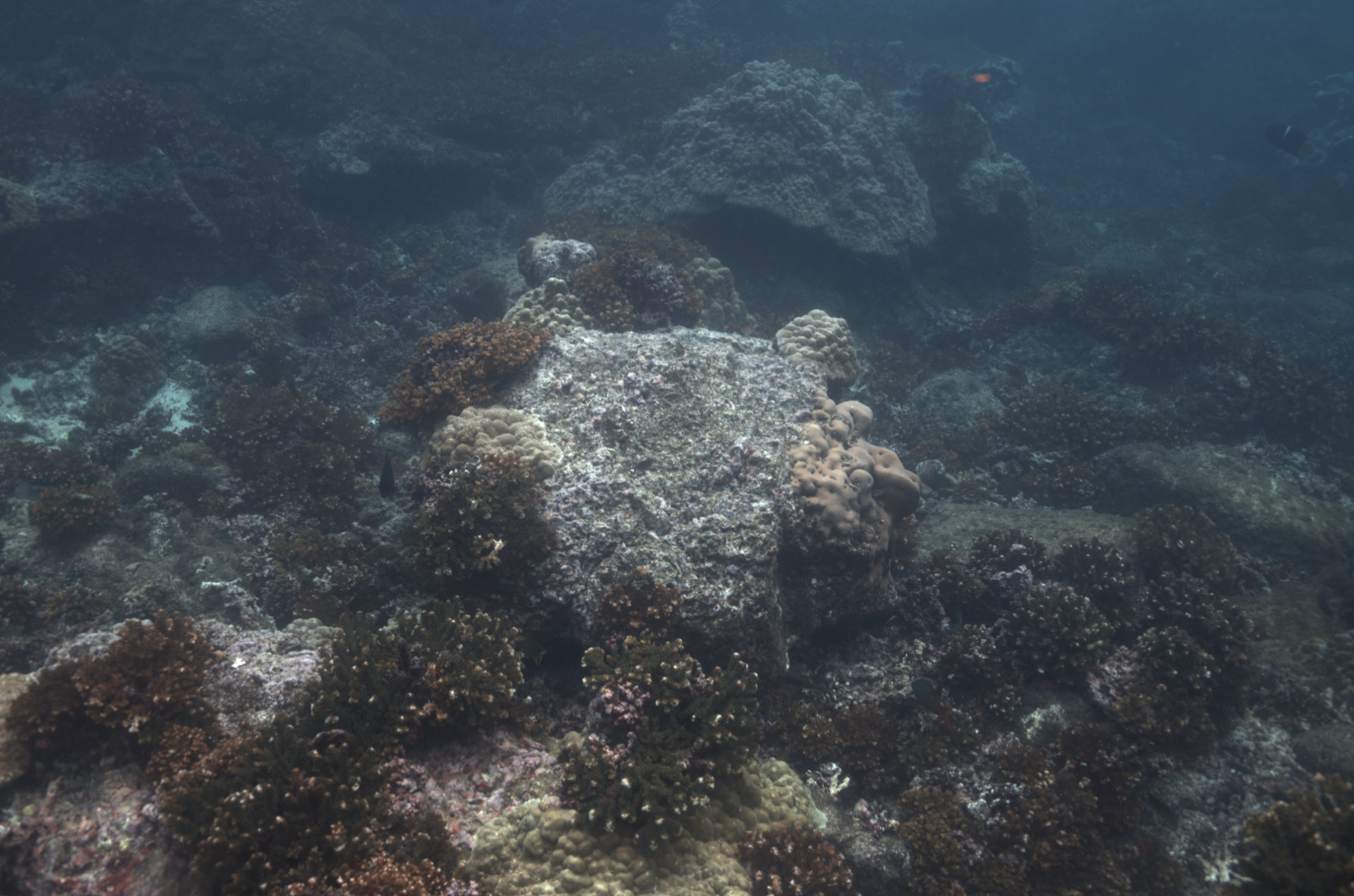}\\
    \small (g)
  \end{minipage}\hfill
  \begin{minipage}[b]{0.32\linewidth}
    \centering
    \includegraphics[height=3.38cm,width=\linewidth]{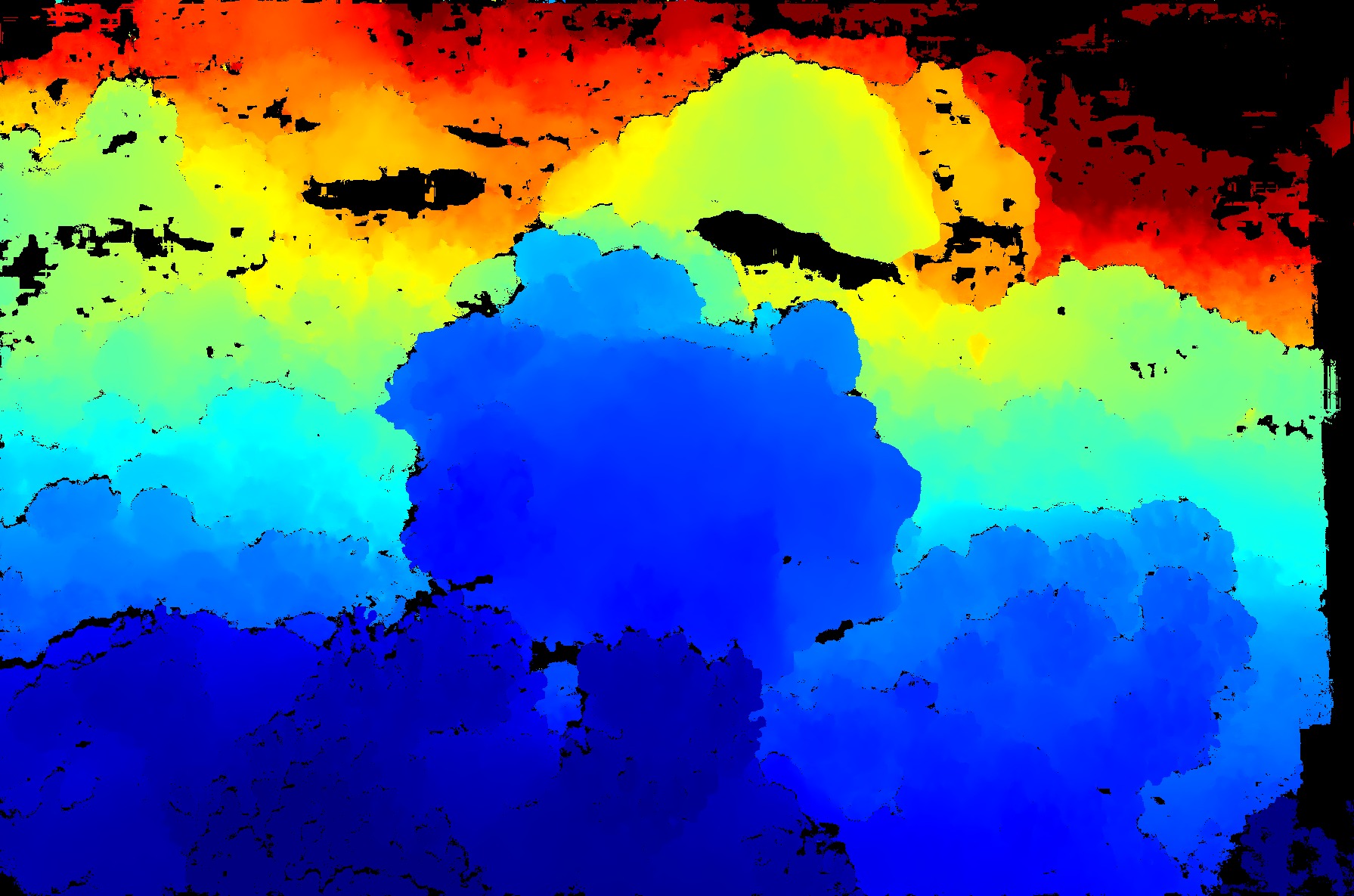}\\
    \small (h)
  \end{minipage}\hfill
  \begin{minipage}[b]{0.32\linewidth}
    \centering
    \includegraphics[height=3.38cm,width=\linewidth]{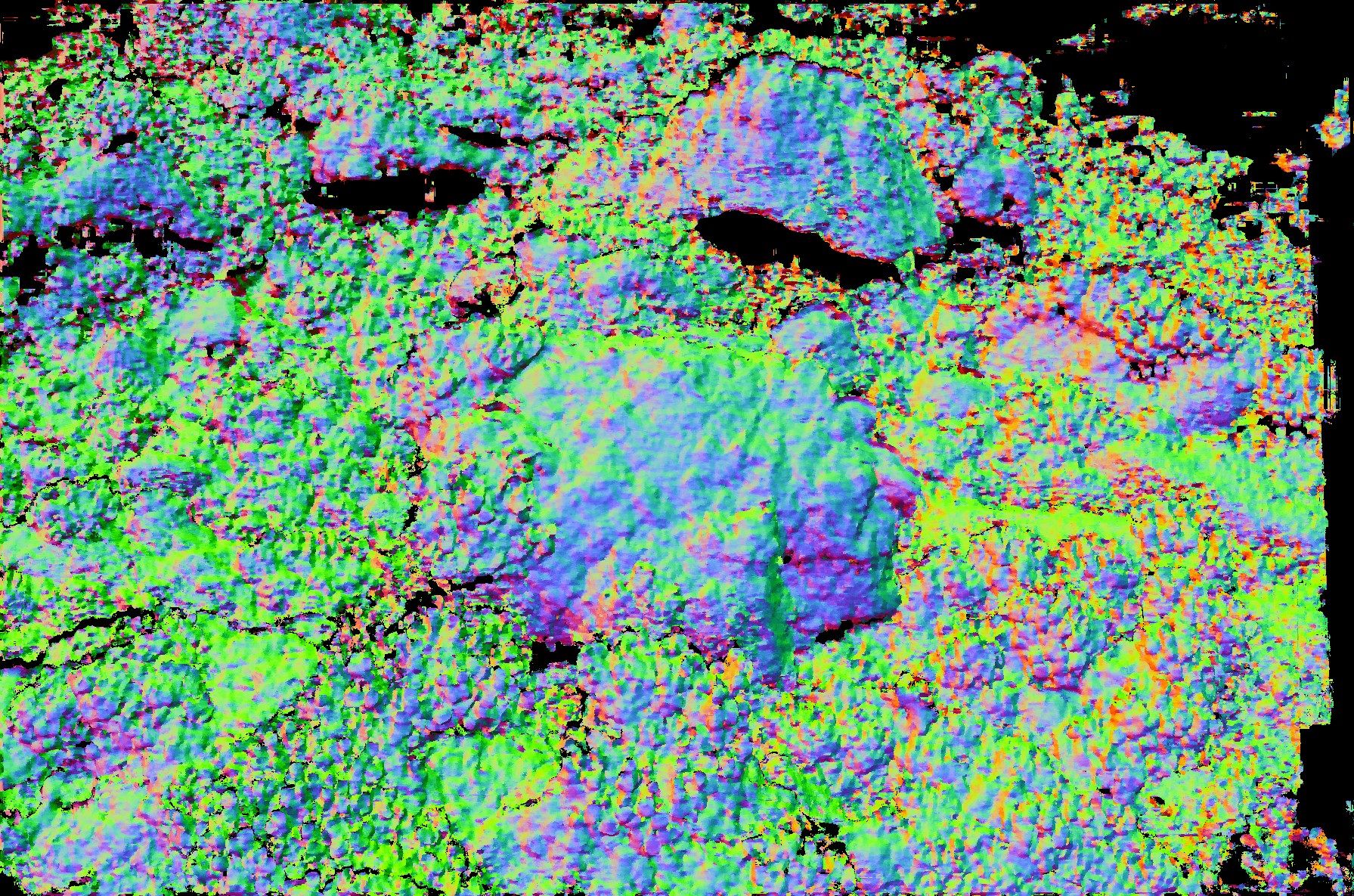}\\
    \small (i)
  \end{minipage}

  \caption{Visualization of input images (a) and their corresponding depth map (b) and normal map estimated with SfM (c). The input images are from SeaThru \citep{levy2023seathru}}
  \label{fig:pure_3d_depth_normal}
\end{figure}

\begin{figure}[t]
    \centering
    \includegraphics[width=\linewidth]{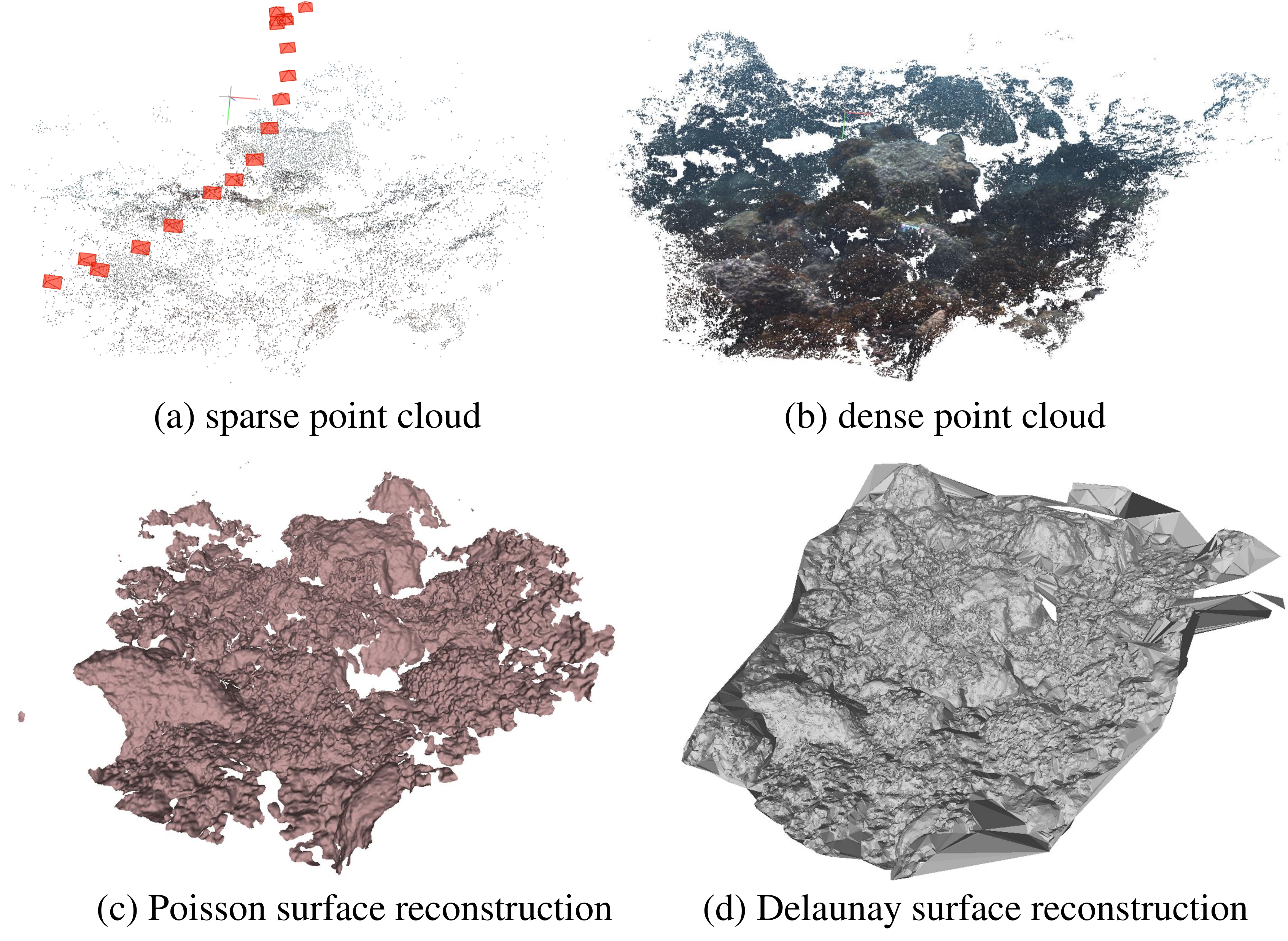}
    \caption{Snapshots of the sparse point cloud (a), dense point cloud (b), Poisson surface reconstruction (c) and Delaunay surface reconstruction (d) of an underwater scene}
    \label{fig:pure_3d_mesh}
\end{figure}

\paragraph{Photogrammetry} We employ COLMAP~\citep{schonberger2016structure} to reconstruct an underwater scene. The input images, along with their corresponding depth and normal maps, are presented in \autoref{fig:pure_3d_depth_normal}. Snapshots of the point clouds and the reconstructed 3D meshes are shown in \autoref{fig:pure_3d_mesh}. 

We observe that when the input images lack sufficient coverage across diverse camera viewpoints, the reconstructed scene exhibits significant missing regions. This issue is particularly evident in Poisson surface reconstruction, where the absence of viewpoints from different angles leads to incomplete and fragmented surfaces. The Delaunay-based reconstruction, while more structurally connected, also suffers from irregularities due to the limited input perspectives. These results highlight the importance of capturing a well-distributed set of images to ensure a more complete and accurate 3D reconstruction.

\begin{figure}[ht]
    \centering
    \includegraphics[width=\linewidth]{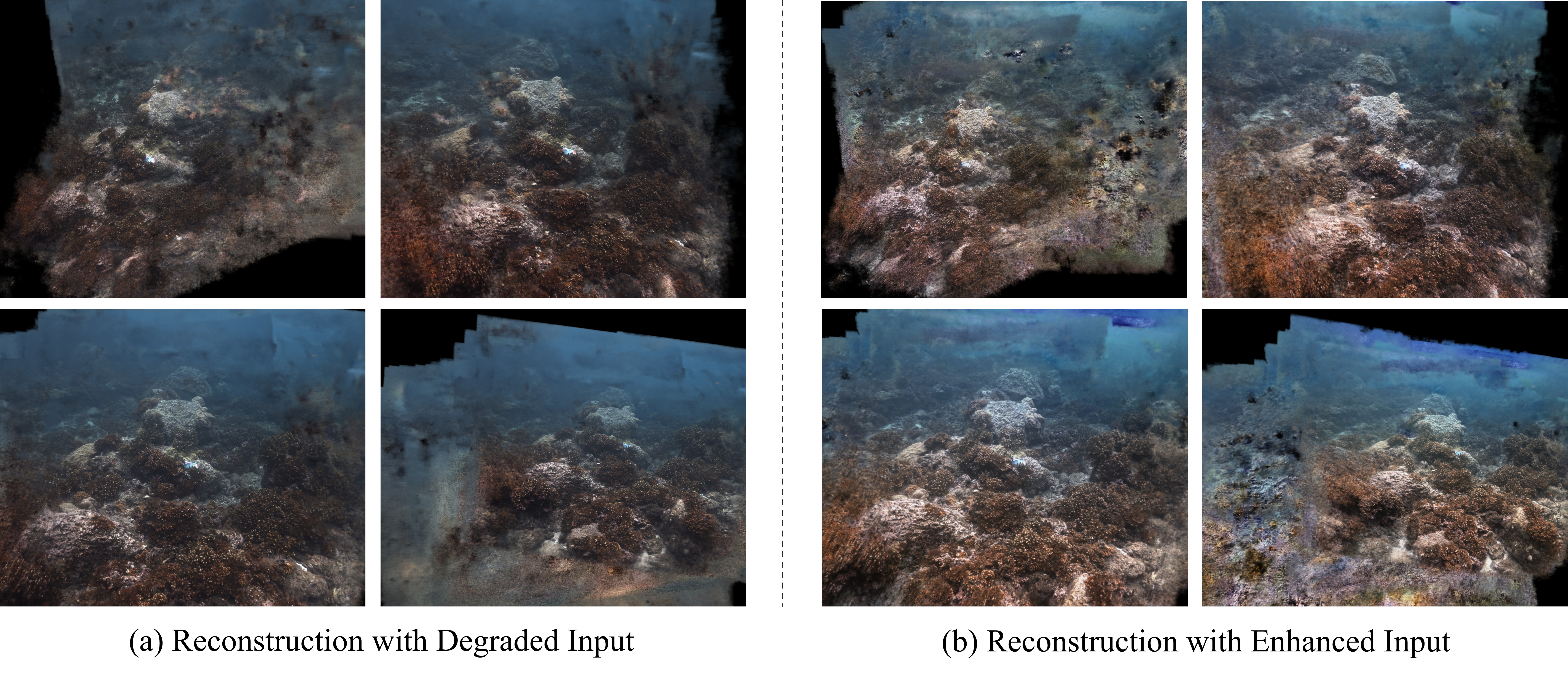}
    \caption{Comparison of reconstructed underwater scenes produced by InstantNGP~\citep{mueller2022instant} without applying any visual enhancement techniques (a) and with enhancement incorporated into the pipeline (b).}
    \label{fig:degradedRe_vs_EnRe}
\end{figure}

\paragraph{NeRF} Instant-NGP~\citep{mueller2022instant} is a NeRF variant optimized for real-time rendering. As shown in \autoref{fig:degradedRe_vs_EnRe}(a), when applied directly to degraded underwater imagery, it outperforms traditional photogrammetry by enabling high-quality, view-dependent novel view synthesis with smooth interpolation and fine detail preservation. However, noticeable floater artifacts appear around scene boundaries, typically caused by inaccuracies in the estimated depth and density fields, particularly in regions where input observations are sparse or inconsistent. Moreover, due to the turbid water medium and light absorption, the reconstructed scene suffers from color cast, diminished visibility of distant objects, and reduced overall brightness.

\begin{figure}[!t]
    \centering
    \includegraphics[width=0.98\linewidth]{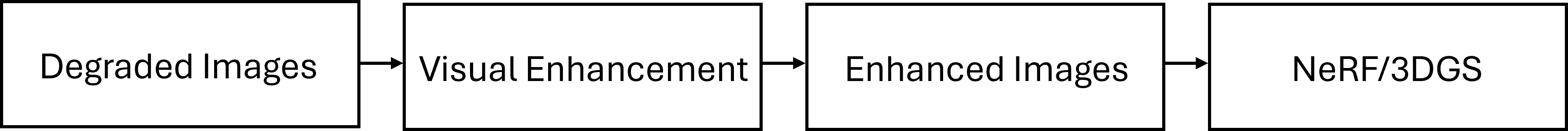}
    \caption{Illustration of two-stage Enhancement + 3D reconstruction pipeline.}
    \label{fig:two_stage_pip}
\end{figure}

\paragraph{Enhancement + 3D Reconstruction}
\label{ssec:2stage}
To mitigate the issues arising from quality degradation in underwater imagery, \citet{Malyugina:beam:2025} first apply image enhancement models to improve the quality of raw underwater inputs--such as removing marine snow--and subsequently use the enhanced images to reconstruct dense 3D scenes. The simplified diagram of this two-stage pipeline is shown in \autoref{fig:two_stage_pip}, which is although not end-to-end, but can effectively improve the visibility on the reconstruction.

\begin{figure}[!t]
    \centering
    \includegraphics[width=\linewidth]{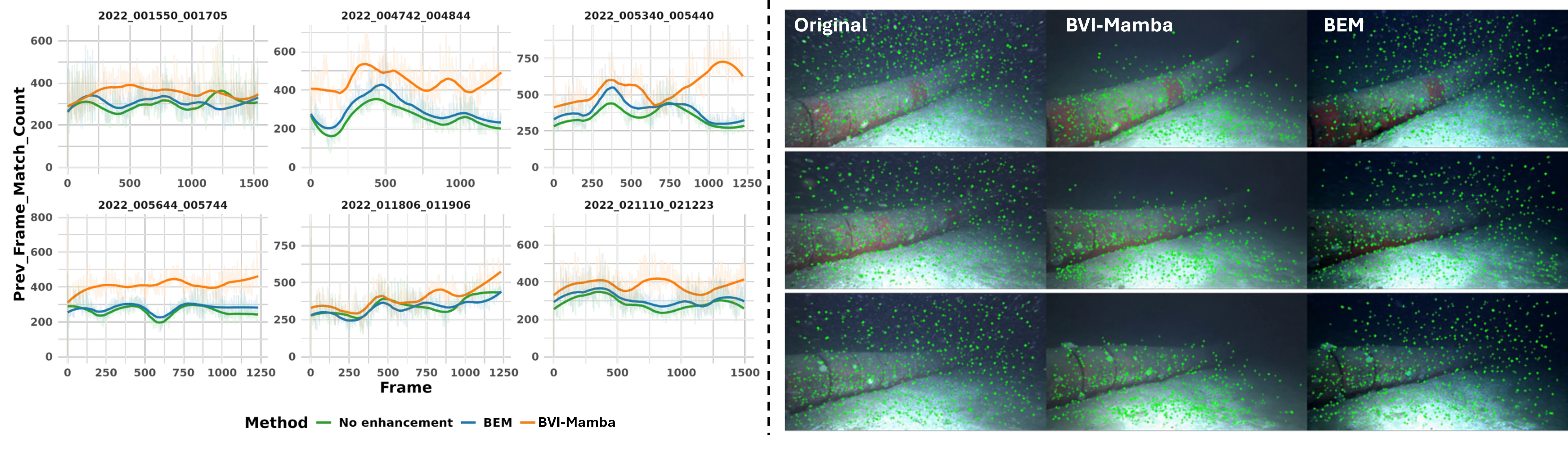}
    \caption{Comparison of feature matching in 3D reconstruction using raw images versus enhanced images produced by BVI-Mamba~\citep{huang2025bvi} and BEM~\citep{huang2026bayesian}. 
(left) Number of frame-to-frame feature matches (higher values indicate more informative features for SLAM); curves are smoothed for clarity. 
(right) Example frames with detected feature points overlaid in green. The images are from \citet{Malyugina:beam:2025}.}
    \label{fig:feature_points}
\end{figure}

{\color{revsienna}
Furthermore, \citet{Malyugina:beam:2025} observed that different enhancement methods can strongly influence both the number and the spatial distribution of features extracted during 3D reconstruction. Ideally, feature points should be concentrated around stable object regions, as illustrated in \autoref{fig:feature_points} (right). More detected features, however, do not automatically imply more geometrically reliable correspondences: strong enhancement can also create unstable gradients, hallucinated texture, or colour inconsistencies across views that fail later geometric verification. In practice, mild structure-preserving enhancement can improve matchability in low-contrast regions, whereas aggressive per-frame enhancement may hurt pose estimation or bundle adjustment even if the visual result appears sharper.}
{\color{revorange}
This observation is consistent with recent bridge studies that explicitly examine how enhancement and visualisation choices affect 3D outcomes. \citet{vrochidis2025underwater3d} report that underwater image enhancement choices measurably influence model quality, while \citet{vrochidis2025colormap} show that even post-reconstruction colour-map selection can change the interpretability of 3D structures. Together, these studies reinforce that enhancement and visualisation are not merely cosmetic add-ons: they influence correspondence quality, surface readability, and ultimately the usefulness of the reconstructed model.
}
% In the initial stage of our two-stage reconstruction process, we employ visual enhancement techniques to address color cast, haze, and other image quality degradations, using the algorithm described in \citep{huang2026bayesian} to enhance visibility and clarify features for 3D reconstruction. We utilize InstanceNGP for the reconstruction, similar to \autoref{ssec:pure3D}. While the original images are used for camera pose estimation to enable comparison with unenhanced pipelines, the enhanced images are employed to optimize network and representation parameters.

By comparing \autoref{fig:degradedRe_vs_EnRe} (a) with (b), we observe that the two-stage reconstruction pipeline can effectively reduce color cast and haze effects compared to pure 3D reconstruction approaches, and in this example there are fewer floater artifacts. However, because the enhancement method is applied to individual images, slight colour differences can still lead to image misalignment or local inaccuracies. This is why some reconstruction pipelines still prefer raw or only lightly corrected images for camera pose estimation, while reserving stronger enhancement for later dense reconstruction or visualisation stages.

Overall, the visual-enhancement-plus-3D-reconstruction paradigm is a viable solution that can address several challenges caused by image degradation. However, it introduces additional development and deployment overhead, and its benefit depends on whether enhancement preserves geometry-relevant cues instead of only producing perceptually stronger images. The generalization capability of enhancement models is often limited by the relatively small size and domain specificity of available training datasets, which means that their effectiveness may vary across different scenarios. This typically requires extra manual effort to select or train enhancement models tailored to each environment.
Looking ahead, this technical pipeline would benefit from addressing two key challenges: 1) developing end-to-end frameworks that jointly perform visual enhancement and 3D reconstruction, and 2) improving the adaptability of enhancement models, e.g., through scene-aware mechanisms or test-time optimization, to enhance generalization and avoid the need for multiple task- or scene-specific enhancement models.

{\color{revsienna}
The practical implication is that low-contrast scenes often benefit most from mild, structure-preserving correction, while pose-estimation-critical stages may still prefer raw or lightly corrected imagery together with explicit physical or refractive modeling. Stronger enhancement is usually safer later in offline dense reconstruction, novel-view rendering, or operator-facing visualization, once camera geometry has already been stabilized.
}

% \begin{figure}[H]
%     \centering
%     \includegraphics[width=\linewidth]{bnn_3d.png}
%     \caption{Input images before and after enhancement}
%     \label{fig:bnn_to_3dre}
% \end{figure}

% \begin{figure}[H]
%     \centering
%     \includegraphics[width=\linewidth]{snapshot_twostage.pdf}
%     \caption{Snapshots of the reconstructed underwater scene using InstanceNGP, with image enhancement using the method proposed by~\citet{huang2026bayesian}}
%     \label{fig:snapshot_twostage}
% \end{figure}

% ====================================================

% ====================================================

\section{Overall challenges and future work}
\label{sec:Overall challenges and future work}

\subsection{\texorpdfstring{{\color{revorange}Cross-cutting Data, Supervision, and Deployment Challenges}}{Cross-cutting Data, Supervision, and Deployment Challenges}}
\label{subsec:cross_cutting_challenges}
{\color{revorange}
Many remaining obstacles are shared by enhancement and reconstruction rather than belonging to one stage alone. The most persistent problem is still the lack of reliable supervision: paired clean underwater targets are scarce, synthetic data reproduce only part of real water behaviour, and domain shift across turbidity, illumination, depth, and site conditions remains severe. Unlike some other restoration settings, underwater data vary strongly with local water properties and dynamic particulate content, which is why models trained in one environment often degrade when transferred to another~\citep{Lin:BVI-RLV:2024,Li:WaterGAN:2017,9880475,Akkaynak:seathrue:2019,Yi:AtlantisGS:2025,wang2024uw}.
}

{\color{revorange}
These shared supervision issues become even more difficult when enhancement, video analysis, and 3D reconstruction must operate together. Multi-view reconstruction requires appearance correction that remains stable across viewpoints, while long-duration video demands temporal consistency rather than isolated frame improvement. Recent methods such as \citet{Malyugina:beam:2025,Liu:Spatiotemporal:2025,huang2025fromrestoration} show that progress increasingly depends on coupling restoration targets with scene dynamics, viewpoint consistency, and field robustness. At the deployment level, long videos, synchronised multi-camera capture, and high-resolution survey data are still expensive to acquire, which helps explain why UIE video methods remain relatively limited~\citep{Li:ZeroTIG:2025} and why underwater 3D reconstruction is still dominated by static-scene settings despite encouraging progress on dynamic reconstruction in clear media~\citep{yang2024deformable3dgs,YANG2025130262}.
}

{\color{revteal}
Reducing the sim-to-real gap will require more than simply enlarging synthetic datasets. Promising strategies include broader randomization of water parameters during simulation, synthetic-to-real curricula, pseudo-paired supervision, and more realistic generative data creation for turbidity, lighting flicker, and marine-snow patterns. WaterGAN-style simulation remains influential for this reason~\citep{Li:WaterGAN:2017}, but future datasets will likely need richer environmental randomization and more realistic dynamic clutter if models are to generalize reliably in field deployment.
}

{\color{revsienna}
The central difficulty is that controlled tank or synthetic captures still under-represent the field conditions that most disrupt deployment, including severe turbidity, non-uniform caustics, drifting particulate clouds, and camera-light interactions that vary along a survey trajectory. Bridging this gap is therefore as much about environmental realism and temporal variability as about the number of training samples.
}

\subsection{Foundation models for underwater imagery}

{\color{revorange}
Beyond these shared bottlenecks, foundation models are best viewed as a forward-looking tool direction rather than another repetition of the challenge discussion above.
}

Foundation models (FMs) are large-scale models trained on diverse datasets, typically through self-supervised learning, and can be adapted or fine-tuned for a wide range of downstream tasks. Their development has been driven by rapid advances in AI-oriented computational power and they appear to be particularly well-suited for domains rich in data but lacking ground-truth annotations, including underwater applications. However, to date, MarineInst \citep{zheng2024marineinst} is the only foundation model developed specifically for marine applications. It's built upon Segment Anything Model (SAM) \citep{Kirillov:SAM:2023}, trained on MarineInst20M, a large-scale dataset constructed from three main sources: (1) existing public marine and underwater datasets, (2) manually collected images from public and private datasets as well as YouTube videos, and (3) publicly available Internet images. MarineInst provides text-image matching (instance captioning) and instance segmentation capabilities. MarineInst shows strong potential as a foundation model. Its downstream tasks, however, perform well mainly on high-level computer vision applications, such as scene understanding and segmentation. Its usability for low-level tasks like enhancement remains uncertain, particularly for 3D reconstruction, as MarineInst was not trained to understand geometry.

Some underwater image enhancement and 3D reconstruction methods leverage foundation models (FMs) trained on natural images and text. These approaches use the output features and prior knowledge from such models, based on the assumption that FMs trained on large datasets have learned rich patterns and representations of visual and textual information, which can potentially generalize to underwater imagery. For example, \citet{Wang:Large:2025} employ SAM to separate foreground and background regions and apply color correction separately. DreamSea \citep{zhang2025infinite} exploits Depth Anything v2 \citep{Yang:depthanything:2024} and DINOv2 \citep{Oquab:DINOv2:2024} to extract depth and other feature information for 3D Gaussian Splatting (3DGS). Similarly, SWAGSplatting uses BLIPo3 \citep{chen2025blip3o} to identify and capture regions of interest. Many approaches also utilize CLIP (Contrastive Language-Image Pre-training) \citep{radford2021learning} to generate captions for downstream tasks such as object detection and scene understanding. This approach could potentially be extended to enhancement and 3D reconstruction, similar to those developed for clear-medium scenes \citep{Zhou_2024_CVPR, WEI2025104222}.

\subsection{\texorpdfstring{{\color{revgreen}Perspectives on Datasets, Modeling Tools, and Evaluation}}{Perspectives on Datasets, Modeling Tools, and Evaluation}}
\label{subsec:future_datasets_models_metrics}

{\color{revgreen}
Future progress will depend not only on better architectures but also on broader task coverage in data collection. Current benchmarks remain dominated by still images or short clips, whereas practical deployments in surveillance, fish monitoring, intervention support, and long-baseline reconstruction require longer videos, richer dynamic content, paired and unpaired video benchmarks, denser task annotations, and more comprehensive coverage across harbours, reefs, lakes, and deep-water domains. Bridging enhancement, video analysis, and 3D reconstruction will therefore require benchmark design that couples restoration targets with motion, semantics, and geometry.
}

{\color{revgreen}
In terms of modelling, graph-based learning is a promising complement to convolutional, transformer, Mamba, and diffusion backbones when labels are sparse or relational structure matters. Recent surveys by \citet{ponzi2025graph} and \citet{sadasivan2025systematic} highlight that graph neural networks are well suited to modelling interactions, non-Euclidean structure, and long-range dependencies with modest annotation budgets. For underwater vision, such priors could be valuable for fish schools, moving-object associations across frames, multi-view correspondence graphs, and cross-sensor fusion, complementing recent graph-based underwater motion analysis methods~\citep{kapoor2024principal,kapoor2025graph}.
}

{\color{revgreen}
Evaluation protocols also require broader reporting. A single ranking score is often insufficient because a method may improve one challenge while degrading another; recent discussions by \citet{pierard2025methodology,pierard2025optimal} likewise caution against relying on one summary statistic alone. Future benchmarks should therefore report restoration quality, temporal consistency, downstream utility, and cross-domain robustness side by side, rather than collapsing all performance into one leaderboard.
}

{\color{revsienna}
For integrated enhancement-and-reconstruction settings, this broader reporting should also include geometry- and trajectory-aware criteria such as Chamfer or Cloud-to-Cloud distance, ATE/RPE, and correspondence-stability indicators wherever possible. Without such multi-domain evaluation, it remains too easy for a method to look strong on image metrics while still degrading mapping or localization reliability.
}

{\color{revteal}
For engineering deployment, the same principle applies to efficiency reporting: latency, memory footprint, and platform suitability should be documented whenever a method is proposed for robotic operation. A model that performs well in offline post-processing may still be unusable for navigation or online inspection if it requires seconds per frame or high-end GPUs.
}

\subsection{Ethical Issues and Bias}

As underwater image enhancement and 3D reconstruction technologies become increasingly sophisticated, they raise important ethical considerations that demand careful attention from researchers, developers, and end users. A fundamental challenge lies in distinguishing legitimate enhancement from misleading manipulation. Aggressive enhancement or reconstruction methods may introduce artifacts, alter colors unnaturally, or generate content that misrepresents actual underwater scenes, risks that are particularly acute when techniques involve generative AI, such as GANs and diffusion models. In scientific applications including marine biology surveys, archaeological documentation, and environmental monitoring, such distortions can compromise research integrity and lead to incorrect conclusions about ecosystem health, species behavior, or site conditions.

These authenticity concerns are compounded by systematic biases in the datasets used to train enhancement algorithms. Most existing methods rely on image formation models trained on limited datasets that fail to represent the full diversity of underwater environments. Current benchmark datasets are predominantly captured in specific geographic regions, water types, and depth ranges, meaning that models trained on coral reef imagery from tropical waters often perform poorly in murky harbor environments, kelp forests, or high-latitude conditions. This geographic and environmental bias can produce enhancements that work well for some scenarios while generating unrealistic or misleading results for others.

Beyond technical limitations, important questions surrounding data ownership, consent, and privacy arise when imagery is collected in protected habitats, cultural heritage sites, or industrial contexts. High-resolution reconstructions of archaeological shipwrecks could inadvertently facilitate looting, while detailed seafloor mapping might reveal sensitive information about underwater infrastructure or ecologically vulnerable areas. Similarly, images captured in marine protected areas or private facilities raise questions about appropriate data sharing and access control.

To address these interconnected challenges, future work should prioritize ethical dataset governance and transparent methodology. This includes detailed documentation of data provenance, balanced representation of environmental diversity across training datasets, standardized disclosure of model limitations and enhancement boundaries, and clear protocols for handling sensitive imagery from protected or culturally significant sites. By proactively addressing these ethical dimensions, the underwater imaging community can ensure that these powerful technologies serve scientific understanding and environmental stewardship while minimizing potential harms.

%BVI-Coral https://zenodo.org/records/11093417

%NPS Submerged Resources Center (https://www.npssubmerged.com/photogrammetry)

% ====================================================
\section{Conclusions}
\label{sec:conclusion}

Underwater imaging is essential for scientific exploration, industrial applications, and environmental conservation, covering a broad range of fields including marine biology, archaeology, geological surveying, and infrastructure inspection. It helps manage resources, monitor ecosystems, and assess the condition of subsea infrastructure like pipelines and offshore platforms. Advancements in technology allow for high-quality images crucial for studying marine life, creating detailed 3D models of submerged archaeological sites, and ensuring the safe operation of industrial facilities under challenging visibility conditions. These efforts are crucial in tracking environmental changes and supporting resource exploration by providing precise mappings of the seafloor, thus minimizing risks and operational costs.

The review begins by addressing the unique challenges of underwater environments and outlines its scope, including discussions on image enhancement and 3D reconstruction pathways. We describe the physics of underwater light propagation and image formation, setting the stage for an exploration of various visual enhancement methods, both traditional and data-driven, and their applicability to underwater scenes. The review further elaborates on different 3D reconstruction techniques tailored for underwater use, including photogrammetry, NeRF, and 3D Gaussian Splatting, discussing their motivations, methodologies, and specific challenges. It concludes with a benchmarking discussion on these methods, emphasizing the need for enhancement integration to achieve accurate underwater 3D reconstructions.

Future research will likely delve deeper into physically correct light-transport modeling, large-scale real-time systems, domain adaptation to address data scarcity, and multi-sensor integration, ultimately broadening underwater exploration, scientific study, and industrial deployment.

\newpage

\bibliography{jrnl}

@inproceedings{schonberger2016structure,
  title={Structure-from-motion revisited},
  author={Schonberger, Johannes L and Frahm, Jan-Michael},
  booktitle={the IEEE/CVF Conference on Computer Vision and Pattern Recognition (CVPR)},
  pages={4104--4113},
  year={2016}
}

@inproceedings{akkaynak2018revised,
  title={A revised underwater image formation model},
  author={Akkaynak, Derya and Treibitz, Tali},
  booktitle={the IEEE/CVF Conference on Computer Vision and Pattern Recognition (CVPR)},
  pages={6723--6732},
  year={2018}
}

@inproceedings{wu20244d,
  title={4{D} gaussian splatting for real-time dynamic scene rendering},
  author={Wu, Guanjun and Yi, Taoran and Fang, Jiemin and Xie, Lingxi and Zhang, Xiaopeng and Wei, Wei and Liu, Wenyu and Tian, Qi and Wang, Xinggang},
  booktitle={Proceedings of the IEEE/CVF conference on computer vision and pattern recognition},
  pages={20310--20320},
  year={2024}
}

@ARTICLE{Liang:GUDCP:2022,
  author={Liang, Zheng and Ding, Xueyan and Wang, Yafei and Yan, Xiaohong and Fu, Xianping},
  journal={IEEE Transactions on Circuits and Systems for Video Technology}, 
  title={{GUDCP: Generalization} of Underwater Dark Channel Prior for Underwater Image Restoration}, 
  year={2022},
  volume={32},
  number={7},
  pages={4879-4884},
  doi={10.1109/TCSVT.2021.3114230}}

@article{Song:Enhancement:2020,
  title = {Enhancement of {{Underwater Images With Statistical Model}} of {{Background Light}} and {{Optimization}} of {{Transmission Map}}},
  author = {Song, Wei and Wang, Yan and Huang, Dongmei and Liotta, Antonio and Perra, Cristian},
  year = {2020},
  month = mar,
  journal = {IEEE Transactions on Broadcasting},
  volume = {66},
  number = {1},
  pages = {153--169},
  issn = {1557-9611},
  doi = {10.1109/TBC.2019.2960942},
  urldate = {2025-02-19}
}

@article{wang2015patch,
  title={A patch-structure representation method for quality assessment of contrast changed images},
  author={Wang, Shiqi and Ma, Kede and Yeganeh, Hojatollah and Wang, Zhou and Lin, Weisi},
  journal={IEEE Signal Processing Letters},
  volume={22},
  number={12},
  pages={2387--2390},
  year={2015},
  publisher={IEEE}
}

@article{wang2002universal,
  title={A universal image quality index},
  author={Wang, Zhou and Bovik, Alan C},
  journal={IEEE signal processing letters},
  volume={9},
  number={3},
  pages={81--84},
  year={2002},
  publisher={IEEE}
}

@inproceedings{mildenhall2020nerf,
  title={NeRF: Representing Scenes as Neural Radiance Fields for View Synthesis},
  author={Mildenhall, Ben and Srinivasan, Pratul P. and Tancik, Matthew and Barron, Jonathan T. and Ramamoorthi, Ravi and Ng, Ren},
  booktitle={Proceedings of the European Conference on Computer Vision (ECCV)},
  pages={405--421},
  year={2020}
}

@inproceedings{lin2024gaussian,
  title={{Gaussian-flow: 4D} reconstruction with dynamic 3d gaussian particle},
  author={Lin, Youtian and Dai, Zuozhuo and Zhu, Siyu and Yao, Yao},
  booktitle={the IEEE/CVF Conference on Computer Vision and Pattern Recognition (CVPR)},
  pages={21136--21145},
  year={2024}
}

@inproceedings{yang2024deformable3dgs,
    title={Deformable 3D Gaussians for High-Fidelity Monocular Dynamic Scene Reconstruction},
    author={Yang, Ziyi and Gao, Xinyu and Zhou, Wen and Jiao, Shaohui and Zhang, Yuqing and Jin, Xiaogang},
    booktitle={Proceedings of the IEEE/CVF Conference on Computer Vision and Pattern Recognition},
    year={2024}
}

@INPROCEEDINGS{10656460,
  author={Tang, Yunkai and Zhu, Chengxuan and Wan, Renjie and Xu, Chao and Shi, Boxin},
  booktitle={2024 IEEE/CVF Conference on Computer Vision and Pattern Recognition (CVPR)}, 
  title={Neural Underwater Scene Representation}, 
  year={2024},
  volume={},
  number={},
  pages={11780-11789},
  doi={10.1109/CVPR52733.2024.01119}}

@INPROCEEDINGS{Sethuraman:WaterNeRF:2023,
  author={Sethuraman, Advaith Venkatramanan and Ramanagopal, Manikandasriram Srinivasan and Skinner, Katherine A.},
  booktitle={OCEANS 2023 - MTS/IEEE U.S. Gulf Coast}, 
  title={{WaterNeRF: Neural} Radiance Fields for Underwater Scenes}, 
  year={2023},
  volume={},
  number={},
  pages={1-7},
  doi={10.23919/OCEANS52994.2023.10336972}}

@inproceedings{kulhanek2024wildgaussians,
author = {Kulhanek, Jonas and Peng, Songyou and Kukelova, Zuzana and Pollefeys, Marc and Sattler, Torsten},
title = {{WildGaussians: 3D} Gaussian splatting in the wild},
year = {2025},
isbn = {9798331314385},
publisher = {Curran Associates Inc.},
address = {Red Hook, NY, USA},
booktitle = {Proceedings of the 38th International Conference on Neural Information Processing Systems},
articleno = {670},
numpages = {18},
location = {Vancouver, BC, Canada},
series = {NIPS '24}
}

@article{huang2025fromrestoration,
  title     = {From Restoration to Reconstruction: Rethinking 3D Gaussian Splatting for Underwater Scenes},
  author    = {Huang, Guoxi and Wang, Haoran and Qi, Zipeng and Lu, Wenjun and Bull, David and Anantrasirichai, Nantheera},
  journal   = {arXiv preprint arXiv:2509.17789},
  year      = {2025},
}

@ARTICLE{Fan:waater:2025,
AUTHOR={Fan, Xinnan  and Wang, Xiaotian  and Ni, Haonan  and Xin, Yuanxue  and Shi, Pengfei },
TITLE={Water-Adapted 3D Gaussian Splatting for precise underwater scene reconstruction},
JOURNAL={Frontiers in Marine Science},    
VOLUME={Volume 12 - 2025},
YEAR={2025},
DOI={10.3389/fmars.2025.1573612}}

@article{yuan2025threeduir,
  title     = {{3D-UIR: 3D} Gaussian for Underwater 3D Scene Reconstruction via Physics-Based Appearance-Medium Decoupling},
  author    = {Yuan, Jieyu and Li, Yujun and Zhang, Yuanlin and Guo, Chunle and Tang, Xiongxin and Wang, Ruixing and Li, Chongyi},
  journal   = {arXiv preprint arXiv:2505.21238},
  year      = {2025},
}

@INPROCEEDINGS{Qiao:RestoreGS:2025,
  author={Qiao, Yuanjian and Shao, Mingwen and Meng, Lingzhuang and Xu, Kai},
  booktitle={2025 IEEE/CVF Conference on Computer Vision and Pattern Recognition (CVPR)}, 
  title={{RestorGS: Depth}-aware Gaussian Splatting for Efficient 3D Scene Restoration}, 
  year={2025},
  volume={},
  number={},
  pages={11177-11186},
  doi={10.1109/CVPR52734.2025.01044}}

@inproceedings{Ancuti:Enhancing:2012,
  title = {Enhancing Underwater Images and Videos by Fusion},
  booktitle = {2012 {{IEEE Conference}} on {{Computer Vision}} and {{Pattern Recognition}}},
  author = {Ancuti, Cosmin and Ancuti, Codruta Orniana and Haber, Tom and Bekaert, Philippe},
  year = {2012},
  month = jun,
  pages = {81--88},
  issn = {1063-6919},
  doi = {10.1109/CVPR.2012.6247661},
  urldate = {2025-02-26}
}

@inproceedings{zhang2024pixel,
  title={Pixel-gs: Density control with pixel-aware gradient for 3d gaussian splatting},
  author={Zhang, Zheng and Hu, Wenbo and Lao, Yixing and He, Tong and Zhao, Hengshuang},
  booktitle={European Conference on Computer Vision},
  pages={326--342},
  year={2024},
  organization={Springer}
}

@inproceedings{levy2023seathru,
  title={SeaThru-NeRF: Neural radiance fields in scattering media},
  author={Levy, Deborah and Peleg, Amit and Pearl, Naama and Rosenbaum, Dan and Akkaynak, Derya and Korman, Simon and Treibitz, Tali},
  booktitle={the IEEE/CVF Conference on Computer Vision and Pattern Recognition (CVPR)},
  pages={56--65},
  year={2023}
}

@article{CHEN2024104025,
title = {SP-SeaNeRF: Underwater Neural Radiance Fields with strong scattering perception},
journal = {Computers \& Graphics},
volume = {123},
pages = {104025},
year = {2024},
doi = {https://doi.org/10.1016/j.cag.2024.104025},
author = {Lifang Chen and Yuchen Xiong and Yanjie Zhang and Ruiyin Yu and Lian Fang and Defeng Liu},
}

@article{Lu:Depth:2017,
  title = {Depth {{Map Reconstruction}} for {{Underwater Kinect Camera Using Inpainting}} and {{Local Image Mode Filtering}}},
  author = {Lu, Huimin and Zhang, Yin and Li, Yujie and Zhou, Quan and Tadoh, Ryunosuke and Uemura, Tomoki and Kim, Hyoungseop and Serikawa, Seiichi},
  year = {2017},
  journal = {IEEE Access},
  volume = {5},
  pages = {7115--7122},
  issn = {2169-3536},
  doi = {10.1109/ACCESS.2017.2690455},
  urldate = {2025-03-05}
}

@inproceedings{park2021nerfies,
  title={Nerfies: Deformable neural radiance fields},
  author={Park, Keunhong and Sinha, Utkarsh and Barron, Jonathan T and Bouaziz, Sofien and Goldman, Dan B and Seitz, Steven M and Martin-Brualla, Ricardo},
  booktitle={International Conference on Computer Vision (ICCV)},
  pages={5865--5874},
  year={2021}
}

@article{he2010single,
  title={Single image haze removal using dark channel prior},
  author={He, Kaiming and Sun, Jian and Tang, Xiaoou},
  journal={IEEE transactions on pattern analysis and machine intelligence},
  volume={33},
  number={12},
  pages={2341--2353},
  year={2010},
  publisher={IEEE}
}

@inproceedings{berman2017diving,
  title={Diving into haze-lines: Color restoration of underwater images},
  author={Berman, Dana and Treibitz, Tali and Avidan, Shai},
  booktitle={Proc. british machine vision conference (BMVC)},
  volume={1},
  pages={2},
  year={2017}
}

@article{lu2015contrast,
  title={Contrast enhancement for images in turbid water},
  author={Lu, Huimin and Li, Yujie and Zhang, Lifeng and Serikawa, Seiichi},
  journal={JOSA A},
  volume={32},
  number={5},
  pages={886--893},
  year={2015},
  publisher={Optica Publishing Group}
}

@inproceedings{schechner2004clear,
  title={Clear underwater vision},
  author={Schechner, Yoav Y and Karpel, Nir},
  booktitle={Proceedings of the 2004 IEEE Computer Society Conference on Computer Vision and Pattern Recognition, 2004. CVPR 2004.},
  volume={1},
  pages={I--I},
  year={2004},
  organization={IEEE}
}

@inproceedings{Narasimhan:Chromatic:2000,
  title = {Chromatic Framework for Vision in Bad Weather},
  booktitle = {Proceedings {{IEEE Conference}} on {{Computer Vision}} and {{Pattern Recognition}}. {{CVPR}} 2000 ({{Cat}}. {{No}}.{{PR00662}})},
  author = {Narasimhan, S.G. and Nayar, S.K.},
  year = {2000},
  month = jun,
  volume = {1},
  pages = {598-605 vol.1},
  issn = {1063-6919},
  doi = {10.1109/CVPR.2000.855874},
  urldate = {2025-02-10}
}

@article{Fattal:Single:2008,
  title = {Single Image Dehazing},
  author = {Fattal, Raanan},
  year = {2008},
  month = aug,
  journal = {ACM Transactions on Graphics},
  volume = {27},
  number = {3},
  pages = {1--9},
  issn = {0730-0301, 1557-7368},
  doi = {10.1145/1360612.1360671},
  urldate = {2025-02-10},
  langid = {english}
}

@inproceedings{Tan:Visibility:2008,
  title = {Visibility in Bad Weather from a Single Image},
  booktitle = {2008 {{IEEE Conference}} on {{Computer Vision}} and {{Pattern Recognition}}},
  author = {Tan, Robby T.},
  year = {2008},
  month = jun,
  pages = {1--8},
  issn = {1063-6919},
  doi = {10.1109/CVPR.2008.4587643},
  urldate = {2025-02-10}
}

@inproceedings{ramazzina2023scatternerf,
  title={Scatternerf: Seeing through fog with physically-based inverse neural rendering},
  author={Ramazzina, Andrea and Bijelic, Mario and Walz, Stefanie and Sanvito, Alessandro and Scheuble, Dominik and Heide, Felix},
  booktitle={International Conference on Computer Vision (ICCV)},
  pages={17957--17968},
  year={2023}
}

@inproceedings{yu2024mip,
  title={Mip-splatting: Alias-free 3d gaussian splatting},
  author={Yu, Zehao and Chen, Anpei and Huang, Binbin and Sattler, Torsten and Geiger, Andreas},
  booktitle={the IEEE/CVF Conference on Computer Vision and Pattern Recognition (CVPR)},
  pages={19447--19456},
  year={2024}
}

@inproceedings{yan2024multi,
  title={Multi-scale 3d gaussian splatting for anti-aliased rendering},
  author={Yan, Zhiwen and Low, Weng Fei and Chen, Yu and Lee, Gim Hee},
  booktitle={the IEEE/CVF Conference on Computer Vision and Pattern Recognition (CVPR)},
  pages={20923--20931},
  year={2024}
}

@article{chen2024deblur,
  title={Deblur-GS: 3D Gaussian Splatting from Camera Motion Blurred Images},
  author={Chen, Wenbo and Liu, Ligang},
  journal={Proceedings of the ACM on Computer Graphics and Interactive Techniques},
  volume={7},
  number={1},
  pages={1--15},
  year={2024},
  publisher={ACM New York, NY, USA}
}

@inproceedings{lee2024deblurring,
  title={Deblurring 3d gaussian splatting},
  author={Lee, Byeonghyeon and Lee, Howoong and Sun, Xiangyu and Ali, Usman and Park, Eunbyung},
  booktitle={European Conference on Computer Vision},
  pages={127--143},
  year={2024},
  organization={Springer}
}

@Article{Prado:3D:2020,
AUTHOR = {Prado, E. and Rodríguez-Basalo, Augusto and Cobo, Adolfo and Ríos, Pilar and Sánchez, Francisco},
TITLE = {3D Fine-scale Terrain Variables from Underwater Photogrammetry},
JOURNAL = {Remote Sensing},
YEAR = {2020},
}

@Article{Nocerino:coral:2020,
AUTHOR = {Nocerino, E. and Menna, Fabio and Gruen, Armin and Troyer, Matthias and Capra, Alessandro and Castagnetti, Cristina and Rossi, Paolo and Brooks, Andrew J. and Schmitt, Russell J. and Holbrook, Sally J.},
TITLE = {Coral Reef Monitoring by Scuba Divers Using Underwater Photogrammetry and Geodetic Surveying},
JOURNAL = {Remote Sensing},
YEAR = {2020},
}

@inproceedings{huang2025bvi,
  title={BVI-Mamba: video enhancement using a visual state-space model for low-light and underwater environments},
  author={Huang, Guoxi and Lin, Ruirui and Li, Yini and Bull, David and Anantrasirichai, Nantheera},
  booktitle={Machine Learning from Challenging Data 2025},
  volume={13460},
  pages={74--81},
  year={2025},
  organization={SPIE}
}

@article{ruan2024vm,
  title={{VM-UNet: Vision} Mamba UNet for Medical Image Segmentation},
  author={Ruan, Jiacheng and Li, Jincheng and Xiang, Suncheng},
  journal={arXiv preprint arXiv:2402.02491},
  year={2024}
}

@article{Storlazzi:vslam:2016,
title = {End of the chain? Rugosity and fine-scale bathymetry from existing underwater digital imagery using structure-from-motion (SfM) technology},
journal = {Coral Reefs},
year = {2016},
author = {Storlazzi, C. and Dartnell, P. and Hatcher, G.A. and Gibbs, A.E.},
}

@ARTICLE{Li:Underwater:2020,
  author={Li, Chongyi and Guo, Chunle and Ren, Wenqi and Cong, Runmin and Hou, Junhui and Kwong, Sam and Tao, Dacheng},
  journal={IEEE Transactions on Image Processing}, 
  title={An Underwater Image Enhancement Benchmark Dataset and Beyond}, 
  year={2020},
  volume={29},
  number={},
  pages={4376-4389}}

@article{Galdran:Automatic:2015,
  title = {Automatic {{Red-Channel}} Underwater Image Restoration},
  author = {Galdran, Adrian and Pardo, David and Pic{\'o}n, Artzai and {Alvarez-Gila}, Aitor},
  year = {2015},
  month = jan,
  journal = {Journal of Visual Communication and Image Representation},
  volume = {26},
  pages = {132--145},
  issn = {10473203},
  doi = {10.1016/j.jvcir.2014.11.006},
  urldate = {2025-02-19},
  langid = {english}
}

@inproceedings{Carlevaris-Bianco:Initial:2010,
  title = {Initial Results in Underwater Single Image Dehazing},
  booktitle = {{{OCEANS}} 2010 {{MTS}}/{{IEEE SEATTLE}}},
  author = {{Carlevaris-Bianco}, Nicholas and Mohan, Anush and Eustice, Ryan M.},
  year = {2010},
  month = sep,
  pages = {1--8},
  issn = {0197-7385},
  doi = {10.1109/OCEANS.2010.5664428},
  urldate = {2025-02-19}
}

@inproceedings{Pizer:Contrastlimited:1990,
  title = {Contrast-Limited Adaptive Histogram Equalization: Speed and Effectiveness},
  shorttitle = {Contrast-Limited Adaptive Histogram Equalization},
  booktitle = {[1990] {{Proceedings}} of the {{First Conference}} on {{Visualization}} in {{Biomedical Computing}}},
  author = {Pizer, S.M. and Johnston, R.E. and Ericksen, J.P. and Yankaskas, B.C. and Muller, K.E.},
  year = {1990},
  month = may,
  pages = {337--345},
  doi = {10.1109/VBC.1990.109340},
  urldate = {2025-02-17}
}

@article{Wang:UIERL:2024,
  title = {{{UIERL}}: {{Internal-External Representation Learning Network}} for {{Underwater Image Enhancement}}},
  shorttitle = {{{UIERL}}},
  author = {Wang, Zhengyong and Shen, Liquan and Yu, Yihan and Hui, Yuan},
  year = {2024},
  journal = {IEEE Transactions on Multimedia},
  volume = {26},
  pages = {9252--9267},
  issn = {1941-0077},
  doi = {10.1109/TMM.2024.3387760},
  urldate = {2025-02-18}
}

@inproceedings{Guo:Underwater:2024,
  title = {Underwater {{Image Restoration}} via {{Polymorphic Large Kernel CNNs}}},
  author = {Guo, Xiaojiao and Dong, Yihang and Chen, Xuhang and Chen, Weiwen and Li, Zimeng and Zheng, FuChen and Pun, Chi-Man},
  booktitle = {2025 IEEE International Conference on Acoustics, Speech and Signal Processing (ICASSP)},
  year = {2025},
  pages = {9731--9735},
  doi = {10.1109/ICASSP49660.2025.10890803},
  url = {https://ieeexplore.ieee.org/document/10890803/}
}

@ARTICLE{Islam:Fast:2020,
  author={Islam, Md Jahidul and Xia, Youya and Sattar, Junaed},
  journal={IEEE Robotics and Automation Letters}, 
  title={Fast Underwater Image Enhancement for Improved Visual Perception}, 
  year={2020},
  volume={5},
  number={2},
  pages={3227-3234},
  doi={10.1109/LRA.2020.2974710}}

@inproceedings{marques2020l2uwe,
  title={L2uwe: A framework for the efficient enhancement of low-light underwater images using local contrast and multi-scale fusion},
  author={Marques, Tunai Porto and Albu, Alexandra Branzan},
  booktitle={Proceedings of the IEEE/CVF conference on computer vision and pattern recognition workshops},
  pages={538--539},
  year={2020}
}

@article{berman2020underwater,
title={Underwater Single Image Color Restoration Using Haze-Lines and a New Quantitative Dataset},
author={Berman, Dana and Levy, Deborah and Avidan, Shai and Treibitz, Tali},
journal={IEEE Transactions on Pattern Analysis and Machine Intelligence},
year={2020}
}

@ARTICLE{Zhou:Underwater:2023,
  author={Zhou, Jingchun and Pang, Lei and Zhang, Dehuan and Zhang, Weishi},
  journal={IEEE Journal of Oceanic Engineering}, 
  title={Underwater Image Enhancement Method via Multi-Interval Subhistogram Perspective Equalization}, 
  year={2023},
  volume={48},
  number={2},
  pages={474-488},
  doi={10.1109/JOE.2022.3223733}}

@ARTICLE{Hou:Benchmarking:2020,
  author={Hou, Guojia and Zhao, Xin and Pan, Zhenkuan and Yang, Huan and Tan, Lu and Li, Jingming},
  journal={IEEE Access}, 
  title={Benchmarking Underwater Image Enhancement and Restoration, and Beyond}, 
  year={2020},
  volume={8},
  number={},
  pages={122078-122091},
  doi={10.1109/ACCESS.2020.3006359}}

@inproceedings{zhang2025infinite,
title={Infinite Leagues Under the Sea: Realistic 3D Underwater Terrain Generation Augmented by Visual Foundation Models},
author={Tianyi Zhang and Weiming Zhi and Matthew Johnson-Roberson},
booktitle={ICLR 2025 Workshop on Foundation Models in the Wild},
year={2025}
}

@inproceedings{zheng2024marineinst,
  title     = {{MarineInst: A} Foundation Model for Marine Image Analysis with Instance Visual Description},
  author    = {Zheng, Zhaoyang and Chen, Yuheng and Zeng, Han and Vu, Thanh-An and Hua, Binh-Son and Yeung, Shing-Yiu Kenneth},
  booktitle = {Proceedings of the European Conference on Computer Vision (ECCV)},
  year      = {2024}
}

@InProceedings{Kirillov:SAM:2023,
    author    = {Kirillov, Alexander and Mintun, Eric and Ravi, Nikhila and Mao, Hanzi and Rolland, Chloe and Gustafson, Laura and Xiao, Tete and Whitehead, Spencer and Berg, Alexander C. and Lo, Wan-Yen and Dollar, Piotr and Girshick, Ross},
    title     = {Segment Anything},
    booktitle = {Proceedings of the IEEE/CVF International Conference on Computer Vision (ICCV)},
    month     = {October},
    year      = {2023},
    pages     = {4015-4026}
}

@INPROCEEDINGS{Wang:Uformer:2022,
  author={Wang, Zhendong and Cun, Xiaodong and Bao, Jianmin and Zhou, Wengang and Liu, Jianzhuang and Li, Houqiang},
  booktitle={2022 IEEE/CVF Conference on Computer Vision and Pattern Recognition (CVPR)}, 
  title={Uformer: A General U-Shaped Transformer for Image Restoration}, 
  year={2022},
  volume={},
  number={},
  pages={17662-17672},}

@article{Oquab:DINOv2:2024,
title={{DINOv2: Learning} Robust Visual Features without Supervision},
author={Maxime Oquab and Timoth{\'e}e Darcet and Th{\'e}o Moutakanni and Huy V. Vo and Marc Szafraniec and Vasil Khalidov and Pierre Fernandez and Daniel HAZIZA and Francisco Massa and Alaaeldin El-Nouby and Mido Assran and Nicolas Ballas and Wojciech Galuba and Russell Howes and Po-Yao Huang and Shang-Wen Li and Ishan Misra and Michael Rabbat and Vasu Sharma and Gabriel Synnaeve and Hu Xu and Herve Jegou and Julien Mairal and Patrick Labatut and Armand Joulin and Piotr Bojanowski},
journal={Transactions on Machine Learning Research},
issn={2835-8856},
year={2024},
url={https://openreview.net/forum?id=a68SUt6zFt},
note={}
}

@Article{kerbl:3Dgaussians:2023,
      author       = {Kerbl, Bernhard and Kopanas, Georgios and Leimk{\"u}hler, Thomas and Drettakis, George},
      title        = {{3D} Gaussian Splatting for Real-Time Radiance Field Rendering},
      journal      = {ACM Transactions on Graphics},
      number       = {4},
      volume       = {42},
      month        = {July},
      year         = {2023},
      url          = {https://repo-sam.inria.fr/fungraph/3d-gaussian-splatting/}
}

@INPROCEEDINGS{Barron:Mip-NeRF360:2022,
  author={Barron, Jonathan T. and Mildenhall, Ben and Verbin, Dor and Srinivasan, Pratul P. and Hedman, Peter},
  booktitle={2022 IEEE/CVF Conference on Computer Vision and Pattern Recognition (CVPR)}, 
  title={{Mip-NeRF 360: Unbounded} Anti-Aliased Neural Radiance Fields}, 
  year={2022},
  volume={},
  number={},
  pages={5460-5469},
  doi={10.1109/CVPR52688.2022.00539}}

@inproceedings{Yang:depthanything:2024,
      title={Depth Anything: Unleashing the Power of Large-Scale Unlabeled Data}, 
      author={Yang, Lihe and Kang, Bingyi and Huang, Zilong and Xu, Xiaogang and Feng, Jiashi and Zhao, Hengshuang},
      booktitle={IEEE/CVF Conference on Computer Vision and Pattern Recognition (CVPR)},
      year={2024}
}

@article{chen2025blip3o,
  title={{BLIP3-o}: A Family of Fully Open Unified Multimodal Models--Architecture, Training and Dataset},
  author={Chen, Jiuhai and Xu, Zhiyang and Pan, Xichen and Hu, Yushi and Qin, Can and Goldstein, Tom and Huang, Lifu and Zhou, Tianyi and Xie, Saining and Savarese, Silvio and Xue, Le and Xiong, Caiming and Xu, Ran},
  journal={arXiv preprint arXiv:2505.09568},
  year={2025}
}

@article{Anwer:Underwater:2017,
  title = {Underwater 3-{{D Scene Reconstruction Using Kinect}} v2 {{Based}} on {{Physical Models}} for {{Refraction}} and {{Time}} of {{Flight Correction}}},
  author = {Anwer, Atif and Azhar Ali, Syed Saad and Khan, Amjad and M{\'e}riaudeau, Fabrice},
  year = {2017},
  journal = {IEEE Access},
  volume = {5},
  pages = {15960--15970},
  issn = {2169-3536},
  doi = {10.1109/ACCESS.2017.2733003},
  urldate = {2025-03-05}
}

@article{Pizer:Adaptive:1987,
  title = {Adaptive {{Histogram Equalization}} and {{Its Variations}}},
  author = {Pizer, Stephen M and Amburn, E Philip and Austin, John D and Cromartie, Robert and Geselowitz, Ari and Greer, Trey and Zuiderveld, Arel},
  year = {1987},
  langid = {english}
}

@inproceedings{Liu:Underwater:2016,
  title = {Underwater Image Restoration Based on Contrast Enhancement},
  booktitle = {2016 {{IEEE International Conference}} on {{Digital Signal Processing}} ({{DSP}})},
  author = {Liu, Hui and Chau, Lap-Pui},
  year = {2016},
  month = oct,
  pages = {584--588},
  issn = {2165-3577},
  doi = {10.1109/ICDSP.2016.7868625},
  urldate = {2025-02-17}
}

@INPROCEEDINGS{Akkaynak:seathrue:2019,  
author={D. {Akkaynak} and T. {Treibitz}},  booktitle={IEEE/CVF Conference on Computer Vision and Pattern Recognition (CVPR)},  title={{Sea-Thru: A} Method for Removing Water From Underwater Images},   year={2019},  volume={},  number={},  pages={1682-1691},}

@InProceedings{wang2024uw,
  title={{UW-GS: Distractor}-Aware 3D Gaussian Splatting for Enhanced Underwater Scene Reconstruction},
  author={Wang, Haoran and Anantrasirichai, Nantheera and Zhang, Fan and Bull, David},
booktitle = {Proceedings of the IEEE/CVF Winter Conference on Applications of Computer Vision (WACV)},
year= {2025},
}

@inproceedings{Peng:Single:2015,
  title = {Single Underwater Image Enhancement Using Depth Estimation Based on Blurriness},
  booktitle = {2015 {{IEEE International Conference}} on {{Image Processing}} ({{ICIP}})},
  author = {Peng, Yan-Tsung and Zhao, Xiangyun and Cosman, Pamela C.},
  year = {2015},
  month = sep,
  pages = {4952--4956},
  doi = {10.1109/ICIP.2015.7351749},
  urldate = {2025-02-24}
}

@article{Zhao:Deriving:2015,
  title = {Deriving Inherent Optical Properties from Background Color and Underwater Image Enhancement},
  author = {Zhao, Xinwei and Jin, Tao and Qu, Song},
  year = {2015},
  month = jan,
  journal = {Ocean Engineering},
  volume = {94},
  pages = {163--172},
  issn = {00298018},
  doi = {10.1016/j.oceaneng.2014.11.036},
  urldate = {2025-02-24},
  langid = {english}
}

@inproceedings{Liu:Spatiotemporal:2025,
author = {Liu, Shaohua and Gao, Ning and Gu, Zuoya and Dou, Hongkun and Deng, Yue and Li, Hongjue},
title = {Spatiotemporal Degradation-Aware 3D Gaussian Splatting for Realistic Underwater Scene Reconstruction},
year = {2025},
doi = {10.1145/3746027.3754888},
booktitle = {Proceedings of the 33rd ACM International Conference on Multimedia},
pages = {141–150},
}

@ARTICLE{Liu:SeaFree:2025,
  author={Liu, Shaohua and Gao, Ning and Fu, Shaowen and Zhong, Xiaoqing and Li, Hongjue},
  journal={IEEE Signal Processing Letters}, 
  title={{SeaFree-GS}: Reconstructing Underwater 3D Scenes With True Appearances}, 
  year={2025},
  volume={32},
  number={},
  pages={2114-2118},
  doi={10.1109/LSP.2025.3567853}}

@article{Lin:BVI-RLV:2024,
  title={{BVI-RLV: A} Fully Registered Dataset and Benchmarks for Low-Light Video Enhancement},
  author={R Lin and N Anantrasirichai and G Huang and J Lin and Q Sun and A Malyugina and DR Bull},
  journal={arXiv preprint arXiv:2407.03535},
  year={2024}
}

@ARTICLE{9880475,
  author={Xiao, Fengqi and Yuan, Fei and Huang, Yifan and Cheng, En},
  journal={IEEE Journal of Oceanic Engineering}, 
  title={Turbid Underwater Image Enhancement Based on Parameter-Tuned Stochastic Resonance}, 
  year={2023},
  volume={48},
  number={1},
  pages={127-146},
  doi={10.1109/JOE.2022.3190517}}

@article{anwar2020diving,
  title={Diving deeper into underwater image enhancement: A survey},
  author={Anwar, Saeed and Li, Chongyi},
  journal={Signal Processing: Image Communication},
  volume={89},
  pages={115978},
  year={2020},
  publisher={Elsevier}
}

@ARTICLE{Wang:Large:2025,
  author={Wang, Hao and K{\"o}ser, Kevin and Ren, Peng},
  journal={IEEE Transactions on Geoscience and Remote Sensing}, 
  title={Large Foundation Model Empowered Discriminative Underwater Image Enhancement}, 
  year={2025},
  volume={63},
  number={},
  pages={1-17},
  doi={10.1109/TGRS.2025.3525962}}

@inproceedings{fabbri2018enhancing,
  title={Enhancing underwater imagery using generative adversarial networks},
  author={Fabbri, Cameron and Islam, Md Jahidul and Sattar, Junaed},
  booktitle={International Conference on Robotics and Automation},
  pages={7159--7165},
  year={2018}
}

@article{panetta2015human,
  title={Human-visual-system-inspired underwater image quality measures},
  author={Panetta, Karen and Gao, Chen and Agaian, Sos},
  journal={IEEE Journal of Oceanic Engineering},
  volume={41},
  number={3},
  pages={541--551},
  year={2015},
  publisher={IEEE}
}

@inproceedings{zhang2018unreasonable,
  title={The unreasonable effectiveness of deep features as a perceptual metric},
  author={Zhang, Richard and Isola, Phillip and Efros, Alexei A and Shechtman, Eli and Wang, Oliver},
  booktitle={Proceedings of the IEEE conference on computer vision and pattern recognition},
  pages={586--595},
  year={2018}
}

@article{wang2009mean,
  title={Mean squared error: Love it or leave it? A new look at signal fidelity measures},
  author={Wang, Zhou and Bovik, Alan C},
  journal={IEEE signal processing magazine},
  volume={26},
  number={1},
  pages={98--117},
  year={2009},
  publisher={IEEE}
}

@article{yang2015underwater,
  title={An underwater color image quality evaluation metric},
  author={Yang, Miao and Sowmya, Arcot},
  journal={IEEE Transactions on Image Processing},
  volume={24},
  number={12},
  pages={6062--6071},
  year={2015},
  publisher={IEEE}
}

@article{chen2021underwater,
  title={Underwater image enhancement based on deep learning and image formation model},
  author={Chen, Xuelei and Zhang, Pin and Quan, Lingwei and Yi, Chao and Lu, Cunyue},
  journal={arXiv preprint arXiv:2101.00991},
  year={2021}
}

@article{bing2023domain,
  title={Domain Adaptation for In-Air to Underwater Image Enhancement via Deep Learning},
  author={Bing, Xuewen and Ren, Wenqi and Tang, Yang and Yen, Gary G and Sun, Qiyu},
  journal={IEEE Transactions on Emerging Topics in Computational Intelligence},
  year={2023},
}

@inproceedings{ye2022underwater,
  title={Underwater light field retention: Neural rendering for underwater imaging},
  author={Ye, Tian and Chen, Sixiang and Liu, Yun and Ye, Yi and Chen, Erkang and Li, Yuche},
  booktitle={IEEE Conference on Computer Vision and Pattern Recognition Workshops},
  pages={488--497},
  year={2022}
}

@article{jiang2022two,
  title={Two-step domain adaptation for underwater image enhancement},
  author={Jiang, Qun and Zhang, Yunfeng and Bao, Fangxun and Zhao, Xiuyang and Zhang, Caiming and Liu, Peide},
  journal={Pattern Recognition},
  volume={122},
  pages={108324},
  year={2022}
}

@article{Liu:RealWorld:2020,
  title = {Real-{{World Underwater Enhancement}}: {{Challenges}}, {{Benchmarks}}, and {{Solutions Under Natural Light}}},
  shorttitle = {Real-{{World Underwater Enhancement}}},
  author = {Liu, Risheng and Fan, Xin and Zhu, Ming and Hou, Minjun and Luo, Zhongxuan},
  year = {2020},
  month = dec,
  journal = {IEEE Transactions on Circuits and Systems for Video Technology},
  volume = {30},
  number = {12},
  pages = {4861--4875},
  issn = {1558-2205},
  doi = {10.1109/TCSVT.2019.2963772},
  urldate = {2025-02-12}
}

@inproceedings{Banerjee:Elimination:2014,
  title = {Elimination of {{Marine Snow}} Effect from Underwater Image - {{An}} Adaptive Probabilistic Approach},
  booktitle = {2014 {{IEEE Students}}' {{Conference}} on {{Electrical}}, {{Electronics}} and {{Computer Science}}},
  author = {Banerjee, Soma and Sanyal, Gautam and Ghosh, Shatadal and Ray, Ranjit and Shome, Sankar Nath},
  year = {2014},
  month = mar,
  pages = {1--4},
  doi = {10.1109/SCEECS.2014.6804438},
  urldate = {2025-02-13}
}

@article{YANG2025130262,
title = {{4D} Gaussian Splatting for high-fidelity dynamic reconstruction of single-view scenes},
journal = {Neurocomputing},
volume = {640},
pages = {130262},
year = {2025},
issn = {0925-2312},
doi = {https://doi.org/10.1016/j.neucom.2025.130262},
author = {Xiaodong Yang and Weixing Xie and Sen Peng and Yihang Fu and Wentao Fan and Baorong Yang and Xiao Dong},
}

@inproceedings{Xie:UVEB:2024,
  title = {{{UVEB}}: {{A Large-scale Benchmark}} and {{Baseline Towards Real-World Underwater Video Enhancement}}},
  author = {Xie, Yaofeng and Kong, Lingwei and Chen, Kai and Zheng, Ziqiang and Yu, Xiao and Yu, Zhibin and Zheng, Bing},
  booktitle = {Proceedings of the {IEEE/CVF} Conference on Computer Vision and Pattern Recognition ({CVPR})},
  year = {2024},
  pages = {22358--22367},
  doi = {10.1109/CVPR52733.2024.02110}
}

@article{peng2023u,
  title={U-shape transformer for underwater image enhancement},
  author={Peng, Lintao and Zhu, Chunli and Bian, Liheng},
  journal={IEEE Transactions on Image Processing},
  year={2023},
  publisher={IEEE}
}

@article{Li:Underwater:2021,
  title = {Underwater {{Image Enhancement}} via {{Medium Transmission-Guided Multi-Color Space Embedding}}},
  author = {Li, Chongyi and Anwar, Saeed and Hou, Junhui and Cong, Runmin and Guo, Chunle and Ren, Wenqi},
  year = {2021},
  journal = {IEEE Transactions on Image Processing},
  volume = {30},
  eprint = {2104.13015},
  primaryclass = {cs},
  pages = {4985--5000},
  issn = {1057-7149, 1941-0042},
  doi = {10.1109/TIP.2021.3076367},
  urldate = {2025-02-07},
  archiveprefix = {arXiv}
}

@article{Jiang:Underwater:2022,
  title = {Underwater {{Image Enhancement With Lightweight Cascaded Network}}},
  author = {Jiang, Nanfeng and Chen, Weiling and Lin, Yuting and Zhao, Tiesong and Lin, Chia-Wen},
  year = {2022},
  journal = {IEEE Transactions on Multimedia},
  volume = {24},
  pages = {4301--4313},
  issn = {1941-0077},
  doi = {10.1109/TMM.2021.3115442},
  urldate = {2025-02-18}
}

@article{Huang:Advanced:2015,
  title = {An {{Advanced Single-Image Visibility Restoration Algorithm}} for {{Real-World Hazy Scenes}}},
  author = {Huang, Shih-Chia and Ye, Jian-Hui and Chen, Bo-Hao},
  year = {2015},
  month = may,
  journal = {IEEE Transactions on Industrial Electronics},
  volume = {62},
  number = {5},
  pages = {2962--2972},
  issn = {1557-9948},
  doi = {10.1109/TIE.2014.2364798},
  urldate = {2025-02-18}
}

@article{Huang:Visibility:2014,
  title = {Visibility {{Restoration}} of {{Single Hazy Images Captured}} in {{Real-World Weather Conditions}}},
  author = {Huang, Shih-Chia and Chen, Bo-Hao and Wang, Wei-Jheng},
  year = {2014},
  month = oct,
  journal = {IEEE Transactions on Circuits and Systems for Video Technology},
  volume = {24},
  number = {10},
  pages = {1814--1824},
  issn = {1558-2205},
  doi = {10.1109/TCSVT.2014.2317854},
  urldate = {2025-02-18}
}

@inproceedings{Chao:Removal:2010,
  title = {Removal of Water Scattering},
  booktitle = {2010 2nd {{International Conference}} on {{Computer Engineering}} and {{Technology}}},
  author = {Chao, Liu and Wang, Meng},
  year = {2010},
  month = apr,
  volume = {2},
  pages = {V2-35-V2-39},
  doi = {10.1109/ICCET.2010.5485339},
  urldate = {2025-02-18}
}

@inproceedings{Khan:Spectroformer:2024,
  title = {Spectroformer: {{Multi-Domain Query Cascaded Transformer Network For Underwater Image Enhancement}}},
  shorttitle = {Spectroformer},
  booktitle = {2024 {{IEEE}}/{{CVF Winter Conference}} on {{Applications}} of {{Computer Vision}} ({{WACV}})},
  author = {Khan, Md Raqib and Mishra, Priyanka and Mehta, Nancy and Phutke, Shruti S. and Vipparthi, Santosh Kumar and Nandi, Sukumar and Murala, Subrahmanyam},
  year = {2024},
  month = jan,
  pages = {1443--1452},
  publisher = {IEEE},
  address = {Waikoloa, HI, USA},
  doi = {10.1109/WACV57701.2024.00148},
  urldate = {2025-02-18},
  copyright = {https://doi.org/10.15223/policy-029},
  isbn = {979-8-3503-1892-0},
  langid = {english}
}

@article{Shen:UDAformer:2023,
  title = {{{UDAformer}}: {{Underwater}} Image Enhancement Based on Dual Attention Transformer},
  shorttitle = {{{UDAformer}}},
  author = {Shen, Zhen and Xu, Haiyong and Luo, Ting and Song, Yang and He, Zhouyan},
  year = {2023},
  month = apr,
  journal = {Computers \& Graphics},
  volume = {111},
  pages = {77--88},
  issn = {00978493},
  doi = {10.1016/j.cag.2023.01.009},
  urldate = {2025-02-18},
  langid = {english}
}

@article{Wang:UIEC^2Net:2021,
  title = {{{UIEC}}{\textasciicircum}2-{{Net}}: {{CNN-based}} Underwater Image Enhancement Using Two Color Space},
  shorttitle = {{{UIEC}}{\textasciicircum}2-{{Net}}},
  author = {Wang, Yudong and Guo, Jichang and Gao, Huan and Yue, Huihui},
  year = {2021},
  month = aug,
  journal = {Signal Processing: Image Communication},
  volume = {96},
  pages = {116250},
  issn = {09235965},
  doi = {10.1016/j.image.2021.116250},
  urldate = {2025-02-18},
  langid = {english}
}

@misc{Guan:WaterMamba:2024,
  title = {{{WaterMamba}}: {{Visual State Space Model}} for {{Underwater Image Enhancement}}},
  shorttitle = {{{WaterMamba}}},
  author = {Guan, Meisheng and Xu, Haiyong and Jiang, Gangyi and Yu, Mei and Chen, Yeyao and Luo, Ting and Song, Yang},
  year = {2024},
  month = may,
  number = {arXiv:2405.08419},
  eprint = {2405.08419},
  primaryclass = {cs},
  publisher = {arXiv},
  doi = {10.48550/arXiv.2405.08419},
  urldate = {2025-02-18},
  archiveprefix = {arXiv}
}

@inproceedings{Gu2024Mamba,
  author    = {A. Gu and T. Dao},
  title     = {{Mamba: Linear}-time sequence modeling with selective state spaces},
  booktitle = {Conference on Language Modeling},
  year      = {2024}
}

@InProceedings{Lin_2024_ACCV,
    author    = {Lin, Wei-Tung and Lin, Yong-Xiang and Chen, Jyun-Wei and Hua, Kai-Lung},
    title     = {PixMamba: Leveraging State Space Models in a Dual-Level Architecture for Underwater Image Enhancement},
    booktitle = {Proceedings of the Asian Conference on Computer Vision (ACCV)},
    month     = {December},
    year      = {2024},
    pages     = {3622-3637}
}

@article{Zhang2024MambaUIE,
  author    = {Song Zhang and Yuqing Duan and Daoliang Li and Ran Zhao},
  title     = {Mamba-UIE: Enhancing Underwater Images with Physical Model Constraint},
  journal   = {arXiv preprint arXiv:2407.19248},
  year      = {2024}
}

@INPROCEEDINGS{Chen:MUIR:2024,
  author={Chen, Liyuan and Li, Weijia and Yang, Qingxia and Tong, Lihan and Chen, Erkang and Huang, Bin and Li, RuiWen},
  booktitle={2024 4th International Conference on Machine Learning and Intelligent Systems Engineering (MLISE)}, 
  title={{MUIR: Mamba} for Underwater Image Rendering}, 
  year={2024},
  volume={},
  number={},
  pages={172-177},
  doi={10.1109/MLISE62164.2024.10674249}}

@article{CONG2021337,
title = {Underwater robot sensing technology: A survey},
journal = {Fundamental Research},
volume = {1},
number = {3},
pages = {337-345},
year = {2021},
issn = {2667-3258},
doi = {https://doi.org/10.1016/j.fmre.2021.03.002},
author = {Yang Cong and Changjun Gu and Tao Zhang and Yajun Gao},
}

@article{Wright2020UnderwaterPhotogrammetry,
  author = {Wright, A.E. and Conlin, D.L. and Shope, S.M.},
  title = {Assessing the Accuracy of Underwater Photogrammetry for Archaeology: A Comparison of Structure from Motion Photogrammetry and Real Time Kinematic Survey at the East Key Construction Wreck},
  journal = {Journal of Marine Science and Engineering},
  year = {2020},
  volume = {8},
  number = {11},
  pages = {849},
  doi = {10.3390/jmse8110849},
  url = {https://doi.org/10.3390/jmse8110849}
}

@INPROCEEDINGS{Yang:comparison:2024,
  author={Yang, Wen and Lin, Yongliang and Lim, Chun Her and Tao, Zewen and Leng, Jianxing},
  booktitle={2024 China Automation Congress (CAC)}, 
  title={Experimental Comparsion Between NeRFs and 3D Gaussian Splatting for Underwater 3D Reconstruction}, 
  year={2024},
  volume={},
  number={},
  pages={6633-6638},
  doi={10.1109/CAC63892.2024.10864941}}

@inproceedings{Khan:Phaseformer:2024,
  title = {Phaseformer: {{Phase-based Attention Mechanism}} for {{Underwater Image Restoration}} and {{Beyond}}},
  author = {Khan, MD Raqib and Negi, Anshul and Kulkarni, Ashutosh and Phutke, Shruti S. and Vipparthi, Santosh Kumar and Murala, Subrahmanyam},
  booktitle = {2025 {IEEE/CVF} Winter Conference on Applications of Computer Vision ({WACV})},
  year = {2025},
  pages = {9618--9629},
  organization = {IEEE},
  doi = {10.1109/WACV61041.2025.00931}
}

@article{li2019fusion,
  title={A fusion adversarial underwater image enhancement network with a public test dataset},
  author={Li, Hanyu and Li, Jingjing and Wang, Wei},
  journal={arXiv preprint arXiv:1906.06819},
  year={2019}
}

@article{liu2019underwater,
  title={Underwater image enhancement with a deep residual framework},
  author={Liu, Peng and Wang, Guoyu and Qi, Hao and Zhang, Chufeng and Zheng, Haiyong and Yu, Zhibin},
  journal={IEEE Access},
  volume={7},
  pages={94614--94629},
  year={2019}
}

@article{Ancuti:Color:2018,
  title = {Color {{Balance}} and {{Fusion}} for {{Underwater Image Enhancement}}},
  author = {Ancuti, Codruta O. and Ancuti, Cosmin and De Vleeschouwer, Christophe and Bekaert, Philippe},
  year = {2018},
  month = jan,
  journal = {IEEE Transactions on Image Processing},
  volume = {27},
  number = {1},
  pages = {379--393},
  issn = {1941-0042},
  doi = {10.1109/TIP.2017.2759252},
  urldate = {2025-02-19}
}

@article{AbdulGhani:Enhancement:2015,
  title = {Enhancement of Low Quality Underwater Image through Integrated Global and Local Contrast Correction},
  author = {Abdul Ghani, Ahmad Shahrizan and Mat Isa, Nor Ashidi},
  year = {2015},
  month = dec,
  journal = {Applied Soft Computing},
  volume = {37},
  pages = {332--344},
  issn = {15684946},
  doi = {10.1016/j.asoc.2015.08.033},
  urldate = {2025-02-26},
  langid = {english}
}

@article{AbdulGhani:Automatic:2017,
  title = {Automatic System for Improving Underwater Image Contrast and Color through Recursive Adaptive Histogram Modification},
  author = {Abdul Ghani, Ahmad Shahrizan and Mat Isa, Nor Ashidi},
  year = {2017},
  month = sep,
  journal = {Computers and Electronics in Agriculture},
  volume = {141},
  pages = {181--195},
  issn = {01681699},
  doi = {10.1016/j.compag.2017.07.021},
  urldate = {2025-02-26},
  langid = {english}
}

@inproceedings{Ghani:Underwater:2014,
  title = {Underwater Image Quality Enhancement through Composition of Dual-Intensity Images and {{Rayleigh-stretching}}},
  booktitle = {2014 {{IEEE Fourth International Conference}} on {{Consumer Electronics Berlin}} ({{ICCE-Berlin}})},
  author = {Ghani, Ahmad Shahrizan Abdul and Isa, Nor Ashidi Mat},
  year = {2014},
  month = sep,
  pages = {219--220},
  issn = {2166-6822},
  doi = {10.1109/ICCE-Berlin.2014.7034265},
  urldate = {2025-02-26}
}

@article{Li:WaterGAN:2017,
  title = {{{WaterGAN}}: {{Unsupervised Generative Network}} to {{Enable Real-Time Color Correction}} of {{Monocular Underwater Images}}},
  shorttitle = {{{WaterGAN}}},
  author = {Li, Jie and Skinner, Katherine A. and Eustice, Ryan M. and Johnson-Roberson, Matthew},
  journal = {IEEE Robotics and Automation Letters},
  volume = {3},
  number = {1},
  pages = {387--394},
  year = {2018},
  month = jan,
  issn = {2377-3766, 2377-3774},
  doi = {10.1109/LRA.2017.2730363}
}

@article{Vasamsetti:Wavelet:2017,
  title = {Wavelet Based Perspective on Variational Enhancement Technique for Underwater Imagery},
  author = {Vasamsetti, Srikanth and Mittal, Neerja and Neelapu, Bala Chakravarthy and Sardana, Harish Kumar},
  year = {2017},
  month = sep,
  journal = {Ocean Engineering},
  volume = {141},
  pages = {88--100},
  issn = {00298018},
  doi = {10.1016/j.oceaneng.2017.06.012},
  urldate = {2025-02-26},
  langid = {english}
}

@article{Garg:Underwater:2018,
  title = {Underwater Image Enhancement Using Blending of {{CLAHE}} and Percentile Methodologies},
  author = {Garg, Diksha and Garg, Naresh Kumar and Kumar, Munish},
  year = {2018},
  month = oct,
  journal = {Multimedia Tools and Applications},
  volume = {77},
  number = {20},
  pages = {26545--26561},
  issn = {1573-7721},
  doi = {10.1007/s11042-018-5878-8},
  urldate = {2025-02-26},
  langid = {english}
}

@article{huang2022underwater,
  title={Underwater image enhancement via adaptive group attention-based multiscale cascade transformer},
  author={Huang, Zhixiong and Li, Jinjiang and Hua, Zhen and Fan, Linwei},
  journal={IEEE Transactions on Instrumentation and Measurement},
  volume={71},
  pages={1--18},
  year={2022}
}

@inproceedings{guo2023underwater,
  title={{Underwater Ranker}: Learn Which Is Better and How to Be Better},
  author={Guo, Chunle and Wu, Ruiqi and Jin, Xin and Han, Linghao and Zhang, Weidong and Chai, Zhi and Li, Chongyi},
  booktitle={AAAI Conference on Artificial Intelligence},
  pages={702--709},
  year={2023}
}

@article{vrochidis2025colormap,
  title={Enhancing Three-Dimensional Reconstruction Through Intelligent Colormap Selection},
  author={Vrochidis, Alexandros and Tzovaras, Dimitrios and Krinidis, Stelios},
  journal={Sensors},
  volume={25},
  number={8},
  pages={2576},
  year={2025},
  doi={10.3390/s25082576}
}

@article{shaker2023impact,
  title   = {The Impact of Image Enhancement and Transfer Learning Techniques on Marine Habitat Mapping},
  author  = {Shaker, E. and Baker, M. R. and Mahmood, Z.},
  journal = {Gazi University Journal of Science},
  year    = {2023},
  volume  = {36},
  number  = {2},
  pages   = {592--606},
  month   = {June},
  doi     = {10.35378/gujs.973082}
}

@ARTICLE{10600694,
  author={Ling, Ziyao and Delnevo, Giovanni and Salomoni, Paola and Mirri, Silvia},
  journal={IEEE Access}, 
  title={Findings on Machine Learning for Identification of Archaeological Ceramics: A Systematic Literature Review}, 
  year={2024},
  volume={12},
  number={},
  pages={100167-100185},
  doi={10.1109/ACCESS.2024.3429623}}

@article{yaqoob2025advancing,
  title   = {Advancing Paleontology: A Survey on Deep Learning Methodologies in Fossil Image Analysis},
  author  = {Yaqoob, M. and Ishaq, M. and Ansari, M. Y. and others},
  journal = {Artificial Intelligence Review},
  year    = {2025},
  volume  = {58},
  pages   = {83},
  doi     = {10.1007/s10462-024-11080-y}
}

@inproceedings{rahnama2025subsea,
  title     = {Subsea Inspection Supported by AI-Driven Computer Vision},
  author    = {Rahnama, R. and Hill, D. and Mills, A.},
  booktitle = {Proceedings of ADIPEC},
  year      = {2025},
  month     = {November},
  doi       = {10.2118/229719-MS}
}

@inproceedings{doersch2023tapir,
  title     = {{TAPIR}: Tracking Any Point with Per-Frame Initialization and Temporal Refinement},
  author    = {Doersch, Carl and Yang, Yi and Vecerik, Mel and Gokay, Dilara and Gupta, Ankush and Aytar, Yusuf and Carreira, Joao and Zisserman, Andrew},
  booktitle = {Proceedings of the IEEE/CVF Conference on Computer Vision and Pattern Recognition (CVPR)},
  year      = {2023}
}

@inproceedings{vrochidis2025underwater3d,
  title={Enhancing {3D} Reconstructions in Underwater Environments: The Impact of Image Enhancement on Model Quality},
  author={Vrochidis, Alexandros and Tzovaras, Dimitrios and Krinidis, Stelios},
  booktitle={The International Archives of the Photogrammetry, Remote Sensing and Spatial Information Sciences},
  volume={XLVIII-2/W10-2025},
  pages={317--324},
  year={2025},
  doi={10.5194/isprs-archives-XLVIII-2-W10-2025-317-2025}
}

@inproceedings{boudiaf2022underwater,
  title={Underwater image enhancement using pre-trained transformer},
  author={Boudiaf, Abderrahmene and Guo, Yuhang and Ghimire, Adarsh and Werghi, Naoufel and De Masi, Giulia and Javed, Sajid and Dias, Jorge},
  booktitle={International Conference on Image Analysis and Processing},
  pages={480--488},
  year={2022}
}

@article{ren2022reinforced,
  title={Reinforced Swin-Convs Transformer for Simultaneous Underwater Sensing Scene Image Enhancement and Super-Resolution},
  author={Ren, Tingdi and Xu, Haiyong and Jiang, Gangyi and Yu, Mei and Zhang, Xiaokang and Wang, Biao and Luo, Ting},
  journal={IEEE Transactions on Geoscience and Remote Sensing},
  volume={60},
  pages={1--16},
  year={2022},
  publisher={IEEE},
  doi={10.1109/TGRS.2022.3144882}
}

@article{wen2024waterformer,
  title={WaterFormer: A Global--Local Transformer for Underwater Image Enhancement With Environment Adaptor},
  author={Wen, Junjie and Cui, Jinqiang and Yang, Guidong and Zhao, Benyun and Zhai, Yu and Gao, Zhi and Dou, Lihua and Chen, Ben M},
  journal={IEEE Robotics \& Automation Magazine},
  volume={31},
  number={1},
  pages={29--40},
  year={2024},
  publisher={IEEE},
  doi={10.1109/MRA.2024.3351487}
}

@article{li2025taformer,
  title={{TAFormer}: A Transmission-Aware Transformer for Underwater Image Enhancement},
  author={Li, Zhiyong and Tang, Cuixiang and Huang, Rihan and Han, Chunyang},
  journal={IEEE Transactions on Circuits and Systems for Video Technology},
  volume={35},
  number={1},
  pages={770--784},
  year={2025},
  publisher={IEEE},
  doi={10.1109/TCSVT.2024.3417165}
}

@article{peng2025histoformer,
  title={Histoformer: Histogram-Based Transformer for Efficient Underwater Image Enhancement},
  author={Peng, Yan-Tsung and Chen, Yen-Rong and Chen, Guan-Rong and Liao, Chun-Jung},
  journal={IEEE Journal of Oceanic Engineering},
  volume={50},
  number={1},
  pages={164--177},
  year={2025},
  publisher={IEEE},
  doi={10.1109/JOE.2024.3474919}
}

@article{zhou2025uiesfiformer,
  title={{UIE-SFIFormer}: Underwater Image Enhancement Based on Physical-Guided Spatial-Frequency Interaction Transformer},
  author={Zhou, Yuan and Xu, Haiyong and Jiang, Gangyi and Ren, Tingdi and Yu, Mei and Chen, Huaizhi},
  journal={IEEE Journal of Oceanic Engineering},
  volume={50},
  number={2},
  pages={727--742},
  year={2025},
  publisher={IEEE},
  doi={10.1109/JOE.2024.3458109}
}

@inproceedings{wen2023syreanet,
  title={SyreaNet: A Physically Guided Underwater Image Enhancement Framework Integrating Synthetic and Real Images},
  author={Wen, Junjie and Cui, Jinqiang and Zhao, Zhenjun and Yan, Ruixin and Gao, Zhi and Dou, Lihua and Chen, Ben M},
  booktitle={IEEE International Conference on Robotics and Automation (ICRA)},
  pages={5177--5183},
  year={2023},
  doi={10.1109/ICRA48891.2023.10161260}
}

@article{wen2025ssduie,
  title={A Semi-Supervised Domain-Adaptive Framework for Real-World Underwater Image Enhancement},
  author={Wen, Junjie and Yang, Guidong and Zhao, Benyun and Huang, Dongyue and Lei, Lei and Zhang, Bo and Gao, Zhi and Chen, Xi and Chen, Ben M},
  journal={IEEE Transactions on Geoscience and Remote Sensing},
  volume={63},
  pages={1--15},
  year={2025},
  doi={10.1109/TGRS.2025.3590798}
}

@inproceedings{yang2025uwmvs,
  title={End-to-End Underwater Multi-View Stereo for Dense Scene Reconstruction},
  author={Yang, Guidong and Wen, Junjie and Zhao, Benyun and Li, Qingxiang and Huang, Yijun and Lei, Lei and Chen, Xi and Lam, Alan H. F. and Chen, Ben M},
  booktitle={IEEE International Conference on Robotics and Automation (ICRA)},
  pages={7616--7623},
  year={2025},
  doi={10.1109/ICRA55743.2025.11128539}
}

@inproceedings{yao2018mvsnet,
  title={MVSNet: Depth Inference for Unstructured Multi-View Stereo},
  author={Yao, Yao and Luo, Zixin and Li, Shiwei and Fang, Tian and Quan, Long},
  booktitle={Proceedings of the European Conference on Computer Vision (ECCV)},
  pages={767--783},
  year={2018}
}

@inproceedings{gu2020casmvsnet,
  title={Cascade Cost Volume for High-Resolution Multi-View Stereo and Stereo Matching},
  author={Gu, Xiaodong and Fan, Zhiwen and Zhu, Siyu and Dai, Zuozhuo and Tan, Feitong and Tan, Ping},
  booktitle={Proceedings of the IEEE/CVF Conference on Computer Vision and Pattern Recognition (CVPR)},
  pages={2495--2504},
  year={2020}
}

@inproceedings{zhang2023geomvsnet,
  title={GeoMVSNet: Learning Multi-View Stereo With Geometry Perception},
  author={Zhang, Zhe and Peng, Rui and Hu, Yuxi and Wang, Ronggang},
  booktitle={Proceedings of the IEEE/CVF Conference on Computer Vision and Pattern Recognition (CVPR)},
  pages={21508--21518},
  year={2023}
}

@misc{dji_terra,
  author={DJI},
  title={DJI Terra},
  howpublished={\url{https://enterprise.dji.com/dji-terra}},
  year={2026},
  note={Accessed: 2026-03-20}
}

@article{rout2024surveillance,
  title={Underwater Visual Surveillance: A Comprehensive Survey},
  author={Rout, Deepak and Ray, Niranjan and Das, Madhabananda},
  journal={Ocean Engineering},
  volume={314},
  pages={119556},
  year={2024},
  doi={10.1016/j.oceaneng.2024.119556}
}

@article{garciagarcia2020background,
  title={Background Subtraction in Real Applications: Challenges, Current Models and Future Directions},
  author={Garcia-Garcia, Belmar and Bouwmans, Thierry and Rosales Silva, Aldo Jonathan},
  journal={Computer Science Review},
  volume={35},
  pages={100204},
  year={2020},
  doi={10.1016/j.cosrev.2019.100204}
}

@inproceedings{helan2006object,
  title={Object Detection in Underwater Images},
  author={Helan, Sebastien and Burie, Jean-Christophe and Bouwmans, Thierry and Bazeille, Stephane},
  booktitle={Proceedings of the SEA TECH WEEK Caract{\'e}risation du Milieu Marin (CMM)},
  address={Brest, France},
  year={2006}
}

@inproceedings{bazeille2006automatic,
  title={Automatic Underwater Image Pre-Processing},
  author={Bazeille, Stephane and Quidu, Isabelle and Jaulin, Luc and Malkasse, Jean-Pierre},
  booktitle={Proceedings of the SEA TECH WEEK Caract{\'e}risation du Milieu Marin (CMM)},
  address={Brest, France},
  year={2006}
}

@incollection{nissar2026human,
  title={A Human Vision Neuroscience-Driven Deep Neural Framework for Change Detection in Underwater Scenes},
  author={Nissar, Mohd and Farhadifard, Fahimeh and Radolko, Martin and von Lukas, Uwe},
  booktitle={Computer Vision, Pattern Recognition, Image Processing, and Graphics},
  series={Communications in Computer and Information Science},
  volume={2522},
  pages={427--438},
  year={2026},
  publisher={Springer}
}

@article{kapoor2025graph,
  title={Graph-Based Moving Object Segmentation for Underwater Videos Using Semi-Supervised Learning},
  author={Kapoor, Meghna and Prummel, Wieke and Giraldo, Jhony H. and Subudhi, Badri Narayan and Zakharova, Anastasia and Bouwmans, Thierry and Bansal, Ankur},
  journal={Computer Vision and Image Understanding},
  volume={252},
  pages={104290},
  year={2025},
  doi={10.1016/j.cviu.2025.104290}
}

@inproceedings{kapoor2024principal,
  title={Principal Graph Neighborhood Aggregation for Underwater Moving Object Detection},
  author={Kapoor, Meghna and Subudhi, Badri Narayan and Jakhetiya, Vinit},
  booktitle={International Conference on Pattern Recognition (ICPR)},
  pages={398--412},
  year={2024}
}

@inproceedings{zhang2024fishtracking,
  title={Underwater Fish Tracking Algorithm Based on {ViBE} Detection and Covariance Matrix},
  author={Zhang, A. and Palaoag, T.},
  booktitle={International Conference on Image, Video and Signal Processing (IVSP)},
  year={2024}
}

@article{spampinato2014typhoon,
  title={Understanding Fish Behavior During Typhoon Events in Real-Life Underwater Environments},
  author={Spampinato, Concetto and Palazzo, Simone and Giordano, Daniela and Kavasidis, Isaak and Lin, Fang-Pang and Lin, Yi-Ting},
  journal={Multimedia Tools and Applications},
  volume={70},
  number={1},
  pages={199--224},
  year={2014},
  doi={10.1007/s11042-012-1228-0}
}

@inproceedings{radolko2016dataset,
  title={Dataset on Underwater Change Detection},
  author={Radolko, Martin and Farhadifard, Fahimeh and von Lukas, Uwe},
  booktitle={OCEANS 2016 MTS/IEEE Monterey},
  year={2016},
  doi={10.1109/OCEANS.2016.7761129}
}

@article{fu2023rethinking,
  title={Rethinking General Underwater Object Detection: Datasets, Challenges, and Solutions},
  author={Fu, Chenping and Liu, Risheng and Fan, Xin and Zhou, Tianyu and Luo, Zijing and Zhang, Lei and Huang, Xiangyu and Luo, Zhongxuan},
  journal={Neurocomputing},
  volume={517},
  pages={243--256},
  year={2023},
  doi={10.1016/j.neucom.2022.10.039}
}

@article{humbert2023octopus,
  title={The Open-Source Camera Trap for Organism Presence and Underwater Surveillance ({OCTOPUS})},
  author={Humbert, J{\'e}r{\^o}me and McMath, Anne and Robin, Alexandre and Espiau, Benoit and Abadie, Arnaud},
  journal={HardwareX},
  volume={15},
  pages={e00472},
  year={2023},
  doi={10.1016/j.ohx.2023.e00472}
}

@article{ponzi2025graph,
  title={Graph Neural Networks: Architectures, Applications, and Future Directions},
  author={Ponzi, Valerio and Napoli, Christian},
  journal={IEEE Access},
  volume={13},
  pages={62870--62891},
  year={2025},
  doi={10.1109/ACCESS.2025.3558752}
}

@article{sadasivan2025systematic,
  title={A Systematic Survey of Graph Convolutional Networks for Artificial Intelligence Applications},
  author={Sadasivan, Amutha and others},
  journal={WIREs Data Mining and Knowledge Discovery},
  volume={15},
  number={2},
  pages={e70012},
  year={2025},
  doi={10.1002/widm.70012}
}

@article{pierard2025methodology,
  title={A Methodology to Evaluate Strategies Predicting Rankings on Unseen Domains},
  author={Pi{\'e}rard, Geoffroy and Van Droogenbroeck, Marc},
  journal={CoRR},
  volume={abs/2505.15595},
  year={2025},
  url={https://arxiv.org/abs/2505.15595}
}

@article{pierard2025optimal,
  title={What Is the Optimal Ranking Score Between Precision and Recall? We Can Always Find It and It Is Rarely {F1}},
  author={Pi{\'e}rard, Geoffroy and Van Droogenbroeck, Marc},
  journal={CoRR},
  volume={abs/2511.22442},
  year={2025},
  url={https://arxiv.org/abs/2511.22442}
}

@inproceedings{Li:ZeroTIG:2025,
  title     = {{Zero-TIG}: Temporal Consistency-Aware Zero-Shot Illumination-Guided Low-light Video Enhancement},
  author    = {Li, Y. and Anantrasirichai, N.},
  booktitle = {Proceedings of the 33rd European Signal Processing Conference (EUSIPCO)},
  year      = {2025}
}

@inproceedings{fu2022uncertainty,
  title={Uncertainty inspired underwater image enhancement},
  author={Fu, Zhenqi and Wang, Wu and Huang, Yue and Ding, Xinghao and Ma, Kai-Kuang},
  booktitle={European Conference on Computer Vision},
  pages={465--482},
  year={2022}
}

@article{liu2022twin,
  title={Twin adversarial contrastive learning for underwater image enhancement and beyond},
  author={Liu, Risheng and Jiang, Zhiying and Yang, Shuzhou and Fan, Xin},
  journal={IEEE Transactions on Image Processing},
  volume={31},
  pages={4922--4936},
  year={2022}
}

@inproceedings{Malyugina:beam:2025,
  title={Marine Snow Removal Using Internally Generated Pseudo Ground Truth},
  author={Alexandra Malyugina and Guoxi Huang and Eduardo Ruiz-Libreros and Ben Leslie and Nantheera Anantrasirichai},
  booktitle={33rd European Signal Processing Conference},
  year={2025}
}

@inproceedings{huang2023contrastive,
  title={Contrastive semi-supervised learning for underwater image restoration via reliable bank},
  author={Huang, Shirui and Wang, Keyan and Liu, Huan and Chen, Jun and Li, Yunsong},
  booktitle={IEEE Conference on Computer Vision and Pattern Recognition},
  pages={18145--18155},
  year={2023}
}

@article{chen2020perceptual,
  title={Perceptual underwater image enhancement with deep learning and physical priors},
  author={Chen, Long and Jiang, Zheheng and Tong, Lei and Liu, Zhihua and Zhao, Aite and Zhang, Qianni and Dong, Junyu and Zhou, Huiyu},
  journal={IEEE Transactions on Circuits and Systems for Video Technology},
  volume={31},
  number={8},
  pages={3078--3092},
  year={2020}
}

@article{chen2022domain,
  title={Domain Adaptation for Underwater Image Enhancement via Content and Style Separation},
  author={Chen, Yu-Wei and Pei, Soo-Chang},
  journal={IEEE Access},
  volume={10},
  pages={90523--90534},
  year={2022}
}

@inproceedings{tang2022autoenhancer,
  title={AutoEnhancer: Transformer on U-Net Architecture Search for Underwater Image Enhancement},
  author={Tang, Yi and Iwaguchi, Takafumi and Kawasaki, Hiroshi and Sagawa, Ryusuke and Furukawa, Ryo},
  booktitle={Proceedings of the Asian Conference on Computer Vision},
  pages={1403--1420},
  year={2022}
}

@inproceedings{uplavikar2019all,
  title={All-in-One Underwater Image Enhancement Using Domain-Adversarial Learning},
  author={Uplavikar, Pritish M and Wu, Zhenyu and Wang, Zhangyang},
  booktitle={IEEE Conference on Computer Vision and Pattern Recognition Workshops},
  pages={1--8},
  year={2019}
}

@article{yin2024unsupervised,
  title={Unsupervised Underwater Image Enhancement Based on Disentangled Representations via Double-Order Contrastive Loss},
  author={Yin, Jiankai and Wang, Yan and Guan, Bowen and Zeng, Xianchao and Guo, Lei},
  journal={IEEE Transactions on Geoscience and Remote Sensing},
  year={2024},
}

@article{jiang2023perception,
  title={Perception-Driven Deep Underwater Image Enhancement without Paired Supervision},
  author={Jiang, Qiuping and Kang, Yaozu and Wang, Zhihua and Ren, Wenqi and Li, Chongyi},
  journal={IEEE Transactions on Multimedia},
  year={2023}
}

@inproceedings{kapoor2023domain,
  title={Domain Adversarial Learning Towards Underwater Image Enhancement},
  author={Kapoor, Meghna and Baghel, Rohan and Subudhi, Badri Narayan and Jakhetiya, Vinit and Bansal, Ankur},
  booktitle={IEEE International Conference on Computer Vision Workshops},
  pages={2241--2251},
  year={2023}
}

@article{yan2023uw,
  title={UW-CycleGAN: Model-driven CycleGAN for Underwater Image Restoration},
  author={Yan, Haorui and Zhang, Zhenwei and Xu, Jing and Wang, Tingting and An, Ping and Wang, Aobo and Duan, Yuping},
  journal={IEEE Transactions on Geoscience and Remote Sensing},
  year={2023}
}

@inproceedings{tang2023underwater,
  title={Underwater image enhancement by transformer-based diffusion model with non-uniform sampling for skip strategy},
  author={Tang, Yi and Kawasaki, Hiroshi and Iwaguchi, Takafumi},
  booktitle={ACM International Conference on Multimedia},
  pages={5419--5427},
  year={2023}
}

@article{zhu2023unsupervised,
  title={Unsupervised underwater image enhancement via content-style representation disentanglement},
  author={Zhu, Pengli and Liu, Yancheng and Wen, Yuanquan and Xu, Minyi and Fu, Xianping and Liu, Siyuan},
  journal={Engineering Applications of Artificial Intelligence},
  volume={126},
  pages={106866},
  year={2023},
}

@inproceedings{Shin:Estimation:2016,
  title = {Estimation of Ambient Light and Transmission Map with Common Convolutional Architecture},
  booktitle = {{{OCEANS}} 2016 {{MTS}}/{{IEEE Monterey}}},
  author = {Shin, Young-Sik and Cho, Younggun and Pandey, Gaurav and Kim, Ayoung},
  year = {2016},
  month = sep,
  pages = {1--7},
  doi = {10.1109/OCEANS.2016.7761342},
  urldate = {2025-02-18}
}

@article{Funt:Retinex:2000,
  title = {Retinex in {{Matlab}}},
  author = {Funt, Brian and Ciurea, Florian and McCann, John},
  year = {2000},
  langid = {english}
}

@article{Zhang:Underwater:2023,
  title = {Underwater {{Image Restoration}} via {{Adaptive Color Correction}} and {{Contrast Enhancement Fusion}}},
  author = {Zhang, Weihong and Li, Xiaobo and Xu, Shuping and Li, Xujin and Yang, Yiguang and Xu, Degang and Liu, Tiegen and Hu, Haofeng},
  year = {2023},
  month = jan,
  journal = {Remote Sensing},
  volume = {15},
  number = {19},
  pages = {4699},
  publisher = {Multidisciplinary Digital Publishing Institute},
  issn = {2072-4292},
  doi = {10.3390/rs15194699},
  urldate = {2025-02-18},
  copyright = {http://creativecommons.org/licenses/by/3.0/},
  langid = {english}
}

@article{rao2023deep,
  author={Rao, Yuan and Liu, Wenjie and Li, Kunqian and Fan, Hao and Wang, Sen and Dong, Junyu},
  journal={IEEE Transactions on Circuits and Systems for Video Technology}, 
  title={Deep Color Compensation for Generalized Underwater Image Enhancement}, 
  year={2024},
  volume={34},
  number={4},
  pages={2577-2590},
  doi={10.1109/TCSVT.2023.3305777}}

@inproceedings{mu2023generalized,
  title={A generalized physical-knowledge-guided dynamic model for underwater image enhancement},
  author={Mu, Pan and Xu, Hanning and Liu, Zheyuan and Wang, Zheng and Chan, Sixian and Bai, Cong},
  booktitle={ACM International Conference on Multimedia},
  pages={7111--7120},
  year={2023}
}

@inproceedings{cong2024underwater,
  title={Underwater Organism Color Fine-Tuning via Decomposition and Guidance},
  author={Cong, Xiaofeng and Gui, Jie and Hou, Junming},
  booktitle={AAAI Conference on Artificial Intelligence},
  volume={38},
  pages={1389--1398},
  year={2024}
}

@inproceedings{mcglamery1980computer,
  title={A computer model for underwater camera systems},
  author={McGlamery, BL},
  booktitle={Ocean Optics VI},
  volume={208},
  pages={221--231},
  year={1980},
  organization={SPIE}
}

@inproceedings{badran2023daut,
  title={DAUT: Underwater Image Enhancement Using Depth Aware U-shape Transformer},
  author={Badran, Mohamed and Torki, Marwan},
  booktitle={IEEE International Conference on Image Processing},
  pages={1830--1834},
  year={2023}
}

@article{liu2021ipmgan,
  title={IPMGAN: Integrating physical model and generative adversarial network for underwater image enhancement},
  author={Liu, Xiaodong and Gao, Zhi and Chen, Ben M},
  journal={Neurocomputing},
  volume={453},
  pages={538--551},
  year={2021}
}

@inproceedings{fu2022unsupervised,
  title={Unsupervised underwater image restoration: From a homology perspective},
  author={Fu, Zhenqi and Lin, Huangxing and Yang, Yan and Chai, Shu and Sun, Liyan and Huang, Yue and Ding, Xinghao},
  booktitle={AAAI Conference on Artificial Intelligence},
  pages={643--651},
  year={2022}
}

@article{DIAMANTI2024105985,
title = {Visual sensing on marine robotics for the 3D documentation of Underwater Cultural Heritage: A review},
journal = {Journal of Archaeological Science},
volume = {166},
pages = {105985},
year = {2024},
issn = {0305-4403},
doi = {https://doi.org/10.1016/j.jas.2024.105985},
url = {https://www.sciencedirect.com/science/article/pii/S0305440324000530},
author = {Eleni Diamanti and Øyvind Ødegård}
}

@inproceedings{Menna:FlatVsDome:2017,
  author = {Menna, Fabio and Nocerino, Erica and Remondino, Fabio},
  title = {Flat Versus Hemispherical Dome Ports in Underwater Photogrammetry},
  booktitle = {The International Archives of the Photogrammetry, Remote Sensing and Spatial Information Sciences},
  volume = {XLII-2/W3},
  pages = {481--487},
  year = {2017},
  doi = {10.5194/isprs-archives-XLII-2-W3-481-2017}
}

@article{She:UnderwaterDomes:2022,
  author = {She, Mengkun and Nakath, David and Song, Yifan and K{\"o}ser, Kevin},
  title = {Refractive Geometry for Underwater Domes},
  journal = {ISPRS Journal of Photogrammetry and Remote Sensing},
  volume = {183},
  pages = {525--540},
  year = {2022},
  doi = {10.1016/j.isprsjprs.2021.11.006}
}

@article{LI2020107038,
title = {Underwater scene prior inspired deep underwater image and video enhancement},
journal = {Pattern Recognition},
volume = {98},
pages = {107038},
year = {2020},
issn = {0031-3203},
doi = {https://doi.org/10.1016/j.patcog.2019.107038},
url = {https://www.sciencedirect.com/science/article/pii/S0031320319303401},
author = {Chongyi Li and Saeed Anwar and Fatih Porikli},
}

@ARTICLE{Zhang:Underwater:2025,
  author={Zhang, Weidong and Liu, Qingmin and Lu, Huimin and Wang, Jianping and Liang, Jing},
  journal={IEEE Transactions on Circuits and Systems for Video Technology}, 
  title={Underwater Image Enhancement via Wavelet Decomposition Fusion of Advantage Contrast}, 
  year={2025},
  volume={},
  number={},
  pages={1-1},
  doi={10.1109/TCSVT.2025.3545595}}

@article{DU2025125271,
title = {{UIEDP: Boosting} underwater image enhancement with diffusion prior},
journal = {Expert Systems with Applications},
volume = {259},
pages = {125271},
year = {2025},
issn = {0957-4174},
doi = {https://doi.org/10.1016/j.eswa.2024.125271},
url = {https://www.sciencedirect.com/science/article/pii/S0957417424021389},
author = {Dazhao Du and Enhan Li and Lingyu Si and Wenlong Zhai and Fanjiang Xu and Jianwei Niu and Fuchun Sun},

}

@article{WITTMANN2024100072,
title = {Robust marker detection and identification using deep learning in underwater images for close range photogrammetry},
journal = {ISPRS Open Journal of Photogrammetry and Remote Sensing},
volume = {13},
pages = {100072},
year = {2024},
issn = {2667-3932},
doi = {https://doi.org/10.1016/j.ophoto.2024.100072},
url = {https://www.sciencedirect.com/science/article/pii/S2667393224000164},
author = {Jost Wittmann and Sangam Chatterjee and Thomas Sure},
}

@Article{Samboko:evaluating:2022,
AUTHOR = {Samboko, H. T. and Schurer, S. and Savenije, H. H. G. and Makurira, H. and Banda, K. and Winsemius, H.},
TITLE = {Evaluating low-cost topographic surveys for computations of conveyance},
JOURNAL = {Geoscientific Instrumentation, Methods and Data Systems},
VOLUME = {11},
YEAR = {2022},
NUMBER = {1},
PAGES = {1--23},
URL = {https://gi.copernicus.org/articles/11/1/2022/},
DOI = {10.5194/gi-11-1-2022}
}

@book{Aubram2013,
  author    = {Daniel Aubram},
  title     = {An arbitrary Lagrangian-Eulerian method for penetration into sand at finite deformation},
  year      = {2013},
  publisher = {Shaker, Aachen},
  doi       = {10.14279/depositonce-3958},
  isbn      = {978-3-8440-2507-1}
}

@article{Istenic2019,
  author    = {Istenič, K. and Gracias, N. and Arnaubec, A. and Escartín, J. and Garcia, R.},
  title     = {Scale Accuracy Evaluation of Image-Based 3D Reconstruction Strategies Using Laser Photogrammetry},
  journal   = {Remote Sensing},
  year      = {2019},
  volume    = {11},
  number    = {18},
  pages     = {2093},
  doi       = {10.3390/rs11182093}
}

@inproceedings{gough2025aquanerf,
  author    = {Luca Gough and Adrian Azzarelli and Fan Zhang and Nantheera Anantrasirichai},
  title     = {AquaNeRF: Neural Radiance Fields in Underwater Media with Distractor Removal},
  booktitle = {IEEE International Symposium on Circuits and Systems},
  year      = {2025},
}

@article{Anantrasirichai2025AI,
  author    = {Nantheera Anantrasirichai and Fan Zhang and David Bull},
  title     = {Artificial Intelligence in Creative Industries: Advances Prior to 2025},
  journal   = {arXiv preprint arXiv:2501.02725},
  year      = {2025}
}

@article{LU2023103926,
title = {Underwater image enhancement method based on denoising diffusion probabilistic model},
journal = {Journal of Visual Communication and Image Representation},
volume = {96},
pages = {103926},
year = {2023},
issn = {1047-3203},
doi = {https://doi.org/10.1016/j.jvcir.2023.103926},
url = {https://www.sciencedirect.com/science/article/pii/S1047320323001761},
author = {Siqi Lu and Fengxu Guan and Hanyu Zhang and Haitao Lai}
}

@article{CHEN2025129274,
title = {{BDMUIE: Underwater} image enhancement based on Bayesian diffusion model},
journal = {Neurocomputing},
volume = {620},
pages = {129274},
year = {2025},
issn = {0925-2312},
doi = {https://doi.org/10.1016/j.neucom.2024.129274},
url = {https://www.sciencedirect.com/science/article/pii/S0925231224020459},
author = {Lingfeng Chen and Zhihan Xu and Chao Wei and Yuanxin Xu}
}

@ARTICLE{Lu:speed:2024,
  author={Lu, Siqi and Guan, Fengxu and Zhang, Hanyu and Lai, Haitao},
  journal={IEEE Transactions on Circuits and Systems for Video Technology}, 
  title={Speed-Up DDPM for Real-Time Underwater Image Enhancement}, 
  year={2024},
  volume={34},
  number={5},
  pages={3576-3588},
  doi={10.1109/TCSVT.2023.3314767}}

@ARTICLE{Xia:patch:2025,
  author={Xia, Haisheng and Bao, Binglei and Liao, Fei and Chen, Jintao and Wang, Binglu and Li, Zhijun},
  journal={IEEE Transactions on Cybernetics}, 
  title={A Patch-Based Method for Underwater Image Enhancement With Denoising Diffusion Models}, 
  year={2025},
  volume={55},
  number={1},
  pages={269-281},
  doi={10.1109/TCYB.2024.3482174}}

@article{yan2023hybrur,
  title={Hybrur: A hybrid physical-neural solution for unsupervised underwater image restoration},
  author={Yan, Shuaizheng and Chen, Xingyu and Wu, Zhengxing and Tan, Min and Yu, Junzhi},
  journal={IEEE Transactions on Image Processing},
  year={2023}
}

@inproceedings{desai2021ruig,
  title={Ruig: Realistic underwater image generation towards restoration},
  author={Desai, Chaitra and Tabib, Ramesh Ashok and Reddy, Sai Sudheer and Patil, Ujwala and Mudenagudi, Uma},
  booktitle={IEEE Conference on Computer Vision and Pattern Recognition Workshops},
  pages={2181--2189},
  year={2021}
}

@article{zhou2024hclr,
  title={HCLR-Net: Hybrid Contrastive Learning Regularization with Locally Randomized Perturbation for Underwater Image Enhancement},
  author={Zhou, Jingchun and Sun, Jiaming and Li, Chongyi and Jiang, Qiuping and Zhou, Man and Lam, Kin-Man and Zhang, Weishi and Fu, Xianping},
  journal={International Journal of Computer Vision},
  pages={1--25},
  year={2024}
}

@article{zhou2023waterhe,
  title={WaterHE-NeRF: Water-ray Matching Neural Radiance Fields for Underwater Scene Reconstruction},
  author={Zhou, Jingchun and Liang, Tianyu and Zhang, Dehuan and Liu, Siyuan and Wang, Junsheng and Wu, Edmond Q.},
  journal={Information Fusion},
  volume={115},
  pages={102770},
  year={2025},
  doi={10.1016/j.inffus.2024.102770}
}

@inproceedings{islam2020simultaneous,
  title={Simultaneous Enhancement and Super-Resolution of Underwater Imagery for Improved Visual Perception},
  author={Islam, Md Jahidul and Luo, Peigen and Sattar, Junaed},
  booktitle={16th Robotics: Science and Systems, RSS 2020},
  year={2020}
}

@article{jaffe1990computer,
  title={Computer modeling and the design of optimal underwater imaging systems},
  author={Jaffe, Jules S},
  journal={IEEE Journal of Oceanic Engineering},
  volume={15},
  number={2},
  pages={101--111},
  year={1990},
  publisher={IEEE}
}

@inproceedings{liu2021swin,
  title={Swin transformer: Hierarchical vision transformer using shifted windows},
  author={Liu, Ze and Lin, Yutong and Cao, Yue and Hu, Han and Wei, Yixuan and Zhang, Zheng and Lin, Stephen and Guo, Baining},
  booktitle={IEEE International Conference on Computer Vision},
  pages={10012--10022},
  year={2021}
}

@article{narasimhan2002vision,
  title={Vision and the atmosphere},
  author={Narasimhan, Srinivasa G and Nayar, Shree K},
  journal={International journal of computer vision},
  volume={48},
  pages={233--254},
  year={2002},
  publisher={Springer}
}

@article{bryson2016true,
  title={True color correction of autonomous underwater vehicle imagery},
  author={Bryson, Mitch and Johnson-Roberson, Matthew and Pizarro, Oscar and Williams, Stefan B},
  journal={Journal of Field Robotics},
  volume={33},
  number={6},
  pages={853--874},
  year={2016},
  publisher={Wiley Online Library}
}

@article{peng2018generalization,
  title={Generalization of the dark channel prior for single image restoration},
  author={Peng, Yan-Tsung and Cao, Keming and Cosman, Pamela C},
  journal={IEEE Transactions on Image Processing},
  volume={27},
  number={6},
  pages={2856--2868},
  year={2018},
  publisher={IEEE}
}

@article{Peng:Underwater:2017,
  title = {Underwater {{Image Restoration Based}} on {{Image Blurriness}} and {{Light Absorption}}},
  author = {Peng, Yan-Tsung and Cosman, Pamela C.},
  year = {2017},
  month = apr,
  journal = {IEEE Transactions on Image Processing},
  volume = {26},
  number = {4},
  pages = {1579--1594},
  issn = {1941-0042},
  doi = {10.1109/TIP.2017.2663846},
  urldate = {2025-02-20}
}

@inproceedings{narasimhan2003interactive,
  title={Interactive (de) weathering of an image using physical models},
  author={Narasimhan, Srinivasa G and Nayar, Shree K},
  booktitle={IEEE Workshop on color and photometric Methods in computer Vision},
  volume={6},
  pages={1},
  year={2003},
  organization={France}
}

@inproceedings{li2016single,
  title={Single underwater image restoration by blue-green channels dehazing and red channel correction},
  author={Li, Chongyi and Quo, Jichang and Pang, Yanwei and Chen, Shanji and Wang, Jian},
  booktitle={IEEE International Conference on Acoustics, Speech and Signal Processing},
  pages={1731--1735},
  year={2016}
}

@article{Wang:Single:2018,
  title = {Single {{Underwater Image Restoration Using Adaptive Attenuation-Curve Prior}}},
  author = {Wang, Yi and Liu, Hui and Chau, Lap-Pui},
  year = {2018},
  month = mar,
  journal = {IEEE Transactions on Circuits and Systems I: Regular Papers},
  volume = {65},
  number = {3},
  pages = {992--1002},
  issn = {1558-0806},
  doi = {10.1109/TCSI.2017.2751671},
  urldate = {2025-02-24}
}

@inproceedings{Fu:retinexbased:2014,
  title = {A Retinex-Based Enhancing Approach for Single Underwater Image},
  booktitle = {2014 {{IEEE International Conference}} on {{Image Processing}} ({{ICIP}})},
  author = {Fu, Xueyang and Zhuang, Peixian and Huang, Yue and Liao, Yinghao and Zhang, Xiao-Ping and Ding, Xinghao},
  year = {2014},
  month = oct,
  pages = {4572--4576},
  issn = {2381-8549},
  doi = {10.1109/ICIP.2014.7025927},
  urldate = {2025-02-18}
}

@inproceedings{DrewsJr:Transmission:2013,
  title = {Transmission {{Estimation}} in {{Underwater Single Images}}},
  booktitle = {2013 {{IEEE International Conference}} on {{Computer Vision Workshops}}},
  author = {Drews Jr, P. and Do Nascimento, E. and Moraes, F. and Botelho, S. and Campos, M.},
  year = {2013},
  month = dec,
  pages = {825--830},
  publisher = {IEEE},
  address = {Sydney, Australia},
  doi = {10.1109/ICCVW.2013.113},
  urldate = {2025-02-07},
  isbn = {978-1-4799-3022-7},
  langid = {english}
}

@article{WEI2025104222,
title = {Leveraging vision-language prompts for real-world image restoration and enhancement},
journal = {Computer Vision and Image Understanding},
volume = {250},
pages = {104222},
year = {2025},
issn = {1077-3142},
doi = {https://doi.org/10.1016/j.cviu.2024.104222},
url = {https://www.sciencedirect.com/science/article/pii/S1077314224003035},
author = {Yanyan Wei and Yilin Zhang and Kun Li and Fei Wang and Shengeng Tang and Zhao Zhang}
}

@InProceedings{Zhou_2024_CVPR,
    author    = {Zhou, Shijie and Chang, Haoran and Jiang, Sicheng and Fan, Zhiwen and Zhu, Zehao and Xu, Dejia and Chari, Pradyumna and You, Suya and Wang, Zhangyang and Kadambi, Achuta},
    title     = {Feature 3DGS: Supercharging 3D Gaussian Splatting to Enable Distilled Feature Fields},
    booktitle = {Proceedings of the IEEE/CVF Conference on Computer Vision and Pattern Recognition (CVPR)},
    month     = {June},
    year      = {2024},
    pages     = {21676-21685}
}

@inproceedings{radford2021learning,
  title={Learning transferable visual models from natural language supervision},
  author={Radford, Alec and Kim, Jong Wook and Hallacy, Chris and Ramesh, Aditya and Goh, Gabriel and Agarwal, Sandhini and Sastry, Girish and Askell, Amanda and Mishkin, Pamela and Clark, Jack and others},
  booktitle={International Conference on Machine Learning},
  pages={8748--8763},
  year={2021},
  organization={PMLR}
}

@article{jiang2020novel,
  title={A novel deep neural network for noise removal from underwater image},
  author={Jiang, Qin and Chen, Yang and Wang, Guoyu and Ji, Tingting},
  journal={Signal Processing: Image Communication},
  volume={87},
  pages={115921},
  year={2020},
  publisher={Elsevier}
}

@inproceedings{Khan:Underwater:2016,
  title={Underwater image enhancement by wavelet based fusion},
  author={Khan, Amjad and Ali, Syed Saad Azhar and Malik, Aamir Saeed and Anwer, Atif and Meriaudeau, Fabrice},
  booktitle={2016 IEEE International Conference on Underwater System Technology: Theory and Applications (USYS)},
  pages={83--88},
  year={2016},
  organization={IEEE}
}

@article{Mertens:Exposure:2009,
  title = {Exposure {{Fusion}}: {{A Simple}} and {{Practical Alternative}} to {{High Dynamic Range Photography}}},
  shorttitle = {Exposure {{Fusion}}},
  author = {Mertens, T. and Kautz, J. and Van Reeth, F.},
  year = {2009},
  journal = {Computer Graphics Forum},
  volume = {28},
  number = {1},
  pages = {161--171},
  issn = {1467-8659},
  doi = {10.1111/j.1467-8659.2008.01171.x},
  urldate = {2025-02-25},
  langid = {english}
}

@article{Ancuti:Single:2013,
  title = {Single {{Image Dehazing}} by {{Multi-Scale Fusion}}},
  author = {Ancuti, Codruta Orniana and Ancuti, Cosmin},
  year = {2013},
  month = aug,
  journal = {IEEE Transactions on Image Processing},
  volume = {22},
  number = {8},
  pages = {3271--3282},
  issn = {1941-0042},
  doi = {10.1109/TIP.2013.2262284},
  urldate = {2025-02-25}
}

@article{Drews:Underwater:2016,
  title = {Underwater {{Depth Estimation}} and {{Image Restoration Based}} on {{Single Images}}},
  author = {Drews, Paulo L.J. and Nascimento, Erickson R. and Botelho, Silvia S.C. and Montenegro Campos, Mario Fernando},
  year = {2016},
  month = mar,
  journal = {IEEE Computer Graphics and Applications},
  volume = {36},
  number = {2},
  pages = {24--35},
  issn = {1558-1756},
  doi = {10.1109/MCG.2016.26},
  urldate = {2025-02-14}
}

@article{Chiang:Underwater:2012,
  title = {Underwater {{Image Enhancement}} by {{Wavelength Compensation}} and {{Dehazing}}},
  author = {Chiang, John Y. and Chen, Ying-Ching},
  year = {2012},
  month = apr,
  journal = {IEEE Transactions on Image Processing},
  volume = {21},
  number = {4},
  pages = {1756--1769},
  issn = {1941-0042},
  doi = {10.1109/TIP.2011.2179666},
  urldate = {2025-02-07}
}

@inproceedings{Tarel:Fast:2009,
  title = {Fast Visibility Restoration from a Single Color or Gray Level Image},
  booktitle = {2009 {{IEEE}} 12th {{International Conference}} on {{Computer Vision}}},
  author = {Tarel, Jean-Philippe and Hautiere, Nicolas},
  year = {2009},
  month = sep,
  pages = {2201--2208},
  publisher = {IEEE},
  address = {Kyoto},
  doi = {10.1109/ICCV.2009.5459251},
  urldate = {2025-02-19},
  isbn = {978-1-4244-4420-5},
  langid = {english}
}

@inproceedings{Emberton:Hierarchical:2015,
  title = {Hierarchical Rank-Based Veiling Light Estimation for Underwater Dehazing},
  booktitle = {Procedings of the {{British Machine Vision Conference}} 2015},
  author = {Emberton, Simon and Chittka, Lars and Cavallaro, Andrea},
  year = {2015},
  pages = {125.1-125.12},
  publisher = {British Machine Vision Association},
  address = {Swansea},
  doi = {10.5244/C.29.125},
  urldate = {2025-02-24},
  isbn = {978-1-901725-53-7},
  langid = {english}
}

@inproceedings{Wen:Single:2013,
  title = {Single Underwater Image Enhancement with a New Optical Model},
  booktitle = {2013 {{IEEE International Symposium}} on {{Circuits}} and {{Systems}} ({{ISCAS}})},
  author = {Wen, Haocheng and Tian, Yonghong and Huang, Tiejun and Gao, Wen},
  year = {2013},
  month = may,
  pages = {753--756},
  issn = {2158-1525},
  doi = {10.1109/ISCAS.2013.6571956},
  urldate = {2025-02-24}
}

@article{Levin:ClosedForm:2008,
  title = {A {{Closed-Form Solution}} to {{Natural Image Matting}}},
  author = {Levin, Anat and Lischinski, Dani and Weiss, Yair},
  year = {2008},
  month = feb,
  journal = {IEEE Transactions on Pattern Analysis and Machine Intelligence},
  volume = {30},
  number = {2},
  pages = {228--242},
  issn = {1939-3539},
  doi = {10.1109/TPAMI.2007.1177},
  urldate = {2025-02-20}
}

@inproceedings{Yang:Low:2011,
  title = {Low {{Complexity Underwater Image Enhancement Based}} on {{Dark Channel Prior}}},
  booktitle = {2011 {{Second International Conference}} on {{Innovations}} in {{Bio-inspired Computing}} and {{Applications}}},
  author = {Yang, Hung-Yu and Chen, Pei-Yin and Huang, Chien-Chuan and Zhuang, Ya-Zhu and Shiau, Yeu-Horng},
  year = {2011},
  month = dec,
  pages = {17--20},
  doi = {10.1109/IBICA.2011.9},
  urldate = {2025-02-20}
}

@article{Gibson:Investigation:2012,
  title = {An {{Investigation}} of {{Dehazing Effects}} on {{Image}} and {{Video Coding}}},
  author = {Gibson, Kristofor B. and Vo, Dung T. and Nguyen, Truong Q.},
  year = {2012},
  month = feb,
  journal = {IEEE Transactions on Image Processing},
  volume = {21},
  number = {2},
  pages = {662--673},
  issn = {1941-0042},
  doi = {10.1109/TIP.2011.2166968},
  urldate = {2025-02-19}
}

@article{Mohan:Underwater:2020,
  title = {Underwater {{Image Enhancement}} Based on {{Histogram Manipulation}} and {{Multiscale Fusion}}},
  author = {Mohan, Sangeetha and Simon, Philomina},
  year = {2020},
  journal = {Procedia Computer Science},
  volume = {171},
  pages = {941--950},
  issn = {18770509},
  doi = {10.1016/j.procs.2020.04.102},
  urldate = {2025-02-18},
  langid = {english}
}

@article{Sethi:Fusion:2019,
  title = {Fusion of {{Underwater Image Enhancement}} and {{Restoration}}},
  author = {Sethi, Rajni and Indu, Sreedevi},
  year = {2019},
  month = jul,
  journal = {International Journal of Pattern Recognition and Artificial Intelligence},
  publisher = {World Scientific Publishing Company},
  doi = {10.1142/S0218001420540075},
  urldate = {2025-02-18},
  langid = {english}
}

@article{Zhang:Single:2019,
  title = {Single {{Image Defogging Based}} on {{Multi-Channel Convolutional MSRCR}}},
  author = {Zhang, Weidong and Dong, Lili and Pan, Xipeng and Zhou, Jingchun and Qin, Li and Xu, Wenhai},
  year = {2019},
  journal = {IEEE Access},
  volume = {7},
  pages = {72492--72504},
  issn = {2169-3536},
  doi = {10.1109/ACCESS.2019.2920403},
  urldate = {2025-02-18}
}

@article{MohdAzmi:Naturalbased:2019,
  title = {Natural-Based Underwater Image Color Enhancement through Fusion of Swarm-Intelligence Algorithm},
  author = {Mohd Azmi, Kamil Zakwan and Abdul Ghani, Ahmad Shahrizan and Md Yusof, Zulkifli and Ibrahim, Zuwairie},
  year = {2019},
  month = dec,
  journal = {Applied Soft Computing},
  volume = {85},
  pages = {105810},
  issn = {15684946},
  doi = {10.1016/j.asoc.2019.105810},
  urldate = {2025-02-18},
  langid = {english}
}

@inproceedings{Joshi:Quantification:2008,
  title = {Quantification of Retinex in Enhancement of Weather Degraded Images},
  booktitle = {2008 {{International Conference}} on {{Audio}}, {{Language}} and {{Image Processing}}},
  author = {Joshi, K. R. and Kamathe, R. S.},
  year = {2008},
  month = jul,
  pages = {1229--1233},
  doi = {10.1109/ICALIP.2008.4590120},
  urldate = {2025-02-18}
}

@article{Zhuang:Bayesian:2021,
  title = {Bayesian Retinex Underwater Image Enhancement},
  author = {Zhuang, Peixian and Li, Chongyi and Wu, Jiamin},
  year = {2021},
  month = may,
  journal = {Engineering Applications of Artificial Intelligence},
  volume = {101},
  pages = {104171},
  issn = {09521976},
  doi = {10.1016/j.engappai.2021.104171},
  urldate = {2025-02-11},
  langid = {english}
}

@inproceedings{zamir2022restormer,
  title={Restormer: Efficient transformer for high-resolution image restoration},
  author={Zamir, Syed Waqas and Arora, Aditya and Khan, Salman and Hayat, Munawar and Khan, Fahad Shahbaz and Yang, Ming-Hsuan},
  booktitle={Proceedings of the IEEE/CVF Conference on Computer Vision and Pattern Recognition (CVPR)},
  pages={5728--5739},
  year={2022}
}

@inproceedings{zhao2024wavelet,
  title={Wavelet-based fourier information interaction with frequency diffusion adjustment for underwater image restoration},
  author={Zhao, Chen and Cai, Weiling and Dong, Chenyu and Hu, Chengwei},
  booktitle={Proceedings of the IEEE/CVF Conference on Computer Vision and Pattern Recognition},
  pages={8281--8291},
  year={2024}
}

@article{guan2023diffwater,
  title={DiffWater: Underwater image enhancement based on conditional denoising diffusion probabilistic model},
  author={Guan, Meisheng and Xu, Haiyong and Jiang, Gangyi and Yu, Mei and Chen, Yeyao and Luo, Ting and Zhang, Xuebo},
  journal={IEEE Journal of Selected Topics in Applied Earth Observations and Remote Sensing},
  volume={17},
  pages={2319--2335},
  year={2023},
  publisher={IEEE}
}

@article{zhang2024dcgf,
  title={DCGF: Diffusion-Color Guided Framework for Underwater Image Enhancement},
  author={Zhang, Yuhan and Yuan, Jieyu and Cai, Zhanchuan},
  journal={IEEE Transactions on Geoscience and Remote Sensing},
  year={2024},
  publisher={IEEE}
}

@article{wang2024uie,
  title={UIE-convformer: Underwater image enhancement based on convolution and feature fusion transformer},
  author={Wang, Biao and Xu, Haiyong and Jiang, Gangyi and Yu, Mei and Ren, Tingdi and Luo, Ting and Zhu, Zhongjie},
  journal={IEEE Transactions on Emerging Topics in Computational Intelligence},
  volume={8},
  number={2},
  pages={1952--1968},
  year={2024},
  publisher={IEEE}
}

@inproceedings{pramanick2024x,
  title={X-caunet: Cross-color channel attention with underwater image-enhancing transformer},
  author={Pramanick, Alik and Sarma, Sandipan and Sur, Arijit},
  booktitle={ICASSP 2024-2024 IEEE International Conference on Acoustics, Speech and Signal Processing (ICASSP)},
  pages={3550--3554},
  year={2024},
  organization={IEEE}
}

@article{zhuang2024globally,
  title={Globally deformable information selection transformer for underwater image enhancement},
  author={Zhuang, Junbin and Zheng, Yan and Guo, Baolong and Yan, Yunyi},
  journal={IEEE Transactions on Circuits and Systems for Video Technology},
  year={2024},
  publisher={IEEE}
}

@article{jiang2023five,
  title={Five A$^{+}$ Network: You Only Need 9K Parameters for Underwater Image Enhancement},
  author={Jiang, Jingxia and Ye, Tian and Bai, Jinbin and Chen, Sixiang and Chai, Wenhao and Jun, Shi and Liu, Yun and Chen, Erkang},
  journal={British Machine Vision Conference (BMVC)},
  year={2023}
}

@article{cong2023pugan,
  title={Pugan: Physical model-guided underwater image enhancement using gan with dual-discriminators},
  author={Cong, Runmin and Yang, Wenyu and Zhang, Wei and Li, Chongyi and Guo, Chun-Le and Huang, Qingming and Kwong, Sam},
  journal={IEEE Transactions on Image Processing},
  volume={32},
  pages={4472--4485},
  year={2023},
  publisher={IEEE}
}

@article{qu2024z,
  title={{Z-Splat: Z-Axis Gaussian Splatting for Camera-Sonar Fusion}},
  author={Qu, Ziyuan and Vengurlekar, Omkar and Qadri, Mohamad and Zhang, Kevin and Kaess, Michael and Metzler, Christopher and Jayasuriya, Suren and Pediredla, Adithya},
  journal={IEEE Transactions on Pattern Analysis and Machine Intelligence},
  year={2024},
  publisher={IEEE}
}

@article{Teague2017Underwater,
  author = {Teague, Jonathan and Scott, Tom},
  title = {Underwater Photogrammetry and 3D Reconstruction of Submerged Objects in Shallow Environments by ROV and Underwater GPS},
  journal = {Journal of Marine Science Research and Technology},
  year = {2017},
  month = {September},
  day = {25}
}

@article{ZHANG2022100510,
title = {Visual {SLAM} for Underwater Vehicles: A Survey},
journal = {Computer Science Review},
volume = {46},
pages = {100510},
year = {2022},
issn = {1574-0137},
doi = {https://doi.org/10.1016/j.cosrev.2022.100510},
url = {https://www.sciencedirect.com/science/article/pii/S1574013722000442},
author = {Song Zhang and Shili Zhao and Dong An and Jincun Liu and He Wang and Yu Feng and Daoliang Li and Ran Zhao}
}

@inproceedings{yang2024seasplat,
  title={{SeaSplat: Representing Underwater Scenes with 3D Gaussian Splatting and a Physically Grounded Image Formation Model}},
  author={Yang, Daniel and Leonard, John J and Girdhar, Yogesh},
  booktitle={International Conference on Robotics and Automation},
  year={2025}
}

@article{li2024watersplatting,
  title={{WaterSplatting: Fast} Underwater 3D Scene Reconstruction Using Gaussian Splatting},
  author={Li, Huapeng and Song, Wenxuan and Xu, Tianao and Elsig, Alexandre and Kulhanek, Jonas},
  journal={International Conference on 3D Vision},
  year={2025}
}

@article{zhang2024recgs,
  title={{RecGS: Removing Water Caustic with Recurrent Gaussian Splatting}},
  author={Zhang, Tianyi and Zhi, Weiming and Meyers, Braden and Durrant, Nelson and Huang, Kaining and Mangelson, Joshua and Barbalata, Corina and Johnson-Roberson, Matthew},
  journal={IEEE Robotics and Automation Letters},
  year={2024},
  publisher={IEEE}
}

@article{Lacka2024,
  author = {Małgorzata Łacka and Jacek Łubczonek},
  title = {Methodology for Creating a Digital Bathymetric Model Using Neural Networks for Combined Hydroacoustic and Photogrammetric Data in Shallow Water Areas},
  journal = {Sensors},
  year = {2024},
  volume = {24},
  number = {1},
  pages = {175},
  doi = {10.3390/s24010175}
}

@article{Cheng2022,
  author = {Cheng, C. and Wang, C. and Yang, D. and Liu, W. and Zhang, F.},
  title = {Underwater Localization and Mapping Based on Multi-Beam Forward Looking Sonar},
  journal = {Frontiers in Neurorobotics},
  volume = {15},
  pages = {801956},
  year = {2022},
  doi = {10.3389/fnbot.2021.801956}
}

@misc{bvi-coral,
  author       = {Anantrasirichai, Nantheera},
  title        = {{BVI-Coral: Underwater scenes for 3D reconstruction}},
  year         = {2024},
  month        = apr,
  day          = {30},
  publisher    = {Zenodo},
  doi          = {10.5281/zenodo.11093417},
  url          = {https://doi.org/10.5281/zenodo.11093417}
}

@INPROCEEDINGS{Rahman:SVIn2:2019,
  author={Rahman, Sharmin and Li, Alberto Quattrini and Rekleitis, Ioannis},
  booktitle={2019 IEEE/RSJ International Conference on Intelligent Robots and Systems (IROS)}, 
  title={{SVIn2: An} Underwater SLAM System using Sonar, Visual, Inertial, and Depth Sensor}, 
  year={2019},
  volume={},
  number={},
  pages={1861-1868},
  doi={10.1109/IROS40897.2019.8967703}}

@ARTICLE{Zhang:Integration:2024,
  author={Zhang, Jiawei and Han, Fenglei and Han, Duanfeng and Yang, Jianfeng and Zhao, Wangyuan and Li, Hansheng},
  journal={IEEE Sensors Journal}, 
  title={Integration of Sonar and Visual–Inertial Systems for SLAM in Underwater Environments}, 
  year={2024},
  volume={24},
  number={10},
  pages={16792-16804},
  doi={10.1109/JSEN.2024.3384301}}

@article{mualem2024gaussian,
  title={Gaussian Splashing: Direct Volumetric Rendering Underwater},
  author={Mualem, Nir and Amoyal, Roy and Freifeld, Oren and Akkaynak, Derya},
  journal={arXiv preprint arXiv:2411.19588},
  year={2024}
}

@article{liu2024aquatic,
  title={Aquatic-GS: A Hybrid 3D Representation for Underwater Scenes},
  author={Liu, Shaohua and Lu, Junzhe and Gu, Zuoya and Li, Jiajun and Deng, Yue},
  journal={arXiv preprint arXiv:2411.00239},
  year={2024}
}

@inproceedings{Skinner:2016ab,
	Author = {Katherine A. Skinner and Eduardo Iscar Ruland and M. Johnson-Roberson},
	Booktitle = {{IEEE} International Conference on Robotics and Automation},
	Title = {Automatic Color Correction for 3D Reconstruction of Underwater Scenes},
	Year = {2017}}

@article{mueller2022instant,
  title = {Instant Neural Graphics Primitives with a Multiresolution Hash Encoding},
  author = {M{\"u}ller, Thomas and Evans, Alex and Schied, Christoph and Keller, Alexander},
  journal = {ACM Transactions on Graphics (Proc. SIGGRAPH)},
  volume = {41},
  number = {4},
  pages = {102:1--102:15},
  year = {2022}
}

@inproceedings{snavely2006photo,
  title = {Photo Tourism: Exploring Photo Collections in 3D},
  author = {Snavely, Noah and Seitz, Steven M. and Szeliski, Richard},
  booktitle = {ACM SIGGRAPH 2006 Papers},
  pages = {835--846},
  year = {2006}
}

@inproceedings{seitz2006mvs,
  title = {A Comparison and Evaluation of Multi-View Stereo Reconstruction Algorithms},
  author = {Seitz, Steven M. and Curless, Brian and Diebel, James and Scharstein, Daniel and Szeliski, Richard},
  booktitle = {Proceedings of the IEEE Computer Vision and Pattern Recognition (CVPR)},
  pages = {519--528},
  year = {2006}
}

@article{woodham1980photometric,
  title = {Photometric method for determining surface orientation from multiple images},
  author = {Woodham, Robert J.},
  journal = {Optical Engineering},
  volume = {19},
  number = {1},
  pages = {139--144},
  year = {1980}
}

@inproceedings{sedlazeck2009rov,
  title = {3D reconstruction based on underwater video from ROV \emph{Kiel 6000} considering underwater imaging conditions},
  author = {Sedlazeck, Axel and Koser, Kevin and Koch, Reinhard},
  booktitle = {Proceedings of OCEANS 2009-Europe},
  pages = {1--10},
  year = {2009}
}

@article{Kajiya:rendering:1986,
  title = {The Rendering Equation},
  author = {Kajiya, James T.},
  year = {1986},
  month = aug,
  journal = {ACM SIGGRAPH Computer Graphics},
  volume = {20},
  number = {4},
  pages = {143--150},
  issn = {0097-8930},
  doi = {10.1145/15886.15902},
  urldate = {2025-02-26},
  langid = {english}
}

@article{chen2022tensorf,
  title={TensoRF: Tensorial Radiance Fields},
  author={Chen, Anpei and Xu, Zexiang and Geiger, Andreas and Yu, Jingyi and Su, Hao},
  journal={ECCV},
  year={2022}
}

@article{huang2026bayesian,
  title={Bayesian Neural Networks for One-to-Many Mapping in Image Enhancement},
  author={Huang, Guoxi and Yang, Qirui and Qi, Zipeng and Lin, RuiRui and Bull, David and Anantrasirichai, Nantheera},
  journal={Proceedings of the AAAI Conference on Artificial Intelligence},
  year={2026}
}

@inproceedings{fridovich2022plenoxels,
  title     = {{Plenoxels: Radiance Fields without Neural Networks}},
  author    = {Fridovich-Keil, Sara and Yu, Alex and Tancik, Matthew and Chen, Qinhong and Recht, Benjamin and Kanazawa, Angjoo},
  booktitle = {CVPR},
  year      = {2022}
}

@inproceedings{martinbrualla2021nerfw,
  title     = {{NeRF in the Wild: Neural Radiance Fields for Unconstrained Photo Collections}},
  author    = {Martin-Brualla, Ricardo and Radwan, Noha and  Sajjadi, Mehdi S. M. and Barron, Jonathan T. and Dosovitskiy, Alexey and Duckworth, Daniel},
  booktitle = {CVPR},
  year      = {2021}
}

@inproceedings{kim2023upnerf,
  title     = {{UP-NeRF: Unconstrained Pose-Prior-Free Neural Radiance Fields}},
  author    = {Kim, Injae and Choi, Minhyuk and Kim, Hyunwoo J.},
  booktitle = {NeurIPS},
  year      = {2023}
}

@inproceedings{pumarola2021dnerf,
  title     = {{D-NeRF: Neural Radiance Fields for Dynamic Scenes}},
  author    = {Pumarola, Albert and  Corona, Enrique and Pons-Moll, Gerard and Moreno-Noguer, Francesc},
  booktitle = {CVPR},
  year      = {2021}
}

@inproceedings{xian2021videonerf,
  title     = {{Space-time Neural Irradiance Fields for Free-Viewpoint Video}},
  author    = {Xian, Wei and Bao, Jianmin and Zhang, Tiancheng and Chen, Dong and Wen, Fang and Guo, Baining},
  booktitle = {CVPR},
  year      = {2021}
}

@inproceedings{du2020neuralflow,
  title     = {Neural Radiance Flow for 4D View Synthesis and Video Processing},
  author    = {Du, Yilun and Zhang, Yinan and Yu, Hong-Xing and Tenenbaum, Joshua B. and Wu, Jiajun},
  booktitle = {Proceedings of the IEEE/CVF International Conference on Computer Vision (ICCV)},
  month     = {October},
  year      = {2021},
  pages     = {14324--14334}
}

@inproceedings{lin2021barf,
  title     = {{BARF: Bundle-Adjusting Neural Radiance Fields}},
  author    = {Lin, Chen-Hsuan and  et al.},
  booktitle = {ICCV},
  year      = {2021}
}

@article{Ju:Towards:2025,
title = {Towards marine snow removal with fusing Fourier information},
journal = {Information Fusion},
volume = {117},
pages = {102810},
year = {2025},
issn = {1566-2535},
doi = {https://doi.org/10.1016/j.inffus.2024.102810},
url = {https://www.sciencedirect.com/science/article/pii/S1566253524005888},
author = {Yakun Ju and Jun Xiao and Cong Zhang and Hao Xie and Anwei Luo and Huiyu Zhou and Junyu Dong and Alex C. Kot},
}

@article{jiang2025swagsplatting,
  title     = {{SWAGSplatting: Semantic}-guided Water-scene Augmented Gaussian Splatting},
  author    = {Jiang, Zhuodong and Wang, Haoran and Huang, Guoxi and Seymour, Brett and Anantrasirichai, Nantheera},
  year      = {2025},
  journal   = {arXiv:2509.00800}, 
}

@inproceedings{jiang2025rusplatting,
  title     = {{RUSplatting: Robust} 3D Gaussian Splatting for Sparse-View Underwater Scene Reconstruction},
  author    = {Jiang, Zhuodong and Wang, Haoran and Huang, Guoxi and Seymour, Brett and Anantrasirichai, Nantheera},
  booktitle = {Proceedings of the British Machine Vision Conference (BMVC)},
  year      = {2025},
  organization = {BMVA},
}

@INPROCEEDINGS{Barron:Zip:2023,
  author={Barron, Jonathan T. and Mildenhall, Ben and Verbin, Dor and Srinivasan, Pratul P. and Hedman, Peter},
  booktitle={2023 IEEE/CVF International Conference on Computer Vision (ICCV)}, 
  title={Zip-NeRF: Anti-Aliased Grid-Based Neural Radiance Fields}, 
  year={2023},
  volume={},
  number={},
  pages={19640-19648},
  doi={10.1109/ICCV51070.2023.01804}}

@inproceedings{Yi:AtlantisGS:2025,
author = {Yi, Jingjun and Bi, Qi and Zheng, Hao and Huang, Huimin and Zhan, Haolan and Shen, Yixian and Ji, Wei and Huang, Yawen and Li, Yuexiang and Wu, Xian and Zheng, Yefeng},
title = {{AtlantisGS: Underwater} Sparse-View Scene Reconstruction via Gaussian Splatting},
year = {2025},
booktitle = {Proceedings of the 33rd ACM International Conference on Multimedia},
pages = {7805--7814},
numpages = {10},
doi = {10.1145/3746027.3755125}
}

\end{document}